\PassOptionsToPackage{table}{xcolor}
\documentclass[nohyperref]{article}

\usepackage{microtype}
\usepackage{graphicx}
\usepackage{subfigure}
\usepackage{booktabs} %
\usepackage{placeins}
\usepackage{multicol,multirow,array,graphicx}

\usepackage[hidelinks]{hyperref}

\usepackage[accepted]{icml2023}

\usepackage{amsmath}
\usepackage{amssymb}
\usepackage{mathtools}
\usepackage{amsthm}
\usepackage{dsfont}
\usepackage{relsize}
\usepackage{scalefnt}
\usepackage{newunicodechar}
\usepackage{enumitem}
\usepackage{fontawesome5}

\usepackage[capitalize,noabbrev]{cleveref}

\theoremstyle{plain}

\theoremstyle{definition}

\theoremstyle{remark}

\usepackage[textsize=tiny]{todonotes}
\usepackage{xcolor,colortbl}
\usepackage{numprint}
\usepackage[normalem]{ulem}
\usepackage{ifthen}
\usepackage{pifont}%
\newcommand{\cmark}{\ding{51}}%
\newcommand{\xmark}{\ding{55}}%

\usepackage{tikz}
\usetikzlibrary{calc}

\usepackage{pgfplots}
\DeclareUnicodeCharacter{2212}{−}
\usepgfplotslibrary{groupplots,dateplot}
\usetikzlibrary{patterns,shapes.arrows}
\pgfplotsset{compat=newest}

\usepackage{xcolor,soul}
\sethlcolor{green}

\usepackage{bm}

\def\tuple#1=(#2 #3){%
    \expandafter\def\csname#1@1\endcsname{#2}
    \expandafter\def\csname#1@2\endcsname{#3}
}
\def\xtuple#1#2{%
    \csname#1@#2\endcsname
}

\def\triple#1=(#2 #3 #4){%
    \expandafter\def\csname#1@1\endcsname{#2}
    \expandafter\def\csname#1@2\endcsname{#3}
    \expandafter\def\csname#1@3\endcsname{#4}
}
\def\xtriple#1#2{%
    \csname#1@#2\endcsname
}

\newcolumntype{C}{>{\centering\arraybackslash}X}

\definecolor{lightgrey}{rgb}{0.9, 0.9, 0.9}
\definecolor{darkgreen}{rgb}{0., 0.5, 0.}
\definecolor{darkred}{rgb}{0.5, 0., 0.}
\definecolor{fade}{rgb}{0.6, 0.6, 0.6}
\definecolor{update}{rgb}{0., 0., 0.}

\definecolor{columnbest}{rgb}{0., 0.4, 0.}

\definecolor{places}{rgb}{0.85, .98, 1.}
\definecolor{textures}{rgb}{0.97, 0.9, 1.0}
\definecolor{species}{rgb}{1.0, 1.0, 0.85}
\definecolor{imageneto}{rgb}{1.0, 0.85, 0.85}

\definecolor{placesline}{rgb}{0.65, .92, .95}
\definecolor{texturesline}{rgb}{0.94, 0.7, 1.0}
\definecolor{speciesline}{rgb}{1.0, 1.0, 0.4}
\definecolor{imagenetoline}{rgb}{1.0, 0.6, 0.6}

\definecolor{INclass}{rgb}{0., .0, 1.}
\newcommand{\INclass}[1]{\textcolor{INclass}{\textit{#1}}}

\definecolor{sourceclass}{rgb}{.4, .2, .}
\newcommand{\sourceclass}[1]{\textcolor{sourceclass}{\textit{#1}}}

\definecolor{OODclass}{rgb}{.75, .0, .55}
\newcommand{\OODclass}[1]{\textcolor{OODclass}{\textit{#1}}}

\definecolor{INTOKclass}{rgb}{0., .0, .5}
\newcommand{\INTOKclass}[1]{\textcolor{INTOKclass}{\textit{#1}}}

\definecolor{OODunderlay}{rgb}{.0, .75, .0}
\definecolor{INunderlay}{rgb}{.88, .1, .0}

\definecolor{green}{rgb}{0., .7, .0}

\DeclareRobustCommand{\okina}{%
  \raisebox{\dimexpr\fontcharht\font`A-\height}{%
    \scalebox{0.8}{`}%
  }%
}
\newunicodechar{ʻ}{\okina}

\newcommand{\FPR}{FPR}
\newcommand{\FPRQ}{FPR@TPRQ}

\newcommand{\dsetname}{NINCO} 
\newcommand{\numoodsamples}{\numprint{5879}}

\newcommand{\includegraphicsmaybe}[1]{\IfFileExists{#1}{\includegraphics{#1}}{\includegraphics{example-image}}}

\newcommand{\belowtablevskip}{-7.0mm}
\newcommand{\belowimgcaptionvskip}{-3.0mm}

\icmltitlerunning{In or Out? Fixing ImageNet OOD Detection Evaluation}

\begin{document}

\twocolumn[
\icmltitle{In or Out? Fixing ImageNet Out-of-Distribution Detection Evaluation}

\icmlsetsymbol{equal}{*}

\begin{icmlauthorlist}
\icmlauthor{Julian Bitterwolf}{equal,yyy}
\icmlauthor{Maximilian Müller}{equal,yyy}
\icmlauthor{Matthias Hein}{yyy}
\end{icmlauthorlist}

\icmlaffiliation{yyy}{University of Tübingen and Tübingen AI Center}

\icmlcorrespondingauthor{Julian Bitterwolf}{julian.bitterwolf@uni-tuebingen.de}

\icmlkeywords{Machine Learning, ICML}

\vskip 0.3in
]

\printAffiliationsAndNotice{\icmlEqualContribution} %
\begin{abstract}
Out-of-distribution (OOD) detection is the problem of identifying inputs which are unrelated to the in-distribution task.
The OOD detection performance when the in-distribution (ID) is ImageNet-1K is commonly being tested on a small range of test OOD datasets.
We find that most of the currently used test OOD datasets, including datasets from the open set recognition (OSR) literature, have severe issues: In some cases more than 50$\%$ of the dataset contains objects belonging to one of the ID classes.
These erroneous samples heavily distort the evaluation of OOD detectors.
As a solution, we introduce with \dsetname{} a novel test OOD dataset, each sample checked to be ID free, which with its fine-grained range of OOD classes allows for a detailed analysis of an OOD detector's strengths and failure modes, particularly when paired with a number of synthetic “OOD unit-tests”.
We provide detailed evaluations across a large set of architectures and OOD detection methods on \dsetname{} and the unit-tests, 
revealing new insights about model weaknesses and the effects of pretraining on OOD detection performance.
We provide code and data at \href{https://github.com/j-cb/NINCO}{\textcolor{blue!50!black}{https://github.com/j-cb/NINCO}}.
\end{abstract}
\section{Introduction}
\label{sec:intro}
While deep learning based models have shown impressive performance on many real world tasks, they often exhibit unforeseen behaviour when confronted with unknown situations like receiving an input that is not related to the task it has been trained on.
Such samples are regarded as out-of-distribution (OOD) and deep neural network classifiers are known to make very confident predictions that those belong to one of the \textbf{in-distribution (ID)} classes \cite{hendrycks2017MSP, HeiAndBit2019}.
This unwanted behaviour is a serious obstacle when applying classifiers in real world applications.
The purpose of OOD detectors is to reject OOD inputs, which depending on the application can mean requesting human intervention, steering towards a safe state, or simply abstaining from making a %
prediction, while at the same time letting ID inputs pass through.

Current OOD detection evaluations in image classification rely on the assumption that there is no ID class present in an OOD test image, not even in the background. 
We follow this definition and consider an input to be \textbf{out-of-distribution (OOD)} if it does not contain any of the in-distribution classes.
However, we show that this assumption is not fulfilled for most of the current test OOD datasets for ImageNet-1K (IN-1K) of \citet{ILSVRC15}.
The closely related task of open set recognition (OSR), which simultaneously demands detection of OOD data and high classification accuracy on the ID data, is evaluated on OOD datasets which have the same requirements as in OOD detection.
We also examine the test OOD datasets that have been used in the OSR literature for IN-1K and find similar issues there.
We demonstrate that occurrences of objects from ID classes in test OOD datasets are often correctly recognized by state-of-the-art OOD detectors, but as an unwarranted consequence held against them as mistakes in OOD detection evaluations (false ``false positive'').
Even in cases where current models struggle to identify ID content, e.g. if ID objects are partially occluded or in the background, OOD datasets containing ID objects are not future proof:
when evaluating on them, one would not realize if a future model correctly predicts the class of a visible ID object.

\def \xoffset {2.25cm}
\def \yoffset {3.cm}
\def \imgwidth {1.0cm}
\def \ximgdist {.05cm}
\def \imgheight {1.0cm}
\def \yline {2.03*\imgheight}
\def \yimgdist {.05cm}
\def \ydsetname {.95cm}
\def \toplabeldist {.55cm}
\def \botlabeldist {.82cm}
\def \tinyscale {.75}

\tikzset{
    pics/imgsquare/.style args={#1/#2/#3/#4/#5/#6/#7/#8}{
      code = {\coordinate (o) at (0,0);
      \coordinate (i1) at ($(o) + (-.5*\imgwidth,0) + (-.5*\ximgdist,0)$);
      \coordinate (i2) at ($(o) + (.5*\imgwidth,0) + (.5*\ximgdist,0)$);
      \coordinate (i3) at ($(o) + (-.5*\imgwidth,-\imgheight) + (-.5*\ximgdist,-\yimgdist)$);
      \coordinate (i4) at ($(o) + (.5*\imgwidth,-\imgheight) + (.5*\ximgdist,-\yimgdist)$);
      \coordinate (t1) at ($(i1) + (0,\toplabeldist)$);
      \coordinate (t2) at ($(i2) + (0,\toplabeldist)$);
      \coordinate (t3) at ($(i3) + (0,-\botlabeldist)$);
      \coordinate (t4) at ($(i4) + (0,-\botlabeldist)$);
  \node[inner sep=0pt] at (i1) {\includegraphics[width=\imgwidth, height=\imgheight]{#1}};
  \node[inner sep=0pt] at (i2) {\includegraphics[width=\imgwidth, height=\imgheight]{#2}};
  \node[inner sep=0pt] at (i3) {\includegraphics[width=\imgwidth, height=\imgheight]{#3}};
  \node[inner sep=0pt] at (i4) {\includegraphics[width=\imgwidth, height=\imgheight]{#4}};
  \draw[anchor=base, align=center] (t1)  node[scale=\tinyscale] {\scriptsize\INclass{\xtuple{#5}{1}}\\[-2.2mm]\tiny\textcolor{sourceclass}{(\xtuple{#5}{2})}};
  \draw[anchor=base, align=center] (t2)  node[scale=\tinyscale] {\scriptsize\INclass{\xtuple{#6}{1}}\\[-2.2mm]\tiny\textcolor{sourceclass}{(\xtuple{#6}{2})}};
  \draw[anchor=base, align=center] (t3)  node[scale=\tinyscale] {\scriptsize\INclass{\xtuple{#7}{1}}\\[-2.2mm]\tiny\textcolor{sourceclass}{(\xtuple{#7}{2})}};
  \draw[anchor=base, align=center] (t4)  node[scale=\tinyscale] {\scriptsize\INclass{\xtuple{#8}{1}}\\[-2.2mm]\tiny\textcolor{sourceclass}{(\xtuple{#8}{2})}};
  }}}


\newcommand{\texturesia}{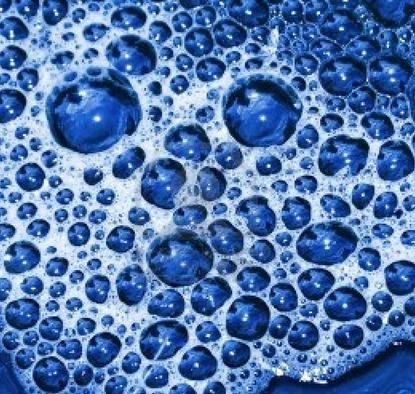}
\newcommand{\texturesib}{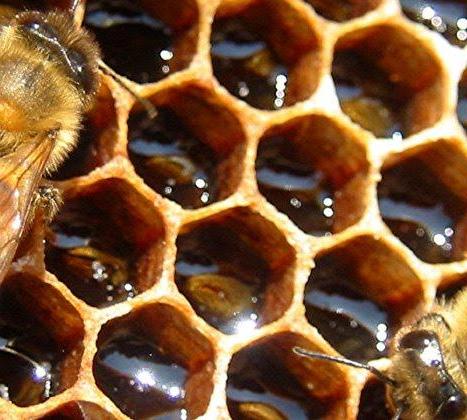}
\newcommand{\texturesic}{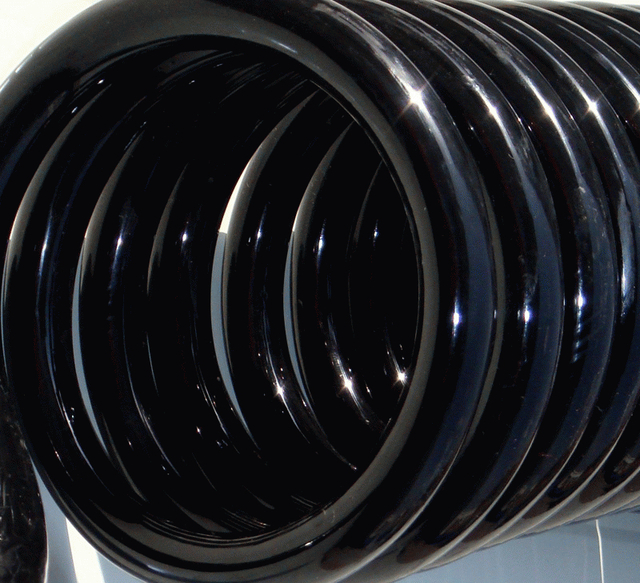}
\newcommand{\texturesid}{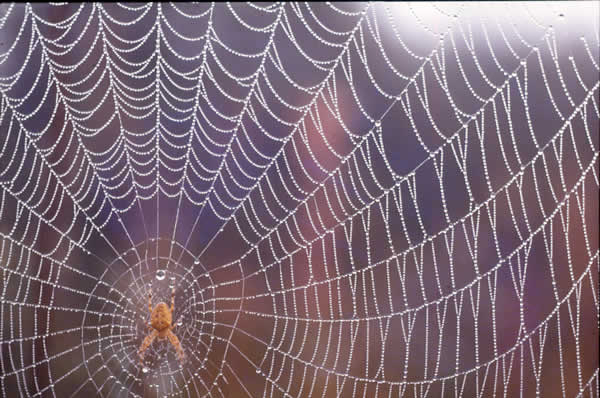}
\tuple texturesta=({bubble} {bubbly})
\tuple texturestb=({honeycomb} {honeycombed})
\tuple texturestc=({coil} {spiralled})
\tuple texturestd=({spider web} {cobwebbed})

\newcommand{\texturesie}{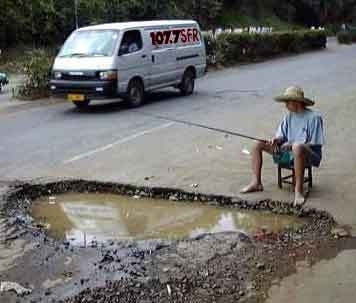}
\newcommand{\texturesif}{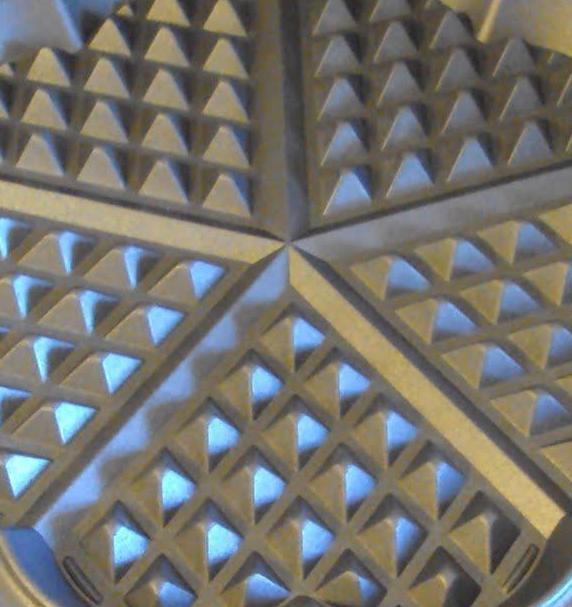}
\newcommand{\texturesig}{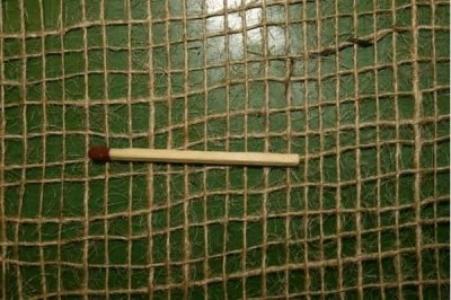}
\newcommand{\texturesih}{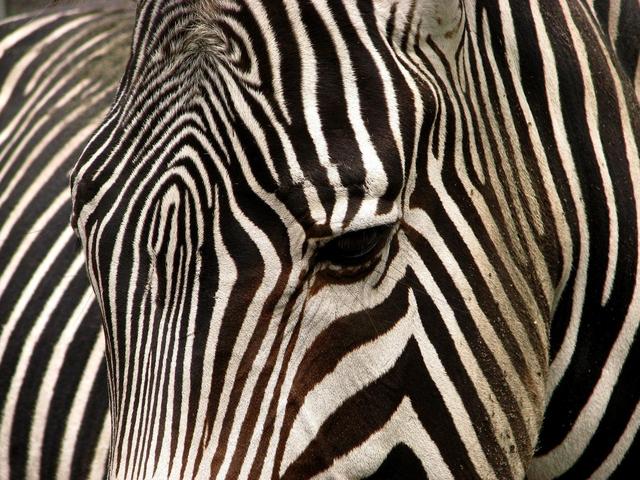}
\tuple textureste=({minibus} {potholed})
\tuple texturestf=({waffle iron} {waffled})
\tuple texturestg=({matchstick} {meshed})
\tuple texturesth=({zebra} {striped})

\newcommand{\placesia}{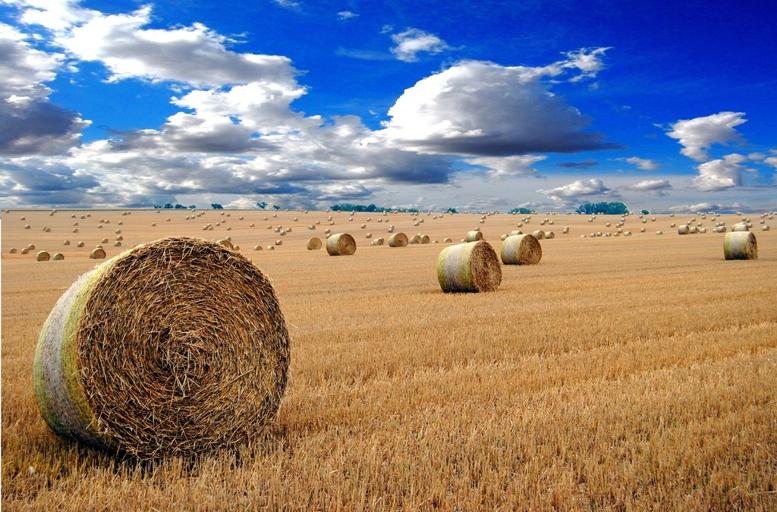}
\newcommand{\placesib}{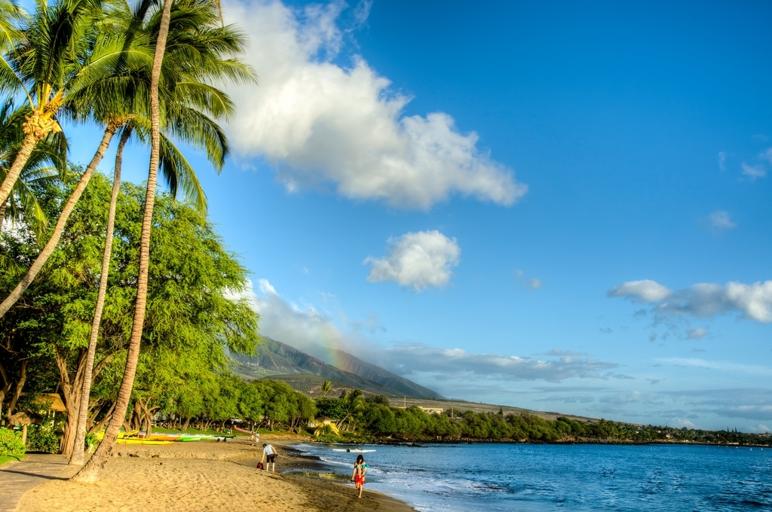}
\newcommand{\placesic}{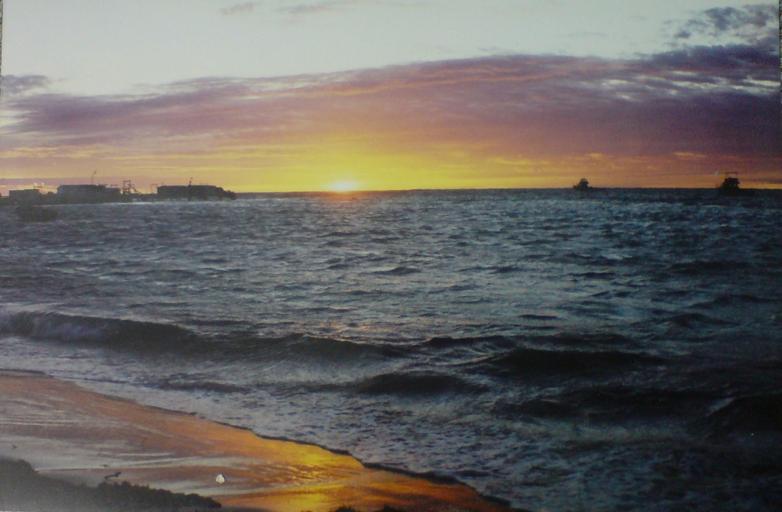}
\newcommand{\placesid}{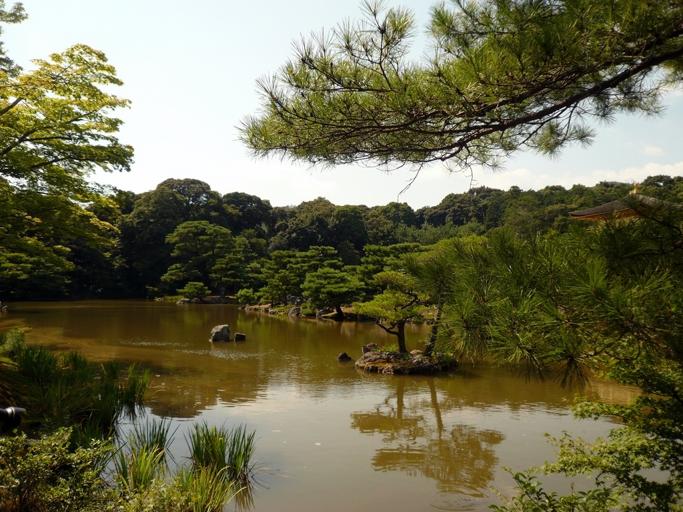}
\tuple placesta=(hay hayfield)
\tuple placestb=(seashore lagoon)
\tuple placestc=(seashore ocean)
\tuple placestd=(lakeside pond)

\newcommand{\placesie}{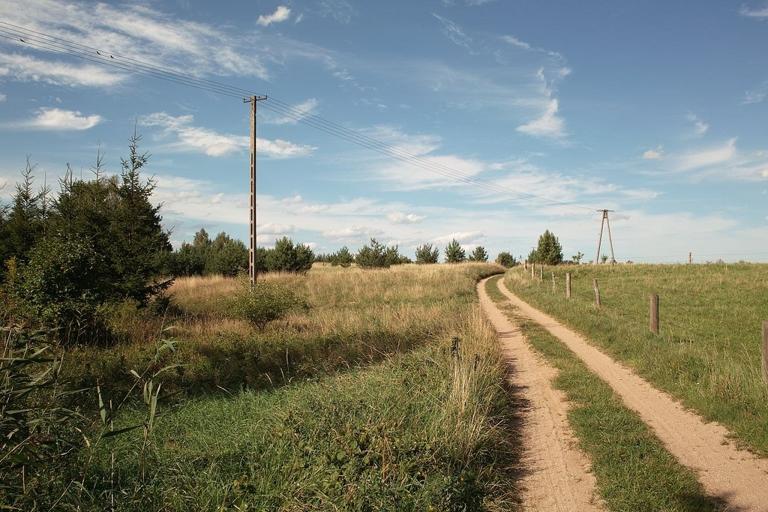}
\newcommand{\placesif}{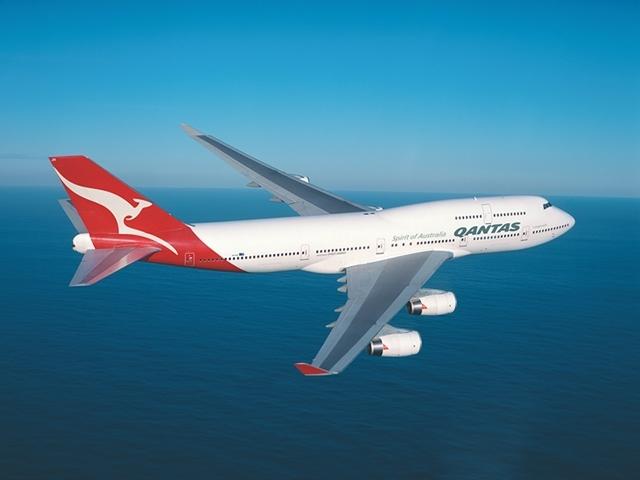}
\newcommand{\placesig}{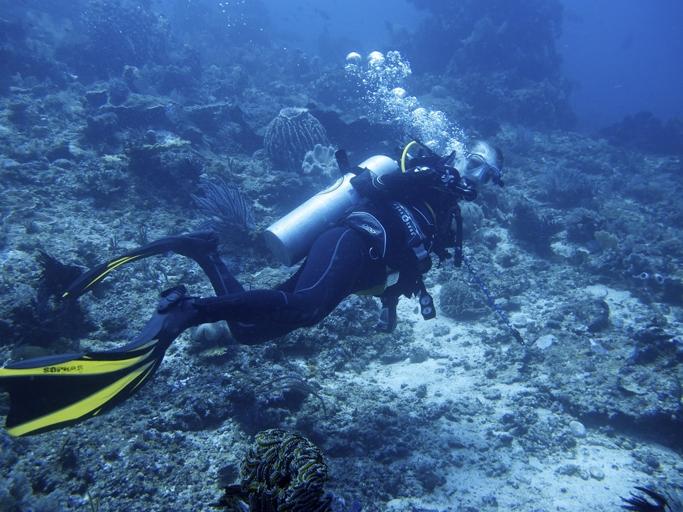}
\newcommand{\placesih}{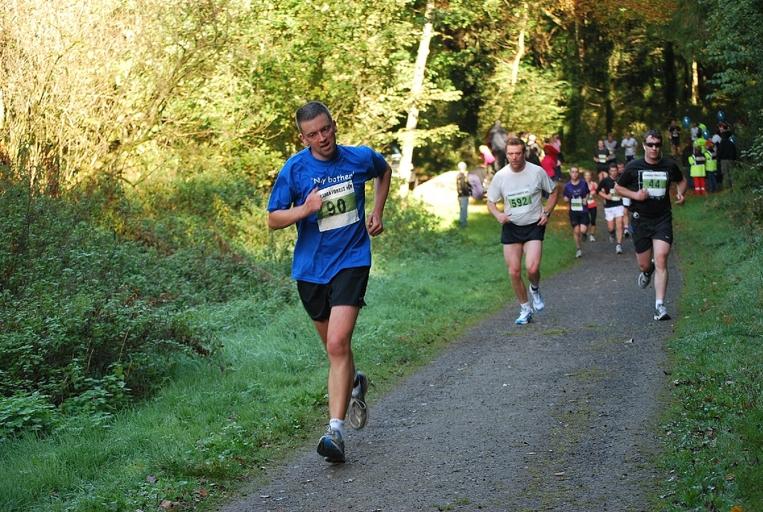}
\tuple placeste=(pole {field road})
\tuple placestf=(airliner sky)
\tuple placestg=({scuba diver} underwater)
\tuple placesth=({running shoe} {forest path})

\newcommand{\speciesia}{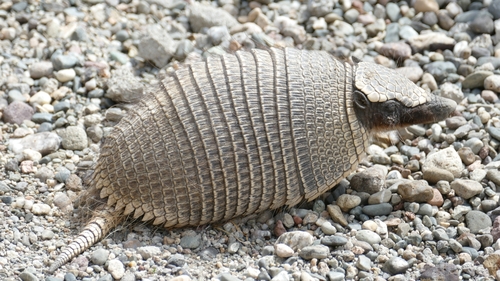}
\newcommand{\speciesib}{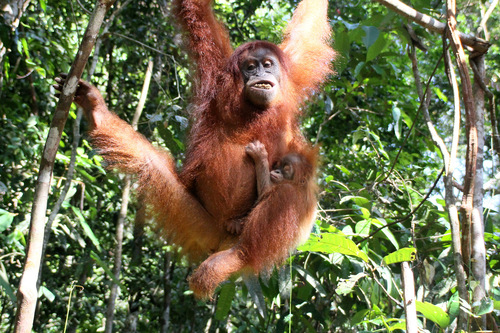}
\newcommand{\speciesic}{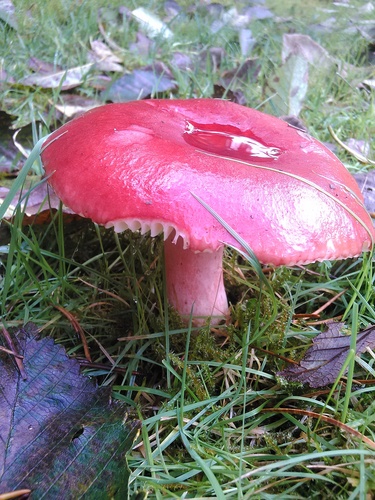}
\newcommand{\speciesid}{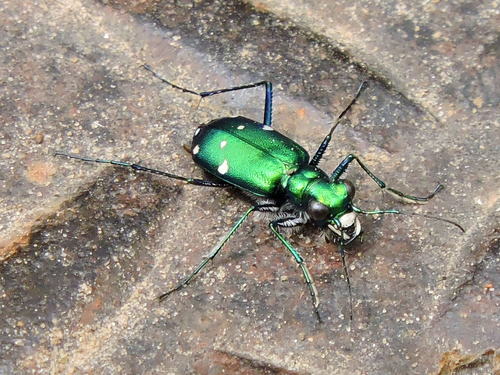}
\tuple speciesta=({armadillo} {dwarf ar.})
\tuple speciestb=({orangutan} {Sumatran or.})
\tuple speciestc=({mushroom} {rosy russula})
\tuple speciestd=({tiger beetle} {six-spotted t.b.})

\newcommand{\speciesie}{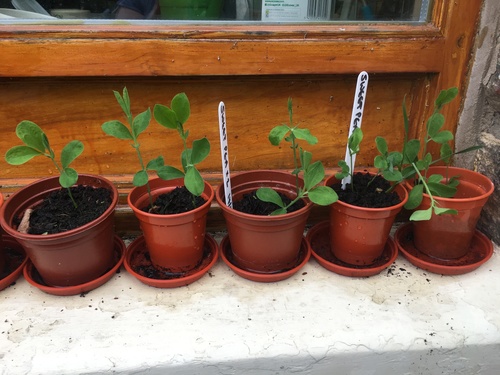}
\newcommand{\speciesif}{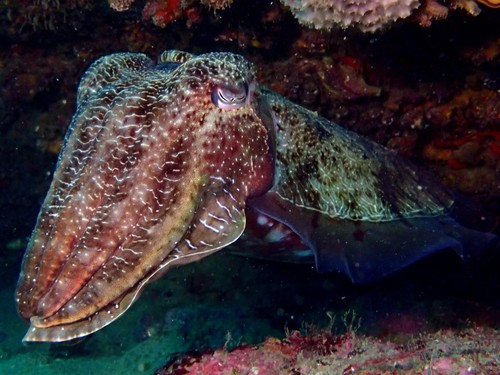}
\newcommand{\speciesig}{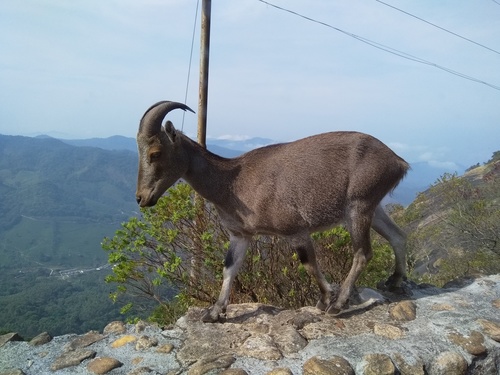}
\newcommand{\speciesih}{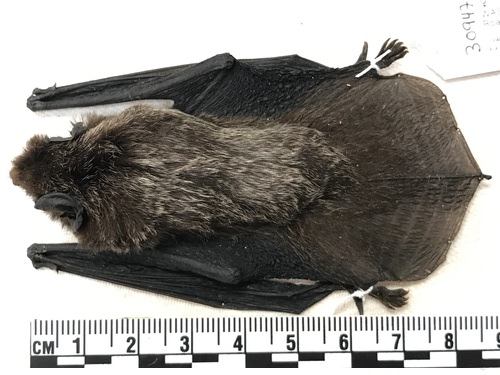}
\tuple specieste=({flower pot} {sweet pea})
\tuple speciestf=({coral reef} {giant cuttlefish})
\tuple speciestg=({pole} {Nilgiri tahr})
\tuple speciesth=({rule} {silver-hair bat}) 

\newcommand{\inOia}{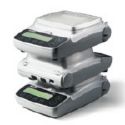}
\newcommand{\inOib}{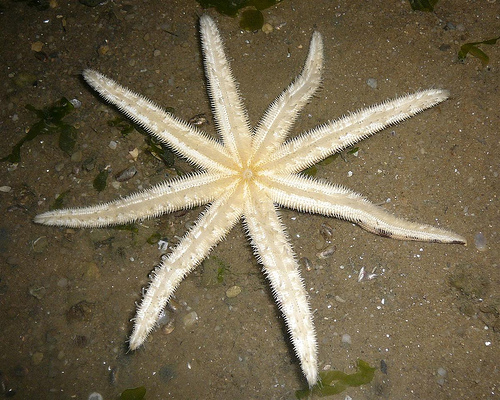}
\newcommand{\inOic}{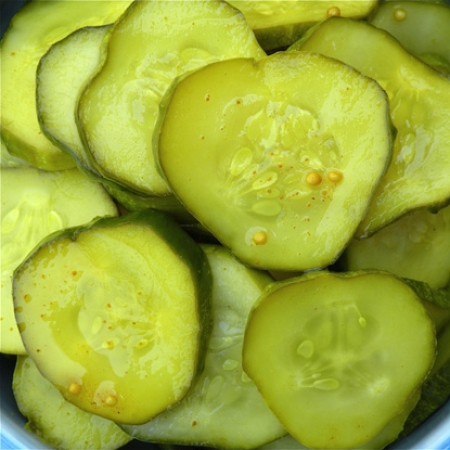}
\newcommand{\inOid}{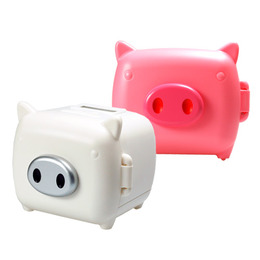}
\tuple inOta=({scale} {analyt. balance})
\tuple inOtb=({starfish} {echinoderm})
\tuple inOtc=({cucumber} {pickle}) 
\tuple inOtd=({piggy bank} {savings bank})

\newcommand{\inOie}{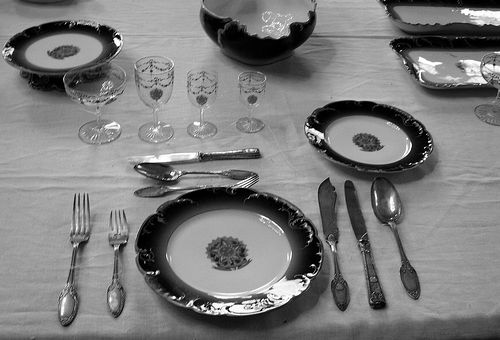}
\newcommand{\inOif}{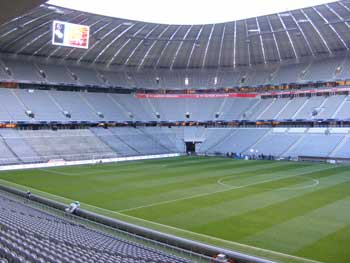}
\newcommand{\inOig}{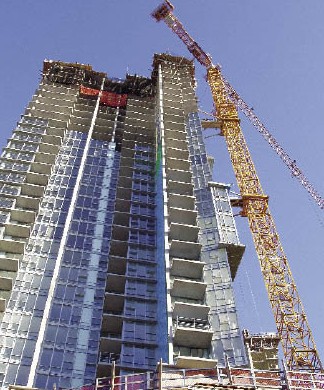}
\newcommand{\inOih}{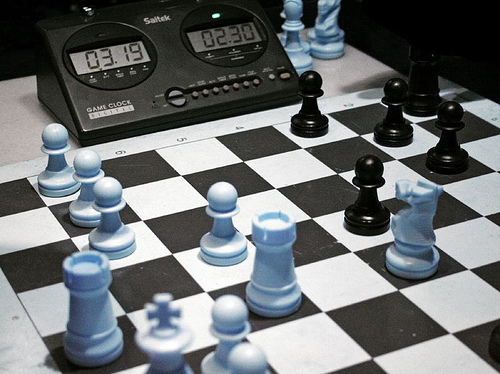}  
\tuple inOte=({plate}  {table knife})
\tuple inOtf=({scoreboard} {ballpark})
\tuple inOtg=({crane} {build. complex})
\tuple inOth=({digital clock} {chessman})



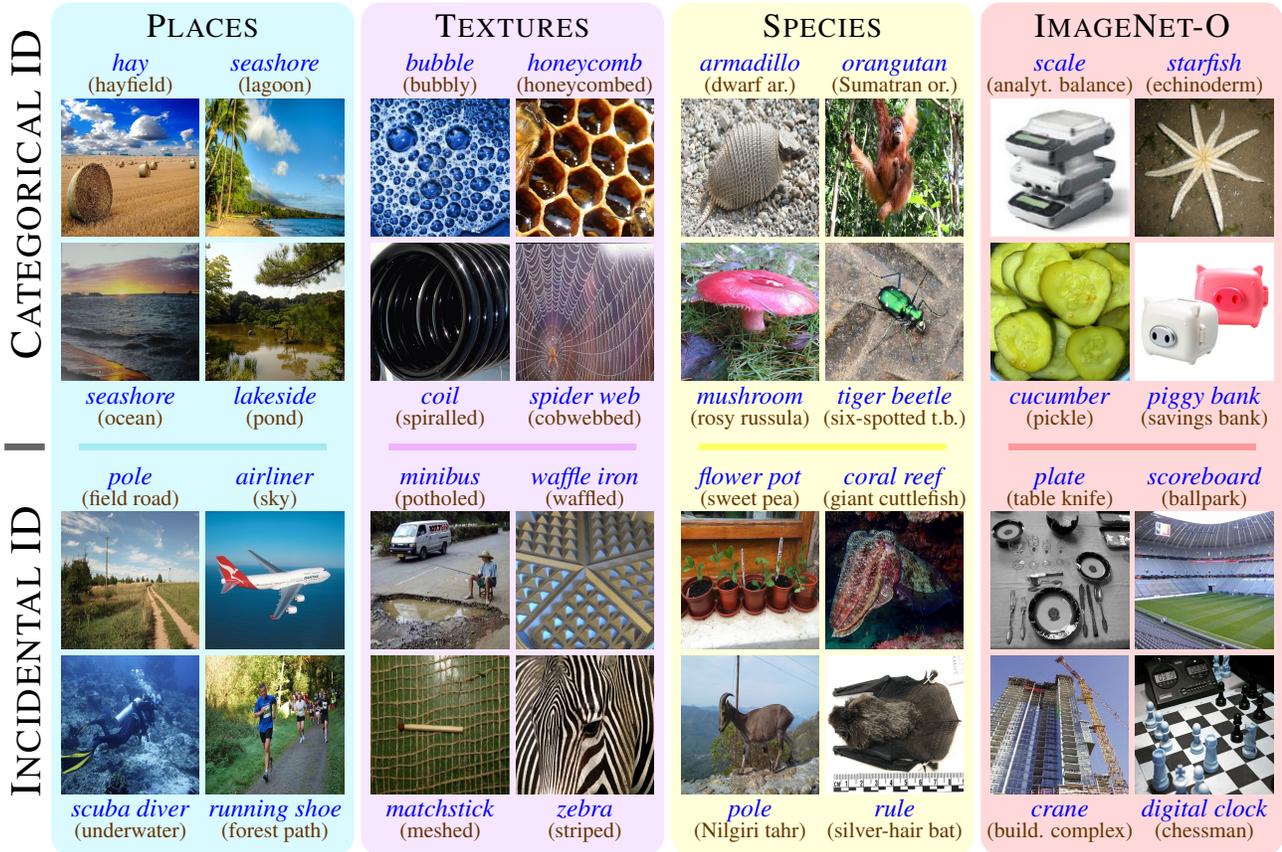
\begin{figure*}[t]
\centering
\pgfmathsetseed{5}
\vskip -1.6mm
\hspace*{-4.5mm}
\resizebox{1.018\textwidth}{!}{
\noindent
\begin{tikzpicture}
  \coordinate (ood1) at (0,0);
  \coordinate (ood2) at (\xoffset,0);
  \coordinate (ood3) at (2*\xoffset,0);
  \coordinate (ood4) at (3*\xoffset,0);
  
  \fill[rounded corners, fill=places, line width=0.mm] ($(ood1) + (-1.1*\imgwidth,1.2*\imgheight)$) rectangle ($(ood1) + (1.1*\imgwidth,-2.*\imgheight-\yoffset)$) {};
  \fill[rounded corners, fill=textures, line width=0.mm] ($(ood2) + (-1.1*\imgwidth,1.2*\imgheight)$) rectangle ($(ood2) + (1.1*\imgwidth,-2.*\imgheight-\yoffset)$) {};
  \fill[rounded corners, fill=species, line width=0.mm] ($(ood3) + (-1.1*\imgwidth,1.2*\imgheight)$) rectangle ($(ood3) + (1.1*\imgwidth,-2.*\imgheight-\yoffset)$) {};
  \fill[rounded corners, fill=imageneto, line width=0.mm] ($(ood4) + (-1.1*\imgwidth,1.2*\imgheight)$) rectangle ($(ood4) + (1.1*\imgwidth,-2.*\imgheight-\yoffset)$) {};

  \draw[color=placesline, line width=0.5mm] ($(ood1) + (-.9*\imgwidth,-\yline)$) -- ($(ood1) + (.9*\imgwidth,-\yline)$) {};
  \draw[color=texturesline, line width=0.5mm] ($(ood2) + (-.9*\imgwidth,-\yline)$) -- ($(ood2) + (.9*\imgwidth,-\yline)$) {};
  \draw[color=speciesline, line width=0.5mm] ($(ood3) + (-.9*\imgwidth,-\yline)$) -- ($(ood3) + (.9*\imgwidth,-\yline)$) {};
  \draw[color=imagenetoline, line width=0.5mm] ($(ood4) + (-.9*\imgwidth,-\yline)$) -- ($(ood4) + (.9*\imgwidth,-\yline)$) {};
  \draw[color=black!60!white, line width=0.5mm] ($(ood1) + (-.64*\xoffset,-\yline)$) -- ($(ood1) + (-.51*\xoffset,-\yline)$) {};

  \pic [local bounding box=A1a] at (ood1) {imgsquare=\placesia/\placesib/\placesic/\placesid/placesta/placestb/placestc/placestd};
  \pic [local bounding box=A1a] at (ood2) {imgsquare=\texturesia/\texturesib/\texturesic/\texturesid/texturesta/texturestb/texturestc/texturestd};
  \pic [local bounding box=A1a] at (ood3) {imgsquare=\speciesia/\speciesib/\speciesic/\speciesid/speciesta/speciestb/speciestc/speciestd};
  \pic [local bounding box=A1a] at (ood4) {imgsquare=\inOia/\inOib/\inOic/\inOid/inOta/inOtb/inOtc/inOtd};
  \pic [local bounding box=A1a] at ($(ood1) + (0,-\yoffset)$) {imgsquare=\placesie/\placesif/\placesig/\placesih/placeste/placestf/placestg/placesth};
  \pic [local bounding box=A1a] at ($(ood2) + (0,-\yoffset)$) {imgsquare=\texturesie/\texturesif/\texturesig/\texturesih/textureste/texturestf/texturestg/texturesth};
  \pic [local bounding box=A1a] at ($(ood3) + (0,-\yoffset)$) {imgsquare=\speciesie/\speciesif/\speciesig/\speciesih/specieste/speciestf/speciestg/speciesth};
  \pic [local bounding box=A1a] at ($(ood4) + (0,-\yoffset)$) {imgsquare=\inOie/\inOif/\inOig/\inOih/inOte/inOtf/inOtg/inOth};
  
  \draw[anchor=base, align=center] ($(ood1) + (0,\ydsetname)$)  node {\scriptsize\textcolor{black}{\textsc{Places}}};
  \draw[anchor=base, align=center] ($(ood2) + (0,\ydsetname)$)  node {\scriptsize\textcolor{black}{\textsc{Textures}}};
  \draw[anchor=base, align=center] ($(ood3) + (0,\ydsetname)$)  node {\scriptsize\textcolor{black}{\textsc{Species}}};
  \draw[anchor=base, align=center] ($(ood4) + (0,\ydsetname)$)  node {\scriptsize\textcolor{black}{\textsc{ImageNet-O}}};

  \draw[anchor=base, align=center] ($(ood1) + (-.57*\xoffset,-1.4*\imgheight)$)  node {\rotatebox{90}{\small\textcolor{black}{\textsc{Categorical ID}}}};
  \draw[anchor=base, align=center] ($(ood1) + (-.57*\xoffset,-1.4*\imgheight) + (0,-1.06*\yoffset)$)  node {\rotatebox{90}{\small\textcolor{black}{\textsc{Incidental ID}}}};
  
\end{tikzpicture}
}
\vskip -2.0mm
\caption{\label{fig:teaser}\textbf{Contamination of OOD test sets with ID samples (ImageNet).} \INclass{Blue:} ImageNet-1K class found in the image.
\textcolor{sourceclass}{(Brown):} Label of the image in the original source dataset.
\textbf{Top:} Samples from classes of the OOD dataset that by class meaning categorically overlap with ImageNet-1K classes.
\textbf{Bottom:} Labels alone do not reveal that the images are ID, but incidental ID objects can be found.
}
\vskip \belowimgcaptionvskip
\end{figure*}

The erroneous occurrences of ID objects in existing OOD datasets can be characterized into two failure modes, which we illustrate in Figure \ref{fig:teaser} and define as follows.
\textbf{Categorical ID contaminations} 
show objects from ID classes which already are classes in a base dataset from which the test OOD dataset has been built. Their label %
coincides with an ID class or semantically designates a subset of an ID class,
e.g. the class \sourceclass{hayfield} from the \textsc{Places} datset and the IN-1k class \INclass{hay}.
\textbf{Incidental ID contaminations} on the other hand occur in images 
which are supposed to belong to an OOD category but which contain an ID object.
The object can be in the background or an aspect of the specific instance of the shown main object, e.g. the IN-1k class \INclass{plane}  in an image of the OOD category \sourceclass{sky}.
We show that ID contaminations strongly impact the conclusions which can be drawn from evaluating OOD detection methods by (1) systematically underestimating the true OOD detection performance and (2) unrightfully punishing stronger OOD detectors.

Probing the true performance of OOD detectors for IN-1K requires a range of OOD classes that are challenging, diverse, and most importantly actually OOD.
Compiling a test OOD dataset is indeed a challenging task, as the 1000 classes of IN-1K cover a fair portion of the images found in general image datasets.
In this paper we introduce the \textbf{\dsetname{}  (No ImageNet Class Objects) dataset} which contains  \numoodsamples{} images that we individually checked not to contain any ID object from the classes in IN-1K.
These images are ordered into 64 OOD classes, which facilitates a specific analysis of the failure modes of an OOD detector.
Additionally, we provide a dataset of \textbf{“OOD unit-tests”}, synthetic images which do not resemble real world photos, but are designed to test specific weaknesses that might have impact in real-world applications (e.g. due to a camera failure). We find that surprisingly many OOD detectors struggle to detect these supposedly easy unit-tests, in particular methods that work well on natural test data.

We provide a detailed OOD detection evaluation on \dsetname{} for a range of eleven OOD detection methods across a large number of architectures and training schemes. 
Surprisingly, it turns out to be difficult for many OOD detectors to improve consistently over the baseline of Maximum Softmax Probability (MSP).
While we confirm the observation that pretraining on larger datasets generally helps OOD detectors and particularly methods explicitly using pre-logit feature-information, we find that the type of pretraining has a strong impact. 

\section{Existing test OOD datasets for ImageNet-1K}
\label{sec:existing}
First, we give an overview of the datasets that have been used to evaluate OOD detection performance for IN-1K as ID.
In the following we use \INclass{blue} for the name of an ImageNet class and \sourceclass{brown} for the category name in the source dataset used for the generation of the test OOD dataset. \\
\textbf{\textsc{iNaturalist OOD Plants}} %
is a subset of \numprint{10000} images curated by \citet{huang2021mos} from 110 OOD plant species of iNat2017
\,\cite{van2018inaturalist} which is sourced from the \href{https://www.inaturalist.org/}{iNaturalist} project.
It is frequently used as test OOD dataset \cite{xia2022usefulness, ming2022delving}.\\
\textbf{\textsc{Places}} is a subset of Places365~\cite{zhou2017places} curated by~\citet{huang2021mos} as “50 categories \textup{[\,\dots]} that are not present in IN-1K”.
\begin{table}[t]
\vskip -0.1in
\centering\caption{\textbf{Percentage of ID samples}, $p = \frac{\text{ID}}{\text{ID} + \text{OOD}}$, \textbf{in commonly used test OOD datasets} found by visual inspection of 400 random samples per dataset. Unclear samples are ignored (which are at most 6.7\% (for \textsc{Places}) of the 400 samples).}
\label{tab:percentage_ID_in_OOD_datasets}
\begin{center}
\begin{tabular}{l@{\hskip -2mm}r|l@{\hskip -2mm}r}
    Dataset     &   ID samples &  
    Dataset     &   ID samples  \\ \hline
    \textsc{Places}      &   59.5\% &
    \textsc{Species}     &   57.0\% \\
    \textsc{ImageNet-O}  &   20.2\% &
    \textsc{Textures}    &   25.6\% \\
    \textsc{iNat. Plants}      &   2.5\%  &
    \textsc{Textures43} &  20.0\% \\
    \textsc{OpenIm.-O} &   4.9\% & 
    \textsc{IN-1K-OOD} & 32.1\% \\
    \textsc{SSB-hard} & 41.6\% &
    \textsc{SSB-easy} & 53.4\% \\
    \textsc{360OpenSet} & 26.9\% & COOD & 38.2\%
    
\end{tabular}
\end{center}
\vskip \belowtablevskip
\end{table}
It is used as test OOD dataset in \cite{huang2021mos, sun2021react, ming2022delving}.
The dataset contains \numprint{9822} images from 50 environment classes. %
We find that several of these classes are either subsets of ID classes, e.g.
\sourceclass{hayfield} (\INclass{hay}),
\sourceclass{cornfield} (\INclass{corn}), \sourceclass{lagoon} (\INclass{seashore} and \INclass{lakeshore}), or contain mostly ID objects, e.g. \sourceclass{underwater} (\INclass{coral reef} and \INclass{scuba diver}), \sourceclass{ocean} (\INclass{seashore}).\\
\textbf{\textsc{Textures}}~\cite{cimpoi2014describing} contains 5640 images of various objects that show one of 47 patterns.
It is used as test OOD dataset in \cite{huang2021mos, sun2021react, wang2021can, xia2022usefulness, ming2022delving} and others.
\citet{wang2022vim}  address the issue of overlap with IN-1K and remove four categorically ID textures (\sourceclass{bubbly} (\INclass{bubble}), \sourceclass{honeycombed} (\INclass{honeycomb}), \sourceclass{cobwebbed} (\INclass{spider web}), \sourceclass{spiralled} (\INclass{spiral})).
We find that even their version (denoted as \textsc{Textures43}) contains about 20\% ID images.
\\
\textbf{\textsc{Species}} %
was proposed in \cite{hendrycks22Scaling} as OOD dataset for IN-21K~\cite{imagenet_cvpr09} and should thus also be OOD for the IN-1K subset.
Sourced from iNaturalist, it consists of \numprint{700000} images from \numprint{1316} species which were selected for not being in IN-21K.
They sort the species into 10 superclasses.
The largest superclass \sourceclass{Fungi} largely coincides with the IN-1K class \INclass{mushroom}, and also many of the remaining species are ID.
Papers evaluating on \textsc{Species} for IN-1K OOD detection include \cite{salehi2021unified, yang2022openood, song2022rankfeat}.\\
\textbf{\textsc{ImageNet-O}}~\cite{hendrycks2021natural} contains \numprint{2000}~images from IN-21K, excluding its subset IN-1K.
To make the dataset challenging it was composed from images where a ResNet-50 classifier for a subset of 200 IN-1K classes attains high confidence.
The samples being OOD relies on the assumption that IN-21K without IN-1K is OOD for IN-1K.
However, this assumption does not hold, due to a significant overlap between ImageNet classes from IN-1K and IN-21K, e.g. \sourceclass{analytical balance}/\INclass{scale} and \sourceclass{pickle}/\INclass{cucumber}, and insufficient filtering for incidental ID objects.
\\
\textbf{\textsc{OpenImage-O}}~\cite{wang2022vim} consists of \numprint{17632}~%
images from the OpenImage-v3~\cite{OpenImages2} test set which their human labellers categorize as OOD.
It is also used in \citet{yang2022openood}.
\\
\textbf{\textsc{360OpenSetClasses}}~\cite{TowardsOSDN} uses those 360 classes (15.000 samples) from ILSVRC2010 which are not part of ILSVRC2012. Like for \textsc{ImageNet-O}, this leads to large semantic overlap, e.g. the class \sourceclass{organ pipe} coinciding with the ID class \INclass{organ}.
\\
\textbf{\textsc{Semantic Shift Benchmark (SSB)}}~\cite{360OSR} contains a \textit{hard} and \textit{easy} OSR benchmark, each consisting of 1000 classes, that were created by regarding the distances between nodes in the WordNet tree. Similar to \textsc{360OpenSetClasses}, we find both categorical and incidental ID contamination, e.g. \sourceclass{rainbow lorikeet}/\INclass{lorikeet}. Papers evaluating on SSB include \cite{Wen2022ASP}.
\\
\textbf{\textsc{ImageNet-1K-OOD}}~\cite{Wang2022IN1KOOD} contains 50.000 images from 1.000 classes randomly sampled from ImageNet-21K, such that those classes don't overlap with ImageNet-1K and ImageNet-LT, another dataset introduced by the authors. Categorical examples include \sourceclass{bobwhite quail}/\INclass{quail} and \sourceclass{king vulture}/\INclass{vulture}.
\\
\textbf{\textsc{COOD-benchmark}}~\cite{anonymous2023COOD} is a general framework for benchmarking ImageNet-1K OOD detection. Their test set consists of ImageNet-21K samples which were filtered by class. It includes severe contamination, including categorical cases like \sourceclass{orange, orange tree}/\INclass{orange}. 
\def \xoffset {2.7cm}
\def \yoffset {2.75cm}
\def \ydsetname {1.75*\yoffset} 
\def \imgwidth {1.cm}
\def \ximgdist {.03cm}
\def \imgheight {1.cm}
\def \yimgdist {.03cm}
\def \toplabeldist {.54cm}
\def \botlabeldist {.88cm}
\def \tinyscale {.62}

\newcommand{\ima}{better_model_positives_images/OOD_Places_MOS_d_desert_road_1421_00001761.jpeg}
\newcommand{\imc}{better_model_positives_images/lasionycteris_noctivagans_523_93651165.jpeg}
\newcommand{\imb}{better_model_positives_images/pteropus_poliocephalus_653_6846499.jpeg}
\newcommand{\imf}{better_model_positives_images/imagenet_o_n03706229_746_speedometer_speed_indicator_0.98283976.jpeg}
\newcommand{\ime}{better_model_positives_images/mirounga_angustirostris_598_6839003.jpeg}
\newcommand{\imd}{better_model_positives_images/perimyotis_subflavus_287_37286487.jpeg}
\newcommand{\img}{better_model_positives_images/openimage_o_hand_8429_7b423365b580594b.jpeg}
\newcommand{\imh}{better_model_positives_images/OOD_Places_MOS_i_igloo_4711_00002880.jpeg}
\triple ta=({53\%\:\!\textit{pole}} {9\%\:\!\textit{horse cart}} {\textsc{Pl}:desert road})
\triple tb=({53\%\:\!\textit{orange}} {21\%\:\!\textit{squirrel monkey}} {\textsc{Sp}:flying fox}) 
\triple tc=({55\%\:\!\textit{chainlink fence}} {20\%\:\!\textit{pelican}} {\textsc{Sp}:silverhair bat})
\triple td=({79\%\:\!\textit{rule}} {9\%\:\!\textit{candle}} {\textsc{Sp}:tricol.\,bat})
\triple te=({46\%\:\!\textit{seashore}} {26\%\:\!\textit{sea lion\ }} {\textsc{Sp}:eleph.\,seal})
\triple tf=({88\%\:\!\textit{odometer}} {36\%\:\!\textit{analog clock}} {\textsc{IO:}speedometer})
\triple tg=({43\%\:\!\textit{bucket}} {13\%\:\!\textit{terrapin}} {\textsc{OO}:hand}) 
\triple th=({48\%\:\!\textit{dome}} {21\%\:\!\textit{megalith}} {\textsc{Pl}:igloo})

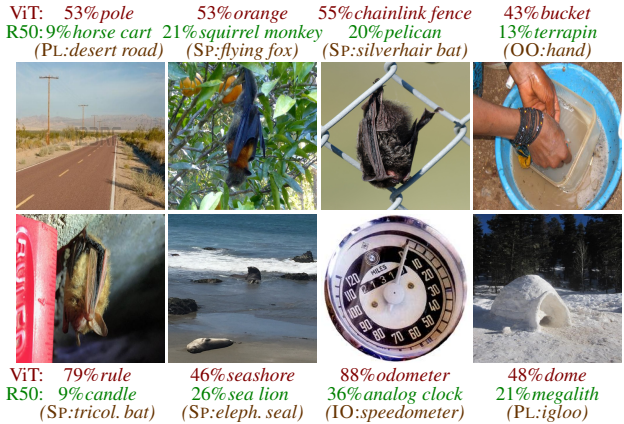
\begin{figure}[t]
\begin{center}
\pgfmathsetseed{5}
\vskip -1.5mm
\hspace*{-3.5mm}
\resizebox{1.05\columnwidth}{!}{
\noindent
\begin{tikzpicture}
\coordinate (o) at (0,0);
      \coordinate (i1) at ($(o) + (-.5*\imgwidth,0) + (-.5*\ximgdist,0)$);
      \coordinate (i2) at ($(o) + (.5*\imgwidth,0) + (.5*\ximgdist,0)$);
      \coordinate (i3) at ($(o) + (1.5*\imgwidth,0) + (1.5*\ximgdist,0)$);
      \coordinate (i4) at ($(o) + (-.5*\imgwidth,-\imgheight) + (-.5*\ximgdist,-\yimgdist)$);
      \coordinate (i5) at ($(o) + (.5*\imgwidth,-\imgheight) + (.5*\ximgdist,-\yimgdist)$);
      \coordinate (i6) at ($(o) + (1.5*\imgwidth,-\imgheight) + (1.5*\ximgdist,-\yimgdist)$);
      \coordinate (i7) at ($(o) + (2.5*\imgwidth,0) + (2.5*\ximgdist,0)$);
      \coordinate (i8) at ($(o) + (2.5*\imgwidth,-\imgheight) + (2.5*\ximgdist,-\yimgdist)$);
      \coordinate (t1) at ($(i1) + (0,\toplabeldist)$);
      \coordinate (t2) at ($(i2) + (0,\toplabeldist)$);
      \coordinate (t3) at ($(i3) + (0,\toplabeldist)$);
      \coordinate (t4) at ($(i4) + (0,-\botlabeldist)$);
      \coordinate (t5) at ($(i5) + (0,-\botlabeldist)$);
      \coordinate (t6) at ($(i6) + (0,-\botlabeldist)$);
      \coordinate (t7) at ($(i7) + (0,\toplabeldist)$);
      \coordinate (t8) at ($(i8) + (0,-\botlabeldist)$);
      \coordinate (t1l) at ($(t1) + (-.44,0)$);
      \coordinate (t1r) at ($(t1) + (0.06,0)$);
      \coordinate (t4l) at ($(t4) + (-.44,0)$);
      \coordinate (t4r) at ($(t4) + (0.05,0)$);
  \node[inner sep=0pt] at (i1) {\includegraphics[width=\imgwidth, height=\imgheight]{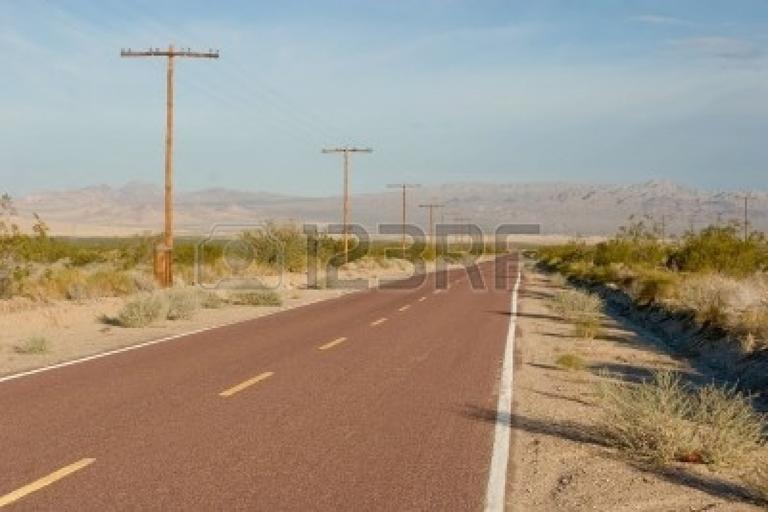}};
  \node[inner sep=0pt] at (i2) {\includegraphics[width=\imgwidth, height=\imgheight]{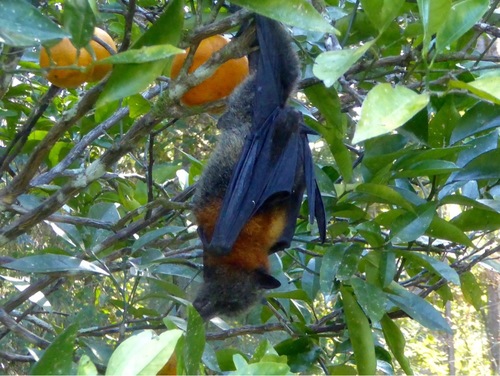}};
  \node[inner sep=0pt] at (i3) {\includegraphics[width=\imgwidth, height=\imgheight]{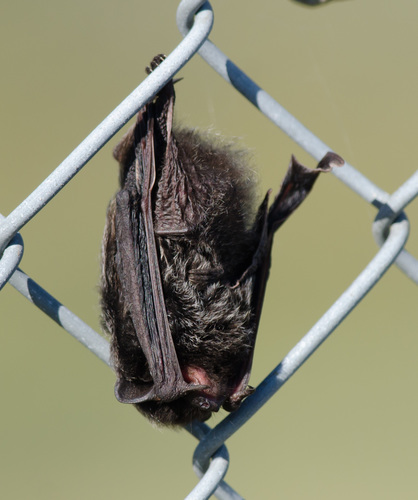}};
  \node[inner sep=0pt] at (i4) {\includegraphics[width=\imgwidth, height=\imgheight]{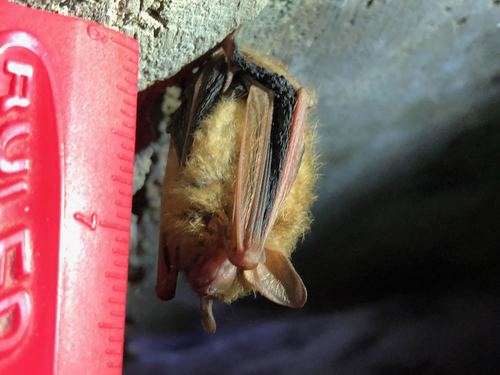}};
  \node[inner sep=0pt] at (i5) {\includegraphics[width=\imgwidth, height=\imgheight]{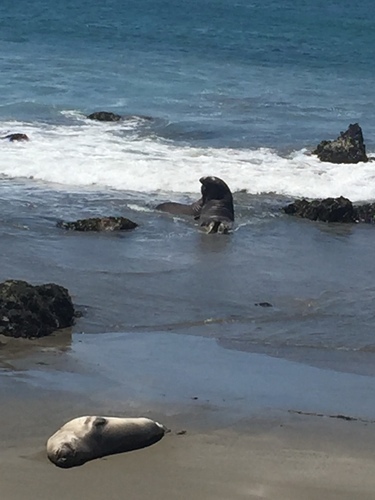}};
  \node[inner sep=0pt] at (i6) {\includegraphics[width=\imgwidth, height=\imgheight]{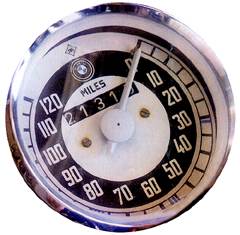}};
  \node[inner sep=0pt] at (i7) {\includegraphics[width=\imgwidth, height=\imgheight]{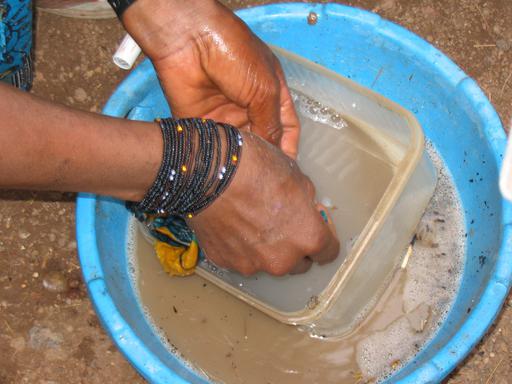}};
  \node[inner sep=0pt] at (i8) {\includegraphics[width=\imgwidth, height=\imgheight]{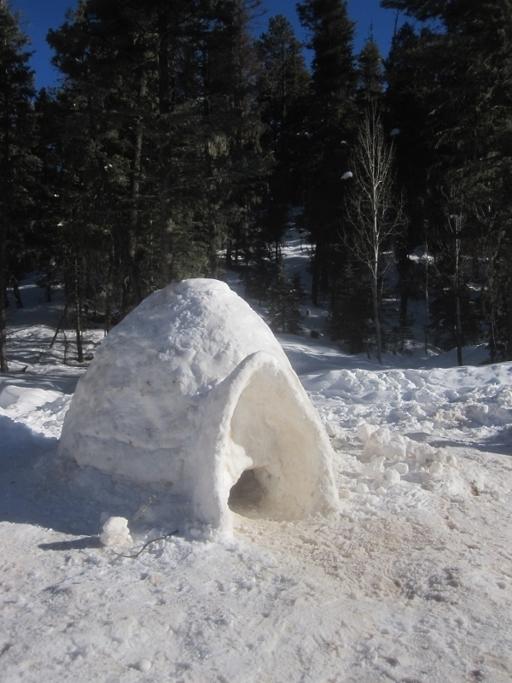}};
  \draw[anchor=base, align=center] (t1r)  node[scale=\tinyscale] {\tiny\textcolor{darkred}{\xtriple{ta}{1}}\\[-2.2mm]\tiny\textcolor{darkgreen}{\xtriple{ta}{2}}\\[-2.2mm]\tiny\sourceclass{(\xtriple{ta}{3})}};
  \draw[anchor=base, align=center] (t2)  node[scale=\tinyscale] {\tiny\textcolor{darkred}{\xtriple{tb}{1}}\\[-2.2mm]\tiny\textcolor{darkgreen}{\xtriple{tb}{2}}\\[-2.2mm]\tiny\sourceclass{(\xtriple{tb}{3})}};
  \draw[anchor=base, align=center] (t3)  node[scale=\tinyscale] {\tiny\textcolor{darkred}{\xtriple{tc}{1}}\\[-2.2mm]\tiny\textcolor{darkgreen}{\xtriple{tc}{2}}\\[-2.2mm]\tiny\sourceclass{(\xtriple{tc}{3})}};
  \draw[anchor=base, align=center] (t4r)  node[scale=\tinyscale] {\tiny\textcolor{darkred}{\xtriple{td}{1}}\\[-2.2mm]\tiny\textcolor{darkgreen}{\xtriple{td}{2}}\\[-2.2mm]\tiny\sourceclass{(\xtriple{td}{3})}};
  \draw[anchor=base, align=center] (t5)  node[scale=\tinyscale] {\tiny\textcolor{darkred}{\xtriple{te}{1}}\\[-2.2mm]\tiny\textcolor{darkgreen}{\xtriple{te}{2}}\\[-2.2mm]\tiny\sourceclass{(\xtriple{te}{3})}};
  \draw[anchor=base, align=center] (t6)  node[scale=\tinyscale] {\tiny\textcolor{darkred}{\xtriple{tf}{1}}\\[-2.2mm]\tiny\textcolor{darkgreen}{\xtriple{tf}{2}}\\[-2.2mm]\tiny\sourceclass{(\xtriple{tf}{3})}};
  \draw[anchor=base, align=center] (t7)  node[scale=\tinyscale] {\tiny\textcolor{darkred}{\xtriple{tg}{1}}\\[-2.2mm]\tiny\textcolor{darkgreen}{\xtriple{tg}{2}}\\[-2.2mm]\tiny\sourceclass{(\xtriple{tg}{3})}};
  \draw[anchor=base, align=center] (t8)  node[scale=\tinyscale] {\tiny\textcolor{darkred}{\xtriple{th}{1}}\\[-2.2mm]\tiny\textcolor{darkgreen}{\xtriple{th}{2}}\\[-2.2mm]\tiny\sourceclass{(\xtriple{th}{3})}};
  \draw[anchor=base, align=right] (t1l)  node[scale=\tinyscale] {\tiny\textcolor{darkred}{ViT:}\\[-2.2mm]\tiny\textcolor{darkgreen}{R50:}\\[-2.2mm]\tiny\textcolor{blue}{\ }};
  \draw[anchor=base, align=right] (t4l)  node[scale=\tinyscale] {\tiny\textcolor{darkred}{ViT:}\\[-2.2mm]\tiny\textcolor{darkgreen}{R50:}\\[-2.2mm]\tiny\textcolor{blue}{\ }};
\end{tikzpicture}
}
\vskip -3.5mm
\caption{\label{fig:better_model_positives}
A \textcolor{darkred}{Vision Transformer} confidently classifies ID objects in samples from popular OOD datasets \sourceclass{(source label in parentheses)} as the correct IN-1K class, but is marked down with false positives in OOD detection evaluation when using MSP (Max Softmax Prob.) as criterion.
The weaker \textcolor{darkgreen}{ResNet-50}, in contrast, doesn't recognize the ID objects and hence the MSP  is low enough to reject all images wrongly as OOD.
This illustrates how \textbf{a better model} (ViT in our case) \textbf{can be unjustly punished when the test OOD dataset 
contains ID objects.}
For both models, the 95\%TPR threshold is at a MSP of 38\%.
Origins of the images:
\textsc{Pl}=\textsc{Places},
\textsc{Sp}=\textsc{Species},
\textsc{OO}=\textsc{OpenImage-O},
\textsc{IO}=\textsc{ImageNet-O}.
}
\end{center}
\vskip \belowimgcaptionvskip
\end{figure}

\subsection{Prevalence of ID samples in popular OOD datasets}\label{sec:prevalence}
Concerningly, several test OOD datasets for IN-1K that are in use by the community contain a substantial fraction of samples that show ID objects.
Figure~\ref{fig:teaser} shows some typical appearances of ID data in supposedly OOD datasets.
The categorical ID failure mode illustrated in the top part %
is the inclusion of samples from explicitly ID classes of the %
source dataset from which the OOD dataset has been built. 
For instance, the class \sourceclass{hayfield} from the \textsc{Places}-dataset overlaps with the IN-1K class \INclass{hay}.
However, also in principally innocuous classes (bottom part), many incidental ID samples can still be found.
Here, the occurring failure modes are numerous: some ID objects happen to be in the background, some are a prominent part of the depicted scene, and some happen to realize both the original class and the ID class.
For instance, the class \sourceclass{table knife} contains samples which also show a \INclass{plate},
and the class \sourceclass{striped} from the \textsc{Textures}-dataset often shows the stripes of a \INclass{zebra}.

In order to quantify the severity of ID objects %
in test OOD datasets, we manually check for ID objects in 400 random samples from each of the most commonly used datasets.
For fair treatment, unclear and ambiguous samples, which we would exclude from \dsetname{} introduced below, are ignored in this survey.
The results in Table~\ref{tab:percentage_ID_in_OOD_datasets} show that for many of these common OOD detection benchmarks, a substantial fraction of samples is actually ID:
For both the \textsc{Places} and \textsc{Species} datasets, it is more  than 50\%.
Only \textsc{iNaturalist OOD Plants} (2.5\% of samples ID) and \textsc{OpenImage-O} (4.9\% ID) contain comparably few ID images.

\subsection{Effect of ID contamination on OOD evaluation}\label{sec:effect_of_cleaning}

In Figure~\ref{fig:better_model_positives}, we show 
how OOD detection evaluation with incidental ID samples can unrightfully punish strong OOD-detectors:
A better model can correctly recognize ID objects with high confidence even if they are in the background of the image, leading to a false ``false positive'' in the evaluation, while a weaker model not recognizing the ID object and providing a low-confidence prediction is ``rewarded'' with a false ``true negative''.
For example, the strong VisionTransformer~(ViT) \cite{dosovitskiy2020vit} identifies the \INclass{pole} besides an otherwise empty desert road, and thus has high confidence on the image where the weaker ResNet-50 does not recognize any ID class with high confidence.
Similarly, in the second example, the ViT is punished with a false ``false  positive'' for recognizing (above the detection threshold) the \INclass{oranges} in the background while ignoring the unknown flying fox (truly OOD), whereas the ResNet-50 even does predict a wrong ID class, namely \INclass{squirrel monkey}, but does so with low confidence (below the detection threshold), and is thus rewarded with a false ``true negative''.

\begin{figure}
    \vskip -0.5mm
    \centering
    \includegraphics[width=1.\linewidth]{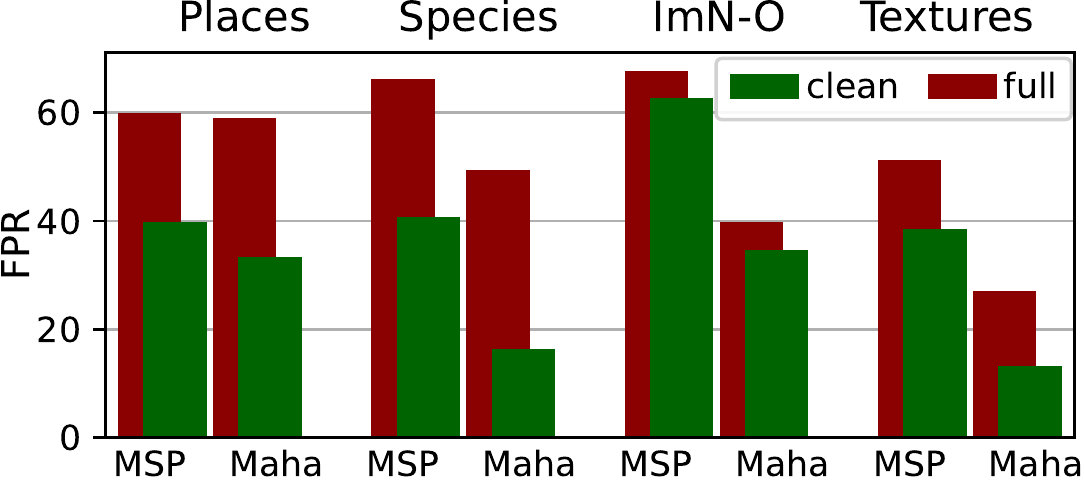}
    \vskip -3.8mm
    \caption{\textbf{OOD-detection %
    before and after removing samples with ID-objects:} We show \FPR{} (lower is better) of two OOD detectors (MSP and Mahalanobis distance) for a ViT, evaluated on cleaned and full subsets of four popular OOD datasets.}
    \label{fig:barplot-subsampled-fpr}
    \vskip \belowimgcaptionvskip
\end{figure}

We quantify the effect of ID contaminations on evaluation results  in customary OOD datasets in Figure \ref{fig:barplot-subsampled-fpr} for the MSP baseline and the Mahalanobis OOD detection method \cite{LeeMahalanobis2018}.
For the test OOD datasets %
which showed a large portion of ID samples in Table~\ref{tab:percentage_ID_in_OOD_datasets},
we report the \FPR{} at 95\% TPR  obtained with a ViT when evaluating on the original 400 samples and our cleaned subsample of it not containing any more ID objects
(detailed results for a range of models and methods can be found in Appendix~\ref{sec:Effect_cleaning_all}).
We find that ID contaminations strongly impact the conclusions which can be drawn from evaluating OOD detection methods on those datasets.
Most clearly, both methods perform substantially \textit{better} after removing the images with ID objects from the OOD datasets, in some cases reducing the FPR by more than 50\%.
This is unsurprising: If a significant fraction of the dataset is actually ID, this fraction should not be detected as OOD by a well-performing method.
Hence, evaluating OOD detection performance with partially ID data leads to a systematic
\textit{overestimation} of the true FPR of the OOD detection method and disadvantages better models as they are more likely to detect ID objects as discussed above.
Additionally, we observe that the differences between OOD detectors become more pronounced. In Figure \ref{fig:barplot-subsampled-fpr} it can be seen that for each dataset, the \FPR{} for the Mahalanobis OOD detector decreases more than for the MSP-baseline. The effect is particularly strong for \textsc{Species} (25.6\% gain of MSP vs. 33.2\% gain of Mahalanobis) and \textsc{Places} (19.6\% gain vs. 26.3\% gain), which are the two datasets we found to contain most ID samples. We further emphasize that due to the presence of large fractions of ID samples in most common benchmarks, even the performance of a perfect detector would saturate significantly above 0\% \FPR. For example with \textsc{Species}, we find that for a strong current detector already more than 85\% of the 'false positives' contain ID objects.

\def \xoffset {2.7cm}
\def \yoffset {2.75cm}
\def \ydsetname {1.75*\yoffset} 
\def \imgwidth {1.cm}
\def \ximgdist {.04cm}
\def \imgheight {1.cm}
\def \yimgdist {.04cm}
\def \toplabeldist {.54cm}
\def \botlabeldist {.88cm}
\def \tinyscale {.62}

\newcommand{\imharda}
{figures/samples/araneus_gemma_0.png}
\newcommand{\imhardc}
{figures/samples/spaghetti_bolognese_0.png}
\newcommand{\imhardb}
{figures/samples/arctocephalus_galapagoensis_1.png}
\newcommand{\imhardf}
{figures/samples/triturus_marmoratus_2.jpeg}
\newcommand{\imharde}
{figures/samples/donuts_0.png}
\newcommand{\imhardd}
{figures/some_examples_per_ood_class_0/glass_of_milk_09_glass_of_milk_0023_flickr.jpeg}
\newcommand{\imhardg}
{figures/samples/chicken_quesadilla_1.png}
\newcommand{\imhardh}
{figures/samples/dendrolagus_lumholtzi_1.png}
\newcommand{\INharda}
{"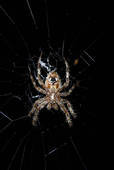"}
\newcommand{\INhardc}
{figures/ImageNet_samples_hard/carbonara_1227993_n07831146_1036.jpeg}
\newcommand{\INhardb}
{figures/ImageNet_samples_hard/sea_lion_194236_n02077923_10221.jpeg}
\newcommand{\INhardf}
{figures/ImageNet_samples_hard/common_newt_33849_n01630670_1000.jpeg}
\newcommand{\INharde}
{figures/ImageNet_samples_hard/bagel_1191859_n07693725_10068.jpeg}
\newcommand{\INhardd}
{figures/ImageNet_samples_hard/eggnog_4.png}
\newcommand{\INhardg}
{figures/ImageNet_samples_hard/burrito_1235833_n07880968_110.jpeg}
\newcommand{\INhardh}
{figures/ImageNet_samples_hard/indri_492845_n02500267_10326.jpeg}

\triple tahard=({\OODclass{cat-faced spider}} {barn spider} {barn spider})
\triple tbhard=({\OODclass{Galáp. fur seal}} {} {sea lion}) 
\triple tchard=({\OODclass{spagh. bolognese}} {} {carbonara})
\triple tdhard=({\OODclass{glass of milk}} {} {eggnog})
\triple tehard=({\OODclass{donut}} {} {bagel})
\triple tfhard=({\OODclass{marbled newt}} {} {common newt})
\triple tghard=({\OODclass{chicken quesadilla}} {} {burrito}) 
\triple thhard=({\OODclass{L. tree-kangaroo}} {} {indri})

\begin{figure*}[t]
\vskip -2.8mm
\begin{center}
\pgfmathsetseed{5}
\hspace*{-2.6mm}
\resizebox{1.018\textwidth}{!}{
\noindent
\begin{tikzpicture}
\coordinate (o) at (0,0);
      \coordinate (i1) at ($(o) + (-.5*\imgwidth,0) + (-.5*\ximgdist,0)$);
      \coordinate (i2) at ($(o) + (0.5*\imgwidth,0) + (0.5*\ximgdist,0)$);
      \coordinate (i3) at ($(o) + (1.5*\imgwidth,0) + (1.5*\ximgdist,0)$);
      \coordinate (i4) at ($(o) + (2.5*\imgwidth,0) + (2.5*\ximgdist,0)$);
      \coordinate (i5) at ($(o) + (3.5*\imgwidth,0) + (3.5*\ximgdist,0)$);
      \coordinate (i6) at ($(o) + (4.5*\imgwidth,0) + (4.5*\ximgdist,0)$);
      \coordinate (i7) at ($(o) + (5.5*\imgwidth,0) + (5.5*\ximgdist,0)$);
      \coordinate (i8) at ($(o) + (6.5*\imgwidth,0) + (6.5*\ximgdist,0)$);
      \coordinate (j1) at ($(i1) + (0,-1*\imgheight) + (0,-1*\yimgdist)$);
      \coordinate (j2) at ($(i2) + (0,-\imgheight) + (0,-\yimgdist)$);
      \coordinate (j3) at ($(i3) + (0,-\imgheight) + (0,-\yimgdist)$);
      \coordinate (j4) at ($(i4) + (0,-\imgheight) + (0,-\yimgdist)$);
      \coordinate (j5) at ($(i5) + (0,-\imgheight) + (0,-\yimgdist)$);
      \coordinate (j6) at ($(i6) + (0,-\imgheight) + (0,-\yimgdist)$);
      \coordinate (j7) at ($(i7) + (0,-\imgheight) + (0,-\yimgdist)$);
      \coordinate (j8) at ($(i8) + (0,-\imgheight) + (0,-\yimgdist)$);
      \coordinate (t1) at ($(i1) + (0,\toplabeldist)$);
      \coordinate (t2) at ($(i2) + (0,\toplabeldist)$);
      \coordinate (t3) at ($(i3) + (0,\toplabeldist)$);
      \coordinate (t4) at ($(i4) + (0,\toplabeldist)$);
      \coordinate (t5) at ($(i5) + (0,\toplabeldist)$);
      \coordinate (t6) at ($(i6) + (0,\toplabeldist)$);
      \coordinate (t7) at ($(i7) + (0,\toplabeldist)$);
      \coordinate (t8) at ($(i8) + (0,\toplabeldist)$);
  \node[inner sep=0pt] at (i1) {\includegraphics[width=\imgwidth, height=\imgheight]{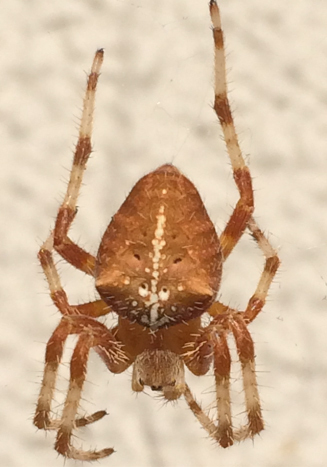}};
  \node[inner sep=0pt] at (i2) {\includegraphics[width=\imgwidth, height=\imgheight]{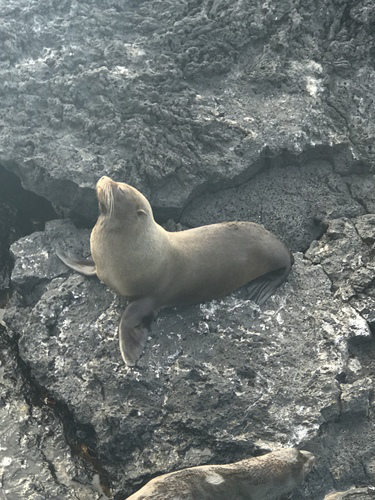}};
  \node[inner sep=0pt] at (i3) {\includegraphics[width=\imgwidth, height=\imgheight]{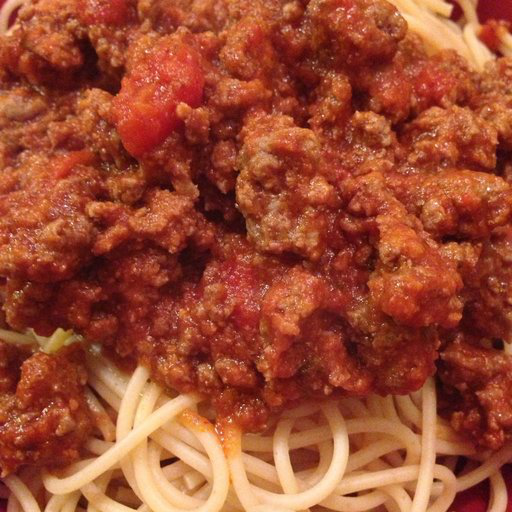}};
  \node[inner sep=0pt] at (i4) {\includegraphics[width=\imgwidth, height=\imgheight]{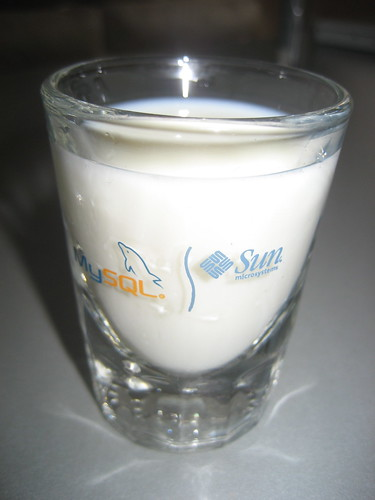}};
  \node[inner sep=0pt] at (i5) {\includegraphics[width=\imgwidth, height=\imgheight]{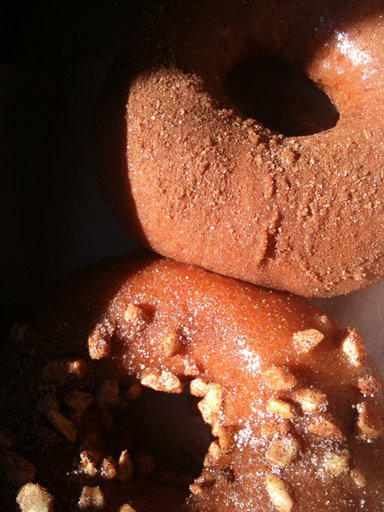}};
  \node[inner sep=0pt] at (i6) {\includegraphics[width=\imgwidth, height=\imgheight]{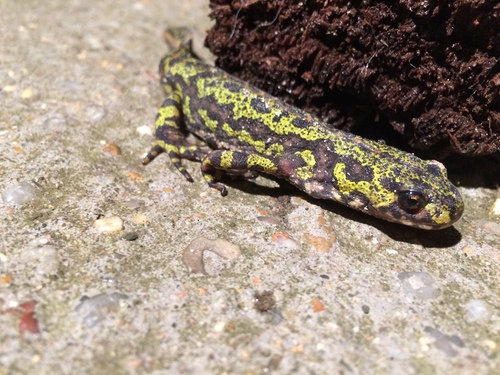}};
  \node[inner sep=0pt] at (i7) {\includegraphics[width=\imgwidth, height=\imgheight]{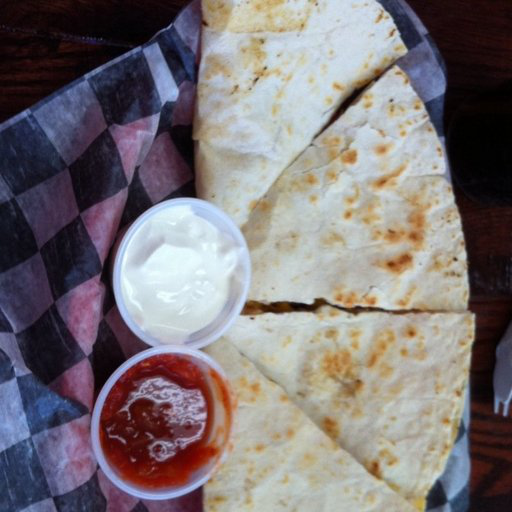}};
  \node[inner sep=0pt] at (i8) {\includegraphics[width=\imgwidth, height=\imgheight]{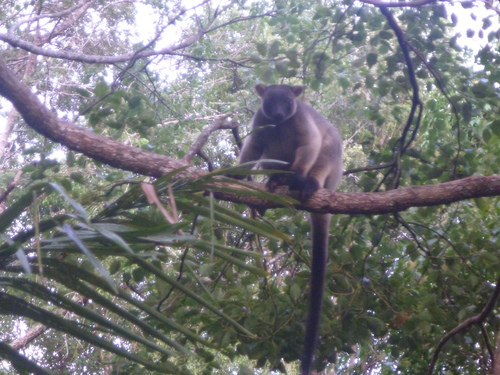}};
  \node[inner sep=0pt] at (j1) {\includegraphics[width=\imgwidth, height=\imgheight]{\INharda}};
  \node[inner sep=0pt] at (j2) {\includegraphics[width=\imgwidth, height=\imgheight]{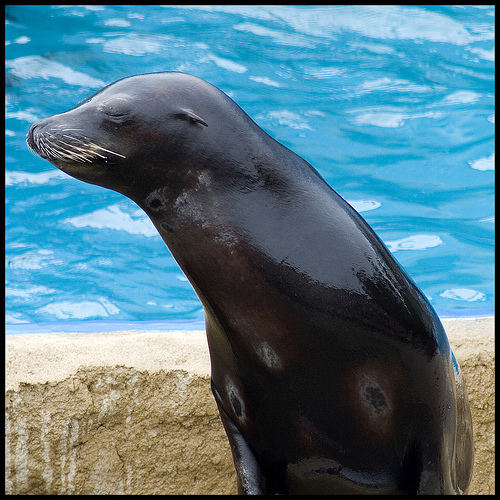}};
  \node[inner sep=0pt] at (j3) {\includegraphics[width=\imgwidth, height=\imgheight]{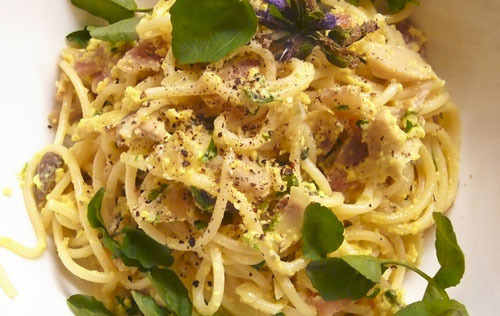}};
  \node[inner sep=0pt] at (j4) {\includegraphics[width=\imgwidth, height=\imgheight]{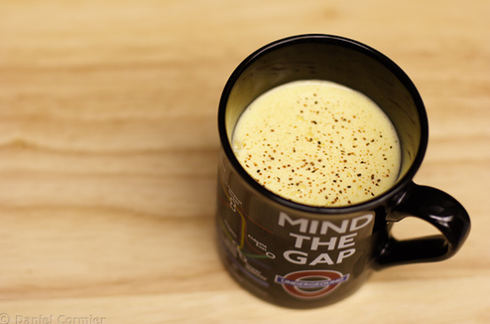}};
  \node[inner sep=0pt] at (j5) {\includegraphics[width=\imgwidth, height=\imgheight]{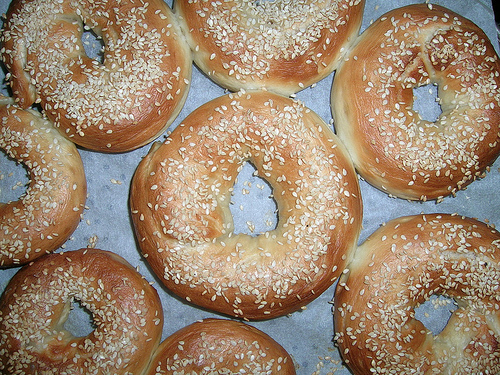}};
  \node[inner sep=0pt] at (j6) {\includegraphics[width=\imgwidth, height=\imgheight]{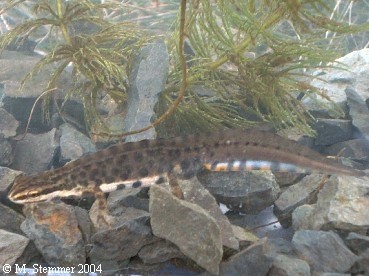}};
  \node[inner sep=0pt] at (j7) {\includegraphics[width=\imgwidth, height=\imgheight]{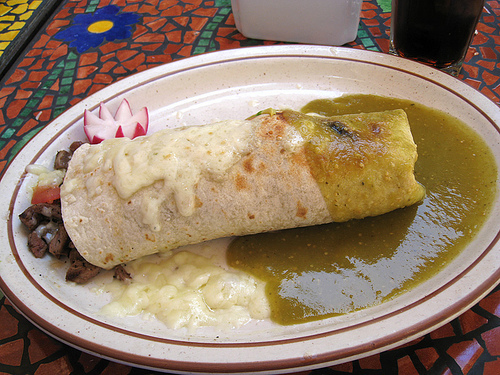}};
  \node[inner sep=0pt] at (j8) {\includegraphics[width=\imgwidth, height=\imgheight]{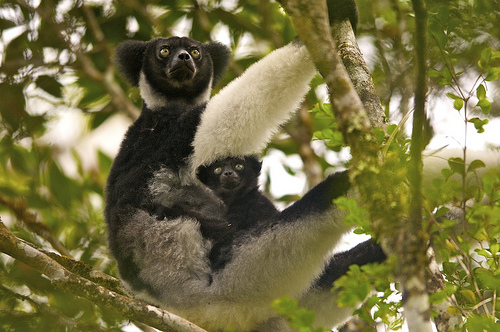}};
  \draw[anchor=base, align=center] (t1)  node[scale=\tinyscale] {\tiny\OODclass{\xtriple{tahard}{1}}\\[-2.2mm]\tiny\INclass{\xtriple{tahard}{3}}};
  \draw[anchor=base, align=center] (t2)  node[scale=\tinyscale] {\tiny\OODclass{\xtriple{tbhard}{1}}\\[-2.2mm]\tiny\INclass{\xtriple{tbhard}{3}}};
  \draw[anchor=base, align=center] (t3)  node[scale=\tinyscale] {\tiny\OODclass{\xtriple{tchard}{1}}\\[-2.2mm]\tiny\INclass{\xtriple{tchard}{3}}};
  \draw[anchor=base, align=center] (t4)  node[scale=\tinyscale] {\tiny\OODclass{\xtriple{tdhard}{1}}\\[-2.2mm]\tiny\INclass{\xtriple{tdhard}{3}}};
  \draw[anchor=base, align=center] (t5)  node[scale=\tinyscale] {\tiny\OODclass{\xtriple{tehard}{1}}\\[-2.2mm]\tiny\INclass{\xtriple{tehard}{3}}};
  \draw[anchor=base, align=center] (t6)  node[scale=\tinyscale] {\tiny\OODclass{\xtriple{tfhard}{1}}\\[-2.2mm]\tiny\INclass{\xtriple{tfhard}{3}}};
  \draw[anchor=base, align=center] (t7)  node[scale=\tinyscale] {\tiny\OODclass{\xtriple{tghard}{1}}\\[-2.2mm]\tiny\INclass{\xtriple{tghard}{3}}};
  \draw[anchor=base, align=center] (t8)  node[scale=\tinyscale] {\tiny\OODclass{\xtriple{thhard}{1}}\\[-2.2mm]\tiny\INclass{\xtriple{thhard}{3}}};
  \draw[color=OODclass, line width=0.4mm] ($(i1) + (-.55*\imgwidth,.47*\imgheight)$) -- ($(i1) + (-.55*\imgwidth,-.47*\imgheight)$) {};
  \draw[color=OODclass, line width=0.4mm] ($(i8) + (.55*\imgwidth,.47*\imgheight)$) -- ($(i8) + (.55*\imgwidth,-.47*\imgheight)$) {};
  \draw[color=INclass, line width=0.4mm] ($(j1) + (-.55*\imgwidth,.47*\imgheight)$) -- ($(j1) + (-.55*\imgwidth,-.47*\imgheight)$) {};
  \draw[color=INclass, line width=0.4mm] ($(j8) + (.55*\imgwidth,.47*\imgheight)$) -- ($(j8) + (.55*\imgwidth,-.47*\imgheight)$) {};
\end{tikzpicture}
}
\vskip -2.0mm
\caption{\label{fig:hard-samples}
\textbf{Difficult OOD classes in \dsetname{}}: Examples of images from some of \dsetname's most difficult (see Table~\ref{tab:all-datasets-fpr}) \OODclass{OOD classes} (first row) and from the \INclass{ImageNet-1K class} (second row) which the \OODclass{OOD class} is most frequently confused for.
}
\end{center}
\vskip \belowimgcaptionvskip
\vskip -2mm
\end{figure*}

\section{A new OOD test set for ImageNet-1K}\label{sec:new}
As discussed in Sec.~\ref{sec:intro}, an \textbf{OOD input for IN-1K} is an image that does not contain an object from one (or several) of the \numprint{1000} IN-1K classes.
These ImageNet classes are based on individual WordNet~\citep{oram2001wordnet} synsets, each consisting of one or more keywords that are synonymous in some context.
During the ImageNet creation process~\cite{imagenet_cvpr09}, images were first collected from the web by using variations of each keyword of a respective class and then verified by humans to fit its synset's definition.

Sourcing OOD test samples for ImageNet-1K from ImageNet-21K (or its subsets) based on class-labels has been leading to highly contaminated datasets (5 of the datasets in Table \ref{tab:percentage_ID_in_OOD_datasets} are sourced from ImageNet-21K and all contain between 20\% and 53\% ID samples and show significant categorical contamination).
This is partly due to the class-structure of those datasets: Both ImageNet-1K and ImageNet-21K contain leaf and internal nodes of the WordNet-tree as classes.
While the internal nodes of ImageNet-1K are not ancestors to other Imagenet-1K classes, ImageNet-21K internal nodes can be ancestors to ImageNet-1K nodes, and vice versa. Moreover, there are ambigous class-definitions in WordNet, like e.g. \sourceclass{police dog}, which is not parent or child of another dog class, but mostly shows a \INclass{german shepherd}, or an \sourceclass{alley cat} showing one of the many cat classes without being parent or child to other cat classes.
Besides, there is significant incidental contamination even for nominally disjoint classes. Since the automation of filtering for challenging OOD data would require a strong detector that already solves the problems that the dataset is meant to pose, we conclude that it is impossible to construct a clean and challenging OOD dataset without manually checking the OOD samples for ID contamination.

In reality, many ImageNet samples fit one but not necessarily \textit{all} keywords of their class label.
This means that to make sure that OOD detectors are treated fairly\footnote{For fair treatment of previous OOD \textit{datasets}, such unclear samples that don't fit all keywords were ignored in Table~\ref{tab:percentage_ID_in_OOD_datasets}.
},
OOD test samples cannot fall into the definition of any keyword of any IN-1K class.
For example, photos of the Sumatran orangutan cannot be considered OOD, since they could be included in the IN-1K class~\INclass{(orangutan, orang, orangutang, Pongo pygmaeus)}, even though \textit{Pongo pygmaeus} only refers to the Bornean orangutan.
To determine what counts as an ID object, we follow the WordNet glosses\footnote{One can look up synsets with glosses \href{http://wordnetweb.princeton.edu/perl/webwn}{\textcolor{blue}{here}}.} as well as dictionary definitions of keywords and source dataset class labels.
For difficult cases, we consult additional sources like Wikipedia.
For example, the species \sourceclass{northern elephant seal} does not fall into the ID class \INclass{sea lion}, among other biological criteria distinguished by the fact that the former do not have ears while the latter do.
An image of an OOD dataset can furthermore not incidentally contain ID objects, to avoid cases as in Figure~\ref{fig:teaser}  
 (bottom) and Figure~\ref{fig:better_model_positives}.

\subsection{\dsetname{} dataset construction}
For each OOD class of our new \dsetname{} dataset, we start by \textbf{choosing a base class} which consists of all samples from a named class of an existing or newly scraped dataset.
The majority of the \dsetname{} base classes are sourced from \textsc{Species}~\cite{hendrycks22Scaling}, which provides images scraped from iNaturalist.
For each base class, we carefully decide, based on WordNet glosses, iNaturalist taxonomy details and Wikipedia, whether it can be included according to the non-permissive interpretation described at the beginning of Section~\ref{sec:new}.
The choice of base classes is not random, since there is no way to randomly sample from the set of concepts that might occur at test time.
Rather, we aim for a variety of classes that are challenging, diverse and, most importantly, not actually categorically ID to begin with.
Then for each base class, we \textbf{individually inspect each image} for ID objects.
To help remembering the 1000 ID classes, we display the 5 top ID classes of a ViT's prediction on each image.
If an ID object is at least partially visible, the corresponding sample is removed.
In cases where it is ambiguous whether we see an ID object in the image, the sample is not included in the cleaned dataset.
As the iNaturalist data (including the \textsc{Species} dataset) has been curated by experts and can be considered very reliable, we generally trust in the main object belonging to the species it is labelled as.
For base classes chosen from the other sources, we consider ourselves competent to verify whether a label is correct.
In addition to samples showing ID objects, we also remove images where no object from the OOD class is visible, e.g. we exclude pictures of animal traces or remains which frequently appear in iNaturalist.
While for most existing datasets, the cleaning has been outsourced to external services like Amazon Mechanical Turk or student labellers.
By researching all OOD classes and visually inspecting all their samples ourselves, we as authors of \dsetname{} were able to do more in-depth research for each ambiguous case and obtain more coherent decisions, which we are positive leads to a higher quality dataset.
Such high data quality is crucial for in-depth evaluations \cite{vasudevan2022when, shankar2021evaluating}, as only being completely in-distribution free allows understanding a detector’s individual mistakes.

The \dsetname{} (\textbf{N}o \textbf{I}mageNet \textbf{C}lass \textbf{O}bjects) dataset consists of 64 OOD classes with a total of \numoodsamples{} samples.
The base classes which we cleaned to obtain \dsetname{} were sourced from \textsc{Species} (35 classes) \cite{hendrycks22Scaling}, \textsc{Places} (3 classes) \cite{zhou2017places}, which both are discussed in Section~\ref{sec:existing},
as well as from the \textsc{Food-101} dataset (7 classes) \cite{BossardFood101}, \textsc{Caltech-101} (4 classes) \cite{caltech101}, \textsc{MyNursingHome} (4 classes) \cite{MyNursingHome}, ImageNet-21k (1 class) and newly scraped from \url{iNaturalist.org} (2 classes) or other websites like \href{https://www.flickr.com/}{Flickr} (8 classes).
Details for all \dsetname{} OOD classes are given in Appendix~\ref{sec:dset_details}.
We show samples from all \dsetname{} classes in Figures~\ref{fig:OODsamples1} and \ref{fig:OODsamples2} in Appendix~\ref{sec:dset_examples}.
In addition to \dsetname{}, we also provide the 2715 OOD images obtained from cleaning 400 samples of eleven test OOD datasets as discussed in Section~\ref{sec:effect_of_cleaning}.
In order to notice ID contaminations potentially biasing the drawn conclusions, we recommend to also evaluate on these cleaned versions when evaluating on those original benchmarks.

\begin{figure*}[t]
    \centering
    \includegraphics[width=1.0\textwidth]{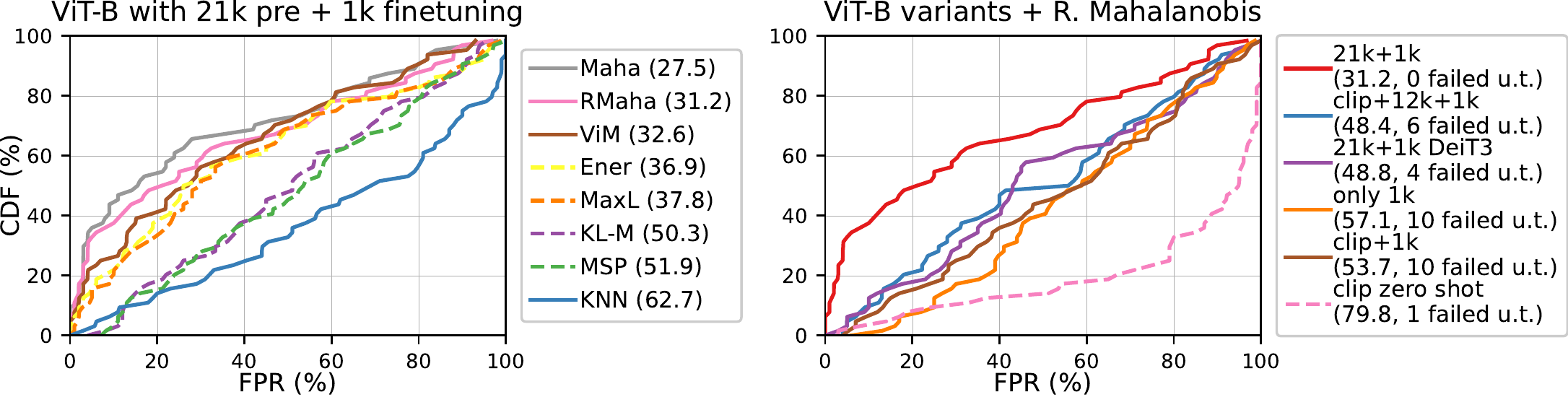}
    \vskip -1.2mm
    \caption{Cumulative distribution of the \% of \dsetname-classes for which an \FPR{} at least as low as a given x-value is achieved.
The area over this curve corresponds to the mean \FPR. The further in the top left corner, the better.
\textbf{The best methods explicitly access pre-logit features (Left):} Different OOD detection methods with a ViT-B pretrained on IN-21k (mean \FPR{} in parentheses, pre-logit feature-accessing methods are solid, others dashed).
    \textbf{Not all pretraining helps (Right):} RMaha applied to ViT-B with different training variants (MCM for CLIP zero-shot is dashed). Only the top model does not fail OOD unit-tests.
    }
    \label{fig:cdf-fpr-vits}
    \vskip \belowimgcaptionvskip
\end{figure*}

\subsection{OOD unit-tests}
Following common practice (e.g. \citet{hendrycks22Scaling}), we argue that evaluating an OOD detector on a range of simple, synthetic classes \textit{besides} the variably challenging natural image classes of an OOD dataset can give additional insights about its OOD detection weaknesses.
Example images and reproducibility details for all 17 pre-existing and newly proposed OOD unit-tests are included in Appendices~\ref{sec:OOD_unit_tests} and~\ref{sec:dset_examples}.
Since these \textbf{OOD unit-tests} do not represent a diverse distribution of photos, but different modes of simple, synthetically generated image inputs which any good OOD detector should be expected to detect, we don't include them in summary metrics or distribution plots.
Instead, we suggest to count an OOD unit test as \textbf{failed} if a method has an \FPR{} above a user-defined threshold, which we suggest setting at 10\%,
and to report the number of \textit{failed} OOD unit-tests (which should be 0 for a strong OOD detector) alongside the aggregate results on a test OOD dataset like \dsetname{}.
For each OOD unit-test, we provide a set of 400 samples in typical ImageNet format, by mirroring the sizes and file formats of random ImageNet samples.
While some OOD unit-tests may appear redundant at first sight, we find that they provide important information as some detectors e.g. mostly pass the \OODclass{monochrome} test but completely fail on \OODclass{black}, which reveals a specific weakness that is %
very realistic to be encountered in practice. 

\subsection{OOD detectors and how to evaluate them}
An \textbf{OOD detector} for inputs from the domain $X$ of possible input images is represented by a score function ${S:X \rightarrow \mathbb{R} \cup \{\pm \infty \}}$ which is generally supposed to be larger on ID inputs than on OOD inputs.
One example is the Maximum Softmax Probability (MSP) or confidence $S_{\text{MSP}}(x) = \max_{k=1,\ldots,K}p_{k}(x)$ of a classifier with output probabilities $p$ for $K$ ID classes.
The MSP is the standard baseline OOD detection method~\cite{hendrycks2017MSP}, since it is intuitively expected to be low on OOD compared to ID inputs.
Observing that standard classifiers are frequently overconfident on OOD inputs, OOD detection research aims at finding detectors that improve on this baseline.
In Appendix~\ref{sec:methods}, we give an overview of a range of OOD detection methods which have been proposed for IN-1K as ID.
An OOD detector is usually obtained by combining such an OOD detection method with a concrete classifier model.
We analyze OOD detectors in terms of the fraction of falsely accepted OOD inputs at a true positive rate of 95\%, short \FPR.
Detailed definitions can be found in Appendix~\ref{sec:fpr_def}.

Different OOD classes (and similarly also different test OOD datasets) represent different probabilistic distributions of inputs that a detector is tested against.
An important arising question is how the collective of individual performance measurements can be interpreted and whether they can be aggregated into one number that can be used to make an informed decision on which OOD detector works best.
Certainly, the notion of ‘best’ may notably vary depending on the application and situation and we often cannot hope to model a ‘true’ out-distribution, or even be sure that it meaningfully exists.
An aggregate number which gives a good overview of an OOD detector's performance on the class based \dsetname{} dataset %
is the \textbf{mean \FPR{}} of the individual FPR values for each of the 64 OOD classes of \dsetname{}.

However, for many applications it is not possible to model the potential OOD inputs that might be encountered at test time with a fixed probability distribution. Thus
a single aggregate number cannot tell the full story, and may hide outliers in the FPR values.
For one, some errors might be \textit{less acceptable} than others,
e.g. a \FPR{} of 20.0\% might be very bad for monochrome inputs, but would lose much significance when subsumed into a mean.
For OOD unit tests, where OOD detectors can be expected to be very robust, we therefore propose regarding pass-fail statistics instead of mean \FPR.
Also, an evaluator might want to be informed about the \textit{concrete failure modes} of the model, e.g. all OOD classes with a particular high \FPR.
An OOD detector showing consistent improvements on most of the OOD classes (instead of only in terms of the mean) can be seen as strong evidence for the method yielding actual improvement, as opposed to the detector overfitting to a limited scope of test OOD data, which \citet{wang2022vim} describe as a form of hackability.
Due to these considerations, and with the OOD data being organized into \textit{OOD classes} as in \dsetname{}, we suggest evaluations of OOD detectors to always provide the \textbf{distribution of results over OOD classes} and additionally to \textbf{make the individual results available}, such that the reader can make an informed comparison based on which types of OOD inputs are most relevant to them.

\section{Evaluation results for OOD Detectors}\label{sec:eval}

We evaluate a range of IN-1K models obtained from the public timm-library \cite{rw2019timm} and state-of-the-art OOD-detection methods on \dsetname.
We focus on transformer architectures and convolutional networks, both with and without pretraining.
While most pretrained models were initially trained on IN-21K, we also include an EfficientNet trained via noisy student \cite{xie2019noisyStudent} on the JFT-300M dataset, and four ViTs with CLIP-pretraining \cite{RadfordClip} and subsequent fine-tuning, as well as a zero-shot CLIP model.
A detailed description of all models can be found in %
Appendix~\ref{sec:models}.
We investigate the following commonly used OOD detection methods, which can be grouped into two categories:
Max-Softmax (MSP) \cite{hendrycks2017MSP}, Max-Logit \cite{hendrycks22Scaling}, Energy \cite{liu2020energy} and KL-Matching \cite{hendrycks22Scaling} derive an OOD-score exclusively from logit outputs, whereas  Mahalanobis distance (Maha) \cite{LeeMahalanobis2018}, Virtual Logit Matching (ViM) \cite{wang2022vim}, ReAct \cite{sun2021react}, Relative Mahalanobis distance (RMaha) \cite{RenRelMaha2021}, and K-Nearest-Neighbours (KNN) \cite{sun2022knnood} also leverage explicit information from the features of the DNN's penultimate (pre-logit) layer.
For the zero-shot evaluation of CLIP, we use Maximum-Concept-Matching (MCM) \cite{ming2022delving} and Cosine-similarity (Cos) \cite{anonymous2023COOD} to class-specific text-embeddings.
Noting that OOD detection based on softmax of a cosine similarity to a specific feature vector has been proposed in different variants (\citet{tack2020csi}, \citet{techapanurak2020hyperparameter} and MCM),we find that using it with classifier class means produces reasonable OOD detection results, marked below as relative cosine class similarity (RCos).
We call those methods which explicitly access the pre-logit feature layer \textbf{\textit{feature-based}} and provide an overview over all methods in Appendix~\ref{sec:methods}.

\begin{figure*}[t]
    \centering
    \includegraphics[width=1.\textwidth]
    {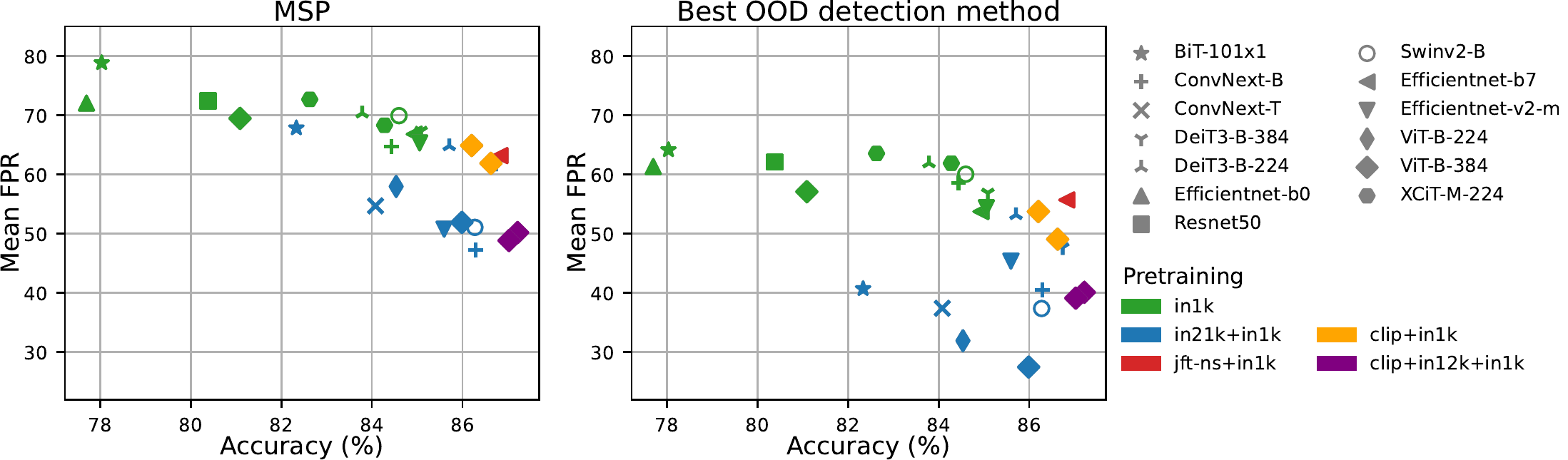}
    \vskip -1.4mm
    \caption{\textbf{IN-21K pretraining boosts feature-based OOD detectors on \dsetname{}:} Mean \FPR\,vs.\,accuracy for MSP and each model's best detector, which (except for the noisy-student model) always explicitly accesses the pre-logit features. OOD detection strongly improves when using models pretrained on IN-21K. %
    Additional CLIP-pretraining or on JFT can yield higher accuracy, but OOD detection need not be better than with IN-21K pretraining.
    }\label{fig:acc-vs-fpr}
    \vskip \belowimgcaptionvskip
\end{figure*}

\subsection{Results on \dsetname}\label{sec:results-on-benchmark}

\textbf{Comparison of OOD detection Methods.}
In Figure~\ref{fig:cdf-fpr-vits} (left), we illustrate the performance of a single ViT %
when combined with a range of OOD-methods.
Overall, most feature-based methods, like Maha, RMaha and ViM, outperform the MSP-baseline by a clear margin.
Notably, MaxLogit and Energy, which do not explicitly access the pre-logit features, are also able to strongly improve over MSP, while KL-Matching performs roughly on par, and KNN much worse.
We observe that while Maha, RMaha and ViM improve over MSP in all \FPR{} ranges, this is different for e.g. MaxLogit:
For large \FPR{}, it is similar to MSP, indicating that the method brings no advantage over MSP for hard test classes, and its improved mean performance is mainly due to lower \FPR{} for the easier OOD classes.
When regarding the mean \FPR{} values of all method-model-combinations shown in Table~\ref{tab:overview-fpr-mean} in Appendix~\ref{app:detailed_results},
we observe that while Maha in combination with a (pretrained) ViT is the single best OOD-detector, 
this method often performs worse when combined with other models.
RMaha, however, yields good results with \textit{all} models, and is together with (Relative) Cosine the only method which can fairly consistently improve over the MSP baseline in terms of mean \FPR{}.
For most models, it is either the best-performing method, or close to the best-performing method, which is somewhat surprising, given its relatively poor performance on the unit-tests.
We further note that for all models (except the noisy-student model), the best-performing method always explicitly accesses the pre-logit features, and that in contrast to e.g. KNN, Energy and ReAct, even the adapted methods based on feature space cosine similarity Cos and MCM/RCos fairly consistently improve over the MSP-baseline.
Each OOD dataset representing a different out-distribution that can be relevant for certain applications, 
we find that results vary on the cleaned subsets of eleven previous benchmarks which we evaluate in Appendix~\ref{sec:Effect_cleaning_all},
while the overall conclusions on the methods and models resemble those on \dsetname.
\begin{figure*}[t]
    \vskip 0in
    \centering
    \includegraphics[width=1.\textwidth]{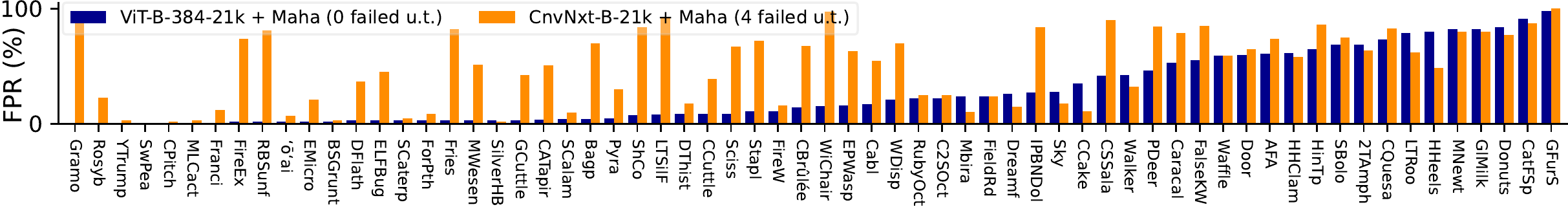}
    \vskip -0.1 in
    \caption{\FPR{} of a pretrained ViT-B and pretrained ConvNext-B for all classes of \dsetname{}.}
    \label{fig:barplot-all-classes}
    \vskip \belowimgcaptionvskip
\end{figure*}

\textbf{Pretraining matters.}
In Figure~\ref{fig:acc-vs-fpr}, we plot the mean \FPR{} on \dsetname{} over the accuracy for all investigated models for both the MSP-baseline (left) and the best-performing OOD detector per model (right).
For MSP, the mean \FPR{} decreases roughly linearly with accuracy.
Since most pretrained models (blue) have higher accuracy, they typically also show better OOD-detection performance,
but also between models of similar accuracy, the pretrained ones achieve better mean \FPR.
For the best-performing OOD detector, improvements can be observed for models both with and without pretraining.
Notably, the linear relation between \FPR{} and accuracy disappears, and all purely 1K models (green) perform roughly on one level.
In comparison, the gains for the majority of models pretrained on IN-21K (blue) are larger.
In particular ViT and BiT benefit strongly from leveraging their respective best method, which as discussed above is always feature-based.
In other words, pretraining helps %
in two ways:
First, it leads to higher ID-performance (accuracy), which benefits methods like the MSP-baseline.
Second, it creates better feature-embeddings for this task, which lead to improvements beyond the accuracy-MSP correlation.
This is most clearly visible for the pretrained BiT-m, which has comparably low accuracy ($82\%$) and hence no outstanding MSP-performance, but outperforms all 1k-models by a significant margin with features leveraging ViM.
However, as we observe in Figure~\ref{fig:cdf-fpr-vits} (right), the benefit of pretraining depends strongly on the specific data and training method:
With RMaha, the ViT with 'traditional' IN-21K pretraining from \cite{steinerhowtotrainyourvit} clearly outperforms models with the distillation-based training of DeiT3 \cite{Touvron2022DeiTIR}, CLIP-pretraining or even CLIP with interjected IN-12K training.
The zero-shot methods for CLIP, despite having shown promising results in \cite{anonymous2023COOD} and \cite{ming2022delving} and performing well on 
the unit tests, are not competitive to IN-1k classifiers on \dsetname.
Regarding all methods, the five models trained with different pretraining strategies (EfficentNet-b7 with noisy student and four ViTs with CLIP-pretraining \cite{RadfordClip} and subsequent fine-tuning) show some of the highest accuracies in our survey, yet, their OOD-detection performance is surprisingly poor.
Overall, we see strong indication that the precise type of pretraining has a large impact on whether it produces a feature space that is beneficial %
for feature based methods.
In Appendix~\ref{sec:overlap21K} we investigate whether IN-21K-pretraining particularly benefits detection of OOD classes that overlap with IN-21K classes, but we notice no substantially different changes between the model with and without pretraining.

\textbf{Analysis of failure cases.}
In Figure~\ref{fig:barplot-all-classes} we plot the individual \FPR{} for each OOD class of \dsetname{} for the combination ViT+Maha, the overall best OOD detector in terms of mean \FPR{}, and contrast it with ConvNext+Maha, which also shows good mean \FPR{}.
Performance varies widely between OOD classes, with both models severely struggling for some classes.
Where the ViT shows large \FPR{}, the ConvNext rarely performs better, while it also fails to detect certain classes like the \OODclass{long-tailed silverfish} where the ViT does well.
We illustrate samples from hard classes in Figure \ref{fig:hard-samples}.
Both models struggle to detect the \OODclass{Galápagos fur seal} (98\% FPR for the ViT), often confused with the IN-1K class \INclass{sea lion}, and \OODclass{cat-faced spider} (confused with \INclass{barn spider}, 91\% FPR).
From a human perspective, those classes are arguably hard to detect. 
We note, however, that it is possible to tell them apart, as a ViT IN-21K-classifier e.g. identifies the \OODclass{Galápagos fur seal} as a \INTOKclass{fur seal} (IN-21K class) in 92\% of samples and misclassifies only 6\% of them as a \INTOKclass{sea lion}.
The networks however also fail for classes more obvious to humans: \OODclass{donut} (84\% FPR ViT, confused with \INclass{bagel}), \OODclass{spaghetti bolognese} (69\% FPR, \INclass{carbonara}) and \OODclass{chicken quesadilla} (73\% FPR, \INclass{burrito}) also confuse both models.

\subsection{Results on the OOD unit-tests}
Auditing OOD detectors on the OOD unit-tests, we find that surprisingly many combinations of models and OOD detection methods struggle to distinguish supposedly easy inputs from ID-data.
While results for all models and methods can be found in Appendix~\ref{app:unit-tests}, we provide some illustrative unit-test results in Table~\ref{tab:unit-test-conv} for a ViT pretrained on IN-21k and a ConvNext both with and without IN-21K pretraining.
In general, most methods fail fewer unit tests when applied to pretrained models, however there are still many severely flawed combinations, often involving methods that would otherwise shine based on their detection of natural OOD data discussed above:
especially the feature-based methods ViM, Maha and RMaha reveal weaknesses, each failing multiple unit-tests on at least 21 of 26 models. %
Many tested OOD detectors are vulnerable to \OODclass{black}, \OODclass{white} and \OODclass{grey}, which is concerning as encountering inputs of this kind could occur in many real-world applications due to camera malfunction or occlusion.
\begin{table}[!h] %
\vskip -2.5mm
    \centering
    \caption{\textbf{Some detectors fail OOD unit-tests:}
    \FPR{} for a ViT and a ConvNext (with and without pretraining) on selected unit-tests. \FPR{} larger than 10\% count as failed and are thus marked red.
    Especially for methods relying on feature representations (like ViM and Maha) the OOD unit-tests reveal difficulties.}
    \label{tab:unit-test-conv}
    \vskip -9.mm
\tabcolsep=0.12cm

\begin{small}

\begin{center}
\begin{tabular}{c c c c c c c c c}
&  method  & \OODclass{bla} & \OODclass{whi} & \OODclass{gre} & \OODclass{hor} & \OODclass{SmN} & \OODclass{Rad} & \OODclass{mon} \\ 
 \hline 
\parbox[t]{2mm}{\multirow{4}{*}{\rotatebox[origin=c]{90}{ViT21k}}}&MSP & 0.0 & 0.0 & 0.0 & 0.2 & 0.5 & 0.0 & 0.0 \\ 
& ViM & 0.0 & $\color{red}{100.0}$ & $\color{red}{46.0}$ & 0.0 & 0.0 & 0.0 & 0.5 \\ 
& Maha & 0.0 & 0.0 & 0.0 & 0.0 & 0.0 & 0.0 & 0.0 \\ 
& Cos & 0.0 & 0.0 & 0.0 & 0.0 & 0.0 & 0.0 & 0.0 \\ 
 \hline 
\parbox[t]{2mm}{\multirow{4}{*}{\rotatebox[origin=c]{90}{Cnv1k}}}&MSP & 0.0 & 0.0 & 0.0 & $\color{red}{60.5}$ & 0.8 & 0.0 & 0.0 \\ 
& ViM & $\color{red}{100.0}$ & $\color{red}{100.0}$ & $\color{red}{100.0}$ & $\color{red}{98.0}$ & $\color{red}{24.5}$ & $\color{red}{100.0}$ & $\color{red}{100.0}$ \\ 
& Maha & $\color{red}{100.0}$ & $\color{red}{100.0}$ & $\color{red}{100.0}$ & $\color{red}{87.5}$ & $\color{red}{27.5}$ & $\color{red}{100.0}$ & $\color{red}{100.0}$ \\ 
& Cos & 0.0 & 0.0 & 0.0 & $\color{red}{27.5}$ & 0.0 & 0.0 & 0.0 \\ 
 \hline 
\parbox[t]{2mm}{\multirow{4}{*}{\rotatebox[origin=c]{90}{Cnv21k}}}&MSP & 0.0 & 0.0 & 0.0 & $\color{red}{13.5}$ & 2.2 & 0.0 & 0.0 \\ 
& ViM & $\color{red}{100.0}$ & $\color{red}{100.0}$ & $\color{red}{100.0}$ & 0.0 & 0.0 & $\color{red}{41.2}$ & 0.5 \\ 
& Maha & $\color{red}{100.0}$ & $\color{red}{100.0}$ & $\color{red}{100.0}$ & 0.0 & 0.0 & $\color{red}{42.5}$ & 2.8 \\ 
& Cos & 0.0 & 0.0 & 0.0 & 0.0 & 0.0 & 0.0 & 0.0 \\ 
 \hline 
\end{tabular}
\end{center}

\end{small}
\vskip -3mm
\end{table}
Here those feature-based methods only provide trustworthy results in combination with ViTs pretrained on IN-21k, the BiT-models and a pretrained EfficientNet-V2.
Methods like Cos (7/26 models fail multiple tests) and MCM/RCos (7/26), originally designed for cosine-trained features as in CLIP, achieve remarkably strong OOD-detection performance on the unit-tests across a broad range of models, both with and without CLIP-pretraining.
While taking note of these general trends, each OOD detector's robustness to the OOD unit-tests should be examined individually.

\section{Conclusions}
\label{sec:conclusions}

We introduce with \dsetname{} a novel, ID-contamination-free and challenging OOD test-dataset for IN-1K with fine-grained class-resolution.
We find that many OOD detectors work better than previously thought, when their recorded number of undetected OOD inputs is not inflated by ID contaminations.
However, most detection methods cannot reliably be applied with arbitrary classifier models, as even OOD unit-tests are failed by many combinations.
We are hopeful for \dsetname{} and the cleaned test OOD subsets to facilitate the more precise development of reliable OOD detectors
which do not try to avoid presumed failures which are actually correct decisions.

\subsection*{Acknowledgements}
\label{sec:acknowledgements}
We thank Vaclav Voracek for helpful discussions and suggesting the cdf plots.
We acknowledge support from the German Federal Ministry of Education and Research (BMBF) through the Tübingen AI Center (FKZ: 01IS18039A) and from the Deutsche Forschungsgemeinschaft (DFG, German Research Foundation) under Germany’s Excellence Strategy (EXC number 2064/1, Project number 390727645), as well as from
the Carl Zeiss Foundation in the project “Certification and Foundations of Safe Machine Learning Systems in Healthcare”.
We also thank the European Laboratory for Learning and Intelligent Systems (ELLIS) for supporting Maximilian Müller.

\FloatBarrier
\bibliography{main.bib}
\bibliographystyle{icml2023}

\newpage
\appendix
\onecolumn
\FloatBarrier
\section{Detailed results on \dsetname{}}\label{app:detailed_results}
A detailed overview over the results on the \dsetname{} benchmark is presented in Table \ref{tab:overview-fpr-mean}, where we show the mean \FPR{} for all models and methods across the dataset's OOD classes. Tables \ref{tab:overview-auroc-mean}-\ref{tab:overview-auprcr-mean} show AUROC, AUPR-S and AUPR-E with the same conclusions.
The best method per model is marked bold, and the difference to the MSP-baseline is shown in green where a model outperforms the MSP-baseline and in red if it performs worse than MSP. %
It is clearly visible that there is no one-fits-all method.
Instead, different models synergize with different methods.
Overall, the two ViT models pretrained only on IN-21K in combination with Mahalanobis distance outperform other models and methods by a clear margin.
This is in line with the observations of previous works \cite{koner2021oodformer, fort2021exploring, anonymous2023COOD}, which also found the ViTs to perform exceptionally well.
In terms of MSP, the ViTs are not better than e.g. the ConvNext, indicating that their improved OOD detection capabilities stem from a favourably structured feature-space.
It is further interesting to see that for models without pretraining, out of all methods only Relative Mahalanobis and the cosine-based methods improve over the MSP-baseline fairly consistently. Apart from KL-Matching and KNN, most methods improve over the MSP-baseline for most pretrained models and the CLIP-methods Cosine and RCos perform comparably well, yielding their best results with models pretrained \textit{both} on CLIP and IN-12k. 
Since CLIP models are trained with cosine-similarity, it is likely that the structure of the feature space after finetuning remains favorable to cosine-based methods, while it might harm the performance of other feature-based methods like Mahanobis compared to models pretrained \textit{only} on IN-21k. 

It has been remarked \cite{hendrycks22Scaling} that the advantage of models pretrained with IN-21K in the OOD detection task CIFAR-10 vs. CIFAR-100 \cite{krizhevsky2009learning} might partially be explained by the CIFAR-100 classes not truly being unseen at train time, as they have a large overlap with IN-21K classes.
We checked each \dsetname{} class for overlap with the \numprint{21843} classes of IN-21K with the help of a ViT classifier for IN-21K, see Table~\ref{tab:OOD_class_info}.
This allows us to test whether the pretrained models have a larger advantage over purely IN-1K-trained models when trying to detect those classes with overlap compared to the classes without overlap.
In Appendix~\ref{sec:overlap21K} notice no substantially different changes between the models with and without pretraining.
We remark, however, that even for several models without pretraining, the subselections of classes show quite different results.

\begin{figure}[htb]
    \vskip -0.5mm
    \centering
    \includegraphics[width=1.\linewidth]{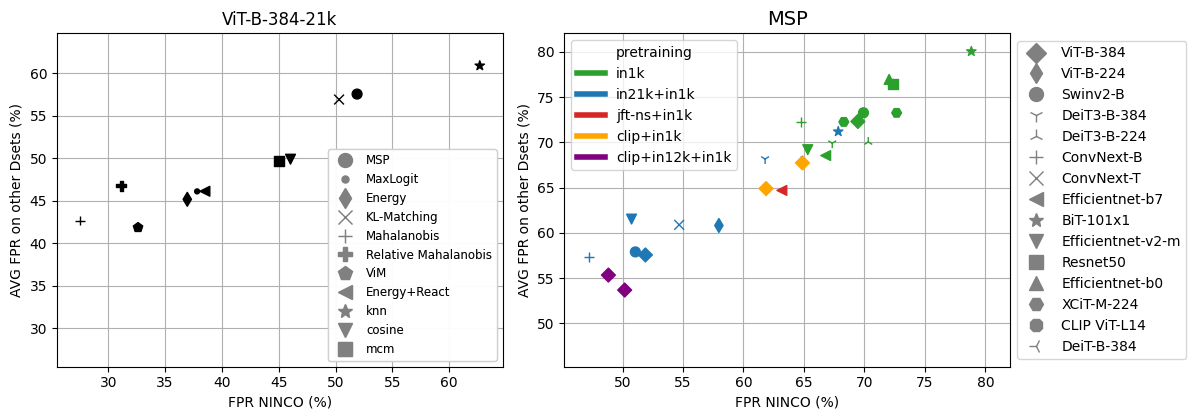}
    \vskip -3.8mm
    \caption{\textbf{Mean \FPR{} on \dsetname{} vs. mean-\FPR{} on previously used contaminated datasetes} with fixed model (left) and fixed method (right). We observe several ranking changes, including the best-performing method and model.}
    \label{fig:NINCO-vs-AVG-subsampled-fpr}
    \vskip \belowimgcaptionvskip
\end{figure}

In Figure \ref{fig:NINCO-vs-AVG-subsampled-fpr} we contrast the results on \dsetname{} with the results from previously used datasets. We show all methods for a pretrained ViT-B-384 and all models for the MSP-baseline. In both cases we observe several ranking changes: For the ViT, the best-performing method changes from ViM to Mahalanobis, and Relative Mahalanobis improves from sixth to second place. For the MSP-baseline, the clip-pretrained ViTs were the strongest OOD detectors on the previously used datasets, but are outperformend by the ConvNext-B on \dsetname{}.

\renewcommand{\arraystretch}{1.}
\tabcolsep=0.10321cm
\begin{table*}[htb]
    \centering
    \caption{\textbf{Mean \FPR{} on our \dsetname{} dataset.} Lower is better. The difference to MSP is shown in red if a method performs worse, and in green if it improves. Bold values mark the best-performing method per model.}
    \label{tab:overview-fpr-mean}
    \small
    \begin{smaller}
\begin{center}
\begin{tabular}{l l l l l l l l l l l l l l }
pre & acc. & model & MSP & MaxL & Ener & KL-M & Maha & RMaha & ViM & E+R & KNN & Cos & MCM/RCos \\ 
 \hline 
{\multirow{9}{*}{{21k}}} & 86.0 & ViT-B-384 & 51.9 & 37.8 ${\textcolor{green}{-14}}$ & 36.9 ${\textcolor{green}{-15}}$ & 50.3 ${\textcolor{green}{-2}}$ & \textbf{27.5} ${\textcolor{green}{-24}}$ & 31.2 ${\textcolor{green}{-21}}$ & 32.6 ${\textcolor{green}{-19}}$ & 38.5 ${\textcolor{green}{-13}}$ & 62.7 ${\textcolor{red}{+11}}$ & 46.0 ${\textcolor{green}{-6}}$ & 45.0 ${\textcolor{green}{-7}}$ \\ 
  & 84.5 & ViT-B-224 & 58.0 & 46.5 ${\textcolor{green}{-12}}$ & 46.1 ${\textcolor{green}{-12}}$ & 57.2 ${\textcolor{green}{-1}}$ & \textbf{31.9} ${\textcolor{green}{-26}}$ & 36.8 ${\textcolor{green}{-21}}$ & 38.4 ${\textcolor{green}{-20}}$ & 49.4 ${\textcolor{green}{-9}}$ & 68.8 ${\textcolor{red}{+11}}$ & 54.7 ${\textcolor{green}{-3}}$ & 54.3 ${\textcolor{green}{-4}}$ \\ 
  & 86.3 & Swinv2-B-256 & 51.1 & 41.1 ${\textcolor{green}{-10}}$ & 40.0 ${\textcolor{green}{-11}}$ & 56.0 ${\textcolor{red}{+5}}$ & 62.8 ${\textcolor{red}{+12}}$ & 53.8 ${\textcolor{red}{+3}}$ & 54.8 ${\textcolor{red}{+4}}$ & \textbf{37.4} ${\textcolor{green}{-14}}$ & 61.9 ${\textcolor{red}{+11}}$ & 51.4 ${\textcolor{red}{+0}}$ & 48.2 ${\textcolor{green}{-3}}$ \\ 
  & 86.7 & Deit3-B-384 & 61.8 & 56.0 ${\textcolor{green}{-6}}$ & 56.3 ${\textcolor{green}{-5}}$ & 60.3 ${\textcolor{green}{-1}}$ & 53.9 ${\textcolor{green}{-8}}$ & 48.8 ${\textcolor{green}{-13}}$ & 56.9 ${\textcolor{green}{-5}}$ & 51.6 ${\textcolor{green}{-10}}$ & 53.4 ${\textcolor{green}{-8}}$ & 48.4 ${\textcolor{green}{-13}}$ & \textbf{47.7} ${\textcolor{green}{-14}}$ \\ 
  & 85.7 & Deit3-B-224 & 64.8 & 59.2 ${\textcolor{green}{-6}}$ & 58.1 ${\textcolor{green}{-7}}$ & 65.2 ${\textcolor{red}{+0}}$ & 60.0 ${\textcolor{green}{-5}}$ & 53.8 ${\textcolor{green}{-11}}$ & 62.5 ${\textcolor{green}{-2}}$ & 55.2 ${\textcolor{green}{-10}}$ & 58.7 ${\textcolor{green}{-6}}$ & 54.2 ${\textcolor{green}{-11}}$ & \textbf{53.2} ${\textcolor{green}{-12}}$ \\ 
  & 86.3 & CnvNxt-B & 47.2 & 41.1 ${\textcolor{green}{-6}}$ & 43.3 ${\textcolor{green}{-4}}$ & 54.9 ${\textcolor{red}{+8}}$ & 49.6 ${\textcolor{red}{+2}}$ & 42.4 ${\textcolor{green}{-5}}$ & 41.5 ${\textcolor{green}{-6}}$ & \textbf{40.5} ${\textcolor{green}{-7}}$ & 51.8 ${\textcolor{red}{+5}}$ & 44.2 ${\textcolor{green}{-3}}$ & 42.6 ${\textcolor{green}{-5}}$ \\ 
  & 84.1 & CnvNxt-T & 54.7 & 47.9 ${\textcolor{green}{-7}}$ & 45.4 ${\textcolor{green}{-9}}$ & 60.7 ${\textcolor{red}{+6}}$ & 46.9 ${\textcolor{green}{-8}}$ & 45.8 ${\textcolor{green}{-9}}$ & \textbf{37.4} ${\textcolor{green}{-17}}$ & 44.1 ${\textcolor{green}{-11}}$ & 56.6 ${\textcolor{red}{+2}}$ & 51.2 ${\textcolor{green}{-4}}$ & 49.2 ${\textcolor{green}{-5}}$ \\ 
  & 82.3 & BiT-m & 67.8 & 62.0 ${\textcolor{green}{-6}}$ & 63.2 ${\textcolor{green}{-5}}$ & 64.9 ${\textcolor{green}{-3}}$ & 50.0 ${\textcolor{green}{-18}}$ & 45.1 ${\textcolor{green}{-23}}$ & \textbf{40.7} ${\textcolor{green}{-27}}$ & 57.1 ${\textcolor{green}{-11}}$ & 58.0 ${\textcolor{green}{-10}}$ & 51.6 ${\textcolor{green}{-16}}$ & 54.4 ${\textcolor{green}{-13}}$ \\ 
  & 85.6 & EffNetv2-M & 50.7 & 48.3 ${\textcolor{green}{-2}}$ & 54.1 ${\textcolor{red}{+3}}$ & 54.6 ${\textcolor{red}{+4}}$ & 62.9 ${\textcolor{red}{+12}}$ & 51.6 ${\textcolor{red}{+1}}$ & 53.5 ${\textcolor{red}{+3}}$ & 89.8 ${\textcolor{red}{+39}}$ & 67.5 ${\textcolor{red}{+17}}$ & \textbf{45.4} ${\textcolor{green}{-5}}$ & 50.6 ${\textcolor{green}{-0}}$ \\ 
 \hline 
{\multirow{12}{*}{{none}}} & 81.1 & ViT-B-384 & 69.5 & 67.7 ${\textcolor{green}{-2}}$ & 68.1 ${\textcolor{green}{-1}}$ & 66.7 ${\textcolor{green}{-3}}$ & 60.0 ${\textcolor{green}{-9}}$ & \textbf{57.1} ${\textcolor{green}{-12}}$ & 69.4 ${\textcolor{green}{-0}}$ & 65.8 ${\textcolor{green}{-4}}$ & 73.6 ${\textcolor{red}{+4}}$ & 68.7 ${\textcolor{green}{-1}}$ & 69.8 ${\textcolor{red}{+0}}$ \\ 
  & 84.6 & Swinv2-B-256 & 69.9 & 67.6 ${\textcolor{green}{-2}}$ & 72.2 ${\textcolor{red}{+2}}$ & 67.5 ${\textcolor{green}{-2}}$ & 63.9 ${\textcolor{green}{-6}}$ & \textbf{60.0} ${\textcolor{green}{-10}}$ & 66.5 ${\textcolor{green}{-3}}$ & 68.8 ${\textcolor{green}{-1}}$ & 69.2 ${\textcolor{green}{-1}}$ & 63.5 ${\textcolor{green}{-6}}$ & 62.0 ${\textcolor{green}{-8}}$ \\ 
  & 85.1 & Deit3-B-384 & 67.3 & 72.8 ${\textcolor{red}{+5}}$ & 87.6 ${\textcolor{red}{+20}}$ & 64.6 ${\textcolor{green}{-3}}$ & 64.0 ${\textcolor{green}{-3}}$ & 59.4 ${\textcolor{green}{-8}}$ & 60.0 ${\textcolor{green}{-7}}$ & 90.2 ${\textcolor{red}{+23}}$ & 74.4 ${\textcolor{red}{+7}}$ & 67.1 ${\textcolor{green}{-0}}$ & \textbf{56.9} ${\textcolor{green}{-10}}$ \\ 
  & 83.8 & Deit3-B-224 & 70.3 & 71.9 ${\textcolor{red}{+2}}$ & 82.3 ${\textcolor{red}{+12}}$ & 68.4 ${\textcolor{green}{-2}}$ & 69.0 ${\textcolor{green}{-1}}$ & 64.3 ${\textcolor{green}{-6}}$ & 63.5 ${\textcolor{green}{-7}}$ & 83.1 ${\textcolor{red}{+13}}$ & 80.4 ${\textcolor{red}{+10}}$ & 73.0 ${\textcolor{red}{+3}}$ & \textbf{61.9} ${\textcolor{green}{-8}}$ \\ 
  & 82.6 & XCiT-M-224 & 72.7 & 73.3 ${\textcolor{red}{+1}}$ & 79.2 ${\textcolor{red}{+6}}$ & 71.8 ${\textcolor{green}{-1}}$ & 66.2 ${\textcolor{green}{-6}}$ & \textbf{63.5} ${\textcolor{green}{-9}}$ & 64.9 ${\textcolor{green}{-8}}$ & 76.4 ${\textcolor{red}{+4}}$ & 71.8 ${\textcolor{green}{-1}}$ & 67.1 ${\textcolor{green}{-6}}$ & 66.0 ${\textcolor{green}{-7}}$ \\ 
  & 84.3 & XCiT-M-224-d & 68.3 & 66.2 ${\textcolor{green}{-2}}$ & 73.1 ${\textcolor{red}{+5}}$ & 66.9 ${\textcolor{green}{-1}}$ & 66.4 ${\textcolor{green}{-2}}$ & \textbf{61.9} ${\textcolor{green}{-6}}$ & 62.3 ${\textcolor{green}{-6}}$ & 72.4 ${\textcolor{red}{+4}}$ & 70.4 ${\textcolor{red}{+2}}$ & 64.6 ${\textcolor{green}{-4}}$ & 62.6 ${\textcolor{green}{-6}}$ \\ 
  & 84.4 & CnvNxt-B & 64.7 & 71.5 ${\textcolor{red}{+7}}$ & 89.1 ${\textcolor{red}{+24}}$ & 68.0 ${\textcolor{red}{+3}}$ & 65.8 ${\textcolor{red}{+1}}$ & 60.6 ${\textcolor{green}{-4}}$ & 65.4 ${\textcolor{red}{+1}}$ & 85.9 ${\textcolor{red}{+21}}$ & 70.5 ${\textcolor{red}{+6}}$ & 61.3 ${\textcolor{green}{-3}}$ & \textbf{58.6} ${\textcolor{green}{-6}}$ \\ 
  & 78.0 & BiT-s & 78.8 & 81.2 ${\textcolor{red}{+2}}$ & 82.9 ${\textcolor{red}{+4}}$ & 68.4 ${\textcolor{green}{-10}}$ & 83.5 ${\textcolor{red}{+5}}$ & \textbf{64.1} ${\textcolor{green}{-15}}$ & 73.5 ${\textcolor{green}{-5}}$ & 77.8 ${\textcolor{green}{-1}}$ & 83.2 ${\textcolor{red}{+4}}$ & 72.1 ${\textcolor{green}{-7}}$ & 84.1 ${\textcolor{red}{+5}}$ \\ 
  & 85.1 & EffNetv2-M & 65.3 & 65.3 ${\textcolor{red}{+0}}$ & 74.5 ${\textcolor{red}{+9}}$ & 62.8 ${\textcolor{green}{-2}}$ & 62.5 ${\textcolor{green}{-3}}$ & 54.9 ${\textcolor{green}{-10}}$ & 72.5 ${\textcolor{red}{+7}}$ & 69.6 ${\textcolor{red}{+4}}$ & 64.4 ${\textcolor{green}{-1}}$ & 59.6 ${\textcolor{green}{-6}}$ & \textbf{54.4} ${\textcolor{green}{-11}}$ \\ 
  & 84.9 & EffNetb7 & 66.8 & 69.0 ${\textcolor{red}{+2}}$ & 81.5 ${\textcolor{red}{+15}}$ & 62.7 ${\textcolor{green}{-4}}$ & 68.1 ${\textcolor{red}{+1}}$ & 54.6 ${\textcolor{green}{-12}}$ & 72.7 ${\textcolor{red}{+6}}$ & 76.3 ${\textcolor{red}{+10}}$ & 66.8 ${\textcolor{red}{+0}}$ & 60.5 ${\textcolor{green}{-6}}$ & \textbf{53.7} ${\textcolor{green}{-13}}$ \\ 
  & 77.7 & EffNet-B0 & 72.0 & 72.4 ${\textcolor{red}{+0}}$ & 79.6 ${\textcolor{red}{+8}}$ & 72.3 ${\textcolor{red}{+0}}$ & 83.3 ${\textcolor{red}{+11}}$ & 74.0 ${\textcolor{red}{+2}}$ & 75.2 ${\textcolor{red}{+3}}$ & 75.1 ${\textcolor{red}{+3}}$ & 86.9 ${\textcolor{red}{+15}}$ & \textbf{61.3} ${\textcolor{green}{-11}}$ & 69.8 ${\textcolor{green}{-2}}$ \\ 
  & 80.4 & ResNet50 & 72.4 & 74.3 ${\textcolor{red}{+2}}$ & 77.9 ${\textcolor{red}{+6}}$ & 69.0 ${\textcolor{green}{-3}}$ & 85.9 ${\textcolor{red}{+13}}$ & 69.5 ${\textcolor{green}{-3}}$ & 78.6 ${\textcolor{red}{+6}}$ & 97.4 ${\textcolor{red}{+25}}$ & 77.9 ${\textcolor{red}{+6}}$ & 63.0 ${\textcolor{green}{-9}}$ & \textbf{62.1} ${\textcolor{green}{-10}}$ \\ 
 \hline 
{\multirow{1}{*}{{JFT}}} & 86.8 & EffNetb7-ns & 63.2 & \textbf{55.7} ${\textcolor{green}{-7}}$ & 61.5 ${\textcolor{green}{-2}}$ & 64.5 ${\textcolor{red}{+1}}$ & 87.4 ${\textcolor{red}{+24}}$ & 68.7 ${\textcolor{red}{+6}}$ & 89.2 ${\textcolor{red}{+26}}$ & 61.7 ${\textcolor{green}{-1}}$ & 73.8 ${\textcolor{red}{+11}}$ & 65.2 ${\textcolor{red}{+2}}$ & 63.7 ${\textcolor{red}{+1}}$ \\ 
 \hline 
{\multirow{2}{*}{{\shortstack[l]{clip\\+12k}}}} & 87.2 & ViT-B-384-l2b & 50.2 & 47.4 ${\textcolor{green}{-3}}$ & 50.3 ${\textcolor{red}{+0}}$ & 52.2 ${\textcolor{red}{+2}}$ & 52.6 ${\textcolor{red}{+2}}$ & 47.3 ${\textcolor{green}{-3}}$ & 45.8 ${\textcolor{green}{-4}}$ & 44.9 ${\textcolor{green}{-5}}$ & 45.4 ${\textcolor{green}{-5}}$ & \textbf{40.1} ${\textcolor{green}{-10}}$ & 40.2 ${\textcolor{green}{-10}}$ \\ 
  & 87.0 & ViT-B-384-oai & 48.8 & 43.7 ${\textcolor{green}{-5}}$ & 44.1 ${\textcolor{green}{-5}}$ & 49.6 ${\textcolor{red}{+1}}$ & 57.7 ${\textcolor{red}{+9}}$ & 48.4 ${\textcolor{green}{-0}}$ & 52.5 ${\textcolor{red}{+4}}$ & 42.2 ${\textcolor{green}{-7}}$ & 45.0 ${\textcolor{green}{-4}}$ & 39.3 ${\textcolor{green}{-10}}$ & \textbf{39.1} ${\textcolor{green}{-10}}$ \\ 
 \hline 
{\multirow{2}{*}{{clip}}} & 86.6 & ViT-B-384-l2b & 61.9 & 61.6 ${\textcolor{green}{-0}}$ & 65.8 ${\textcolor{red}{+4}}$ & 57.5 ${\textcolor{green}{-4}}$ & 52.7 ${\textcolor{green}{-9}}$ & 50.5 ${\textcolor{green}{-11}}$ & 51.7 ${\textcolor{green}{-10}}$ & 63.2 ${\textcolor{red}{+1}}$ & 57.0 ${\textcolor{green}{-5}}$ & 50.8 ${\textcolor{green}{-11}}$ & \textbf{49.1} ${\textcolor{green}{-13}}$ \\ 
  & 86.2 & ViT-B-384-oai & 64.9 & 64.9 ${\textcolor{red}{+0}}$ & 69.7 ${\textcolor{red}{+5}}$ & 61.8 ${\textcolor{green}{-3}}$ & 55.7 ${\textcolor{green}{-9}}$ & \textbf{53.7} ${\textcolor{green}{-11}}$ & 56.9 ${\textcolor{green}{-8}}$ & 67.3 ${\textcolor{red}{+2}}$ & 61.4 ${\textcolor{green}{-4}}$ & 56.6 ${\textcolor{green}{-8}}$ & 54.3 ${\textcolor{green}{-11}}$ \\ 
 \hline 
{\multirow{2}{*}{\shortstack[l]{clip\\z. shot}}} & 74.3 & clip-ViT-L-336 & ---- & ---- & ---- & ---- & ---- & ---- & ---- & ---- & ---- & 72.5 & \textbf{67.1} \\ 
  & 66.6 & clip-ViT-B-224 & ---- & ---- & ---- & ---- & ---- & ---- & ---- & ---- & ---- & \textbf{79.1} & 79.8 \\ 
 \hline 
\end{tabular}
\end{center}

\end{smaller}
\vskip \belowtablevskip
\end{table*}

\renewcommand{\arraystretch}{1.}
\tabcolsep=0.1cm
\begin{table*}[htb]
    \centering
    \caption{\textbf{Mean AUROC on our \dsetname{} dataset.} Higher is better. The difference to MSP is shown in red if a method performs worse, and in green if it improves. Bold values mark the best-performing method per model.}
    \label{tab:overview-auroc-mean}
    \small
\begin{smaller}
\begin{center}
\begin{tabular}{l l l l l l l l l l l l l l }
pre & acc. & model & MSP & MaxL & Ener & KL-M & Maha & RMaha & ViM & E+R & KNN & Cos & MCM/RCos \\ 
 \hline 
{\multirow{9}{*}{{21k}}} & 86.0 & ViT-B-384 & 87.2 & 92.5 ${\textcolor{green}{+5}}$ & 92.7 ${\textcolor{green}{+5}}$ & 86.9 ${\textcolor{red}{-0}}$ & \textbf{95.0} ${\textcolor{green}{+8}}$ & 94.0 ${\textcolor{green}{+7}}$ & 94.0 ${\textcolor{green}{+7}}$ & 92.5 ${\textcolor{green}{+5}}$ & 85.9 ${\textcolor{red}{-1}}$ & 91.5 ${\textcolor{green}{+4}}$ & 91.7 ${\textcolor{green}{+4}}$ \\ 
  & 84.5 & ViT-B-224 & 85.5 & 90.6 ${\textcolor{green}{+5}}$ & 90.8 ${\textcolor{green}{+5}}$ & 85.2 ${\textcolor{red}{-0}}$ & \textbf{94.0} ${\textcolor{green}{+8}}$ & 92.8 ${\textcolor{green}{+7}}$ & 92.5 ${\textcolor{green}{+7}}$ & 90.1 ${\textcolor{green}{+5}}$ & 82.6 ${\textcolor{red}{-3}}$ & 89.5 ${\textcolor{green}{+4}}$ & 89.4 ${\textcolor{green}{+4}}$ \\ 
  & 86.3 & Swinv2-B-256 & 86.3 & 87.0 ${\textcolor{green}{+1}}$ & 85.8 ${\textcolor{red}{-0}}$ & 86.1 ${\textcolor{red}{-0}}$ & 88.0 ${\textcolor{green}{+2}}$ & 89.0 ${\textcolor{green}{+3}}$ & \textbf{89.9} ${\textcolor{green}{+4}}$ & 88.8 ${\textcolor{green}{+3}}$ & 84.3 ${\textcolor{red}{-2}}$ & 89.1 ${\textcolor{green}{+3}}$ & 89.8 ${\textcolor{green}{+4}}$ \\ 
  & 86.7 & Deit3-B-384 & 81.1 & 77.7 ${\textcolor{red}{-3}}$ & 74.9 ${\textcolor{red}{-6}}$ & 83.6 ${\textcolor{green}{+2}}$ & 89.7 ${\textcolor{green}{+9}}$ & 90.0 ${\textcolor{green}{+9}}$ & 89.0 ${\textcolor{green}{+8}}$ & 80.7 ${\textcolor{red}{-0}}$ & 87.0 ${\textcolor{green}{+6}}$ & 90.1 ${\textcolor{green}{+9}}$ & \textbf{90.4} ${\textcolor{green}{+9}}$ \\ 
  & 85.7 & Deit3-B-224 & 80.3 & 77.3 ${\textcolor{red}{-3}}$ & 74.8 ${\textcolor{red}{-6}}$ & 82.4 ${\textcolor{green}{+2}}$ & 88.3 ${\textcolor{green}{+8}}$ & 88.8 ${\textcolor{green}{+9}}$ & 87.6 ${\textcolor{green}{+7}}$ & 79.5 ${\textcolor{red}{-1}}$ & 85.2 ${\textcolor{green}{+5}}$ & 88.8 ${\textcolor{green}{+8}}$ & \textbf{89.0} ${\textcolor{green}{+9}}$ \\ 
  & 86.3 & CnvNxt-B & 87.9 & 87.6 ${\textcolor{red}{-0}}$ & 85.8 ${\textcolor{red}{-2}}$ & 88.0 ${\textcolor{green}{+0}}$ & 90.9 ${\textcolor{green}{+3}}$ & 91.8 ${\textcolor{green}{+4}}$ & \textbf{92.5} ${\textcolor{green}{+5}}$ & 87.9 ${\textcolor{red}{-0}}$ & 87.5 ${\textcolor{red}{-0}}$ & 91.5 ${\textcolor{green}{+4}}$ & 91.8 ${\textcolor{green}{+4}}$ \\ 
  & 84.1 & CnvNxt-T & 85.1 & 86.0 ${\textcolor{green}{+1}}$ & 85.3 ${\textcolor{green}{+0}}$ & 85.2 ${\textcolor{green}{+0}}$ & 89.8 ${\textcolor{green}{+5}}$ & 89.5 ${\textcolor{green}{+4}}$ & \textbf{92.5} ${\textcolor{green}{+7}}$ & 86.0 ${\textcolor{green}{+1}}$ & 86.0 ${\textcolor{green}{+1}}$ & 89.1 ${\textcolor{green}{+4}}$ & 89.6 ${\textcolor{green}{+5}}$ \\ 
  & 82.3 & BiT-m & 82.2 & 83.3 ${\textcolor{green}{+1}}$ & 82.5 ${\textcolor{green}{+0}}$ & 82.8 ${\textcolor{green}{+1}}$ & 90.6 ${\textcolor{green}{+8}}$ & 90.0 ${\textcolor{green}{+8}}$ & \textbf{92.2} ${\textcolor{green}{+10}}$ & 85.8 ${\textcolor{green}{+4}}$ & 86.4 ${\textcolor{green}{+4}}$ & 89.6 ${\textcolor{green}{+7}}$ & 88.5 ${\textcolor{green}{+6}}$ \\ 
  & 85.6 & EffNetv2-M & 86.3 & 85.1 ${\textcolor{red}{-1}}$ & 82.7 ${\textcolor{red}{-4}}$ & 87.3 ${\textcolor{green}{+1}}$ & 87.6 ${\textcolor{green}{+1}}$ & 88.9 ${\textcolor{green}{+3}}$ & 87.9 ${\textcolor{green}{+2}}$ & 73.0 ${\textcolor{red}{-13}}$ & 83.8 ${\textcolor{red}{-3}}$ & \textbf{90.1} ${\textcolor{green}{+4}}$ & 88.8 ${\textcolor{green}{+3}}$ \\ 
 \hline 
{\multirow{12}{*}{{none}}} & 81.1 & ViT-B-384 & 81.4 & 84.2 ${\textcolor{green}{+3}}$ & 84.2 ${\textcolor{green}{+3}}$ & 80.7 ${\textcolor{red}{-1}}$ & 86.5 ${\textcolor{green}{+5}}$ & \textbf{87.3} ${\textcolor{green}{+6}}$ & 82.6 ${\textcolor{green}{+1}}$ & 84.6 ${\textcolor{green}{+3}}$ & 79.7 ${\textcolor{red}{-2}}$ & 84.4 ${\textcolor{green}{+3}}$ & 84.1 ${\textcolor{green}{+3}}$ \\ 
  & 84.6 & Swinv2-B-256 & 80.4 & 77.8 ${\textcolor{red}{-3}}$ & 75.0 ${\textcolor{red}{-5}}$ & 81.9 ${\textcolor{green}{+1}}$ & 86.2 ${\textcolor{green}{+6}}$ & \textbf{86.7} ${\textcolor{green}{+6}}$ & 81.1 ${\textcolor{green}{+1}}$ & 80.2 ${\textcolor{red}{-0}}$ & 81.9 ${\textcolor{green}{+1}}$ & 85.8 ${\textcolor{green}{+5}}$ & 86.3 ${\textcolor{green}{+6}}$ \\ 
  & 85.1 & Deit3-B-384 & 81.7 & 76.4 ${\textcolor{red}{-5}}$ & 66.6 ${\textcolor{red}{-15}}$ & 83.5 ${\textcolor{green}{+2}}$ & 86.9 ${\textcolor{green}{+5}}$ & \textbf{88.0} ${\textcolor{green}{+6}}$ & 84.8 ${\textcolor{green}{+3}}$ & 61.8 ${\textcolor{red}{-20}}$ & 80.4 ${\textcolor{red}{-1}}$ & 85.5 ${\textcolor{green}{+4}}$ & 87.0 ${\textcolor{green}{+5}}$ \\ 
  & 83.8 & Deit3-B-224 & 81.0 & 78.8 ${\textcolor{red}{-2}}$ & 74.8 ${\textcolor{red}{-6}}$ & 82.3 ${\textcolor{green}{+1}}$ & 85.5 ${\textcolor{green}{+5}}$ & \textbf{86.9} ${\textcolor{green}{+6}}$ & 84.0 ${\textcolor{green}{+3}}$ & 74.6 ${\textcolor{red}{-6}}$ & 77.6 ${\textcolor{red}{-3}}$ & 83.8 ${\textcolor{green}{+3}}$ & 85.6 ${\textcolor{green}{+5}}$ \\ 
  & 82.6 & XCiT-M-224 & 77.9 & 72.2 ${\textcolor{red}{-6}}$ & 64.5 ${\textcolor{red}{-13}}$ & 81.2 ${\textcolor{green}{+3}}$ & 85.1 ${\textcolor{green}{+7}}$ & 85.7 ${\textcolor{green}{+8}}$ & \textbf{86.0} ${\textcolor{green}{+8}}$ & 73.3 ${\textcolor{red}{-5}}$ & 80.9 ${\textcolor{green}{+3}}$ & 84.7 ${\textcolor{green}{+7}}$ & 85.0 ${\textcolor{green}{+7}}$ \\ 
  & 84.3 & XCiT-M-224-d & 82.7 & 80.2 ${\textcolor{red}{-3}}$ & 74.1 ${\textcolor{red}{-9}}$ & 82.9 ${\textcolor{green}{+0}}$ & 85.5 ${\textcolor{green}{+3}}$ & \textbf{86.8} ${\textcolor{green}{+4}}$ & 85.3 ${\textcolor{green}{+3}}$ & 78.6 ${\textcolor{red}{-4}}$ & 81.4 ${\textcolor{red}{-1}}$ & 85.8 ${\textcolor{green}{+3}}$ & 86.1 ${\textcolor{green}{+3}}$ \\ 
  & 84.4 & CnvNxt-B & 81.1 & 76.2 ${\textcolor{red}{-5}}$ & 64.6 ${\textcolor{red}{-16}}$ & 83.4 ${\textcolor{green}{+2}}$ & 85.2 ${\textcolor{green}{+4}}$ & 86.6 ${\textcolor{green}{+6}}$ & 82.5 ${\textcolor{green}{+1}}$ & 72.9 ${\textcolor{red}{-8}}$ & 81.3 ${\textcolor{green}{+0}}$ & 85.8 ${\textcolor{green}{+5}}$ & \textbf{86.9} ${\textcolor{green}{+6}}$ \\ 
  & 78.0 & BiT-s & 80.1 & 77.3 ${\textcolor{red}{-3}}$ & 75.6 ${\textcolor{red}{-5}}$ & 82.3 ${\textcolor{green}{+2}}$ & 71.2 ${\textcolor{red}{-9}}$ & \textbf{84.9} ${\textcolor{green}{+5}}$ & 77.9 ${\textcolor{red}{-2}}$ & 76.2 ${\textcolor{red}{-4}}$ & 68.8 ${\textcolor{red}{-11}}$ & 78.3 ${\textcolor{red}{-2}}$ & 69.8 ${\textcolor{red}{-10}}$ \\ 
  & 85.1 & EffNetv2-M & 81.8 & 78.3 ${\textcolor{red}{-3}}$ & 71.8 ${\textcolor{red}{-10}}$ & 84.0 ${\textcolor{green}{+2}}$ & 86.5 ${\textcolor{green}{+5}}$ & \textbf{88.9} ${\textcolor{green}{+7}}$ & 80.1 ${\textcolor{red}{-2}}$ & 79.0 ${\textcolor{red}{-3}}$ & 83.4 ${\textcolor{green}{+2}}$ & 87.3 ${\textcolor{green}{+6}}$ & 88.1 ${\textcolor{green}{+6}}$ \\ 
  & 84.9 & EffNetb7 & 79.6 & 72.8 ${\textcolor{red}{-7}}$ & 64.6 ${\textcolor{red}{-15}}$ & 84.2 ${\textcolor{green}{+5}}$ & 84.5 ${\textcolor{green}{+5}}$ & \textbf{88.6} ${\textcolor{green}{+9}}$ & 81.6 ${\textcolor{green}{+2}}$ & 71.8 ${\textcolor{red}{-8}}$ & 82.5 ${\textcolor{green}{+3}}$ & 86.9 ${\textcolor{green}{+7}}$ & 87.9 ${\textcolor{green}{+8}}$ \\ 
  & 77.7 & EffNet-B0 & 80.8 & 78.5 ${\textcolor{red}{-2}}$ & 74.9 ${\textcolor{red}{-6}}$ & 81.9 ${\textcolor{green}{+1}}$ & 76.7 ${\textcolor{red}{-4}}$ & 82.7 ${\textcolor{green}{+2}}$ & 81.6 ${\textcolor{green}{+1}}$ & 79.1 ${\textcolor{red}{-2}}$ & 76.2 ${\textcolor{red}{-5}}$ & \textbf{85.0} ${\textcolor{green}{+4}}$ & 82.3 ${\textcolor{green}{+1}}$ \\ 
  & 80.4 & ResNet50 & 81.5 & 81.5 ${\textcolor{green}{+0}}$ & 81.2 ${\textcolor{red}{-0}}$ & 79.3 ${\textcolor{red}{-2}}$ & 75.8 ${\textcolor{red}{-6}}$ & 85.0 ${\textcolor{green}{+4}}$ & 81.3 ${\textcolor{red}{-0}}$ & 64.6 ${\textcolor{red}{-17}}$ & 76.3 ${\textcolor{red}{-5}}$ & 84.9 ${\textcolor{green}{+3}}$ & \textbf{85.5} ${\textcolor{green}{+4}}$ \\ 
 \hline 
{\multirow{1}{*}{{JFT}}} & 86.8 & EffNetb7-ns & 83.6 & 82.5 ${\textcolor{red}{-1}}$ & 78.6 ${\textcolor{red}{-5}}$ & 83.1 ${\textcolor{red}{-0}}$ & 78.1 ${\textcolor{red}{-5}}$ & \textbf{86.6} ${\textcolor{green}{+3}}$ & 74.6 ${\textcolor{red}{-9}}$ & 81.1 ${\textcolor{red}{-2}}$ & 79.2 ${\textcolor{red}{-4}}$ & 85.2 ${\textcolor{green}{+2}}$ & 85.0 ${\textcolor{green}{+1}}$ \\ 
 \hline 
{\multirow{2}{*}{{\shortstack[l]{clip\\+12k}}}} & 87.2 & ViT-B-384-l2b & 86.1 & 82.6 ${\textcolor{red}{-4}}$ & 78.1 ${\textcolor{red}{-8}}$ & 88.8 ${\textcolor{green}{+3}}$ & 90.5 ${\textcolor{green}{+4}}$ & 91.1 ${\textcolor{green}{+5}}$ & 91.9 ${\textcolor{green}{+6}}$ & 83.4 ${\textcolor{red}{-3}}$ & 89.5 ${\textcolor{green}{+3}}$ & \textbf{92.2} ${\textcolor{green}{+6}}$ & 92.1 ${\textcolor{green}{+6}}$ \\ 
  & 87.0 & ViT-B-384-oai & 87.2 & 85.8 ${\textcolor{red}{-1}}$ & 84.2 ${\textcolor{red}{-3}}$ & 88.4 ${\textcolor{green}{+1}}$ & 89.6 ${\textcolor{green}{+2}}$ & 91.1 ${\textcolor{green}{+4}}$ & 90.8 ${\textcolor{green}{+4}}$ & 86.5 ${\textcolor{red}{-1}}$ & 89.9 ${\textcolor{green}{+3}}$ & 92.5 ${\textcolor{green}{+5}}$ & \textbf{92.5} ${\textcolor{green}{+5}}$ \\ 
 \hline 
{\multirow{2}{*}{{clip}}} & 86.6 & ViT-B-384-l2b & 81.1 & 73.5 ${\textcolor{red}{-8}}$ & 68.5 ${\textcolor{red}{-13}}$ & 85.9 ${\textcolor{green}{+5}}$ & 89.1 ${\textcolor{green}{+8}}$ & 89.1 ${\textcolor{green}{+8}}$ & 88.8 ${\textcolor{green}{+8}}$ & 71.7 ${\textcolor{red}{-9}}$ & 86.1 ${\textcolor{green}{+5}}$ & 89.8 ${\textcolor{green}{+9}}$ & \textbf{90.0} ${\textcolor{green}{+9}}$ \\ 
  & 86.2 & ViT-B-384-oai & 78.8 & 70.4 ${\textcolor{red}{-8}}$ & 65.0 ${\textcolor{red}{-14}}$ & 84.4 ${\textcolor{green}{+6}}$ & 88.6 ${\textcolor{green}{+10}}$ & 88.5 ${\textcolor{green}{+10}}$ & 88.3 ${\textcolor{green}{+10}}$ & 68.1 ${\textcolor{red}{-11}}$ & 84.6 ${\textcolor{green}{+6}}$ & 88.6 ${\textcolor{green}{+10}}$ & \textbf{89.2} ${\textcolor{green}{+10}}$ \\ 
 \hline 
{\multirow{2}{*}{\shortstack[l]{clip\\z. shot}}} & 74.3 & clip-ViT-L-336 & ---- & ---- & ---- & ---- & ---- & ---- & ---- & ---- & ---- & 79.7 & \textbf{81.1} \\ 
  & 66.6 & clip-ViT-B-224 & ---- & ---- & ---- & ---- & ---- & ---- & ---- & ---- & ---- & 74.0 & \textbf{74.9} \\ 
 \hline 
\end{tabular}
\end{center}

\end{smaller}
\vskip \belowtablevskip
\end{table*}

\renewcommand{\arraystretch}{1.}
\tabcolsep=0.1cm
\begin{table*}[htb]
    \centering
    \caption{\textbf{Mean AUPR-S on our \dsetname{} dataset.} Higher is better. The difference to MSP is shown in red if a method performs worse, and in green if it improves. Bold values mark the best-performing method per model.}
    \label{tab:overview-auprc-mean}
    \small
\begin{smaller}
\begin{center}
\begin{tabular}{l l l l l l l l l l l l l l }
pre & acc. & model & MSP & MaxL & Ener & KL-M & Maha & RMaha & ViM & E+R & KNN & Cos & MCM/RCos \\ 
 \hline 
{\multirow{9}{*}{{21k}}} & 86.0 & ViT-B-384 & 97.2 & 98.4 ${\textcolor{green}{+1}}$ & 98.4 ${\textcolor{green}{+1}}$ & 96.6 ${\textcolor{red}{-1}}$ & \textbf{99.0} ${\textcolor{green}{+2}}$ & 98.8 ${\textcolor{green}{+2}}$ & 98.7 ${\textcolor{green}{+2}}$ & 98.4 ${\textcolor{green}{+1}}$ & 97.0 ${\textcolor{red}{-0}}$ & 98.2 ${\textcolor{green}{+1}}$ & 98.3 ${\textcolor{green}{+1}}$ \\ 
  & 84.5 & ViT-B-224 & 96.7 & 97.9 ${\textcolor{green}{+1}}$ & 98.0 ${\textcolor{green}{+1}}$ & 96.1 ${\textcolor{red}{-1}}$ & \textbf{98.7} ${\textcolor{green}{+2}}$ & 98.5 ${\textcolor{green}{+2}}$ & 98.4 ${\textcolor{green}{+2}}$ & 97.8 ${\textcolor{green}{+1}}$ & 96.1 ${\textcolor{red}{-1}}$ & 97.8 ${\textcolor{green}{+1}}$ & 97.8 ${\textcolor{green}{+1}}$ \\ 
  & 86.3 & Swinv2-B-256 & 96.2 & 95.8 ${\textcolor{red}{-0}}$ & 95.2 ${\textcolor{red}{-1}}$ & 96.6 ${\textcolor{green}{+0}}$ & 97.4 ${\textcolor{green}{+1}}$ & 97.5 ${\textcolor{green}{+1}}$ & 97.9 ${\textcolor{green}{+2}}$ & 96.6 ${\textcolor{green}{+0}}$ & 96.5 ${\textcolor{green}{+0}}$ & 97.7 ${\textcolor{green}{+2}}$ & \textbf{97.9} ${\textcolor{green}{+2}}$ \\ 
  & 86.7 & Deit3-B-384 & 94.1 & 91.7 ${\textcolor{red}{-2}}$ & 90.4 ${\textcolor{red}{-4}}$ & 95.9 ${\textcolor{green}{+2}}$ & 97.9 ${\textcolor{green}{+4}}$ & 97.9 ${\textcolor{green}{+4}}$ & 97.6 ${\textcolor{green}{+4}}$ & 93.1 ${\textcolor{red}{-1}}$ & 97.1 ${\textcolor{green}{+3}}$ & 97.9 ${\textcolor{green}{+4}}$ & \textbf{98.0} ${\textcolor{green}{+4}}$ \\ 
  & 85.7 & Deit3-B-224 & 94.1 & 91.7 ${\textcolor{red}{-2}}$ & 90.6 ${\textcolor{red}{-3}}$ & 95.5 ${\textcolor{green}{+1}}$ & 97.5 ${\textcolor{green}{+3}}$ & 97.6 ${\textcolor{green}{+3}}$ & 97.3 ${\textcolor{green}{+3}}$ & 92.7 ${\textcolor{red}{-1}}$ & 96.6 ${\textcolor{green}{+3}}$ & 97.6 ${\textcolor{green}{+4}}$ & \textbf{97.6} ${\textcolor{green}{+4}}$ \\ 
  & 86.3 & CnvNxt-B & 96.8 & 96.3 ${\textcolor{red}{-1}}$ & 95.7 ${\textcolor{red}{-1}}$ & 97.2 ${\textcolor{green}{+0}}$ & 98.1 ${\textcolor{green}{+1}}$ & 98.2 ${\textcolor{green}{+1}}$ & \textbf{98.4} ${\textcolor{green}{+2}}$ & 96.4 ${\textcolor{red}{-0}}$ & 97.2 ${\textcolor{green}{+0}}$ & 98.2 ${\textcolor{green}{+1}}$ & 98.3 ${\textcolor{green}{+1}}$ \\ 
  & 84.1 & CnvNxt-T & 96.0 & 95.9 ${\textcolor{red}{-0}}$ & 95.6 ${\textcolor{red}{-0}}$ & 96.4 ${\textcolor{green}{+0}}$ & 97.7 ${\textcolor{green}{+2}}$ & 97.6 ${\textcolor{green}{+2}}$ & \textbf{98.4} ${\textcolor{green}{+2}}$ & 95.8 ${\textcolor{red}{-0}}$ & 96.9 ${\textcolor{green}{+1}}$ & 97.6 ${\textcolor{green}{+2}}$ & 97.8 ${\textcolor{green}{+2}}$ \\ 
  & 82.3 & BiT-m & 95.7 & 95.7 ${\textcolor{red}{-0}}$ & 95.5 ${\textcolor{red}{-0}}$ & 95.3 ${\textcolor{red}{-0}}$ & 98.0 ${\textcolor{green}{+2}}$ & 97.7 ${\textcolor{green}{+2}}$ & \textbf{98.3} ${\textcolor{green}{+3}}$ & 96.6 ${\textcolor{green}{+1}}$ & 97.0 ${\textcolor{green}{+1}}$ & 97.7 ${\textcolor{green}{+2}}$ & 97.5 ${\textcolor{green}{+2}}$ \\ 
  & 85.6 & EffNetv2-M & 96.3 & 95.7 ${\textcolor{red}{-1}}$ & 95.0 ${\textcolor{red}{-1}}$ & 97.0 ${\textcolor{green}{+1}}$ & 97.3 ${\textcolor{green}{+1}}$ & 97.5 ${\textcolor{green}{+1}}$ & 97.2 ${\textcolor{green}{+1}}$ & 93.6 ${\textcolor{red}{-3}}$ & 96.5 ${\textcolor{green}{+0}}$ & \textbf{97.8} ${\textcolor{green}{+1}}$ & 97.5 ${\textcolor{green}{+1}}$ \\ 
 \hline 
{\multirow{12}{*}{{none}}} & 81.1 & ViT-B-384 & 95.5 & 96.1 ${\textcolor{green}{+1}}$ & 96.2 ${\textcolor{green}{+1}}$ & 94.6 ${\textcolor{red}{-1}}$ & 96.9 ${\textcolor{green}{+1}}$ & \textbf{97.1} ${\textcolor{green}{+2}}$ & 95.7 ${\textcolor{green}{+0}}$ & 96.1 ${\textcolor{green}{+1}}$ & 95.2 ${\textcolor{red}{-0}}$ & 96.4 ${\textcolor{green}{+1}}$ & 96.3 ${\textcolor{green}{+1}}$ \\ 
  & 84.6 & Swinv2-B-256 & 94.5 & 92.6 ${\textcolor{red}{-2}}$ & 91.4 ${\textcolor{red}{-3}}$ & 95.2 ${\textcolor{green}{+1}}$ & 96.9 ${\textcolor{green}{+2}}$ & \textbf{97.0} ${\textcolor{green}{+2}}$ & 94.9 ${\textcolor{green}{+0}}$ & 94.0 ${\textcolor{red}{-1}}$ & 95.8 ${\textcolor{green}{+1}}$ & 96.9 ${\textcolor{green}{+2}}$ & 96.9 ${\textcolor{green}{+2}}$ \\ 
  & 85.1 & Deit3-B-384 & 95.2 & 92.9 ${\textcolor{red}{-2}}$ & 89.6 ${\textcolor{red}{-6}}$ & 95.7 ${\textcolor{green}{+1}}$ & 97.1 ${\textcolor{green}{+2}}$ & \textbf{97.4} ${\textcolor{green}{+2}}$ & 96.1 ${\textcolor{green}{+1}}$ & 88.2 ${\textcolor{red}{-7}}$ & 95.4 ${\textcolor{green}{+0}}$ & 96.8 ${\textcolor{green}{+2}}$ & 96.8 ${\textcolor{green}{+2}}$ \\ 
  & 83.8 & Deit3-B-224 & 95.1 & 94.1 ${\textcolor{red}{-1}}$ & 93.0 ${\textcolor{red}{-2}}$ & 95.5 ${\textcolor{green}{+0}}$ & 96.8 ${\textcolor{green}{+2}}$ & \textbf{97.1} ${\textcolor{green}{+2}}$ & 95.9 ${\textcolor{green}{+1}}$ & 93.2 ${\textcolor{red}{-2}}$ & 94.8 ${\textcolor{red}{-0}}$ & 96.3 ${\textcolor{green}{+1}}$ & 96.6 ${\textcolor{green}{+2}}$ \\ 
  & 82.6 & XCiT-M-224 & 93.7 & 90.7 ${\textcolor{red}{-3}}$ & 87.6 ${\textcolor{red}{-6}}$ & 95.3 ${\textcolor{green}{+2}}$ & 96.5 ${\textcolor{green}{+3}}$ & \textbf{96.8} ${\textcolor{green}{+3}}$ & 96.7 ${\textcolor{green}{+3}}$ & 91.6 ${\textcolor{red}{-2}}$ & 95.4 ${\textcolor{green}{+2}}$ & 96.5 ${\textcolor{green}{+3}}$ & 96.5 ${\textcolor{green}{+3}}$ \\ 
  & 84.3 & XCiT-M-224-d & 95.8 & 94.1 ${\textcolor{red}{-2}}$ & 92.1 ${\textcolor{red}{-4}}$ & 95.6 ${\textcolor{red}{-0}}$ & 96.7 ${\textcolor{green}{+1}}$ & \textbf{97.0} ${\textcolor{green}{+1}}$ & 96.5 ${\textcolor{green}{+1}}$ & 93.9 ${\textcolor{red}{-2}}$ & 95.6 ${\textcolor{red}{-0}}$ & 96.8 ${\textcolor{green}{+1}}$ & 96.9 ${\textcolor{green}{+1}}$ \\ 
  & 84.4 & CnvNxt-B & 94.9 & 93.0 ${\textcolor{red}{-2}}$ & 89.8 ${\textcolor{red}{-5}}$ & 95.8 ${\textcolor{green}{+1}}$ & 96.7 ${\textcolor{green}{+2}}$ & 96.9 ${\textcolor{green}{+2}}$ & 95.6 ${\textcolor{green}{+1}}$ & 92.8 ${\textcolor{red}{-2}}$ & 95.5 ${\textcolor{green}{+1}}$ & 96.8 ${\textcolor{green}{+2}}$ & \textbf{97.0} ${\textcolor{green}{+2}}$ \\ 
  & 78.0 & BiT-s & 95.3 & 94.7 ${\textcolor{red}{-1}}$ & 94.3 ${\textcolor{red}{-1}}$ & 95.3 ${\textcolor{green}{+0}}$ & 92.7 ${\textcolor{red}{-3}}$ & \textbf{96.5} ${\textcolor{green}{+1}}$ & 94.8 ${\textcolor{red}{-0}}$ & 94.4 ${\textcolor{red}{-1}}$ & 92.0 ${\textcolor{red}{-3}}$ & 94.7 ${\textcolor{red}{-1}}$ & 92.1 ${\textcolor{red}{-3}}$ \\ 
  & 85.1 & EffNetv2-M & 95.1 & 92.9 ${\textcolor{red}{-2}}$ & 90.4 ${\textcolor{red}{-5}}$ & 96.0 ${\textcolor{green}{+1}}$ & 97.0 ${\textcolor{green}{+2}}$ & \textbf{97.6} ${\textcolor{green}{+3}}$ & 94.9 ${\textcolor{red}{-0}}$ & 93.8 ${\textcolor{red}{-1}}$ & 96.2 ${\textcolor{green}{+1}}$ & 97.2 ${\textcolor{green}{+2}}$ & 97.1 ${\textcolor{green}{+2}}$ \\ 
  & 84.9 & EffNetb7 & 94.0 & 90.7 ${\textcolor{red}{-3}}$ & 87.9 ${\textcolor{red}{-6}}$ & 96.2 ${\textcolor{green}{+2}}$ & 96.5 ${\textcolor{green}{+3}}$ & \textbf{97.5} ${\textcolor{green}{+3}}$ & 95.6 ${\textcolor{green}{+2}}$ & 90.9 ${\textcolor{red}{-3}}$ & 95.9 ${\textcolor{green}{+2}}$ & 97.0 ${\textcolor{green}{+3}}$ & 97.3 ${\textcolor{green}{+3}}$ \\ 
  & 77.7 & EffNet-B0 & 95.1 & 94.1 ${\textcolor{red}{-1}}$ & 93.1 ${\textcolor{red}{-2}}$ & 95.4 ${\textcolor{green}{+0}}$ & 94.6 ${\textcolor{red}{-0}}$ & 96.1 ${\textcolor{green}{+1}}$ & 95.9 ${\textcolor{green}{+1}}$ & 94.6 ${\textcolor{red}{-1}}$ & 94.6 ${\textcolor{red}{-1}}$ & \textbf{96.5} ${\textcolor{green}{+1}}$ & 95.8 ${\textcolor{green}{+1}}$ \\ 
  & 80.4 & ResNet50 & 95.5 & 95.6 ${\textcolor{green}{+0}}$ & 95.5 ${\textcolor{red}{-0}}$ & 94.2 ${\textcolor{red}{-1}}$ & 94.3 ${\textcolor{red}{-1}}$ & 96.7 ${\textcolor{green}{+1}}$ & 95.6 ${\textcolor{green}{+0}}$ & 91.7 ${\textcolor{red}{-4}}$ & 94.1 ${\textcolor{red}{-1}}$ & 96.5 ${\textcolor{green}{+1}}$ & \textbf{96.7} ${\textcolor{green}{+1}}$ \\ 
 \hline 
{\multirow{1}{*}{{JFT}}} & 86.8 & EffNetb7-ns & 95.8 & 94.7 ${\textcolor{red}{-1}}$ & 92.9 ${\textcolor{red}{-3}}$ & 95.8 ${\textcolor{green}{+0}}$ & 95.0 ${\textcolor{red}{-1}}$ & \textbf{97.2} ${\textcolor{green}{+1}}$ & 94.1 ${\textcolor{red}{-2}}$ & 94.4 ${\textcolor{red}{-1}}$ & 95.3 ${\textcolor{red}{-1}}$ & 96.8 ${\textcolor{green}{+1}}$ & 96.6 ${\textcolor{green}{+1}}$ \\ 
 \hline 
{\multirow{2}{*}{{\shortstack[l]{clip\\+12k}}}} & 87.2 & ViT-B-384-l2b & 95.9 & 94.0 ${\textcolor{red}{-2}}$ & 92.6 ${\textcolor{red}{-3}}$ & 97.5 ${\textcolor{green}{+2}}$ & 98.0 ${\textcolor{green}{+2}}$ & 98.1 ${\textcolor{green}{+2}}$ & 98.3 ${\textcolor{green}{+2}}$ & 94.5 ${\textcolor{red}{-1}}$ & 97.7 ${\textcolor{green}{+2}}$ & 98.4 ${\textcolor{green}{+2}}$ & \textbf{98.4} ${\textcolor{green}{+3}}$ \\ 
  & 87.0 & ViT-B-384-oai & 96.6 & 95.5 ${\textcolor{red}{-1}}$ & 94.9 ${\textcolor{red}{-2}}$ & 97.3 ${\textcolor{green}{+1}}$ & 97.8 ${\textcolor{green}{+1}}$ & 98.1 ${\textcolor{green}{+2}}$ & 98.1 ${\textcolor{green}{+2}}$ & 95.9 ${\textcolor{red}{-1}}$ & 97.8 ${\textcolor{green}{+1}}$ & \textbf{98.5} ${\textcolor{green}{+2}}$ & 98.4 ${\textcolor{green}{+2}}$ \\ 
 \hline 
{\multirow{2}{*}{{clip}}} & 86.6 & ViT-B-384-l2b & 94.2 & 90.8 ${\textcolor{red}{-3}}$ & 89.2 ${\textcolor{red}{-5}}$ & 96.6 ${\textcolor{green}{+2}}$ & 97.6 ${\textcolor{green}{+3}}$ & 97.5 ${\textcolor{green}{+3}}$ & 97.4 ${\textcolor{green}{+3}}$ & 90.2 ${\textcolor{red}{-4}}$ & 96.8 ${\textcolor{green}{+3}}$ & 97.8 ${\textcolor{green}{+4}}$ & \textbf{97.8} ${\textcolor{green}{+4}}$ \\ 
  & 86.2 & ViT-B-384-oai & 93.1 & 89.6 ${\textcolor{red}{-3}}$ & 88.0 ${\textcolor{red}{-5}}$ & 96.1 ${\textcolor{green}{+3}}$ & 97.5 ${\textcolor{green}{+4}}$ & 97.5 ${\textcolor{green}{+4}}$ & 97.4 ${\textcolor{green}{+4}}$ & 89.2 ${\textcolor{red}{-4}}$ & 96.4 ${\textcolor{green}{+3}}$ & 97.5 ${\textcolor{green}{+4}}$ & \textbf{97.7} ${\textcolor{green}{+5}}$ \\ 
 \hline 
{\multirow{2}{*}{\shortstack[l]{clip\\z. shot}}} & 74.3 & clip-ViT-L-336 & ---- & ---- & ---- & ---- & ---- & ---- & ---- & ---- & ---- & 95.2 & \textbf{95.5} \\ 
  & 66.6 & clip-ViT-B-224 & ---- & ---- & ---- & ---- & ---- & ---- & ---- & ---- & ---- & 93.6 & \textbf{93.9} \\ 
 \hline 
\end{tabular}
\end{center}

\end{smaller}
\vskip \belowtablevskip
\end{table*}

\renewcommand{\arraystretch}{1.}
\tabcolsep=0.1cm
\begin{table*}[htb]
    \centering
    \caption{\textbf{Mean AUPR-E on our \dsetname{} dataset.} Higher is better. The difference to MSP is shown in red if a method performs worse, and in green if it improves. Bold values mark the best-performing method per model.}
    \label{tab:overview-auprcr-mean}
    \small
\begin{smaller}
\begin{center}
\begin{tabular}{l l l l l l l l l l l l l l }
pre & acc. & model & MSP & MaxL & Ener & KL-M & Maha & RMaha & ViM & E+R & KNN & Cos & MCM/RCos \\ 
 \hline 
{\multirow{9}{*}{{21k}}} & 86.0 & ViT-B-384 & 60.2 & 69.8 ${\textcolor{green}{+10}}$ & 71.3 ${\textcolor{green}{+11}}$ & 60.4 ${\textcolor{green}{+0}}$ & \textbf{78.9} ${\textcolor{green}{+19}}$ & 74.9 ${\textcolor{green}{+15}}$ & 74.2 ${\textcolor{green}{+14}}$ & 69.4 ${\textcolor{green}{+9}}$ & 53.8 ${\textcolor{red}{-6}}$ & 65.3 ${\textcolor{green}{+5}}$ & 65.3 ${\textcolor{green}{+5}}$ \\ 
  & 84.5 & ViT-B-224 & 56.1 & 64.9 ${\textcolor{green}{+9}}$ & 65.4 ${\textcolor{green}{+9}}$ & 56.4 ${\textcolor{green}{+0}}$ & \textbf{75.3} ${\textcolor{green}{+19}}$ & 72.4 ${\textcolor{green}{+16}}$ & 71.3 ${\textcolor{green}{+15}}$ & 62.0 ${\textcolor{green}{+6}}$ & 48.2 ${\textcolor{red}{-8}}$ & 59.5 ${\textcolor{green}{+3}}$ & 59.6 ${\textcolor{green}{+3}}$ \\ 
  & 86.3 & Swinv2-B-256 & 61.9 & 68.2 ${\textcolor{green}{+6}}$ & 69.8 ${\textcolor{green}{+8}}$ & 54.7 ${\textcolor{red}{-7}}$ & 54.7 ${\textcolor{red}{-7}}$ & 59.9 ${\textcolor{red}{-2}}$ & 59.8 ${\textcolor{red}{-2}}$ & \textbf{71.6} ${\textcolor{green}{+10}}$ & 52.6 ${\textcolor{red}{-9}}$ & 60.1 ${\textcolor{red}{-2}}$ & 63.0 ${\textcolor{green}{+1}}$ \\ 
  & 86.7 & Deit3-B-384 & 54.1 & 55.8 ${\textcolor{green}{+2}}$ & 54.7 ${\textcolor{green}{+1}}$ & 53.0 ${\textcolor{red}{-1}}$ & 59.2 ${\textcolor{green}{+5}}$ & 62.0 ${\textcolor{green}{+8}}$ & 57.2 ${\textcolor{green}{+3}}$ & 60.3 ${\textcolor{green}{+6}}$ & 59.0 ${\textcolor{green}{+5}}$ & 62.4 ${\textcolor{green}{+8}}$ & \textbf{63.4} ${\textcolor{green}{+9}}$ \\ 
  & 85.7 & Deit3-B-224 & 50.4 & 53.5 ${\textcolor{green}{+3}}$ & 53.3 ${\textcolor{green}{+3}}$ & 49.6 ${\textcolor{red}{-1}}$ & 54.9 ${\textcolor{green}{+4}}$ & 58.2 ${\textcolor{green}{+8}}$ & 53.8 ${\textcolor{green}{+3}}$ & 56.7 ${\textcolor{green}{+6}}$ & 55.8 ${\textcolor{green}{+5}}$ & 59.2 ${\textcolor{green}{+9}}$ & \textbf{60.0} ${\textcolor{green}{+10}}$ \\ 
  & 86.3 & CnvNxt-B & 64.0 & 68.3 ${\textcolor{green}{+4}}$ & 67.0 ${\textcolor{green}{+3}}$ & 57.1 ${\textcolor{red}{-7}}$ & 65.3 ${\textcolor{green}{+1}}$ & 68.2 ${\textcolor{green}{+4}}$ & 69.6 ${\textcolor{green}{+6}}$ & \textbf{70.2} ${\textcolor{green}{+6}}$ & 59.5 ${\textcolor{red}{-5}}$ & 67.2 ${\textcolor{green}{+3}}$ & 68.1 ${\textcolor{green}{+4}}$ \\ 
  & 84.1 & CnvNxt-T & 58.3 & 63.4 ${\textcolor{green}{+5}}$ & 65.1 ${\textcolor{green}{+7}}$ & 52.4 ${\textcolor{red}{-6}}$ & 64.8 ${\textcolor{green}{+7}}$ & 65.3 ${\textcolor{green}{+7}}$ & \textbf{71.4} ${\textcolor{green}{+13}}$ & 65.0 ${\textcolor{green}{+7}}$ & 55.3 ${\textcolor{red}{-3}}$ & 60.3 ${\textcolor{green}{+2}}$ & 62.6 ${\textcolor{green}{+4}}$ \\ 
  & 82.3 & BiT-m & 47.7 & 52.5 ${\textcolor{green}{+5}}$ & 51.8 ${\textcolor{green}{+4}}$ & 51.7 ${\textcolor{green}{+4}}$ & 63.5 ${\textcolor{green}{+16}}$ & 66.2 ${\textcolor{green}{+18}}$ & \textbf{68.8} ${\textcolor{green}{+21}}$ & 55.7 ${\textcolor{green}{+8}}$ & 56.6 ${\textcolor{green}{+9}}$ & 62.1 ${\textcolor{green}{+14}}$ & 60.2 ${\textcolor{green}{+13}}$ \\ 
  & 85.6 & EffNetv2-M & 60.9 & 62.3 ${\textcolor{green}{+1}}$ & 58.7 ${\textcolor{red}{-2}}$ & 56.5 ${\textcolor{red}{-4}}$ & 54.5 ${\textcolor{red}{-6}}$ & 59.6 ${\textcolor{red}{-1}}$ & 58.6 ${\textcolor{red}{-2}}$ & 31.0 ${\textcolor{red}{-30}}$ & 49.7 ${\textcolor{red}{-11}}$ & \textbf{65.6} ${\textcolor{green}{+5}}$ & 61.5 ${\textcolor{green}{+1}}$ \\ 
 \hline 
{\multirow{12}{*}{{none}}} & 81.1 & ViT-B-384 & 47.1 & 49.6 ${\textcolor{green}{+2}}$ & 50.3 ${\textcolor{green}{+3}}$ & 49.6 ${\textcolor{green}{+2}}$ & 56.6 ${\textcolor{green}{+9}}$ & \textbf{58.5} ${\textcolor{green}{+11}}$ & 48.3 ${\textcolor{green}{+1}}$ & 51.8 ${\textcolor{green}{+5}}$ & 43.9 ${\textcolor{red}{-3}}$ & 49.7 ${\textcolor{green}{+3}}$ & 49.2 ${\textcolor{green}{+2}}$ \\ 
  & 84.6 & Swinv2-B-256 & 46.8 & 46.9 ${\textcolor{green}{+0}}$ & 42.7 ${\textcolor{red}{-4}}$ & 48.7 ${\textcolor{green}{+2}}$ & 52.1 ${\textcolor{green}{+5}}$ & \textbf{54.8} ${\textcolor{green}{+8}}$ & 48.3 ${\textcolor{green}{+1}}$ & 46.0 ${\textcolor{red}{-1}}$ & 46.3 ${\textcolor{red}{-1}}$ & 51.7 ${\textcolor{green}{+5}}$ & 53.6 ${\textcolor{green}{+7}}$ \\ 
  & 85.1 & Deit3-B-384 & 49.8 & 43.1 ${\textcolor{red}{-7}}$ & 29.0 ${\textcolor{red}{-21}}$ & 51.4 ${\textcolor{green}{+2}}$ & 51.9 ${\textcolor{green}{+2}}$ & 55.5 ${\textcolor{green}{+6}}$ & 53.5 ${\textcolor{green}{+4}}$ & 25.4 ${\textcolor{red}{-24}}$ & 43.1 ${\textcolor{red}{-7}}$ & 49.3 ${\textcolor{red}{-0}}$ & \textbf{57.0} ${\textcolor{green}{+7}}$ \\ 
  & 83.8 & Deit3-B-224 & 47.0 & 45.2 ${\textcolor{red}{-2}}$ & 35.7 ${\textcolor{red}{-11}}$ & 48.5 ${\textcolor{green}{+2}}$ & 49.4 ${\textcolor{green}{+2}}$ & 51.9 ${\textcolor{green}{+5}}$ & 50.9 ${\textcolor{green}{+4}}$ & 35.5 ${\textcolor{red}{-11}}$ & 38.5 ${\textcolor{red}{-9}}$ & 45.8 ${\textcolor{red}{-1}}$ & \textbf{52.7} ${\textcolor{green}{+6}}$ \\ 
  & 82.6 & XCiT-M-224 & 42.9 & 40.1 ${\textcolor{red}{-3}}$ & 33.5 ${\textcolor{red}{-9}}$ & 46.8 ${\textcolor{green}{+4}}$ & 52.6 ${\textcolor{green}{+10}}$ & \textbf{54.2} ${\textcolor{green}{+11}}$ & 52.6 ${\textcolor{green}{+10}}$ & 38.6 ${\textcolor{red}{-4}}$ & 44.6 ${\textcolor{green}{+2}}$ & 50.3 ${\textcolor{green}{+7}}$ & 50.3 ${\textcolor{green}{+7}}$ \\ 
  & 84.3 & XCiT-M-224-d & 49.0 & 48.4 ${\textcolor{red}{-1}}$ & 41.7 ${\textcolor{red}{-7}}$ & 50.2 ${\textcolor{green}{+1}}$ & 51.3 ${\textcolor{green}{+2}}$ & \textbf{53.7} ${\textcolor{green}{+5}}$ & 53.1 ${\textcolor{green}{+4}}$ & 43.6 ${\textcolor{red}{-5}}$ & 46.0 ${\textcolor{red}{-3}}$ & 51.8 ${\textcolor{green}{+3}}$ & 52.6 ${\textcolor{green}{+4}}$ \\ 
  & 84.4 & CnvNxt-B & 49.6 & 43.7 ${\textcolor{red}{-6}}$ & 27.7 ${\textcolor{red}{-22}}$ & 48.8 ${\textcolor{red}{-1}}$ & 51.4 ${\textcolor{green}{+2}}$ & \textbf{55.5} ${\textcolor{green}{+6}}$ & 50.3 ${\textcolor{green}{+1}}$ & 33.3 ${\textcolor{red}{-16}}$ & 45.6 ${\textcolor{red}{-4}}$ & 53.1 ${\textcolor{green}{+3}}$ & 55.5 ${\textcolor{green}{+6}}$ \\ 
  & 78.0 & BiT-s & 40.5 & 37.5 ${\textcolor{red}{-3}}$ & 36.5 ${\textcolor{red}{-4}}$ & 49.1 ${\textcolor{green}{+9}}$ & 33.6 ${\textcolor{red}{-7}}$ & \textbf{51.5} ${\textcolor{green}{+11}}$ & 42.1 ${\textcolor{green}{+2}}$ & 39.4 ${\textcolor{red}{-1}}$ & 32.8 ${\textcolor{red}{-8}}$ & 43.1 ${\textcolor{green}{+3}}$ & 33.1 ${\textcolor{red}{-7}}$ \\ 
  & 85.1 & EffNetv2-M & 50.1 & 48.6 ${\textcolor{red}{-1}}$ & 39.0 ${\textcolor{red}{-11}}$ & 52.3 ${\textcolor{green}{+2}}$ & 53.8 ${\textcolor{green}{+4}}$ & 58.8 ${\textcolor{green}{+9}}$ & 45.3 ${\textcolor{red}{-5}}$ & 45.0 ${\textcolor{red}{-5}}$ & 52.2 ${\textcolor{green}{+2}}$ & 55.1 ${\textcolor{green}{+5}}$ & \textbf{59.1} ${\textcolor{green}{+9}}$ \\ 
  & 84.9 & EffNetb7 & 48.3 & 43.9 ${\textcolor{red}{-4}}$ & 31.5 ${\textcolor{red}{-17}}$ & 53.8 ${\textcolor{green}{+5}}$ & 49.5 ${\textcolor{green}{+1}}$ & 59.2 ${\textcolor{green}{+11}}$ & 45.5 ${\textcolor{red}{-3}}$ & 39.4 ${\textcolor{red}{-9}}$ & 49.8 ${\textcolor{green}{+2}}$ & 54.0 ${\textcolor{green}{+6}}$ & \textbf{60.3} ${\textcolor{green}{+12}}$ \\ 
  & 77.7 & EffNet-B0 & 45.3 & 44.1 ${\textcolor{red}{-1}}$ & 37.9 ${\textcolor{red}{-7}}$ & 45.9 ${\textcolor{green}{+1}}$ & 36.3 ${\textcolor{red}{-9}}$ & 45.1 ${\textcolor{red}{-0}}$ & 43.0 ${\textcolor{red}{-2}}$ & 43.0 ${\textcolor{red}{-2}}$ & 33.7 ${\textcolor{red}{-12}}$ & \textbf{54.2} ${\textcolor{green}{+9}}$ & 48.1 ${\textcolor{green}{+3}}$ \\ 
  & 80.4 & ResNet50 & 44.8 & 44.7 ${\textcolor{red}{-0}}$ & 42.2 ${\textcolor{red}{-3}}$ & 48.0 ${\textcolor{green}{+3}}$ & 34.2 ${\textcolor{red}{-11}}$ & 48.4 ${\textcolor{green}{+4}}$ & 41.5 ${\textcolor{red}{-3}}$ & 22.0 ${\textcolor{red}{-23}}$ & 39.4 ${\textcolor{red}{-5}}$ & 53.1 ${\textcolor{green}{+8}}$ & \textbf{54.3} ${\textcolor{green}{+9}}$ \\ 
 \hline 
{\multirow{1}{*}{{JFT}}} & 86.8 & EffNetb7-ns & 52.7 & \textbf{56.3} ${\textcolor{green}{+4}}$ & 50.6 ${\textcolor{red}{-2}}$ & 49.3 ${\textcolor{red}{-3}}$ & 36.1 ${\textcolor{red}{-17}}$ & 50.2 ${\textcolor{red}{-3}}$ & 33.1 ${\textcolor{red}{-20}}$ & 50.6 ${\textcolor{red}{-2}}$ & 43.1 ${\textcolor{red}{-10}}$ & 50.4 ${\textcolor{red}{-2}}$ & 51.5 ${\textcolor{red}{-1}}$ \\ 
 \hline 
{\multirow{2}{*}{{\shortstack[l]{clip\\+12k}}}} & 87.2 & ViT-B-384-l2b & 61.0 & 61.4 ${\textcolor{green}{+0}}$ & 58.3 ${\textcolor{red}{-3}}$ & 58.2 ${\textcolor{red}{-3}}$ & 62.4 ${\textcolor{green}{+1}}$ & 65.2 ${\textcolor{green}{+4}}$ & 66.9 ${\textcolor{green}{+6}}$ & 63.4 ${\textcolor{green}{+2}}$ & 64.2 ${\textcolor{green}{+3}}$ & \textbf{68.2} ${\textcolor{green}{+7}}$ & 68.0 ${\textcolor{green}{+7}}$ \\ 
  & 87.0 & ViT-B-384-oai & 62.2 & 64.9 ${\textcolor{green}{+3}}$ & 64.5 ${\textcolor{green}{+2}}$ & 60.5 ${\textcolor{red}{-2}}$ & 59.5 ${\textcolor{red}{-3}}$ & 64.2 ${\textcolor{green}{+2}}$ & 62.3 ${\textcolor{green}{+0}}$ & 66.8 ${\textcolor{green}{+5}}$ & 63.8 ${\textcolor{green}{+2}}$ & 69.4 ${\textcolor{green}{+7}}$ & \textbf{69.7} ${\textcolor{green}{+7}}$ \\ 
 \hline 
{\multirow{2}{*}{{clip}}} & 86.6 & ViT-B-384-l2b & 51.9 & 48.7 ${\textcolor{red}{-3}}$ & 43.8 ${\textcolor{red}{-8}}$ & 56.1 ${\textcolor{green}{+4}}$ & 61.0 ${\textcolor{green}{+9}}$ & 61.3 ${\textcolor{green}{+9}}$ & 60.2 ${\textcolor{green}{+8}}$ & 46.3 ${\textcolor{red}{-6}}$ & 55.9 ${\textcolor{green}{+4}}$ & 60.7 ${\textcolor{green}{+9}}$ & \textbf{63.0} ${\textcolor{green}{+11}}$ \\ 
  & 86.2 & ViT-B-384-oai & 50.2 & 44.6 ${\textcolor{red}{-6}}$ & 39.5 ${\textcolor{red}{-11}}$ & 53.9 ${\textcolor{green}{+4}}$ & 59.8 ${\textcolor{green}{+10}}$ & 59.8 ${\textcolor{green}{+10}}$ & 58.9 ${\textcolor{green}{+9}}$ & 42.1 ${\textcolor{red}{-8}}$ & 53.5 ${\textcolor{green}{+3}}$ & 58.3 ${\textcolor{green}{+8}}$ & \textbf{60.2} ${\textcolor{green}{+10}}$ \\ 
 \hline 
{\multirow{2}{*}{\shortstack[l]{clip\\z. shot}}} & 74.3 & clip-ViT-L-336 & ---- & ---- & ---- & ---- & ---- & ---- & ---- & ---- & ---- & 44.2 & \textbf{48.5} \\ 
  & 66.6 & clip-ViT-B-224 & ---- & ---- & ---- & ---- & ---- & ---- & ---- & ---- & ---- & 37.5 & \textbf{38.0} \\ 
 \hline 
\end{tabular}
\end{center}

\end{smaller}
\vskip \belowtablevskip
\end{table*}

\renewcommand{\arraystretch}{1.3}
\tabcolsep=0.05751cm
\begin{table*}[htb]
    \centering
    \caption{\FPR{} on all classes of \dsetname{} (lower is better) for a pretrained ViT-B and a pretrained ConvNext-B.}
    \label{tab:all-datasets-fpr}
    \small
    \begin{tiny}
\begin{center}
\rowcolors{2}{white}{gray!25}
\begin{tabular}{c | c c c c c c c c c c c | c c c c c c c c c c c }
 & \multicolumn{11}{c}{ViT-B-384-21k}&  \multicolumn{11}{c}{CnvNxt-B-21k} \\ 
  Dataset &MSP & MaxL & Ener & KL-M & Maha & RMaha & ViM & E+R & KNN & Cos & RCos & MSP & MaxL & Ener & KL-M & Maha & RMaha & ViM & E+R & KNN & Cos & RCos \\ 
 \hline 
\OODclass{Caracal} & 79.0 & 75.0 & 73.0 & 68.0 & 53.0 & 47.0 & 82.0 & 88.0 & 87.0 & 83.0 & 81.0 & 77.0 & 74.0 & 67.0 & 79.0 & 79.0 & 76.0 & 86.0 & 67.0 & 87.0 & 77.0 & 77.0 \\ 
 \OODclass{2TAmph} & 75.6 & 90.9 & 91.5 & 70.5 & 68.8 & 54.0 & 92.0 & 93.8 & 95.5 & 86.9 & 84.7 & 79.5 & 85.8 & 91.5 & 69.3 & 63.6 & 54.5 & 72.7 & 86.9 & 83.5 & 72.2 & 71.0 \\ 
 \OODclass{AFA} & 95.7 & 87.0 & 82.6 & 93.5 & 60.9 & 71.7 & 65.2 & 80.4 & 80.4 & 84.8 & 82.6 & 89.1 & 82.6 & 69.6 & 89.1 & 73.9 & 80.4 & 45.7 & 65.2 & 82.6 & 80.4 & 80.4 \\ 
 \OODclass{CatFSp} & 90.0 & 96.0 & 96.0 & 94.0 & 91.0 & 88.0 & 91.0 & 97.0 & 100.0 & 97.0 & 97.0 & 78.0 & 77.0 & 86.0 & 92.0 & 87.0 & 81.0 & 88.0 & 86.0 & 98.0 & 96.0 & 95.0 \\ 
 \OODclass{GFurS} & 100.0 & 98.9 & 96.7 & 95.6 & 97.8 & 96.7 & 75.8 & 100.0 & 98.9 & 100.0 & 100.0 & 98.9 & 98.9 & 81.3 & 96.7 & 100.0 & 100.0 & 100.0 & 87.9 & 100.0 & 100.0 & 100.0 \\ 
 \OODclass{Bagp} & 8.2 & 2.1 & 2.1 & 12.4 & 4.1 & 4.1 & 4.1 & 1.0 & 11.3 & 6.2 & 4.1 & 16.5 & 5.2 & 2.1 & 37.1 & 70.1 & 20.6 & 18.6 & 3.1 & 12.4 & 8.2 & 7.2 \\ 
 \OODclass{CSSala} & 76.0 & 86.0 & 89.0 & 76.0 & 42.0 & 60.0 & 57.0 & 93.0 & 100.0 & 94.0 & 94.0 & 78.0 & 71.0 & 66.0 & 79.0 & 90.0 & 70.0 & 92.0 & 66.0 & 97.0 & 86.0 & 82.0 \\ 
 \OODclass{Cabl} & 46.6 & 15.9 & 14.8 & 56.8 & 17.0 & 42.0 & 11.4 & 21.6 & 45.5 & 48.9 & 38.6 & 35.2 & 36.4 & 48.9 & 52.3 & 54.5 & 60.2 & 29.5 & 48.9 & 23.9 & 34.1 & 31.8 \\ 
 \OODclass{CQuesa} & 86.0 & 95.0 & 97.0 & 81.0 & 73.0 & 77.0 & 82.0 & 88.0 & 99.0 & 98.0 & 98.0 & 85.0 & 87.0 & 93.0 & 85.0 & 83.0 & 78.0 & 81.0 & 92.0 & 96.0 & 89.0 & 87.0 \\ 
 \OODclass{DThist} & 52.0 & 22.0 & 19.0 & 48.0 & 9.0 & 13.0 & 6.0 & 20.0 & 51.0 & 34.0 & 32.0 & 41.0 & 42.0 & 49.0 & 46.0 & 18.0 & 27.0 & 10.0 & 46.0 & 41.0 & 34.0 & 34.0 \\ 
 \OODclass{CBrûlée} & 53.5 & 33.3 & 32.3 & 66.7 & 14.1 & 24.2 & 23.2 & 28.3 & 79.8 & 62.6 & 54.5 & 30.3 & 19.2 & 24.2 & 58.6 & 67.7 & 45.5 & 36.4 & 19.2 & 45.5 & 41.4 & 39.4 \\ 
 \OODclass{LTSilF} & 59.0 & 28.0 & 26.0 & 51.0 & 8.0 & 4.0 & 36.0 & 40.0 & 79.0 & 44.0 & 40.0 & 35.0 & 17.0 & 11.0 & 68.0 & 92.0 & 63.0 & 94.0 & 11.0 & 98.0 & 75.0 & 66.0 \\ 
 \OODclass{CCake} & 77.5 & 62.5 & 55.0 & 71.2 & 35.0 & 83.8 & 18.8 & 56.2 & 88.8 & 51.2 & 46.2 & 81.2 & 85.0 & 93.8 & 66.2 & 11.2 & 40.0 & 10.0 & 88.8 & 12.5 & 17.5 & 20.0 \\ 
 \OODclass{CPitch} & 23.0 & 5.0 & 6.0 & 14.0 & 0.0 & 0.0 & 0.0 & 1.0 & 51.0 & 14.0 & 15.0 & 22.0 & 18.0 & 26.0 & 30.0 & 2.0 & 6.0 & 3.0 & 22.0 & 18.0 & 10.0 & 10.0 \\ 
 \OODclass{LTRoo} & 68.0 & 88.0 & 95.0 & 72.0 & 79.0 & 68.0 & 93.0 & 97.0 & 100.0 & 96.0 & 98.0 & 62.0 & 70.0 & 84.0 & 60.0 & 62.0 & 39.0 & 84.0 & 84.0 & 93.0 & 80.0 & 78.0 \\ 
 \OODclass{Donuts} & 81.0 & 79.0 & 80.0 & 86.0 & 84.0 & 88.0 & 74.0 & 77.0 & 97.0 & 86.0 & 86.0 & 76.0 & 70.0 & 71.0 & 82.0 & 77.0 & 82.0 & 68.0 & 71.0 & 82.0 & 82.0 & 82.0 \\ 
 \OODclass{Door} & 54.0 & 26.0 & 26.0 & 56.0 & 60.0 & 77.0 & 29.0 & 25.0 & 67.0 & 49.0 & 52.0 & 30.0 & 29.0 & 39.0 & 43.0 & 65.0 & 60.0 & 33.0 & 32.0 & 40.0 & 34.0 & 32.0 \\ 
 \OODclass{WDisp} & 52.5 & 36.4 & 35.4 & 44.4 & 21.2 & 35.4 & 24.2 & 38.4 & 61.6 & 28.3 & 30.3 & 58.6 & 51.5 & 54.5 & 59.6 & 69.7 & 60.6 & 62.6 & 55.6 & 49.5 & 37.4 & 36.4 \\ 
 \OODclass{EMicro} & 35.0 & 24.0 & 23.0 & 26.0 & 2.0 & 1.0 & 22.0 & 23.0 & 44.0 & 25.0 & 19.0 & 47.0 & 39.0 & 45.0 & 48.0 & 21.0 & 17.0 & 24.0 & 38.0 & 45.0 & 34.0 & 28.0 \\ 
 \OODclass{Franci} & 26.0 & 9.0 & 3.0 & 18.0 & 0.0 & 0.0 & 0.0 & 4.0 & 81.0 & 15.0 & 16.0 & 28.0 & 10.0 & 13.0 & 30.0 & 12.0 & 18.0 & 5.0 & 11.0 & 44.0 & 24.0 & 24.0 \\ 
 \OODclass{FieldRd} & 26.0 & 17.7 & 15.6 & 33.3 & 24.0 & 25.0 & 21.9 & 18.8 & 50.0 & 22.9 & 25.0 & 29.2 & 26.0 & 41.7 & 33.3 & 24.0 & 29.2 & 20.8 & 33.3 & 37.5 & 19.8 & 20.8 \\ 
 \OODclass{ForPth} & 24.0 & 5.0 & 3.0 & 38.0 & 3.0 & 11.0 & 3.0 & 4.0 & 29.0 & 19.0 & 18.0 & 13.0 & 11.0 & 18.0 & 21.0 & 9.0 & 18.0 & 4.0 & 12.0 & 20.0 & 15.0 & 15.0 \\ 
 \OODclass{MLCact} & 13.0 & 0.0 & 0.0 & 15.0 & 0.0 & 0.0 & 0.0 & 0.0 & 9.0 & 1.0 & 0.0 & 32.0 & 20.0 & 17.0 & 44.0 & 3.0 & 5.0 & 2.0 & 14.0 & 2.0 & 9.0 & 9.0 \\ 
 \OODclass{FireEx} & 16.0 & 2.8 & 0.9 & 13.2 & 1.9 & 2.8 & 0.9 & 0.0 & 5.7 & 1.9 & 1.9 & 5.7 & 3.8 & 4.7 & 10.4 & 73.6 & 26.4 & 41.5 & 3.8 & 0.9 & 1.9 & 1.9 \\ 
 \OODclass{FireW} & 50.0 & 30.0 & 29.0 & 45.0 & 11.0 & 10.0 & 28.0 & 28.0 & 32.0 & 26.0 & 24.0 & 58.0 & 62.0 & 69.0 & 47.0 & 16.0 & 22.0 & 19.0 & 68.0 & 45.0 & 35.0 & 35.0 \\ 
 \OODclass{Fries} & 38.0 & 30.0 & 39.0 & 37.0 & 3.0 & 1.0 & 22.0 & 6.0 & 99.0 & 70.0 & 76.0 & 67.0 & 53.0 & 56.0 & 57.0 & 82.0 & 55.0 & 73.0 & 53.0 & 96.0 & 82.0 & 73.0 \\ 
 \OODclass{GlMilk} & 83.1 & 64.0 & 55.1 & 86.5 & 82.0 & 89.9 & 46.1 & 68.5 & 77.5 & 86.5 & 86.5 & 61.8 & 56.2 & 46.1 & 73.0 & 79.8 & 82.0 & 49.4 & 47.2 & 65.2 & 77.5 & 75.3 \\ 
 \OODclass{Gramo} & 7.1 & 1.8 & 0.0 & 16.1 & 0.0 & 0.0 & 1.8 & 0.0 & 0.0 & 0.0 & 0.0 & 0.0 & 0.0 & 0.0 & 16.1 & 87.5 & 67.9 & 60.7 & 0.0 & 0.0 & 0.0 & 0.0 \\ 
 \OODclass{BSGrunt} & 49.0 & 20.8 & 12.5 & 51.0 & 2.1 & 4.2 & 2.1 & 10.4 & 56.2 & 17.7 & 20.8 & 31.2 & 26.0 & 29.2 & 42.7 & 3.1 & 7.3 & 1.0 & 20.8 & 24.0 & 18.8 & 15.6 \\ 
 \OODclass{HHeels} & 46.5 & 35.4 & 32.3 & 64.6 & 79.8 & 87.9 & 40.4 & 70.7 & 83.8 & 78.8 & 48.5 & 41.4 & 35.4 & 45.5 & 64.6 & 48.5 & 54.5 & 17.2 & 41.4 & 63.6 & 45.5 & 44.4 \\ 
 \OODclass{HinTp} & 88.2 & 84.3 & 82.4 & 82.4 & 64.7 & 56.9 & 76.5 & 80.4 & 98.0 & 88.2 & 90.2 & 68.6 & 54.9 & 41.2 & 80.4 & 86.3 & 86.3 & 68.6 & 39.2 & 92.2 & 86.3 & 88.2 \\ 
 \OODclass{HHClam} & 93.5 & 93.5 & 90.3 & 93.5 & 61.3 & 67.7 & 77.4 & 93.5 & 96.8 & 96.8 & 93.5 & 77.4 & 54.8 & 51.6 & 74.2 & 58.1 & 67.7 & 41.9 & 45.2 & 71.0 & 71.0 & 71.0 \\ 
 \OODclass{SilverHB} & 12.1 & 2.0 & 2.0 & 8.1 & 3.0 & 2.0 & 4.0 & 2.0 & 6.1 & 3.0 & 3.0 & 11.1 & 10.1 & 16.2 & 15.2 & 2.0 & 3.0 & 2.0 & 11.1 & 6.1 & 4.0 & 4.0 \\ 
 \OODclass{SwPea} & 10.0 & 1.0 & 1.0 & 12.0 & 0.0 & 1.0 & 1.0 & 0.0 & 33.0 & 2.0 & 2.0 & 19.0 & 14.0 & 19.0 & 29.0 & 0.0 & 1.0 & 0.0 & 9.0 & 1.0 & 1.0 & 1.0 \\ 
 \OODclass{RBSunf} & 76.0 & 12.0 & 5.0 & 69.0 & 2.0 & 2.0 & 4.0 & 1.0 & 65.0 & 14.0 & 9.0 & 36.0 & 13.0 & 7.0 & 53.0 & 81.0 & 57.0 & 30.0 & 7.0 & 44.0 & 35.0 & 18.0 \\ 
 \OODclass{ELFBug} & 37.0 & 26.0 & 49.0 & 38.0 & 3.0 & 3.0 & 44.0 & 71.0 & 98.0 & 67.0 & 83.0 & 35.0 & 20.0 & 17.0 & 66.0 & 45.0 & 29.0 & 22.0 & 16.0 & 67.0 & 41.0 & 36.0 \\ 
 \OODclass{Mbira} & 28.4 & 19.4 & 20.9 & 25.4 & 23.9 & 16.4 & 32.8 & 20.9 & 17.9 & 19.4 & 17.9 & 40.3 & 40.3 & 71.6 & 29.9 & 10.4 & 9.0 & 11.9 & 61.2 & 6.0 & 10.4 & 9.0 \\ 
 \OODclass{MWesen} & 97.0 & 33.3 & 18.2 & 90.9 & 3.0 & 30.3 & 3.0 & 9.1 & 57.6 & 57.6 & 39.4 & 78.8 & 51.5 & 30.3 & 87.9 & 51.5 & 78.8 & 6.1 & 33.3 & 81.8 & 84.8 & 84.8 \\ 
 \OODclass{C2SOct} & 40.0 & 49.0 & 58.0 & 36.0 & 22.0 & 17.0 & 61.0 & 62.0 & 87.0 & 54.0 & 52.0 & 40.0 & 48.0 & 58.0 & 45.0 & 25.0 & 13.0 & 34.0 & 51.0 & 59.0 & 34.0 & 32.0 \\ 
 \OODclass{RubyOct} & 42.0 & 48.0 & 48.0 & 39.0 & 22.0 & 14.0 & 55.0 & 48.0 & 88.0 & 54.0 & 54.0 & 30.0 & 34.0 & 44.0 & 32.0 & 25.0 & 13.0 & 32.0 & 40.0 & 50.0 & 28.0 & 28.0 \\ 
 \OODclass{PDeer} & 81.7 & 61.0 & 59.8 & 89.0 & 46.3 & 58.5 & 52.4 & 85.4 & 91.5 & 80.5 & 80.5 & 64.6 & 34.1 & 31.7 & 93.9 & 84.1 & 73.2 & 75.6 & 29.3 & 95.1 & 81.7 & 80.5 \\ 
 \OODclass{DFlath} & 58.0 & 33.0 & 32.0 & 52.0 & 3.0 & 18.0 & 13.0 & 32.0 & 62.0 & 31.0 & 33.0 & 62.0 & 58.0 & 53.0 & 68.0 & 37.0 & 39.0 & 26.0 & 47.0 & 64.0 & 52.0 & 49.0 \\ 
 \OODclass{EPWasp} & 73.0 & 55.0 & 56.0 & 64.0 & 16.0 & 29.0 & 29.0 & 63.0 & 90.0 & 75.0 & 72.0 & 39.0 & 27.0 & 24.0 & 73.0 & 63.0 & 43.0 & 45.0 & 24.0 & 76.0 & 65.0 & 59.0 \\ 
 \OODclass{FalseKW} & 80.6 & 74.6 & 74.6 & 74.6 & 55.2 & 55.2 & 59.7 & 86.6 & 98.5 & 91.0 & 89.6 & 73.1 & 53.7 & 53.7 & 74.6 & 85.1 & 77.6 & 86.6 & 56.7 & 97.0 & 85.1 & 83.6 \\ 
 \OODclass{Pyra} & 11.0 & 5.0 & 6.0 & 12.0 & 5.0 & 6.0 & 7.0 & 6.0 & 11.0 & 7.0 & 5.0 & 21.0 & 14.0 & 19.0 & 26.0 & 30.0 & 9.0 & 12.0 & 15.0 & 13.0 & 11.0 & 10.0 \\ 
 \OODclass{Sky} & 22.1 & 23.5 & 25.0 & 22.1 & 27.9 & 25.0 & 38.2 & 29.4 & 44.1 & 14.7 & 16.2 & 20.6 & 25.0 & 64.7 & 17.6 & 17.6 & 23.5 & 25.0 & 54.4 & 25.0 & 13.2 & 13.2 \\ 
 \OODclass{Dreamf} & 60.0 & 44.0 & 45.0 & 63.0 & 26.0 & 31.0 & 30.0 & 42.0 & 69.0 & 48.0 & 50.0 & 64.0 & 63.0 & 67.0 & 65.0 & 15.0 & 29.0 & 11.0 & 59.0 & 65.0 & 56.0 & 53.0 \\ 
 \OODclass{YTrump} & 14.0 & 1.0 & 0.0 & 4.0 & 0.0 & 0.0 & 0.0 & 0.0 & 54.0 & 10.0 & 14.0 & 14.0 & 7.0 & 5.0 & 23.0 & 3.0 & 1.0 & 0.0 & 1.0 & 3.0 & 0.0 & 0.0 \\ 
 \OODclass{Sciss} & 29.0 & 9.0 & 10.0 & 42.0 & 9.0 & 12.0 & 11.0 & 11.0 & 19.0 & 22.0 & 26.0 & 27.0 & 24.0 & 24.0 & 44.0 & 67.0 & 64.0 & 31.0 & 22.0 & 16.0 & 31.0 & 26.0 \\ 
 \OODclass{GCuttle} & 30.3 & 10.1 & 8.1 & 33.3 & 3.0 & 4.0 & 15.2 & 10.1 & 37.4 & 14.1 & 14.1 & 35.4 & 30.3 & 35.4 & 47.5 & 42.4 & 11.1 & 62.6 & 35.4 & 42.4 & 18.2 & 15.2 \\ 
 \OODclass{CCuttle} & 34.0 & 23.0 & 24.0 & 24.0 & 9.0 & 8.0 & 27.0 & 25.0 & 44.0 & 22.0 & 20.0 & 22.0 & 22.0 & 30.0 & 34.0 & 39.0 & 8.0 & 57.0 & 27.0 & 36.0 & 16.0 & 15.0 \\ 
 \OODclass{SCalam} & 21.2 & 11.1 & 12.1 & 21.2 & 4.0 & 4.0 & 13.1 & 8.1 & 40.4 & 15.2 & 15.2 & 29.3 & 25.3 & 27.3 & 35.4 & 10.1 & 5.1 & 11.1 & 22.2 & 34.3 & 18.2 & 18.2 \\ 
 \OODclass{ShCo} & 58.2 & 4.5 & 4.5 & 43.3 & 7.5 & 1.5 & 37.3 & 7.5 & 3.0 & 1.5 & 1.5 & 13.4 & 7.5 & 6.0 & 64.2 & 83.6 & 17.9 & 76.1 & 6.0 & 10.4 & 10.4 & 10.4 \\ 
 \OODclass{SCaterp} & 11.0 & 13.0 & 19.0 & 11.0 & 3.0 & 3.0 & 14.0 & 15.0 & 81.0 & 28.0 & 29.0 & 31.0 & 21.0 & 27.0 & 29.0 & 5.0 & 8.0 & 4.0 & 21.0 & 22.0 & 13.0 & 12.0 \\ 
 \OODclass{SBolo} & 67.2 & 47.8 & 49.3 & 83.6 & 68.7 & 79.1 & 50.7 & 22.4 & 100.0 & 100.0 & 100.0 & 71.6 & 76.1 & 82.1 & 82.1 & 74.6 & 61.2 & 73.1 & 82.1 & 100.0 & 94.0 & 91.0 \\ 
 \OODclass{Stapl} & 34.0 & 14.0 & 12.0 & 31.0 & 11.0 & 21.0 & 13.0 & 18.0 & 20.0 & 17.0 & 19.0 & 27.0 & 23.0 & 24.0 & 30.0 & 72.0 & 59.0 & 38.0 & 22.0 & 22.0 & 27.0 & 26.0 \\ 
 \OODclass{Rosyb} & 65.0 & 28.0 & 17.0 & 45.0 & 0.0 & 2.0 & 2.0 & 11.0 & 86.0 & 36.0 & 37.0 & 75.0 & 72.0 & 70.0 & 72.0 & 23.0 & 40.0 & 23.0 & 65.0 & 71.0 & 53.0 & 53.0 \\ 
 \OODclass{CATapir} & 13.0 & 10.0 & 11.0 & 12.0 & 4.0 & 3.0 & 15.0 & 15.0 & 23.0 & 13.0 & 13.0 & 29.0 & 32.0 & 39.0 & 48.0 & 51.0 & 10.0 & 74.0 & 36.0 & 58.0 & 24.0 & 24.0 \\ 
 \OODclass{MNewt} & 90.0 & 94.0 & 95.0 & 91.0 & 82.0 & 83.0 & 80.0 & 96.0 & 99.0 & 98.0 & 98.0 & 90.0 & 93.0 & 95.0 & 93.0 & 80.0 & 74.0 & 86.0 & 95.0 & 99.0 & 92.0 & 91.0 \\ 
 \OODclass{IPBNDol} & 57.0 & 47.0 & 45.0 & 56.0 & 27.0 & 29.0 & 47.0 & 64.0 & 90.0 & 64.0 & 69.0 & 47.0 & 36.0 & 39.0 & 64.0 & 84.0 & 59.0 & 92.0 & 41.0 & 94.0 & 72.0 & 69.0 \\ 
 \OODclass{ʻōʻai} & 63.0 & 28.0 & 23.0 & 51.0 & 2.0 & 5.0 & 3.0 & 18.0 & 71.0 & 33.0 & 32.0 & 54.0 & 47.0 & 35.0 & 50.0 & 7.0 & 14.0 & 2.0 & 25.0 & 30.0 & 19.0 & 19.0 \\ 
 \OODclass{Waffle} & 57.4 & 54.1 & 59.0 & 55.7 & 59.0 & 49.2 & 60.7 & 52.5 & 83.6 & 52.5 & 57.4 & 70.5 & 72.1 & 80.3 & 55.7 & 59.0 & 62.3 & 59.0 & 75.4 & 54.1 & 62.3 & 62.3 \\ 
 \OODclass{Walker} & 77.8 & 52.5 & 46.5 & 53.5 & 42.4 & 57.6 & 44.4 & 35.4 & 56.6 & 56.6 & 52.5 & 34.3 & 17.2 & 15.2 & 35.4 & 32.3 & 22.2 & 18.2 & 14.1 & 12.1 & 12.1 & 11.1 \\ 
 \OODclass{WiChair} & 97.2 & 40.8 & 25.4 & 90.1 & 15.5 & 32.4 & 12.7 & 43.7 & 31.0 & 32.4 & 25.4 & 81.7 & 49.3 & 25.4 & 94.4 & 97.2 & 95.8 & 84.5 & 28.2 & 95.8 & 91.5 & 88.7 \\ 
 \hline 
mean & 51.9 & 37.8 & 36.9 & 50.3 & 27.5 & 31.2 & 32.6 & 38.5 & 62.7 & 46.0 & 45.0 & 47.2 & 41.1 & 43.3 & 54.9 & 49.6 & 42.4 & 41.5 & 40.5 & 51.8 & 44.2 & 42.6 \\ 
 \hline 
\end{tabular}
\end{center}

\end{tiny}
\vskip \belowtablevskip
\end{table*}
\FloatBarrier

\section{Models}\label{sec:models}
In Table \ref{tab:models-overview} we give an overview over the evaluated models. 
All model implementation and model weights were taken from the publicly available timm-repository \cite{rw2019timm}, except for the BiT-s weights, which can be obtained via the \href{https://github.com/google-research/big_transfer}{\underline{github repository}} of \cite{kolesnikov2020bit}, and the zero-shot CLIP models, which are also available via \href{https://github.com/openai/CLIP}{\underline{github}}. For the ViTs finetuned from CLIP and the ViT without pretraining we used the timm-version \textit{0.8.0dev0}, for all other models version \textit{0.6.12}. IN-12k (\href{https://web.archive.org/web/20230201122707/https://twitter.com/wightmanr/status/1616908317196169217}{\underline{description}} and \href{https://github.com/rwightman/pytorch-image-models/blob/main/results/imagenet12k_synsets.txt}{\underline{defining synsets}}) is a subset of IN-21k, for which the classes with few samples are excluded, leading to an overlap of roughly 85\%.
\tabcolsep=0.1cm
\renewcommand{\arraystretch}{1.}
\begin{table}[htb]
    \caption{Overview over the evaluated models.}
    \label{tab:models-overview}
    \centering
\begin{center}
\begin{tabular}{c c c c c }
model & pretraining & top-1 acc. & params & timm name  \\ 
\hline 
ViT-B-384-l2b-12k & laion2b + IN-12k & 87.2 & 87M & vit{\textunderscore}base{\textunderscore}patch16{\textunderscore}clip{\textunderscore}384.laion2b{\textunderscore}ft{\textunderscore}in12k{\textunderscore}in1k \\ 
 ViT-B-384-oai-12k & openai + IN-12k & 87.0 & 87M & vit{\textunderscore}base{\textunderscore}patch16{\textunderscore}clip{\textunderscore}384.openai{\textunderscore}ft{\textunderscore}in12k{\textunderscore}in1k \\ 
 ViT-B-384-l2b & laion2b & 86.6 & 87M & vit{\textunderscore}base{\textunderscore}patch16{\textunderscore}clip{\textunderscore}384.laion2b{\textunderscore}ft{\textunderscore}in1k \\ 
 ViT-B-384-oai & openai & 86.2 & 87M & vit{\textunderscore}base{\textunderscore}patch16{\textunderscore}clip{\textunderscore}384.openai{\textunderscore}ft{\textunderscore}in1k \\ 
 ViT-B-384-21k & IN-21k & 86.0 & 87M & vit{\textunderscore}base{\textunderscore}patch16{\textunderscore}384 \\ 
 ViT-B-224-21k & IN-21k & 84.5 & 87M & vit{\textunderscore}base{\textunderscore}patch16{\textunderscore}224 \\ 
 Swinv2-B-256-21k & IN-21k & 86.3 & 88M & swinv2{\textunderscore}base{\textunderscore}window12to16{\textunderscore}192to256{\textunderscore}22kft1k \\ 
 Deit3-B-384-21k & IN-21k & 86.7 & 87M & deit3{\textunderscore}base{\textunderscore}patch16{\textunderscore}384{\textunderscore}in21ft1k \\ 
 Deit3-B-224-21k & IN-21k & 85.7 & 87M & deit3{\textunderscore}base{\textunderscore}patch16{\textunderscore}224{\textunderscore}in21ft1k \\ 
 CnvNxt-B-21k & IN-21k & 86.3 & 89M & convnext{\textunderscore}base{\textunderscore}in22ft1k \\ 
 CnvNxt-T-21k & IN-21k & 84.1 & 29M & convnext{\textunderscore}tiny{\textunderscore}384{\textunderscore}in22ft1k \\ 
 BiT-m & IN-21k & 82.3 & 45M & resnetv2{\textunderscore}101x1{\textunderscore}bitm \\ 
 EffNetv2-M-21k & IN-21k & 85.6 & 54M & tf{\textunderscore}efficientnetv2{\textunderscore}m{\textunderscore}in21ft1k \\ 
 EffNetb7-ns & JFT - noisy student & 86.8 & 66M & tf{\textunderscore}efficientnet{\textunderscore}b7{\textunderscore}ns \\ 
 ViT-B-384 & --- & 81.1 & 87M & vit{\textunderscore}base{\textunderscore}patch16{\textunderscore}384.augreg{\textunderscore}in1k \\ 
 Swinv2-B-256 & --- & 84.6 & 88M & swinv2{\textunderscore}base{\textunderscore}window16{\textunderscore}256 \\ 
 Deit3-B-384 & --- & 85.1 & 87M & deit3{\textunderscore}base{\textunderscore}patch16{\textunderscore}384 \\ 
 Deit3-B-224 & --- & 83.8 & 87M & deit3{\textunderscore}base{\textunderscore}patch16{\textunderscore}224 \\ 
 XCiT-M-224 & --- & 82.6 & 84M & xcit{\textunderscore}medium{\textunderscore}24{\textunderscore}p16{\textunderscore}224 \\ 
 XCiT-M-224-d & --- & 84.3 & 84M & xcit{\textunderscore}medium{\textunderscore}24{\textunderscore}p16{\textunderscore}224{\textunderscore}dist \\ 
 CnvNxt-B & --- & 84.4 & 89M & convnext{\textunderscore}base \\ 
 BiT-s & --- & 78.0 & 45M & resnetv2{\textunderscore}101x1{\textunderscore}bitm \\ 
 EffNetv2-M & --- & 85.0 & 54M & tf{\textunderscore}efficientnetv2{\textunderscore}m \\ 
 EffNetb7 & --- & 84.9 & 66M & tf{\textunderscore}efficientnet{\textunderscore}b7 \\ 
 EffNet-B0 & --- & 77.7 & 5M & efficientnet{\textunderscore}b0 \\ 
 ResNet50 & --- & 80.4 & 26M & resnet50 \\ 
 \hline 
CLIP-ViT-B16 & openai & 66.6 & 150M & --- \\ 
 CLIP-ViT-B16 & openai & 74.2 & 428M & --- \\ 
 \end{tabular}
\end{center}

\vskip \belowtablevskip
\end{table}
\FloatBarrier

\clearpage
\section{Methods}\label{sec:methods}
Here we give an overview over the evaluated OOD detection methods.
For clarity, %
we denote vectors in bold and lowercase letters and matrices in bold an uppercase letters. We write neural networks as functions $n$, which are parametrized by weights $\mathbf{\theta}$, take an input sample $\mathbf{x}$ and produce an output vector $\mathbf{o}$ of size $C$, where $C$ is typically the number of classes in a classification task (1000 in the case of IN-1K). We refer to $\mathbf{o}$ as the logits of $\mathbf{x}$, which can be transformed to a probability vector $\mathbf{p}$ (also of size $C$) via the softmax function: $p_{i}=\exp{(o_{i})/\sum_{c}\exp(o_{c})}$. The network $n$ can be decomposed into a feature extractor $h$ and the networks last layer $g$: \begin{equation*}
\mathbf{o}=n(\mathbf{x})=g(h (\mathbf{x})),
\end{equation*}
where $g$ is a fully connected, linear layer, i.e. $g(\mathbf{h})=\mathbf{W}^{T}\mathbf{h}+\mathbf{b}$ with weight $\mathbf{W}$ and bias $\mathbf{b}$. We refer to $\mathbf{h}=h(\mathbf{x})$ as the \textit{features} or the \textit{embeddings} of $\mathbf{x}$ w.r.t. the network $n$.
As presented in Section~\ref{sec:eval}, for each sample $\mathbf{x}$, a method returns an OOD-score $s=f(\mathbf{x})$, a scalar value which is supposed to be larger for ID data and smaller for OOD data.
Methods accessing $h(\bm{x})$ directly in order to compute the OOD-score are referred to as feature-based methods, in contrast to methods that derive their OOD-score from the logits $\bm{o}$ (even though obviously the logits implicitly also depend on these features). In the following, we will describe how each method computes the score $s$ for a test input $\mathbf{x}$.

\textbf{MSP} \cite{hendrycks2017MSP}: The most popular OOD-detection baseline uses the confidence, i.e. the max softmax probability of a models probability output vector:
\begin{equation*}
s=\max_{c}({p}_{c})
\end{equation*}
\newline
\textbf{Max-Logit} \cite{hendrycks22Scaling}: Similar to MSP, Max-Logit returns the largest entry of the logit-vector $\mathbf{o}$, i.e.  
\begin{equation*}
s=\max_{c}({o}_{c})
\end{equation*}\newline
\textbf{Energy} \cite{liu2020energy}:
The Energy based OOD detection method uses the denominator of the softmax-function as OOD-score:
\begin{equation*}
s=\log\sum_{c}^{C}\exp{(o_{c})}
\end{equation*}
\textbf{KL-Matching} \cite{hendrycks22Scaling}: KL-Matching computes a mean probability vector $\mathbf{d}_c$ for each of the $C$ classes. For a test input, the KL-distances of all $\mathbf{d}_c$ vectors to its probability vector $\mathbf{p}$ are computed, and the OOD-score is the negative of the smallest of those distances:
\begin{equation*}
s=-\min_{c}\text{KL}[\mathbf{p}||\mathbf{d}_{c}]
\end{equation*}
In the original paper by \cite{hendrycks22Scaling}, the average for $\mathbf{d}_{c}$ is computed over an additional validation set. Since none of the other methods leverages extra data and we are interested in fair comparison, we deploy KL-Matching like in \cite{wang2022vim,yang2022openood}, where the average is computed over the train set.

\textbf{KNN} \cite{sun2022knnood}: KNN is a non-parametric method that computes distances in the feature-space. Specifically, the feature vector of a test input is normalized to $\mathbf{z}=\mathbf{h}/||\mathbf{h}||_{2}$ and the pairwise distances $r_{i}(\mathbf{z})=||\mathbf{z}-\mathbf{z}_{i}||_{2}$ to the normalized features $\mathcal{Z}=\{\mathbf{z}_{1},...,\mathbf{z}_{N}\}$ of all samples of the training set are computed.
The distances $r_{i}(\mathbf{z})$ are then sorted according to their magnitude and the $K^{\text{th}}$ smallest distance, denoted $r^{K}(\mathbf{z})$ is used as negative OOD-score:
\begin{equation*}
s=-r^{K}(\mathbf{z})
\end{equation*}
Like suggested in \cite{sun2022knnood}, we use $K=1000$.\newline
\textbf{Mahalanobis distance} \cite{LeeMahalanobis2018}: This popular method fits a class-conditional Gaussian with shared covariance matrix to the train set, i.e. computes
\begin{equation*}
\hat{\mu}_{c}=\frac{1}{N_{c}}\sum_{i:y_{i}=c}\mathbf{h}_{i},\hspace{15pt} \hat{\mathbf{\Sigma}}=\frac{1}{N}\sum_{c}\sum_{i:y_{i}=c}(\mathbf{h}_{i}-\hat{\mu}_{c})(\mathbf{h}_{i}-\hat{\mu}_{c})^{T}
\end{equation*}
where $N_{c}$ is the number of train samples in class $c$ and $N$ is the total number of train samples. The OOD-score of a test sample is then the Mahalanobis distance induced by $\hat{\mathbf{\Sigma}}$ between its feature $\mathbf{h}$ and the closest class mean:
\begin{equation*}
s=-\min_{c}(\mathbf{h}-\hat{\mu}_{c})\hat{\mathbf{\Sigma}}^{-1}(\mathbf{h}-\hat{\mu}_{c})^{T}
\end{equation*}

\textbf{Relative Mahalanobis distance} \cite{RenRelMaha2021}: A modification of the Mahalanobis distance method, thought to improve near-OOD detection, is to additionally fit a global Gaussian distribution to the train set without taking class-information into account: 
\begin{equation*}
\hat{\mu}_{\text{global}}=\frac{1}{N}\sum_{i}\mathbf{h}_{i},\hspace{15pt} \hat{\mathbf{\Sigma}}_{\text{global}}=\frac{1}{N}\sum_{i}(\mathbf{h}_{i}-\hat{\mu}_{\text{global}})(\mathbf{h}_{i}-\hat{\mu}_{\text{global}})^{T}
\end{equation*}
The OOD-score is then defined as the difference between the original Mahalanobis distance and the Mahalanobis distance w.r.t. the global Gaussian distribution:
\begin{equation*}
s=-\min_{c}\left( (\mathbf{h}-\hat{\mu}_{c})\hat{\mathbf{\Sigma}}^{-1}(\mathbf{h}-\hat{\mu}_{c})^{T}-(\mathbf{h}-\hat{\mu}_{\text{global}})\hat{\mathbf{\Sigma}}_{\text{global}}^{-1}(\mathbf{h}-\hat{\mu}_{\text{global}})^{T}\right)
\end{equation*}

\textbf{ReAct} \cite{sun2021react}: The authors propose to perform a truncation of the feature vector, $\mathbf{\Bar{h}}=\min(\mathbf{h},r)$, where the $\min$ operation is to be understood element-wise and $r$ is the truncation threshold.  The truncated features can then be converted to so-called rectified logits via $\mathbf{\Bar{o}}=g(\mathbf{\Bar{h}})=\mathbf{W}^{T}\mathbf{\Bar{h}}+\mathbf{b}$. While the rectified logits can now be used with a variety of existing detection methods, we follow \cite{sun2021react} and use the rectified Energy as OOD-score:
\begin{equation*}
s=\log\sum_{c}^{C}\exp{(\Bar{o}_{c})}
\end{equation*}
As suggested in \cite{wang2022vim}, we set the threshold $r$ such that 1\% of the activations from the train set would be truncated.\newline

\textbf{Virtual Logit Matching} \cite{wang2022vim}: The idea behind ViM is that meaningful features are thought to lie in a low-dimensional manifold, called the principal space $P$, whereas features from OOD-samples should also lie in $P^{\perp}$, the space orthogonal to $P$. $P$ is the $D$-dimensional subspace spanned by the eigenvectors with the largest $D$ eigenvalues of the matrix $\mathbf{F}^{T}\mathbf{F}$, where $\mathbf{F}$ is the matrix of all train features offsetted by $\mathbf{u}=-(\mathbf{W^{T}})^{+}\mathbf{b}$ ($+$ denotes the Moore-Penrose inverse). A sample with feature vector $\mathbf{h}$ is then also offset to $\mathbf{\Tilde{h}}=\mathbf{h}-\mathbf{u}$ and can be decomposed into $\mathbf{\Tilde{h}}=\mathbf{\Tilde{h}}^{P}+\mathbf{\Tilde{h}}^{P^{\perp}}$, and $\mathbf{\Tilde{h}}^{P^{\perp}}$ is referred to as the \textit{Residual} of $\mathbf{h}$. ViM leverages the Residual and converts it to a virtual logit $o_{0}=\alpha||\mathbf{\Tilde{h}}^{P^{\perp}}||_{2}$, where 
\begin{equation*}
\alpha=\frac{\sum_{i=1}^{N}\max_{c}o_{i}^{c}}{\sum_{i=1}^{N}||\mathbf{h}^{P^{\perp}}_{i}||_{2}}
\end{equation*}
is designed to match the scale of the virtual logit to the scale of the real train logits. The virtual logit is then appended to the original logits of the test sample, i.e. to $\mathbf{o}$, and a new probability vector is computed via the softmax function. The probability corresponding to the virtual logit is then the final OOD-score:
\begin{equation*}\label{vim}
s=-\frac{\exp{(o_0)}}{\sum_{c=1}^{C}\exp{(o_{c})}+\exp{(o_0)}}
\end{equation*}
Like suggested in \cite{wang2022vim}, we use $D=1000$ if the dimensionality of the feature space $d$ is $d\geq2048$, $D=512$ if $2048\geq d\geq 768$, and $D=d/2$ rounded to integers otherwise. 

\textbf{Cosine} \cite{tack2020csi,anonymous2023COOD}: This method computes the maximum cosine-similarity between the features of a test-sample and embedding vectors $\Tilde{\mathbf{u}}_{c}$ (sometimes also called concept-vector):
\begin{equation}
    s=\max_{c}\Tilde{\mathbf{u}}_{c}^{T}\mathbf{h}/||\Tilde{\mathbf{u}}_{c}^{T}||_{2}
\end{equation}
For zero-shot CLIP, $\Tilde{\mathbf{u}}_{c}$ can be obtained by creating text-embeddings from the ImageNet class names. Encoding \textit{'A photo of a ...'} yields an embedding from the corresponding class. For classifiers, we use the class-wise train means $\hat{\mu}_c$, that are also used for Mahalanobis distance. 

\textbf{MCM/RCos} \cite{ming2022delving,techapanurak2020hyperparameter}: Maximum-Concept-Matching was recently introduced as a zero-shot OOD detection method for CLIP and applies additional softmax-scaling to the cosine-similarities of the \textit{Cosine} method, potentially with a temperature scaling (which we omit, following \cite{ming2022delving}). Again, we extend this method to work with conventional classifiers by using the class-means $\hat{\mu}_c$ like they are used for Mahalanobis distance as embedding/concept vectors. We then refer to it as relative cosine (short: MCM/RCos or just RCos) in order to distinguish it from CLIPs zero-shot method.

\FloatBarrier

\section{Definitions of OOD detection metrics}\label{sec:fpr_def}
The performance of OOD detectors is commonly reported in terms of the \textit{false positive rate at a fixed true positive rate} Q, denoted as \textbf{\FPRQ}, short \textbf{\FPR}.%
This means that the detector is interpreted as %
making the decision to \textit{accept} an unknown input $x$ \textit{if} $S(x) \geq \tau$, for a threshold $\tau$ that is chosen such that Q\% of ID inputs are accepted, and \textit{rejecting} the input as OOD \textit{if} $S(x) < \tau$.
The \FPRQ{} counts the fraction of falsely accepted OOD inputs under this decision scheme.
This means the \textit{lower} the \FPRQ{}, the \textit{better} the OOD detection performance.
In the OOD detection literature, the most commonly used value for Q is 95\%, which we too use throughout this paper.
We also report results in terms of the mean \textit{area under the receiver-operator characteristic curve}, short \textbf{AUROC} in Table~\ref{tab:overview-auroc-mean}.
It represents the probability that an ID input receives a higher score (equal scores counted half) than an OOD input when both are drawn randomly from their respective evaluation datasets~\cite{bitterwolf2022breaking}.
Like for the \FPR, the mean AUROC corresponds to first uniformly drawing an OOD class and then drawing a sample from that class.

\section{Illustrative examples from the cleaning process}\label{app:cleaning-process}
\def \imgwidth {1.05cm}
\def \ximgdist {.1cm}
\def \imgheight {1.05cm}
\def \yline {2.03*\imgheight}
\def \yimgdist {.16cm}
\def \ydsetname {.85cm}
\def \toplabeldist {.08cm}
\def \botlabeldist {.2cm}
\def \tinyscale {.83}
\def \underlayscale {.538}
\def \xoffset {2.35cm} 

\tikzset{
    pics/imgsquare/.style args={#1/#2/#3/#4/#5/#6/#7/#8}{
      code = {\coordinate (o) at (0,0);
      \coordinate (i1) at ($(o) + (-.5*\imgwidth,0) + (-.5*\ximgdist,0)$);
      \coordinate (i2) at ($(o) + (.5*\imgwidth,0) + (.5*\ximgdist,0)$);
      \coordinate (i3) at ($(o) + (-.5*\imgwidth,-\imgheight) + (-.5*\ximgdist,-\yimgdist)$);
      \coordinate (i4) at ($(o) + (.5*\imgwidth,-\imgheight) + (.5*\ximgdist,-\yimgdist)$);
      \coordinate (t1) at ($(i1) + (0,.5*\imgheight) + (0,\toplabeldist)$);
      \coordinate (t2) at ($(i2) + (0,.5*\imgheight) + (0,\toplabeldist)$);
      \coordinate (t3) at ($(i3) + (0,-.5*\imgheight) + (0,-\botlabeldist)$);
      \coordinate (t4) at ($(i4) + (0,-.5*\imgheight) + (0,-\botlabeldist)$);
  \fill[fill=INunderlay, line width=0.mm] ($(i1) + (-\underlayscale*\imgwidth,\underlayscale*\imgheight)$) rectangle ($(i1) + (\underlayscale*\imgwidth,-\underlayscale
  *\imgheight)$) {};
  \fill[fill=INunderlay, line width=0.mm] ($(i2) + (-\underlayscale*\imgwidth,\underlayscale*\imgheight)$) rectangle ($(i2) + (\underlayscale*\imgwidth,-\underlayscale
  *\imgheight)$) {};  
  \fill[fill=OODunderlay, line width=0.mm] ($(i3) + (-\underlayscale*\imgwidth,\underlayscale*\imgheight)$) rectangle ($(i3) + (\underlayscale*\imgwidth,-\underlayscale
  *\imgheight)$) {};
  \fill[fill=OODunderlay, line width=0.mm] ($(i4) + (-\underlayscale*\imgwidth,\underlayscale*\imgheight)$) rectangle ($(i4) + (\underlayscale*\imgwidth,-\underlayscale
  *\imgheight)$) {};
  \node[inner sep=0pt] at (i1) {\includegraphics[width=\imgwidth, height=\imgheight]{#1}};
  \node[inner sep=0pt] at (i2) {\includegraphics[width=\imgwidth, height=\imgheight]{#2}};
  \node[inner sep=0pt] at (i3) {\includegraphics[width=\imgwidth, height=\imgheight]{#3}};
  \node[inner sep=0pt] at (i4) {\includegraphics[width=\imgwidth, height=\imgheight]{#4}};
  \draw[anchor=base, align=center] (t1)  node[scale=\tinyscale] {\tiny \textcolor{INunderlay}\xmark{}\,\INclass{(#5)}};;
  \draw[anchor=base, align=center] (t2)  node[scale=\tinyscale] {\tiny \textcolor{INunderlay}\xmark{}\,\INclass{(#6)}};
  \draw[anchor=base, align=center] (t3)  node[scale=\tinyscale] {\tiny \textcolor{OODunderlay}\cmark{}};
  \draw[anchor=base, align=center] (t4)  node[scale=\tinyscale] {\tiny \textcolor{OODunderlay}\cmark{}};
  \draw[anchor=base, align=center] ($(o) + (0,\ydsetname)$)  node {\tiny \OODclass{#7}};
  }}}

\newcommand{\hamIDa}{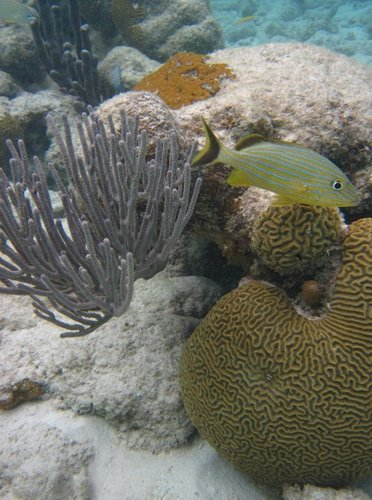}
\newcommand{\hamIDb}{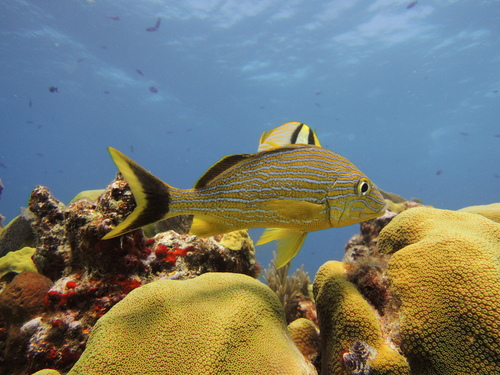}
\newcommand{\hamcleana}{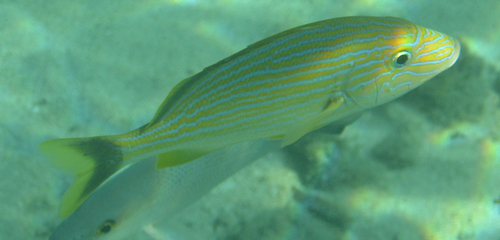}
\newcommand{\hamcleanb}{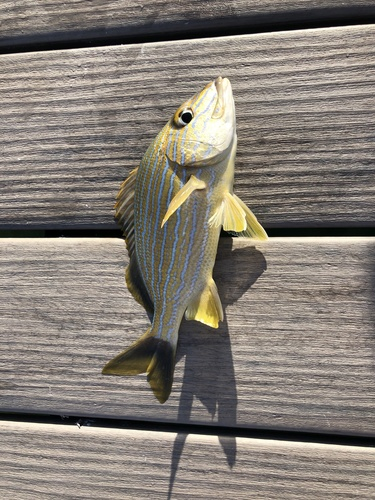}
\newcommand{\hamcleanc}{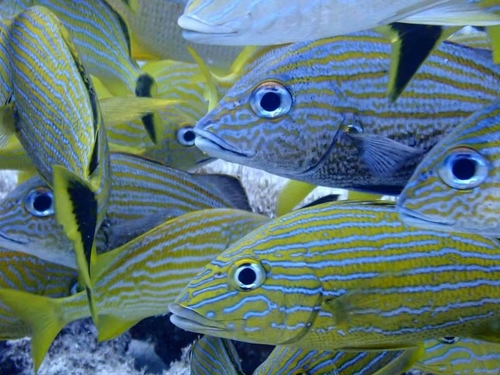}

\newcommand{\araIDa}{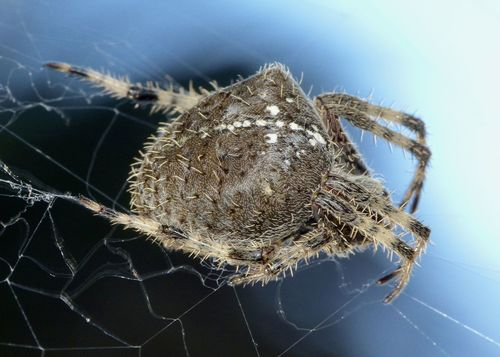}
\newcommand{\araIDb}{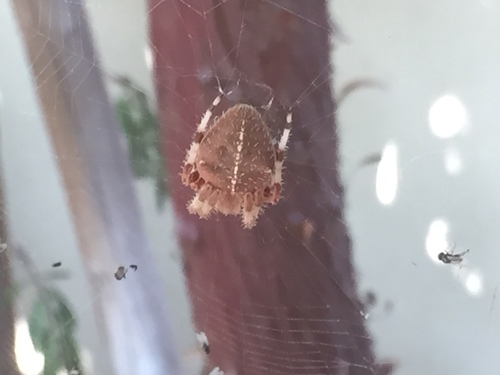}
\newcommand{\araIDc}{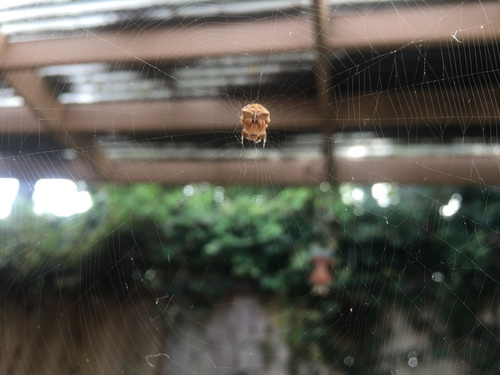}
\newcommand{\araIDd}{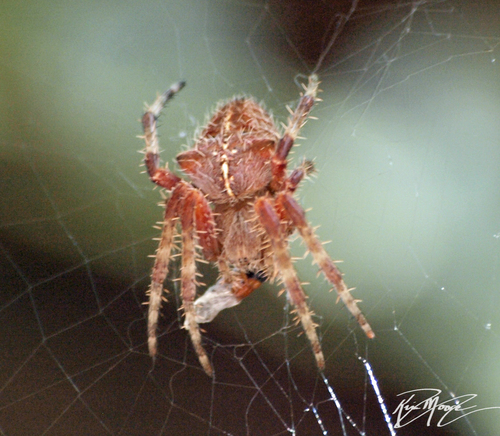}
\newcommand{\araca}{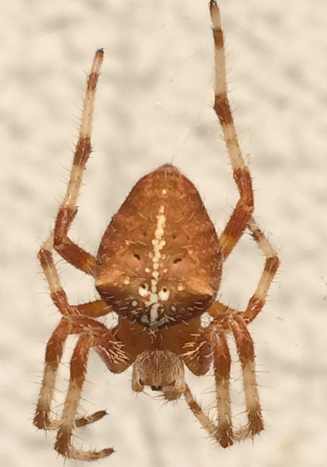}
\newcommand{\aracb}{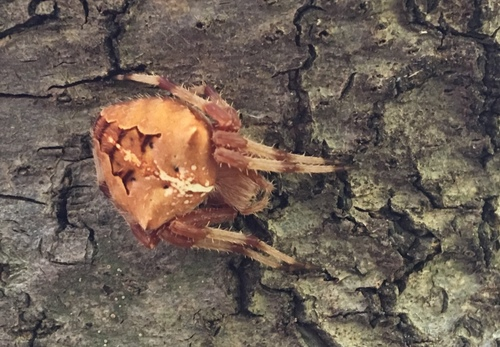}

\newcommand{\skyIDa}{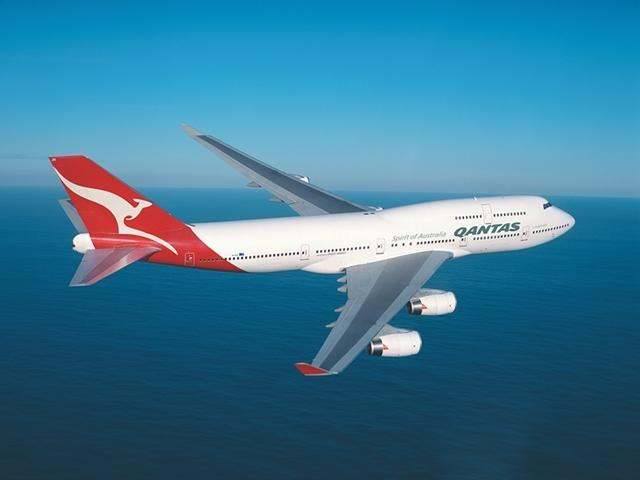}
\newcommand{\skyIDb}{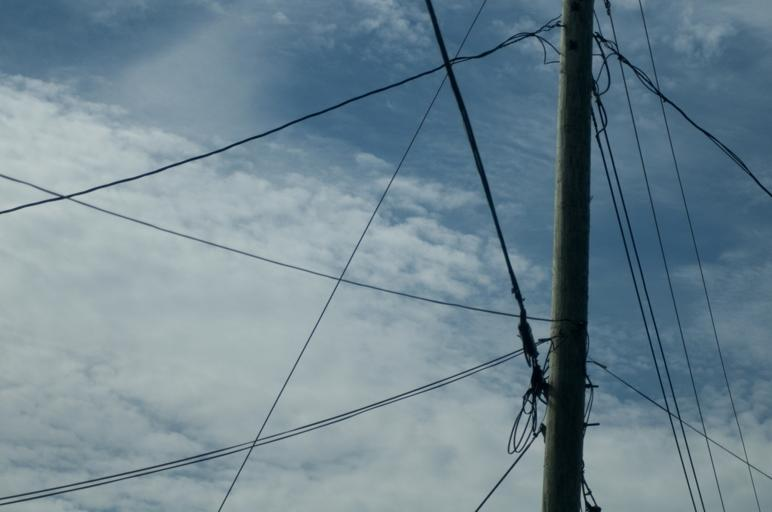}
\newcommand{\skycleana}{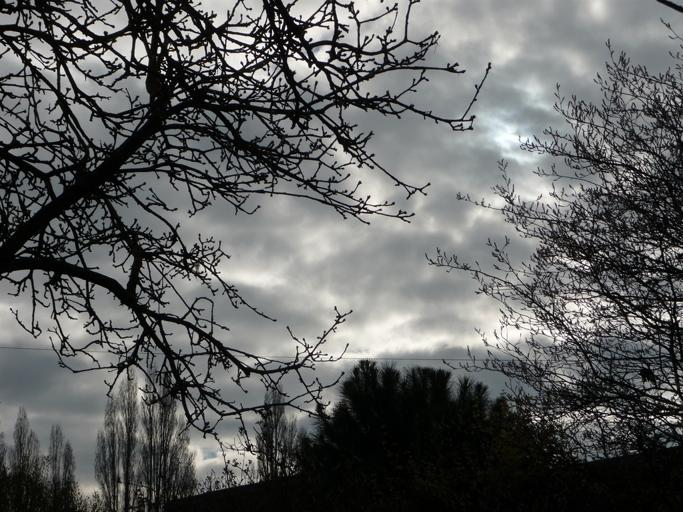}
\newcommand{\skycleanb}{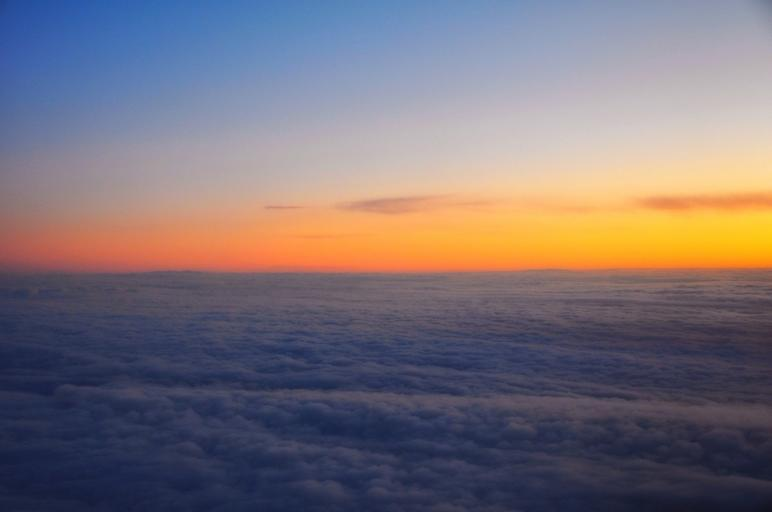}

\newcommand{\waffIDa}{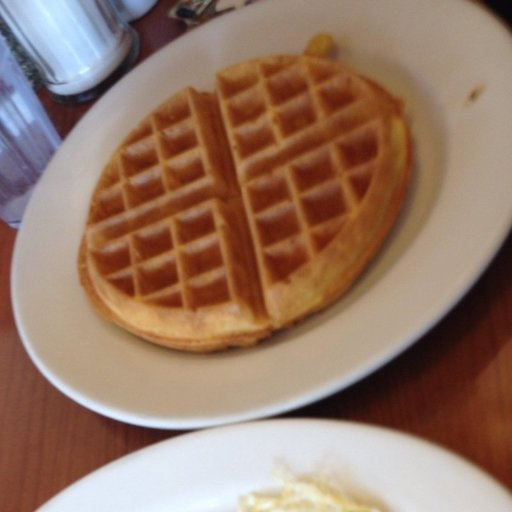}
\newcommand{\waffIDb}{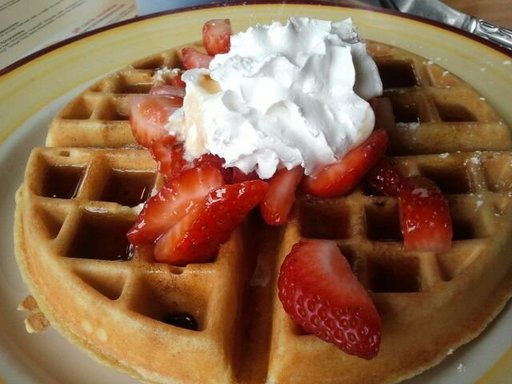}
\newcommand{\waffcleana}{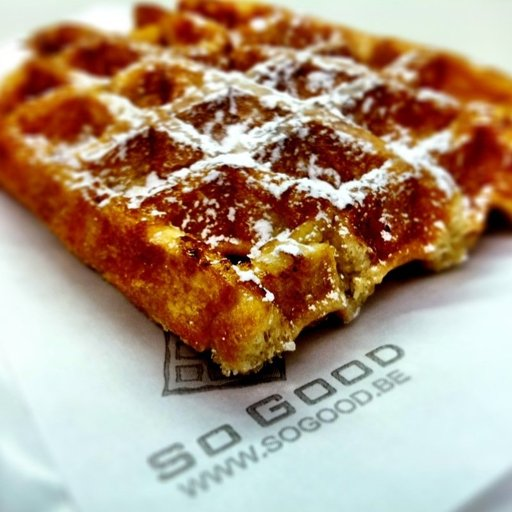}
\newcommand{\waffcleanb}{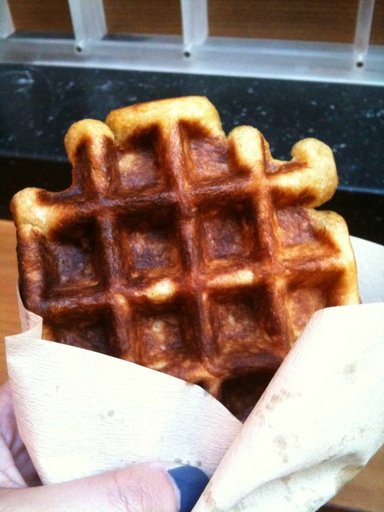}

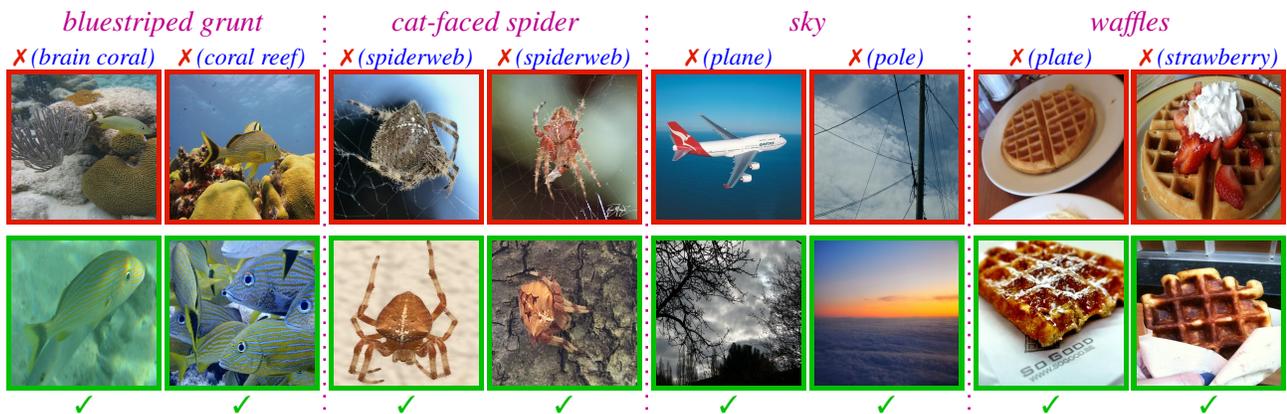
\begin{figure*}[h]
\centering
\pgfmathsetseed{5}
\hspace*{-4mm}
\resizebox{1.02\textwidth}{!}{
\noindent
\begin{tikzpicture}
  \coordinate (ood1) at (0,0);
  \coordinate (ood2) at (\xoffset,0);
  \coordinate (ood3) at (2.*\xoffset,0);
  \coordinate (ood4) at (3.*\xoffset,0);
  \pic [local bounding box=A1a] at (ood1) {imgsquare=\hamIDa/\hamIDb/\hamcleana/\hamcleanc/brain coral/coral reef/bluestriped grunt/bluestriped grunt};
  \pic [local bounding box=A1a] at (ood2) {imgsquare=\araIDa/\araIDd/\araca/\aracb/spiderweb/spiderweb/cat-faced spider/cat-faced spider};
  \pic [local bounding box=A1a] at (ood3) {imgsquare=\skyIDa/\skyIDb/\skycleana/\skycleanb/plane/pole/sky/sky};
  \pic [local bounding box=A1a] at (ood4) {imgsquare=\waffIDa/\waffIDb/\waffcleana/\waffcleanb/plate/strawberry/waffles/waffles};
  \draw[dotted, color=OODclass, line width=0.19mm] ($(ood1) + (1.*\imgwidth+\ximgdist+.027cm,.96cm)$) -- ($(ood1) + (1.*\imgwidth+\ximgdist+.027cm,-2.0cm)$) {};
  \draw[dotted, color=OODclass, line width=0.19mm] ($(ood2) + (1.*\imgwidth+\ximgdist+.027cm,.96cm)$) -- ($(ood2) + (1.*\imgwidth+\ximgdist+.027cm,-2.0cm)$) {};
  \draw[dotted, color=OODclass, line width=0.19mm] ($(ood3) + (1.*\imgwidth+\ximgdist+.027cm,.96cm)$) -- ($(ood3) + (1.*\imgwidth+\ximgdist+.027cm,-2.0cm)$) {};
\end{tikzpicture}
}
\vskip -3mm
\caption{\label{fig:cleanded-dataset}\textbf{Cleaning the OOD classes.} \textbf{Top:} Samples that were excluded due to overlap with ID classes.
\textbf{Bottom:} Samples from the same OOD class that were included in the cleaned datasets.
}
\end{figure*}

\FloatBarrier

\newpage

\section{Details of the \dsetname{} dataset.}
\label{sec:dset_details}
\newlength{\mytabcolsep}
\setlength{\mytabcolsep}{\tabcolsep}
\tabcolsep=0.38cm
\renewcommand{\arraystretch}{1.}
\begin{table}[!h]
\centering
\caption{Detailed information for each OOD class. For determining overlap with classes of IN-21K, we checked the 8 most common predictions of a ViT classifier for IN-21K on the \dsetname{} OOD class.}
\label{tab:OOD_class_info}
\begin{center}
\smaller
\rowcolors{2}{white}{gray!25}
\resizebox{0.9\textwidth}{!}{%
\begin{tabular}{l l r c l}
OOD class name & shortname & \# samples & source dataset & 
ImageNet-21K overlap
\\ 
\hline 
\OODclass{AFA (cyanobacterium)}
  &  \OODclass{AFA}
  &  46
  &  \textsc{Species}
  &  \INTOKclass{microorganism}
\\
\OODclass{bagpipe}
  &  \OODclass{Bagp}
  &  97
  &  Imagenet-21k
  &  \INTOKclass{bagpipe}
\\
\OODclass{bluestriped grunt}
  &  \OODclass{BSGrunt}
  &  96
  &  \textsc{Species}
  &  \INTOKclass{grunt}
\\
\OODclass{cable}
  &  \OODclass{Cabl}
  &  88
  &  scraped
  &  \INTOKclass{cable television}
\\
\OODclass{California pitcher plant}
  &  \OODclass{CPitch}
  &  100
  &  \textsc{Species}
  &  \INTOKclass{pitcher plant}
\\
\OODclass{California slender salamander}
  &  \OODclass{CSSala}
  &  100
  &  \textsc{Species}
  &  \INTOKclass{slender salamander}
\\
\OODclass{California two-spot octopus}
  &  \OODclass{C2SOct}
  &  100
  &  \textsc{Species}
  &  \INTOKclass{octopus}
\\
\OODclass{caracal}
  &  \OODclass{Caracal}
  &  100
  &  iNat. Download
  &  \INTOKclass{caracal}
\\
\OODclass{cat-faced spider}
  &  \OODclass{CatFSp}
  &  100
  &  \textsc{Species}
  &  \textit{unclear/very broad class}
\\
\OODclass{Central American tapir}
  &  \OODclass{CATapir}
  &  100
  &  \textsc{Species}
  &  \INTOKclass{tapir}
\\
\OODclass{chicken quesadilla}
  &  \OODclass{CQuesa}
  &  100
  &  \textsc{Food-101}
  &  \textbf{-}
\\
\OODclass{common cuttlefish}
  &  \OODclass{CCuttle}
  &  100
  &  \textsc{Species}
  &  \INTOKclass{cuttlefish}
\\
\OODclass{crème brûlée}
  &  \OODclass{CBrûlée}
  &  99
  &  \textsc{Food-101}
  &  \INTOKclass{creme brulee}
\\
\OODclass{cupcakes}
  &  \OODclass{CCake}
  &  80
  &  \textsc{Food-101}
  &  \textbf{-}
\\
\OODclass{donuts}
  &  \OODclass{Donuts}
  &  100
  &  \textsc{Food-101}
  &  \INTOKclass{doughnut}
\\
\OODclass{door}
  &  \OODclass{Door}
  &  100
  &  MyNursingHome
  &  \INTOKclass{interior door}
\\
\OODclass{dreamfish}
  &  \OODclass{Dreamf}
  &  100
  &  \textsc{Species}
  &  \INTOKclass{sea bream}
\\
\OODclass{dune thistle}
  &  \OODclass{DThist}
  &  100
  &  \textsc{Species}
  &  \INTOKclass{creme brulee}
\\
\OODclass{dusky flathead (fish)}
  &  \OODclass{DFlath}
  &  100
  &  \textsc{Species}
  &  \INTOKclass{flathead}
\\
\OODclass{E. micromeris (cactus)}
  &  \OODclass{EMicro}
  &  100
  &  \textsc{Species}
  &  \textbf{-}
\\
\OODclass{Eastern leaf-footed bug}
  &  \OODclass{ELFBug}
  &  100
  &  \textsc{Species}
  &  \INTOKclass{leaf-footed bug}
\\
\OODclass{European paper wasp}
  &  \OODclass{EPWasp}
  &  100
  &  \textsc{Species}
  &  \INTOKclass{paper wasp}
\\
\OODclass{false killer whale}
  &  \OODclass{FalseKW}
  &  67
  &  \textsc{Species}
  &  \textit{unclear/very broad class}
\\
\OODclass{field road}
  &  \OODclass{FieldRd}
  &  96
  &  \textsc{Places}
  &  \INTOKclass{byway}
\\
\OODclass{fire extinguisher}
  &  \OODclass{FireEx}
  &  106
  &  MyNursingHome
  &  \INTOKclass{fire extinguisher}
\\
\OODclass{fireworks}
  &  \OODclass{FireW}
  &  100
  &  scraped
  &  \textbf{-}
\\
\OODclass{forest path}
  &  \OODclass{ForPth}
  &  100
  &  \textsc{Places}
  &  \textit{unclear/very broad class}
\\
\OODclass{Franciscan wallflower}
  &  \OODclass{Franci}
  &  100
  &  \textsc{Species}
  &  \INTOKclass{wallflower}
\\
\OODclass{French fries}
  &  \OODclass{Fries}
  &  100
  &  \textsc{Food-101}
  &  \INTOKclass{french fries}
\\
\OODclass{Galápagos fur seal}
  &  \OODclass{GFurS}
  &  91
  &  \textsc{Species}
  &  \INTOKclass{arcella}
\\
\OODclass{giant cuttlefish}
  &  \OODclass{GCuttle}
  &  99
  &  \textsc{Species}
  &  \INTOKclass{cuttlefish}
\\
\OODclass{glass of milk}
  &  \OODclass{GlMilk}
  &  89
  &  scraped
  &  \INTOKclass{milk}
\\
\OODclass{gramophone}
  &  \OODclass{Gramo}
  &  56
  &  scraped
  &  \INTOKclass{gramophone}
\\
\OODclass{high heels}
  &  \OODclass{HHeels}
  &  99
  &  scraped
  &  \textbf{-}
\\
\OODclass{Hindu temple}
  &  \OODclass{HinTp}
  &  51
  &  scraped
  &  \textit{unclear/very broad class}
\\
\OODclass{Horse Hoof clam}
  &  \OODclass{HHClam}
  &  31
  &  \textsc{Species}
  &  \INTOKclass{seashell}
\\
\OODclass{Indo-Pacific bottlenose dolphin}
  &  \OODclass{IPBNDol}
  &  100
  &  \textsc{Species}
  &  \INTOKclass{dolphin}
\\
\OODclass{long-tailed silverfish}
  &  \OODclass{LTSilF}
  &  100
  &  \textsc{Species}
  &  \INTOKclass{silverfish}
\\
\OODclass{Lumholtz's tree-kangaroo}
  &  \OODclass{LTRoo}
  &  100
  &  \textsc{Species}
  &  \INTOKclass{tree wallaby}
\\
\OODclass{M. wesenbergii (cyanobacterium)}
  &  \OODclass{MWesen}
  &  33
  &  \textsc{Species}
  &  \INTOKclass{microorganism}
\\
\OODclass{marbled newt}
  &  \OODclass{MNewt}
  &  100
  &  \textsc{Species}
  &  \INTOKclass{newt}
\\
\OODclass{mbira}
  &  \OODclass{Mbira}
  &  67
  &  scraped
  &  \textbf{-}
\\
\OODclass{Mexican lime cactus}
  &  \OODclass{MLCact}
  &  100
  &  \textsc{Species}
  &  \INTOKclass{barrel cactus}
\\
\OODclass{Pampas deer}
  &  \OODclass{PDeer}
  &  82
  &  \textsc{Species}
  &  \INTOKclass{buck}
\\
\OODclass{pyramid}
  &  \OODclass{Pyra}
  &  100
  &  caltech-101
  &  \INTOKclass{Cheops}
\\
\OODclass{redbreast sunfish}
  &  \OODclass{RBSunf}
  &  100
  &  \textsc{Species}
  &  \INTOKclass{sunfish}
\\
\OODclass{rosybells (flowering plant)}
  &  \OODclass{Rosyb}
  &  100
  &  \textsc{Species}
  &  \textbf{-}
\\
\OODclass{ruby octopus}
  &  \OODclass{RubyOct}
  &  100
  &  \textsc{Species}
  &  \INTOKclass{octopus}
\\
\OODclass{scissors}
  &  \OODclass{Sciss}
  &  100
  &  caltech-101
  &  \INTOKclass{scissors}
\\
\OODclass{shuttlecock}
  &  \OODclass{ShCo}
  &  67
  &  scraped
  &  \INTOKclass{shuttlecock}
\\
\OODclass{silver-haired bat}
  &  \OODclass{SilverHB}
  &  99
  &  \textsc{Species}
  &  \INTOKclass{bat}
\\
\OODclass{skipper caterpillar}
  &  \OODclass{SCaterp}
  &  100
  &  iNat. Download
  &  \INTOKclass{caterpillar}
\\
\OODclass{sky}
  &  \OODclass{Sky}
  &  68
  &  \textsc{Places}
  &  \INTOKclass{sky}
\\
\OODclass{southern calamari}
  &  \OODclass{SCalam}
  &  99
  &  \textsc{Species}
  &  \INTOKclass{squid}
\\
\OODclass{spaghetti bolognese}
  &  \OODclass{SBolo}
  &  67
  &  \textsc{Food-101}
  &  \INTOKclass{spaghetti}
\\
\OODclass{stapler}
  &  \OODclass{Stapl}
  &  100
  &  caltech-101
  &  \INTOKclass{stapler}
\\
\OODclass{sweet pea}
  &  \OODclass{SwPea}
  &  100
  &  \textsc{Species}
  &  \textit{unclear/very broad class}
\\
\OODclass{two-toed amphiuma (salamander)}
  &  \OODclass{2TAmph}
  &  176
  &  \textsc{Species}
  &  \INTOKclass{amphiuma}
\\
\OODclass{waffles}
  &  \OODclass{Waffle}
  &  61
  &  \textsc{Food-101}
  &  \textbf{-}
\\
\OODclass{walker}
  &  \OODclass{Walker}
  &  99
  &  MyNursingHome
  &  \INTOKclass{walker}
\\
\OODclass{water dispenser (jugless)}
  &  \OODclass{WDisp}
  &  100
  &  MyNursingHome
  &  \INTOKclass{water cooler}
\\
\OODclass{Windsor chair}
  &  \OODclass{WiChair}
  &  71
  &  caltech-101
  &  \INTOKclass{Windsor chair}
\\
\OODclass{yellow trumpets}
  &  \OODclass{YTrump}
  &  100
  &  \textsc{Species}
  &  \INTOKclass{yellow trumpet}
\\
\OODclass{ʻōhelo ʻai (flowering plant)}
  &  \OODclass{ʻōʻai}
  &  100
  &  \textsc{Species}
  &  \textbf{-}
\\
 \end{tabular}
 }
\end{center}
\vskip \belowtablevskip
\end{table}

\setlength{\tabcolsep}{\mytabcolsep}
\clearpage
\FloatBarrier
\newpage

\section{Details and recipes for OOD unit-tests}\label{sec:OOD_unit_tests}
We provide 400 samples for each of 17 OOD unit-tests, mirroring the sizes and file formats of random ImageNet samples.
Their reproducible definitions are given as follows:
\begin{itemize}
    \item \textbf{uniform noise~\cite{hendrycks2017MSP}:}
        Each RBG colour channel of each pixel is independently sampled uniformly between $0.0$ or $1.0$.
    \item \textbf{Gaussian noise~\cite{hendrycks2017MSP}:}
        For each image, first $\sigma$ is chosen randomly between $(0.05, 0.075, 0.1, 0.15, 0.2, 0.3, 0.5)$.\\
        Then each RBG colour channel of each pixel is independently sampled from $\mathcal{N}(0.5, \sigma)$.
    \item \textbf{Rademacher noise~\cite{HenMazDie2019}:}
        Then each RBG colour channel of each pixel is independently set to $0.0$ or $1.0$ with 50\% probability.
    \item \textbf{IN pixel permutations~\cite{HeiAndBit2019}:}
        We choose a random IN-1K validation image and randomly shuffle its pixels (no remixing of colours).
    \item \textbf{black:}
        All colour channels are set to 0.0.
    \item \textbf{white:}
        All colour channels are set to 1.0.
    \item \textbf{shades of grey:}
        All colour channels are set to the same value, sampled uniformly between $0.0$ or $1.0$.
    \item \textbf{monochrome:}
        All pixels are set to a uniformly random RGB-colour (sampled uniformly from $[0.0, 1.0]^3$).
    \item \textbf{tricolour:}
        The image is split into three stripes of equal size, vertically or horizontally with probability 50\%.\\
        Each stripe is set to an independent uniformly random RGB-colour.
    \item \textbf{primary tricolour:}
        The image is split into three stripes of equal size, vertically or horizontally with probability 50\%.\\
        Each stripe is set to a colour where each RGB-channel value is chosen randomly as either $0.0$ or $1.0$.
    \item \textbf{horizontal stripes:}
        The image is split into a random number chosen between $(4, 5, 7, 10, 15, 20)$ of horizontal stripes of equal size.\\
        Each stripe is set to an independent uniformly random RGB-colour.
    \item \textbf{vertical stripes:}
        The image is split into a random number chosen between $(4, 5, 7, 10, 15, 20)$ of vertical stripes of equal size.\\
        Each stripe is set to an independent uniformly random RGB-colour.
    \item \textbf{smooth noise~\cite{HeiAndBit2019, bitterwolf2020certifiably, meinke2022provably}:}
        For each image, first $\sigma$ is chosen randomly between $(10, 15, 25, 40, 60, 85)$.\\
        A uniform noise image is sampled.\\
        Then we apply a Gaussian filter with a size of $\sigma$ pixels.\\
        Finally, the pixel values are scaled linearly such that the minimum brightness over all channels and pixels is $0.0$ and the maximum is $1.0$.
    \item \textbf{smooth noise+:}
        For each image, first $\sigma$ is chosen randomly between $(10, 15, 25, 40, 60, 85)$.\\
        A uniform noise image is sampled.\\
        Then we apply a Gaussian filter with a size of $\sigma$ pixels.\\
        Finally, each RGB channel is scaled linearly such that its minimum brightness over all pixels is $0.0$ and the maximum is $1.0$.
    \item \textbf{smooth color:}
        For each image, first $\sigma$ is chosen randomly between $(10, 15, 25, 40, 60, 85)$, $\delta$ uniformly between $0.1$ and $0.3$, and a uniformly random RGB-colour $c$.\\
        A uniform noise image is sampled.\\
        Then we apply a Gaussian filter with a size of $\sigma$ pixels.\\
        Finally, each RGB channel is scaled linearly such that $c-\delta$ is the 2.5th quantile of its values and $c+\delta$ the 97.5th.
    \item \textbf{smooth IN pixel permutations~\cite{HeiAndBit2019}:}
        For each image, first $\sigma$ is chosen randomly between $(1, 1.5, 2, 3, 4, 6, 8)$.\\
        An IN pixel permutations image is sampled.\\
        Then we apply a Gaussian filter with a size of $\sigma$ pixels.
    \item \textbf{blobs~\cite{HenMazDie2019}:}
        For each image, first $\sigma$ is chosen randomly between $(1.5, 2, 2.5, 3, 3.5, 4)$.\\
        Each RBG colour channel of each pixel is independently set to $1.0$ with 70\% probability or $0.0$ with 30\%.\\
        Then we apply a Gaussian filter with a size of $\sigma$ pixels.\\
        Finally, all channel values below $0.75$ are set to $0.0$.
\end{itemize}
Where necessary, the resulting channel values are clipped to $[0,1]$.
We show samples of each unit-test in the following Appendix~\ref{sec:dset_examples} in Figure~\ref{fig:unitsamples}.
\FloatBarrier
\newpage

\section{Examples images from each OOD class in \dsetname{} and from OOD unit-tests}
\label{sec:dset_examples}
\def \xoffset {2.11cm}
\def \yoffset {1.4cm}
\def \imgwidth {1.0cm}
\def \ximgdist {.03cm}
\def \imgheight {1.0cm}
\def \toplabeldist {.55cm}
\def \botlabeldist {.16cm}
\def \tinyscale {.75}


\tikzset{
    pics/oodsamples0/.style args={#1/#2/#3}{
      code = {\coordinate (o) at (0,0);
      \coordinate (i1) at ($(o) + (-.5*\imgwidth,0) + (-.5*\ximgdist,0)$);
      \coordinate (i2) at ($(o) + (.5*\imgwidth,0) + (.5*\ximgdist,0)$);
      \coordinate (t1) at ($(o)+ (0,-.5*\imgheight) + (0,-\botlabeldist)$);
  \node[inner sep=0pt, anchor=center] at (i1) {\IfFileExists{figures/some_examples_per_ood_class_0/#2}{\includegraphics[width=\imgwidth, height=\imgheight]{figures/some_examples_per_ood_class_0/#2}}{\includegraphics[width=\imgwidth, height=\imgheight]{example-image-a}}};
  \node[inner sep=0pt, anchor=center] at (i2) {\IfFileExists{figures/some_examples_per_ood_class_0/#3}{\includegraphics[width=\imgwidth, height=\imgheight]{figures/some_examples_per_ood_class_0/#3}}{\includegraphics[width=\imgwidth, height=\imgheight]{example-image-b}}};
  \draw[anchor=base, align=center] (t1)  node[scale=\tinyscale] {\tiny\OODclass{#1}};
  }}}

\triple oodclass00=({AFA} {"aphanizomenon_flosaquae_00_10050623.jpeg"} {"aphanizomenon_flosaquae_04_12588356.jpeg"})
\triple oodclass01=({bagpipe} {"bagpipe_00_n02775483_10038.jpeg"} {"bagpipe_04_n02775483_10125.jpeg"})
\triple oodclass02=({bluestriped grunt} {"haemulon_sciurus_00_10072587.jpeg"} {"haemulon_sciurus_04_1144627.jpeg"})
\triple oodclass03=({cable} {"cable_00_0002.png"} {"cable_04_0008.png"})
\triple oodclass10=({California pitcher plant} {"darlingtonia_californica_00_10001598.jpeg"} {"darlingtonia_californica_04_10238374.jpeg"})
\triple oodclass11=({California slender salamander} {"batrachoseps_attenuatus_00_1025296.jpeg"} {"batrachoseps_attenuatus_04_1052916.jpeg"})
\triple oodclass12=({California two-spot octopus} {"octopus_bimaculoides_00_10038001.jpeg"} {"octopus_bimaculoides_04_10385567.jpeg"})
\triple oodclass13=({caracal} {"Caracal caracal caracal_00_000-10896690.jpeg"} {"Caracal caracal caracal_04_005-15119962.jpeg"})
\triple oodclass20=({cat-faced spider} {"araneus_gemma_00_1045412.jpeg"} {"araneus_gemma_04_15246572.jpeg"})
\triple oodclass21=({Central American tapir} {"tapirus_bairdii_00_10103742.jpeg"} {"tapirus_bairdii_04_10441784.jpeg"})
\triple oodclass22=({chicken quesadilla} {"chicken_quesadilla_00_2380604.jpeg"} {"chicken_quesadilla_04_1826455.jpeg"})
\triple oodclass23=({common cuttlefish} {"sepia_officinalis_00_10291413.jpeg"} {"sepia_officinalis_04_11310877.jpeg"})
\triple oodclass30=({crème brûlée} {"creme_brulee_00_1253807.jpeg"} {"creme_brulee_04_2318273.jpeg"})
\triple oodclass31=({cupcakes} {"cup_cakes_00_2920467.jpeg"} {"cup_cakes_04_630654.jpeg"})
\triple oodclass32=({donuts} {"donuts_00_2842.jpeg"} {"donuts_04_580038.jpeg"})
\triple oodclass33=({door} {"door_00_00001.jpeg"} {"door_04_00005.jpeg"})
\triple oodclass40=({dreamfish} {"sarpa_salpa_00_10230766.jpeg"} {"sarpa_salpa_04_10988258.jpeg"})
\triple oodclass41=({dune thistle} {"cirsium_pitcheri_00_12378705.png"} {"cirsium_pitcheri_04_13099322.jpeg"})
\triple oodclass42=({dusky flathead} {"platycephalus_fuscus_00_10064587.jpeg"} {"platycephalus_fuscus_04_10792114.jpeg"})
\triple oodclass43=({E. micromeris} {"epithelantha_micromeris_00_10020633.jpeg"} {"epithelantha_micromeris_04_10756018.jpeg"})
\triple oodclass50=({Eastern leaf-footed bug} {"leptoglossus_phyllopus_00_1024124.jpeg"} {"leptoglossus_phyllopus_04_1098402.jpeg"})
\triple oodclass51=({European paper wasp} {"polistes_dominula_00_101572.jpeg"} {"polistes_dominula_04_107269.jpeg"})
\triple oodclass52=({false killer whale} {"pseudorca_crassidens_00_10149236.jpeg"} {"pseudorca_crassidens_04_14033485.jpeg"})
\triple oodclass53=({field road} {"f_field_road_00_00000092.jpeg"} {"f_field_road_04_00000292.jpeg"})
\triple oodclass60=({fire extinguisher} {"fire_extinguisher_00_00001.jpeg"} {"fire_extinguisher_04_00005.jpeg"})
\triple oodclass61=({fireworks} {"fireworks_00_0000.png"} {"fireworks_04_0006.png"})
\triple oodclass62=({forest path} {"f_forest_path_00_00000011.jpeg"} {"f_forest_path_04_00000296.jpeg"})
\triple oodclass63=({Franciscan wallflower} {"erysimum_franciscanum_00_10000824.jpeg"} {"erysimum_franciscanum_04_10086385.jpeg"})
\triple oodclass70=({French fries} {"french_fries_00_2382614.jpeg"} {"french_fries_04_292607.jpeg"})
\triple oodclass71=({Galápagos fur seal} {"arctocephalus_galapagoensis_00_10364938.jpeg"} {"arctocephalus_galapagoensis_04_12789420.jpeg"})
\triple oodclass72=({giant cuttlefish} {"sepia_apama_00_10002106.jpeg"} {"sepia_apama_04_10181823.jpeg"})
\triple oodclass73=({glass of milk} {"glass_of_milk_00_glass_of_milk_0002_flickr.jpeg"} {"glass_of_milk_04_glass_of_milk_0011_flickr.jpeg"})
\triple oodclass80=({gramophone} {"gramophone_00_image_0001.jpeg"} {"gramophone_04_image_0005.jpeg"})
\triple oodclass81=({high heels} {"high heels_00_101.jpeg"} {"high heels_04_109.jpeg"})
\triple oodclass82=({Hindu temple} {"hindu_temple_00_10.jpeg"} {"hindu_temple_04_15.jpeg"})
\triple oodclass83=({Horse Hoof clam} {"hippopus_hippopus_00_17586984.jpeg"} {"hippopus_hippopus_04_31028820.jpeg"})
\triple oodclass90=({Indo-Pacific bottlenose dolphin} {"tursiops_aduncus_00_1047859.jpeg"} {"tursiops_aduncus_04_10954905.jpeg"})
\triple oodclass91=({long-tailed silverfish} {"ctenolepisma_longicaudata_00_10007377.jpeg"} {"ctenolepisma_longicaudata_04_10269995.jpeg"})
\triple oodclass92=({Lumholtz's tree-kangaroo} {"dendrolagus_lumholtzi_00_11323300.jpeg"} {"dendrolagus_lumholtzi_04_17275614.jpeg"})
\triple oodclass93=({M. wesenbergii} {"microcystis_wesenbergii_00_10050422.jpeg"} {"microcystis_wesenbergii_04_16433465.jpeg"})
\triple oodclass100=({marbled newt} {"triturus_marmoratus_00_10615360.jpeg"} {"triturus_marmoratus_04_1129138.jpeg"})
\triple oodclass101=({mbira} {"mbira_00_0.png"} {"mbira_04_103.png"})
\triple oodclass102=({Mexican lime cactus} {"ferocactus_pilosus_00_10000962.jpeg"} {"ferocactus_pilosus_04_10342896.jpeg"})
\triple oodclass103=({Pampas deer} {"ozotoceros_bezoarticus_00_18557909.jpeg"} {"ozotoceros_bezoarticus_04_20255841.jpeg"})
\triple oodclass110=({pyramid} {"pyramid_00_image_0002.jpeg"} {"pyramid_09_image_0012.jpeg"})
\triple oodclass111=({redbreast sunfish} {"lepomis_auritus_00_10033361.jpeg"} {"lepomis_auritus_04_10312959.jpeg"})
\triple oodclass112=({rosybells} {"streptopus_lanceolatus_00_10209695.jpeg"} {"streptopus_lanceolatus_04_11457266.jpeg"})
\triple oodclass113=({ruby octopus} {"octopus_rubescens_00_10041096.jpeg"} {"octopus_rubescens_04_10321376.jpeg"})
\triple oodclass120=({scissors} {"scissors_00_image_0001.jpeg"} {"scissors_04_image_0005.jpeg"})
\triple oodclass121=({shuttlecock} {"shuttlecock_00_0000.png"} {"shuttlecock_04_0006.png"})
\triple oodclass122=({silver-haired bat} {"lasionycteris_noctivagans_00_10418492.jpeg"} {"lasionycteris_noctivagans_04_11506657.jpeg"})
\triple oodclass123=({skipper caterpillar} {"skipper_caterpillar_00_004-2385220.jpeg"} {"skipper_caterpillar_04_008-4927169.png"})
\triple oodclass130=({sky} {"s_sky_00_00000043.jpeg"} {"s_sky_04_00000225.jpeg"})
\triple oodclass131=({southern calamari} {"sepioteuthis_australis_00_10096976.jpeg"} {"sepioteuthis_australis_04_11342444.jpeg"})
\triple oodclass132=({spaghetti bolognese} {"spaghetti_bolognese_00_3109585.jpeg"} {"spaghetti_bolognese_04_2773286.jpeg"})
\triple oodclass133=({stapler} {"stapler_00_image_0003.jpeg"} {"stapler_04_image_0008.jpeg"})
\triple oodclass140=({sweet pea} {"lathyrus_odoratus_00_1001576.jpeg"} {"lathyrus_odoratus_04_12191481.jpeg"})
\triple oodclass141=({two-toed amphiuma} {"amphiuma_means_00_10045958.jpeg"} {"amphiuma_means_04_10720687.jpeg"})
\triple oodclass142=({waffles} {"waffles_00_903753.jpeg"} {"waffles_04_743722.jpeg"})
\triple oodclass143=({walker} {"walker_00_00001.jpeg"} {"walker_04_00007.jpeg"})
\triple oodclass150=({water dispenser} {"empty_water_dispencer_00_00001.jpeg"} {"empty_water_dispencer_04_00013.jpeg"})
\triple oodclass151=({Windsor chair} {"windsor_chair_00_image_0001.jpeg"} {"windsor_chair_04_image_0005.jpeg"})
\triple oodclass152=({yellow trumpets} {"sarracenia_alata_00_10078270.jpeg"} {"sarracenia_alata_04_10299356.jpeg"})
\triple oodclass153=({ʻōhelo ʻai} {"vaccinium_reticulatum_00_10014724.jpeg"} {"vaccinium_reticulatum_04_11413487.jpeg"})

\begin{figure*}[!h]
\centering
\pgfmathsetseed{5}
\hspace*{-2.5mm}
\resizebox{1.022\textwidth}{!}{
\noindent
\begin{tikzpicture}
  \coordinate (ood1) at (0,0);
  \foreach \y in {0,...,5}{
  \foreach \x in {0,...,3}{
  \pic [local bounding box=A1a] at ($(ood1) + (\x*\xoffset,-\y*\yoffset)$) {oodsamples0=\xtriple{oodclass\y\x}{1}/\xtriple{oodclass\y\x}{2}/\xtriple{oodclass\y\x}{3}};
  }
  }
\end{tikzpicture}
}
\caption{\label{fig:OODsamples1}Samples of each class of the \dsetname{} dataset (1/3).
}
\end{figure*}
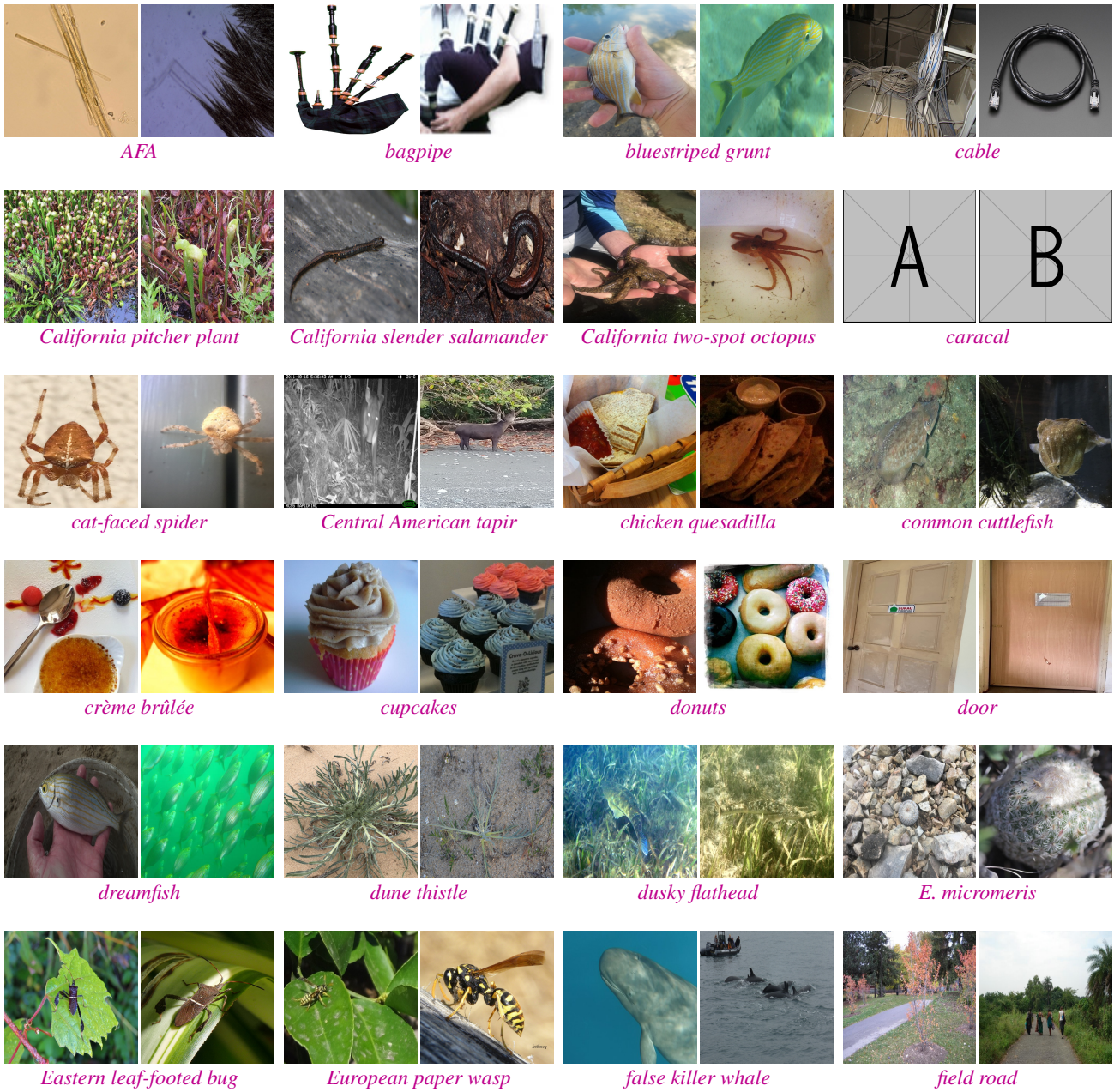

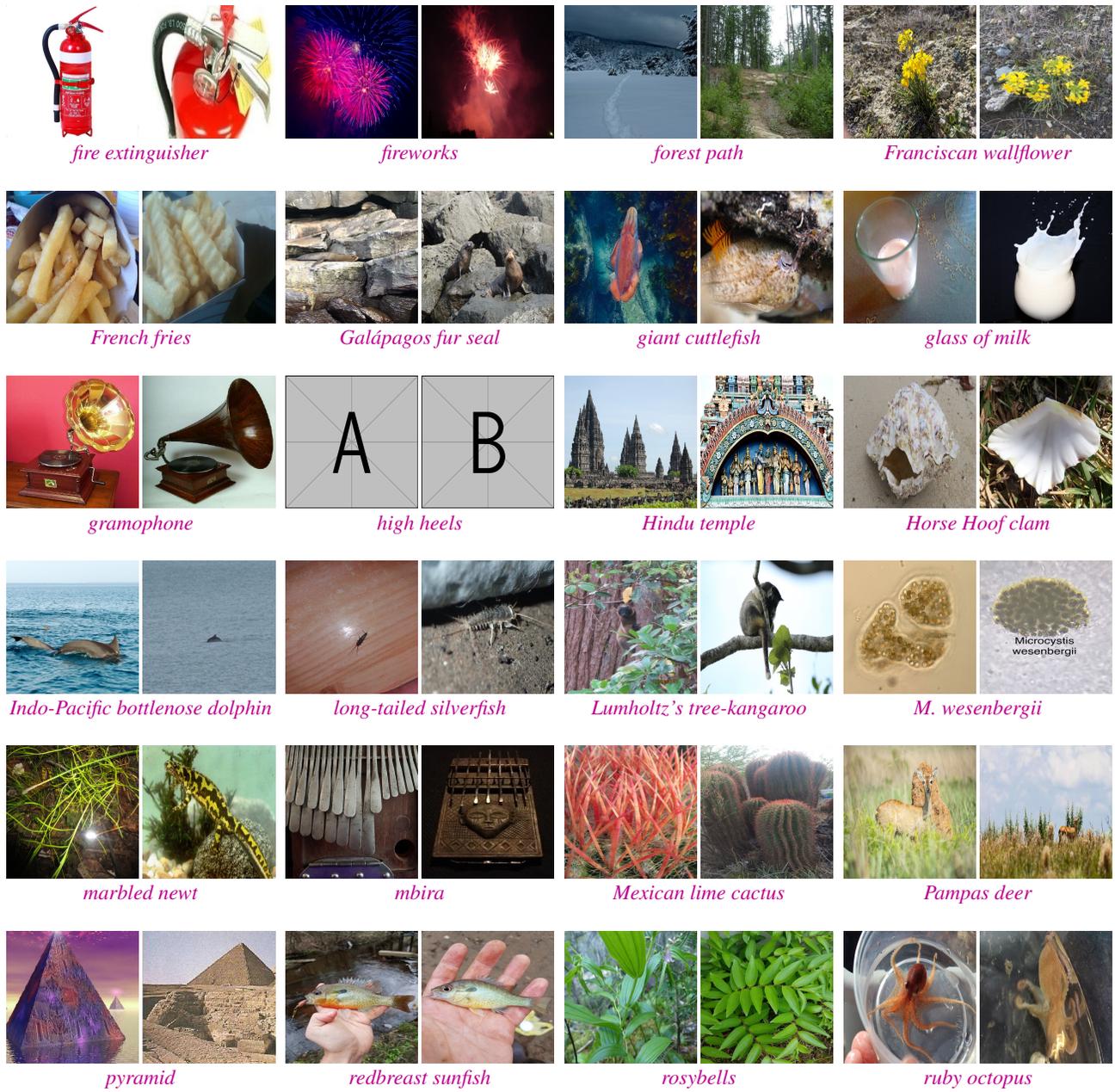
\begin{figure*}[t]
\centering
\pgfmathsetseed{5}
\hspace*{-2.5mm}
\resizebox{1.025\textwidth}{!}{
\noindent
\begin{tikzpicture}
  \coordinate (ood1) at (0,0);
  \foreach \y in {6,...,11}{
  \foreach \x in {0,...,3}{
  \pic [local bounding box=A1a] at ($(ood1) + (\x*\xoffset,-\y*\yoffset)$) {oodsamples0=\xtriple{oodclass\y\x}{1}/\xtriple{oodclass\y\x}{2}/\xtriple{oodclass\y\x}{3}};
  }
  }
\end{tikzpicture}
}
\caption{\label{fig:OODsamples2}Samples of each class of the \dsetname{} dataset (2/3).
}
\end{figure*}

\begin{figure*}[t]
\centering
\pgfmathsetseed{5}
\hspace*{-2.5mm}
\resizebox{1.025\textwidth}{!}{
\noindent
\begin{tikzpicture}
  \coordinate (ood1) at (0,0);
  \foreach \y in {12,...,15}{
  \foreach \x in {0,...,3}{
  \pic [local bounding box=A1a] at ($(ood1) + (\x*\xoffset,-\y*\yoffset)$) {oodsamples0=\xtriple{oodclass\y\x}{1}/\xtriple{oodclass\y\x}{2}/\xtriple{oodclass\y\x}{3}};
  }
  }
\end{tikzpicture}
}
\caption{\label{fig:OODsamples3}Samples of each class of the \dsetname{} dataset (3/3).
}
\end{figure*}
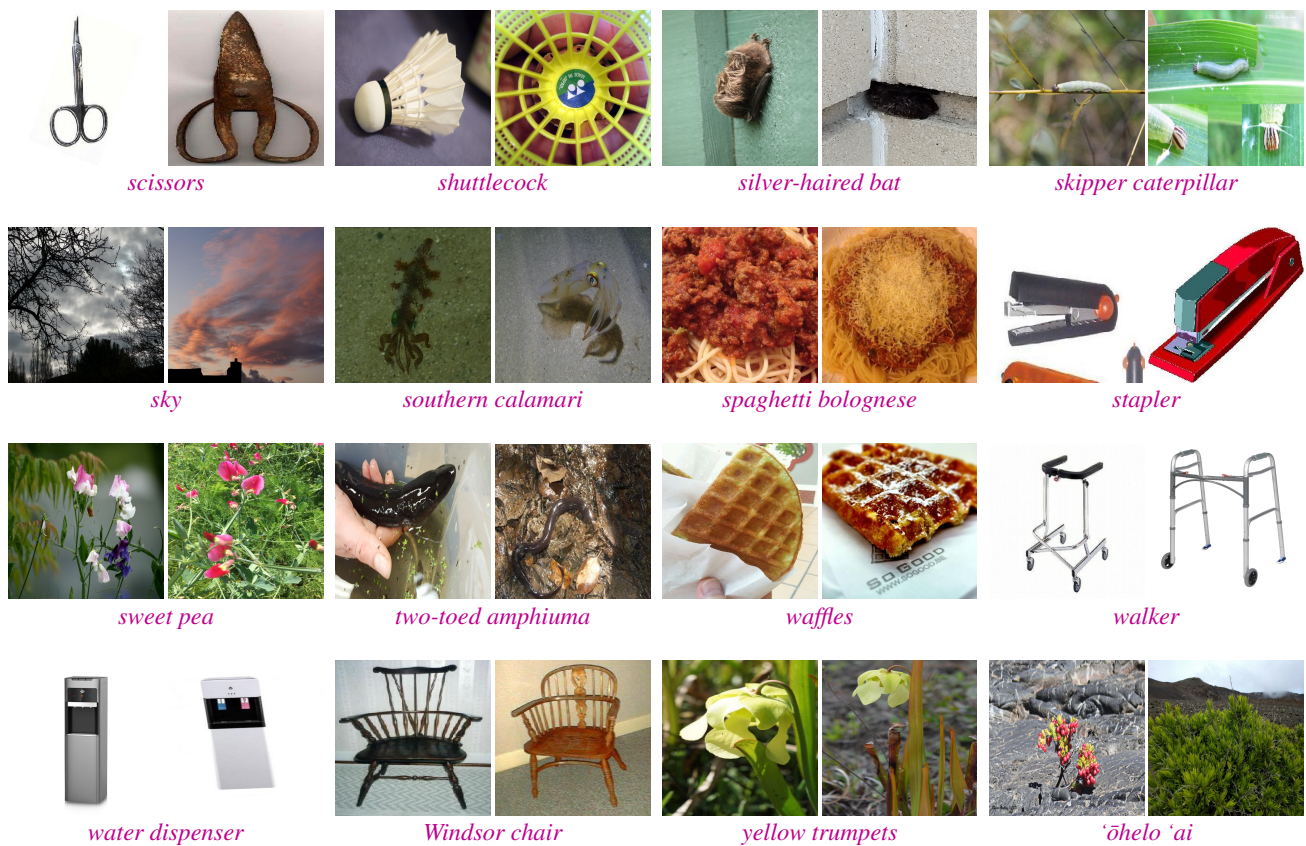

\triple unittestclass00=({uniform noise} {"unit_tests/uni_00_uni_00112.png"} {"unit_tests/uni_04_uni_00408.png"})
\triple unittestclass01=({Gaussian noise} {"unit_tests/Gaussian_00_Gaussian_00217.jpeg"} {"unit_tests/Gaussian_04_Gaussian_00452.png"})
\triple unittestclass02=({Rademacher noise} {"unit_tests/Rademacher_00_Rademacher_00101.jpeg"} {"unit_tests/Rademacher_04_Rademacher_00780.png"})
\triple unittestclass03=({IN pixel permutations} {"unit_tests/pixel_perm_00_pixel_perm_00076.png"} {"unit_tests/pixel_perm_04_pixel_perm_00302.png"})
\triple unittestclass11=({white} {"unit_tests/white_00_white_00020.png"} {"unit_tests/white_04_white_00601.jpeg"})
\triple unittestclass10=({black} {"unit_tests/black_00_black_00048.png"} {"unit_tests/black_04_black_00540.png"})
\triple unittestclass12=({shades of grey} {"unit_tests/grey_00_grey_00004.png"} {"unit_tests/grey_04_grey_00519.jpeg"})
\triple unittestclass13=({monochrome} {"unit_tests/monochrome_00_monochrome_00204.png"} {"unit_tests/monochrome_04_monochrome_01110.png"})
\triple unittestclass20=({tricolour} {"unit_tests/tricolour_00_tricolour_00039.jpeg"} {"unit_tests/tricolour_04_tricolour_00559.jpeg"})
\triple unittestclass21=({primary tricolour} {"unit_tests/tricolour_primary_09_tricolour_primary_01694.png"} {"unit_tests/tricolour_primary_04_tricolour_primary_01009.jpeg"})
\triple unittestclass22=({horizontal stripes} {"unit_tests/horizontal_stripes_00_horizontal_stripes_00389.jpeg"} {"unit_tests/horizontal_stripes_04_horizontal_stripes_00585.jpeg"})
\triple unittestclass23=({vertical stripes} {"unit_tests/vertical_stripes_00_vertical_stripes_00655.jpeg"} {"unit_tests/vertical_stripes_04_vertical_stripes_01085.jpeg"})
\triple unittestclass30=({smooth noise} {"unit_tests/low_freq_00_low_freq_00075.jpeg"} {"unit_tests/low_freq_09_low_freq_02465.jpeg"})
\triple unittestclass31=({smooth noise+} {"unit_tests/low_freq_channelfullscale_00_low_freq_channelfullscale_00054.png"} {"unit_tests/low_freq_channelfullscale_09_low_freq_channelfullscale_00977.jpeg"})
\triple unittestclass32=({smooth color} {"unit_tests/low_freq_colorrange_00_low_freq_colorrange_00228.png"} {"unit_tests/low_freq_colorrange_14_low_freq_colorrange_01943.jpeg"})
\triple unittestclass33=({smooth IN pixel permutations} {"unit_tests/low_freq_pixel_perm_00_low_freq_pixel_perm_00086.png"} {"unit_tests/low_freq_pixel_perm_04_low_freq_pixel_perm_00473.jpeg"})
\triple unittestclass40=({blobs} {"unit_tests/blobs_00_blobs_00055.jpeg"} {"unit_tests/blobs_04_blobs_00542.png"})

\begin{figure*}[t]
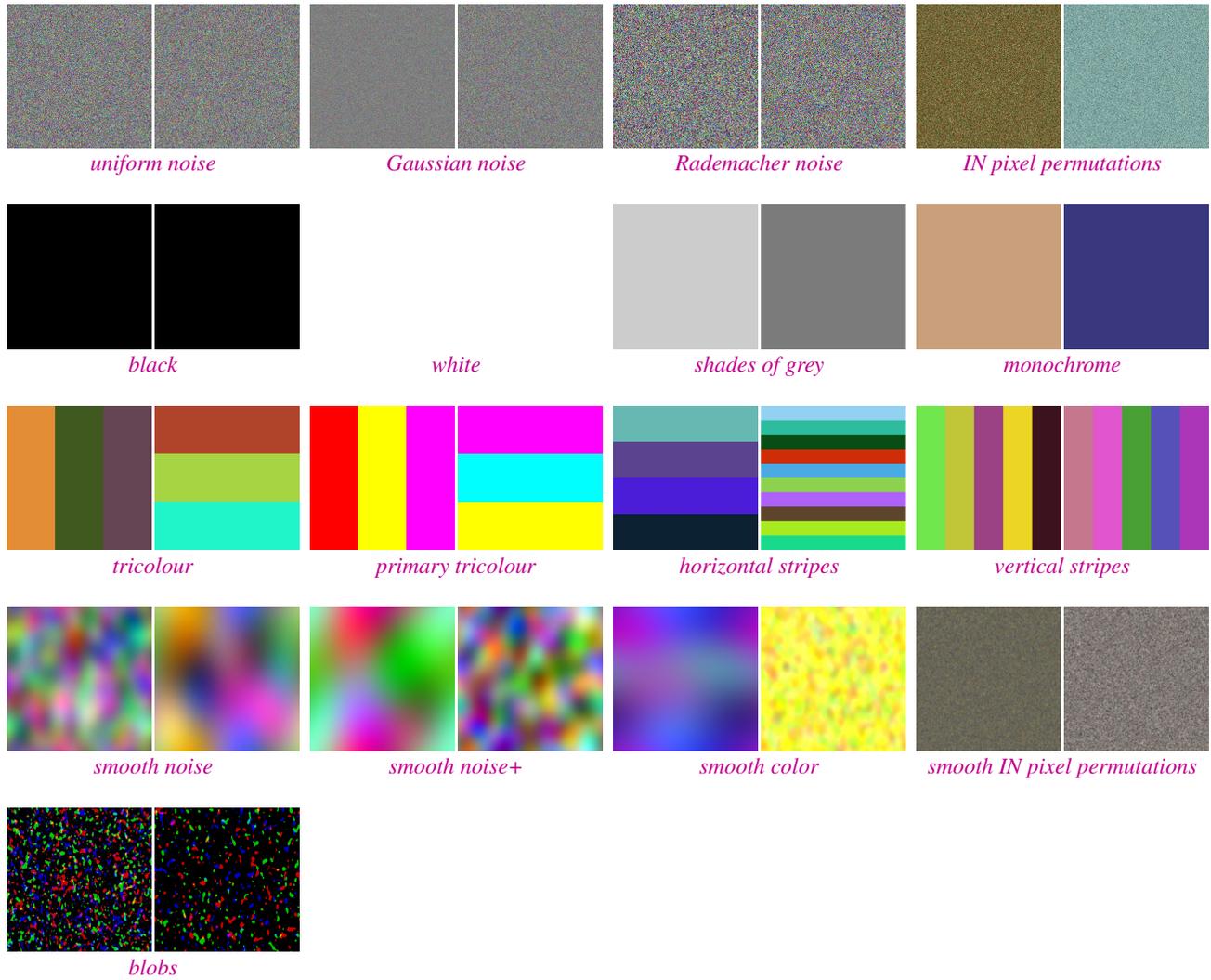

\centering
\pgfmathsetseed{5}
\hspace*{-2.5mm}
\resizebox{1.025\textwidth}{!}{
\noindent

}
\caption{\label{fig:unitsamples}Samples of each OOD unit-test.
}
\end{figure*}

\FloatBarrier
\newpage

\section{Tables - Unit tests}
\label{app:unit-tests}

\renewcommand{\arraystretch}{.5}
\tabcolsep=0.02951cm
\begin{table}[!ht]
    \caption{\FPR{} of pretrained transformers for unit-tests. The ViTs pretrained \textit{only} on ImageNet-21k fail only few unit tests, the other models struggle often with feature-based methods.}
    \label{tab:unit-tests-pre-fpr-vit}
    \centering
    \tiny
\begin{center}
\begin{tabular}{c c c c c c c c c c c c c c c c c c c c c c }
model &acc. &  method & \# fails & max & \OODclass{Gauss} & \OODclass{Rade} & \OODclass{Black} & \OODclass{Blob} & \OODclass{Grey} & \OODclass{Hor} & \OODclass{SmN} & \OODclass{SmN+} & \OODclass{SmCol} & \OODclass{SmPxPerm} & \OODclass{Mono} & \OODclass{PxPerm} & \OODclass{Tri} & \OODclass{PrTri} & \OODclass{Uni} & \OODclass{Ver} & \OODclass{White} \\ 
 \hline 
 &  & MSP & 0 & 4.2 & 0.5 & 0.0 & 0.0 & 4.2 & 0.0 & 0.0 & 0.0 & 0.0 & 0.0 & 1.5 & 0.2 & 0.0 & 0.5 & 0.0 & 1.0 & 3.0 & 0.8 \\ 
  &  & MaxLogit & 0 & 0.0 & 0.0 & 0.0 & 0.0 & 0.0 & 0.0 & 0.0 & 0.0 & 0.0 & 0.0 & 0.0 & 0.0 & 0.0 & 0.0 & 0.0 & 0.0 & 0.0 & 0.0 \\ 
  &  & ViM & 2 & 100.0 & 0.0 & 0.0 & 0.0 & 0.0 & $\color{red}{100.0}$ & 0.0 & $\color{red}{46.0}$ & 0.5 & 0.0 & 5.0 & 0.0 & 0.0 & 0.0 & 0.0 & 0.5 & 0.0 & 0.0 \\ 
  &  & Mahalanobis & 0 & 0.0 & 0.0 & 0.0 & 0.0 & 0.0 & 0.0 & 0.0 & 0.0 & 0.0 & 0.0 & 0.0 & 0.0 & 0.0 & 0.0 & 0.0 & 0.0 & 0.0 & 0.0 \\ 
  &  & Energy+React & 0 & 0.0 & 0.0 & 0.0 & 0.0 & 0.0 & 0.0 & 0.0 & 0.0 & 0.0 & 0.0 & 0.0 & 0.0 & 0.0 & 0.0 & 0.0 & 0.0 & 0.0 & 0.0 \\ 
 ViT-B-384-21k & 86.0 & Energy & 0 & 0.0 & 0.0 & 0.0 & 0.0 & 0.0 & 0.0 & 0.0 & 0.0 & 0.0 & 0.0 & 0.0 & 0.0 & 0.0 & 0.0 & 0.0 & 0.0 & 0.0 & 0.0 \\ 
  &  & KL-Matching & 0 & 4.8 & 0.0 & 0.0 & 0.0 & 3.8 & 0.0 & 0.0 & 0.0 & 0.0 & 0.0 & 0.5 & 1.2 & 0.0 & 0.0 & 0.0 & 0.0 & 4.8 & 0.0 \\
\hline 
\end{tabular}
\end{center}

\vskip \belowtablevskip
\end{table}

\renewcommand{\arraystretch}{.5}
\tabcolsep=0.02951cm
\begin{table}[htb]
    \caption{\FPR{} of pretrained convolutional networks for OOD unit-tests.}
    \label{tab:unit-tests-pre-fpr-conv}
    \centering
    \tiny
\begin{center}
\begin{tabular}{c c c c c c c c c c c c c c c c c c c c c c }
model &acc. &  method & \# fails & max & \OODclass{Gauss} & \OODclass{Rade} & \OODclass{Black} & \OODclass{Blob} & \OODclass{Grey} & \OODclass{Hor} & \OODclass{SmN} & \OODclass{SmN+} & \OODclass{SmCol} & \OODclass{SmPxPerm} & \OODclass{Mono} & \OODclass{PxPerm} & \OODclass{Tri} & \OODclass{PrTri} & \OODclass{Uni} & \OODclass{Ver} & \OODclass{White} \\ 
 \hline 
 &  & MSP & 1 & 13.5 & 0.0 & 0.0 & 0.0 & 0.2 & 0.0 & $\color{red}{13.5}$ & 2.2 & 5.5 & 2.0 & 5.8 & 0.0 & 0.5 & 2.5 & 8.2 & 0.2 & 0.2 & 0.0 \\ 
  &  & MaxLogit & 0 & 6.0 & 0.0 & 0.0 & 0.0 & 0.0 & 0.0 & 4.5 & 2.5 & 6.0 & 1.0 & 2.8 & 0.0 & 0.0 & 0.2 & 4.0 & 0.2 & 0.2 & 0.0 \\ 
  &  & Energy & 1 & 10.8 & 0.0 & 0.0 & 0.0 & 0.0 & 0.0 & 2.5 & 5.2 & $\color{red}{10.8}$ & 2.2 & 2.8 & 0.0 & 0.0 & 0.0 & 3.0 & 0.0 & 0.2 & 0.0 \\ 
  &  & KL-Matching & 2 & 29.2 & 0.0 & 0.0 & 0.0 & 2.0 & 0.0 & $\color{red}{29.2}$ & 2.5 & 2.0 & 2.5 & 6.2 & 0.0 & 1.2 & 4.2 & $\color{red}{10.8}$ & 0.2 & 0.0 & 0.0 \\ 
  &  & Mahalanobis & 4 & 100.0 & 0.8 & $\color{red}{42.5}$ & $\color{red}{100.0}$ & 0.0 & $\color{red}{100.0}$ & 0.0 & 0.0 & 0.0 & 0.0 & 2.0 & 2.8 & 0.0 & 0.0 & 5.0 & 0.2 & 0.0 & $\color{red}{100.0}$ \\ 
 CnvNxt-B-21k & 86.3 & Relative Mahalanobis & 5 & 100.0 & 0.5 & $\color{red}{41.0}$ & $\color{red}{100.0}$ & 0.0 & $\color{red}{100.0}$ & 1.0 & 0.0 & 0.0 & 0.0 & $\color{red}{13.0}$ & 1.5 & 0.0 & 0.2 & 9.8 & 0.0 & 0.0 & $\color{red}{100.0}$ \\ 
  &  & ViM & 4 & 100.0 & 0.5 & $\color{red}{41.2}$ & $\color{red}{100.0}$ & 0.0 & $\color{red}{100.0}$ & 0.0 & 0.0 & 0.0 & 0.0 & 0.0 & 0.5 & 0.0 & 0.0 & 2.0 & 0.0 & 0.0 & $\color{red}{100.0}$ \\
\hline 
\end{tabular}
\end{center}

\vskip \belowtablevskip
\end{table}

\renewcommand{\arraystretch}{.5}
\tabcolsep=0.02951cm
\begin{table}[htb]
    \caption{\FPR{} of transformers without pretraining for OOD unit-tests. }
    \label{tab:unit-tests-1k-fpr-vit}
    \centering
    \tiny
\begin{center}
\begin{tabular}{c c c c c c c c c c c c c c c c c c c c c c }
model &acc. &  method & \# fails & max & \OODclass{Gauss} & \OODclass{Rade} & \OODclass{Black} & \OODclass{Blob} & \OODclass{Grey} & \OODclass{Hor} & \OODclass{SmN} & \OODclass{SmN+} & \OODclass{SmCol} & \OODclass{SmPxPerm} & \OODclass{Mono} & \OODclass{PxPerm} & \OODclass{Tri} & \OODclass{PrTri} & \OODclass{Uni} & \OODclass{Ver} & \OODclass{White} \\ 
 \hline 
 &  & MSP & 7 & 43.8 & 0.0 & 0.0 & 0.0 & $\color{red}{15.8}$ & 0.0 & $\color{red}{26.2}$ & $\color{red}{11.8}$ & $\color{red}{11.2}$ & $\color{red}{13.2}$ & 5.5 & 0.0 & $\color{red}{43.8}$ & 0.5 & $\color{red}{10.5}$ & 1.0 & 4.5 & 0.0 \\ 
  &  & MaxLogit & 2 & 19.2 & 0.0 & 0.0 & 0.0 & 4.0 & 0.0 & $\color{red}{16.8}$ & 5.2 & 6.2 & 6.5 & 1.0 & 0.0 & $\color{red}{19.2}$ & 0.2 & 7.8 & 0.0 & 2.2 & 0.0 \\ 
  &  & Energy & 2 & 17.0 & 3.0 & 0.0 & 0.0 & 1.2 & 0.0 & $\color{red}{17.0}$ & 2.8 & 2.8 & 5.8 & 0.5 & 0.0 & $\color{red}{12.2}$ & 0.8 & 8.8 & 0.0 & 2.8 & 0.0 \\ 
  &  & KL-Matching & 6 & 44.8 & 0.0 & 0.0 & 0.0 & $\color{red}{13.8}$ & 6.2 & $\color{red}{28.0}$ & 8.2 & 8.2 & 9.0 & 4.8 & 3.0 & $\color{red}{44.8}$ & $\color{red}{10.2}$ & $\color{red}{12.2}$ & 0.2 & $\color{red}{27.5}$ & 0.0 \\ 
  &  & Mahalanobis & 4 & 80.8 & $\color{red}{19.0}$ & 0.5 & 0.0 & 0.0 & $\color{red}{80.8}$ & $\color{red}{22.0}$ & 0.8 & 0.2 & 0.2 & 1.2 & 8.8 & $\color{red}{22.8}$ & 9.8 & 3.2 & 8.0 & 1.0 & 0.0 \\ 
 ViT-B-384 & 81.1 & Relative Mahalanobis & 10 & 100.0 & 6.8 & 0.0 & $\color{red}{100.0}$ & 0.8 & $\color{red}{100.0}$ & $\color{red}{26.8}$ & 1.0 & 1.0 & 1.8 & $\color{red}{15.2}$ & $\color{red}{49.8}$ & $\color{red}{24.0}$ & $\color{red}{49.8}$ & $\color{red}{17.0}$ & 0.2 & $\color{red}{12.2}$ & $\color{red}{100.0}$ \\ 
  &  & ViM & 5 & 94.8 & $\color{red}{74.5}$ & $\color{red}{12.2}$ & 0.0 & 0.0 & 0.0 & $\color{red}{28.0}$ & 4.8 & 4.0 & 0.5 & 2.2 & 0.0 & $\color{red}{27.5}$ & 0.5 & 1.8 & $\color{red}{94.8}$ & 0.0 & 0.0 \\
\hline 
\end{tabular}
\end{center}

\vskip \belowtablevskip
\end{table}

\renewcommand{\arraystretch}{.015}
\tabcolsep=0.02951cm
\begin{table}[htb]
    \caption{\FPR{} of convolutional networks without pretraining for OOD unit-tests.}
    \label{tab:unit-tests-1k-fpr-conv}
    \centering
    \tiny
\begin{center}
\begin{tabular}{c c c c c c c c c c c c c c c c c c c c c c }
model &acc. &  method & \# fails & max & \OODclass{Gauss} & \OODclass{Rade} & \OODclass{Black} & \OODclass{Blob} & \OODclass{Grey} & \OODclass{Hor} & \OODclass{SmN} & \OODclass{SmN+} & \OODclass{SmCol} & \OODclass{SmPxPerm} & \OODclass{Mono} & \OODclass{PxPerm} & \OODclass{Tri} & \OODclass{PrTri} & \OODclass{Uni} & \OODclass{Ver} & \OODclass{White} \\ 
 \hline 
 &  & MSP & 3 & 48.0 & 0.0 & 0.0 & 0.0 & 5.2 & 0.0 & $\color{red}{48.0}$ & 1.8 & 4.0 & 3.2 & 2.8 & 0.0 & $\color{red}{11.2}$ & $\color{red}{12.0}$ & 5.2 & 0.0 & 0.2 & 0.0 \\ 
  &  & MaxLogit & 1 & 40.2 & 0.0 & 0.0 & 0.0 & 2.2 & 0.0 & $\color{red}{40.2}$ & 2.0 & 4.2 & 1.5 & 1.0 & 0.0 & 6.0 & 6.5 & 3.2 & 0.0 & 0.0 & 0.0 \\ 
  &  & Energy & 4 & 56.0 & 0.0 & 0.0 & 0.0 & 1.0 & 0.0 & $\color{red}{56.0}$ & $\color{red}{18.2}$ & $\color{red}{32.2}$ & 4.0 & 1.8 & 0.0 & $\color{red}{10.5}$ & 5.5 & 6.5 & 0.0 & 3.0 & 0.0 \\ 
  &  & KL-Matching & 4 & 52.8 & 0.5 & 0.0 & 0.0 & 8.0 & 0.0 & $\color{red}{52.8}$ & 6.5 & 8.2 & 6.8 & 9.8 & 0.0 & $\color{red}{14.8}$ & $\color{red}{37.0}$ & $\color{red}{18.2}$ & 0.0 & 1.2 & 0.0 \\ 
  &  & Mahalanobis & 16 & 100.0 & $\color{red}{100.0}$ & $\color{red}{99.8}$ & $\color{red}{100.0}$ & $\color{red}{91.5}$ & $\color{red}{100.0}$ & $\color{red}{100.0}$ & $\color{red}{13.5}$ & $\color{red}{11.0}$ & $\color{red}{43.0}$ & $\color{red}{57.2}$ & $\color{red}{100.0}$ & 8.5 & $\color{red}{99.8}$ & $\color{red}{98.5}$ & $\color{red}{99.8}$ & $\color{red}{51.2}$ & $\color{red}{100.0}$ \\ 
 XCiT-M-224 & 82.6 & Relative Mahalanobis & 15 & 100.0 & $\color{red}{99.5}$ & $\color{red}{99.2}$ & $\color{red}{100.0}$ & $\color{red}{82.5}$ & $\color{red}{100.0}$ & $\color{red}{100.0}$ & $\color{red}{12.5}$ & 10.0 & $\color{red}{40.8}$ & $\color{red}{54.0}$ & $\color{red}{100.0}$ & 8.5 & $\color{red}{99.8}$ & $\color{red}{98.2}$ & $\color{red}{99.8}$ & $\color{red}{52.0}$ & $\color{red}{100.0}$ \\
\hline 
\end{tabular}
\end{center}

\vskip \belowtablevskip
\end{table}
\FloatBarrier
\section{Effect of ID contamination on all models}\label{sec:Effect_cleaning_all}

In Table~\ref{tab:overview-fpr-subsampled} we show the \FPR{} values averaged across the cleaned subsampled datasets on which Table~\ref{tab:percentage_ID_in_OOD_datasets} in the main paper is based.

Detailed results on the individual datasets are shown in Tables~\ref{tab:clean-vs-pre-vit}-\ref{tab:clean-vs-all-1k-conv}.
There, we show results on the uncleaned full (-f) and cleaned subsampled (-c) datasets: \textsc{Places} (Pl), \textsc{Species} (Spc), \textsc{ImageNet-O} (IN), \textsc{Textures} (txt) \& \textsc{Textures43}, \textsc{OpenImage-O} (OpO), \textsc{iNaturalist OOD Plants} (iNat), \textsc{ImageNet-1K-OOD} (IN1K), \textsc{360OpenSetClasses} (OS), \textsc{Semantic Shift Benchmark easy} (SBe) \& \textsc{hard} (SBh), and \textsc{COOD} (CO).

Since \textsc{Textures} and \textsc{iNaturalist} are fairly easy test OOD datasets, the \FPR{} values of most models in Table~\ref{tab:overview-fpr-subsampled} are lower than on \dsetname. In general, the results allow similar conclusions:
Feature-based methods outperform methods not explicitly accessing pre-logit feature-information, yet still fail for some models, and pretraining only on IN-21k yields the best OOD-detectors. Again, Cosine and MCM/RCos improve fairly consistently over MSP, and are in some cases even the best-performing method.
\renewcommand{\arraystretch}{1.}
\tabcolsep=0.10321cm
\begin{table*}[htb]
    \centering
    \caption{Mean \FPR{} on subsampled datasets (averaged).}
    \small
    \begin{smaller}
\begin{center}
\begin{tabular}{l l l l l l l l l l l l l l }
pre & acc. & model & MSP & MaxL & Ener & KL-M & Maha & RMaha & ViM & E+R & KNN & Cos & MCM/RCos \\ 
 \hline 
{\multirow{9}{*}{{21k}}} & 86.0 & ViT-B-384 & 39.7 & 27.0 ${\textcolor{green}{-13}}$ & 25.7 ${\textcolor{green}{-14}}$ & 38.4 ${\textcolor{green}{-1}}$ & 22.4 ${\textcolor{green}{-17}}$ & 25.5 ${\textcolor{green}{-14}}$ & \textbf{22.4} ${\textcolor{green}{-17}}$ & 27.5 ${\textcolor{green}{-12}}$ & 48.2 ${\textcolor{red}{+8}}$ & 30.6 ${\textcolor{green}{-9}}$ & 30.4 ${\textcolor{green}{-9}}$ \\ 
  & 84.5 & ViT-B-224 & 43.3 & 30.8 ${\textcolor{green}{-13}}$ & 29.3 ${\textcolor{green}{-14}}$ & 42.7 ${\textcolor{green}{-1}}$ & \textbf{23.8} ${\textcolor{green}{-19}}$ & 28.2 ${\textcolor{green}{-15}}$ & 24.7 ${\textcolor{green}{-19}}$ & 32.6 ${\textcolor{green}{-11}}$ & 53.3 ${\textcolor{red}{+10}}$ & 37.0 ${\textcolor{green}{-6}}$ & 36.1 ${\textcolor{green}{-7}}$ \\ 
  & 86.3 & Swinv2-B-256 & 41.9 & 32.3 ${\textcolor{green}{-10}}$ & 31.5 ${\textcolor{green}{-10}}$ & 46.4 ${\textcolor{red}{+4}}$ & 47.4 ${\textcolor{red}{+5}}$ & 40.4 ${\textcolor{green}{-2}}$ & 37.5 ${\textcolor{green}{-4}}$ & \textbf{27.8} ${\textcolor{green}{-14}}$ & 43.1 ${\textcolor{red}{+1}}$ & 35.5 ${\textcolor{green}{-6}}$ & 34.2 ${\textcolor{green}{-8}}$ \\ 
  & 86.7 & Deit3-B-384 & 53.4 & 45.4 ${\textcolor{green}{-8}}$ & 46.4 ${\textcolor{green}{-7}}$ & 52.5 ${\textcolor{green}{-1}}$ & 40.8 ${\textcolor{green}{-13}}$ & 37.8 ${\textcolor{green}{-16}}$ & 41.2 ${\textcolor{green}{-12}}$ & 39.9 ${\textcolor{green}{-13}}$ & 40.1 ${\textcolor{green}{-13}}$ & 36.3 ${\textcolor{green}{-17}}$ & \textbf{36.0} ${\textcolor{green}{-17}}$ \\ 
  & 85.7 & Deit3-B-224 & 55.1 & 46.9 ${\textcolor{green}{-8}}$ & 47.2 ${\textcolor{green}{-8}}$ & 56.1 ${\textcolor{red}{+1}}$ & 46.6 ${\textcolor{green}{-9}}$ & 42.6 ${\textcolor{green}{-12}}$ & 47.5 ${\textcolor{green}{-8}}$ & 42.0 ${\textcolor{green}{-13}}$ & 45.1 ${\textcolor{green}{-10}}$ & 41.4 ${\textcolor{green}{-14}}$ & \textbf{40.4} ${\textcolor{green}{-15}}$ \\ 
  & 86.3 & CnvNxt-B & 38.6 & 32.9 ${\textcolor{green}{-6}}$ & 35.3 ${\textcolor{green}{-3}}$ & 43.6 ${\textcolor{red}{+5}}$ & 36.3 ${\textcolor{green}{-2}}$ & 30.5 ${\textcolor{green}{-8}}$ & 29.9 ${\textcolor{green}{-9}}$ & 31.1 ${\textcolor{green}{-8}}$ & 37.0 ${\textcolor{green}{-2}}$ & 30.0 ${\textcolor{green}{-9}}$ & \textbf{29.5} ${\textcolor{green}{-9}}$ \\ 
  & 84.1 & CnvNxt-T & 44.1 & 37.6 ${\textcolor{green}{-7}}$ & 35.7 ${\textcolor{green}{-8}}$ & 50.7 ${\textcolor{red}{+7}}$ & 36.2 ${\textcolor{green}{-8}}$ & 37.0 ${\textcolor{green}{-7}}$ & \textbf{27.7} ${\textcolor{green}{-16}}$ & 34.0 ${\textcolor{green}{-10}}$ & 44.1 ${\textcolor{green}{-0}}$ & 40.2 ${\textcolor{green}{-4}}$ & 38.9 ${\textcolor{green}{-5}}$ \\ 
  & 82.3 & BiT-m & 59.9 & 52.0 ${\textcolor{green}{-8}}$ & 52.6 ${\textcolor{green}{-7}}$ & 55.3 ${\textcolor{green}{-5}}$ & 30.9 ${\textcolor{green}{-29}}$ & 32.7 ${\textcolor{green}{-27}}$ & \textbf{26.9} ${\textcolor{green}{-33}}$ & 46.3 ${\textcolor{green}{-14}}$ & 37.2 ${\textcolor{green}{-23}}$ & 32.9 ${\textcolor{green}{-27}}$ & 38.2 ${\textcolor{green}{-22}}$ \\ 
  & 85.6 & EffNetv2-M & 43.4 & 42.5 ${\textcolor{green}{-1}}$ & 49.7 ${\textcolor{red}{+6}}$ & 46.3 ${\textcolor{red}{+3}}$ & 43.7 ${\textcolor{red}{+0}}$ & 41.1 ${\textcolor{green}{-2}}$ & 37.0 ${\textcolor{green}{-6}}$ & 89.0 ${\textcolor{red}{+46}}$ & 50.2 ${\textcolor{red}{+7}}$ & \textbf{32.4} ${\textcolor{green}{-11}}$ & 38.5 ${\textcolor{green}{-5}}$ \\ 
 \hline 
{\multirow{12}{*}{{none}}} & 81.1 & ViT-B-384 & 63.5 & 59.4 ${\textcolor{green}{-4}}$ & 58.8 ${\textcolor{green}{-5}}$ & 59.6 ${\textcolor{green}{-4}}$ & 49.1 ${\textcolor{green}{-14}}$ & \textbf{48.2} ${\textcolor{green}{-15}}$ & 61.4 ${\textcolor{green}{-2}}$ & 55.4 ${\textcolor{green}{-8}}$ & 64.0 ${\textcolor{red}{+0}}$ & 59.1 ${\textcolor{green}{-4}}$ & 60.9 ${\textcolor{green}{-3}}$ \\ 
  & 84.6 & Swinv2-B-256 & 63.5 & 63.0 ${\textcolor{green}{-1}}$ & 68.6 ${\textcolor{red}{+5}}$ & 60.9 ${\textcolor{green}{-3}}$ & 49.4 ${\textcolor{green}{-14}}$ & \textbf{46.0} ${\textcolor{green}{-17}}$ & 52.0 ${\textcolor{green}{-11}}$ & 60.5 ${\textcolor{green}{-3}}$ & 57.0 ${\textcolor{green}{-6}}$ & 52.1 ${\textcolor{green}{-11}}$ & 50.4 ${\textcolor{green}{-13}}$ \\ 
  & 85.1 & Deit3-B-384 & 60.0 & 64.8 ${\textcolor{red}{+5}}$ & 83.2 ${\textcolor{red}{+23}}$ & 57.8 ${\textcolor{green}{-2}}$ & 51.2 ${\textcolor{green}{-9}}$ & 48.5 ${\textcolor{green}{-11}}$ & 44.9 ${\textcolor{green}{-15}}$ & 89.2 ${\textcolor{red}{+29}}$ & 65.6 ${\textcolor{red}{+6}}$ & 57.2 ${\textcolor{green}{-3}}$ & \textbf{43.8} ${\textcolor{green}{-16}}$ \\ 
  & 83.8 & Deit3-B-224 & 60.4 & 62.2 ${\textcolor{red}{+2}}$ & 76.1 ${\textcolor{red}{+16}}$ & 58.9 ${\textcolor{green}{-1}}$ & 57.6 ${\textcolor{green}{-3}}$ & 52.8 ${\textcolor{green}{-8}}$ & \textbf{48.9} ${\textcolor{green}{-11}}$ & 80.4 ${\textcolor{red}{+20}}$ & 73.7 ${\textcolor{red}{+13}}$ & 64.4 ${\textcolor{red}{+4}}$ & 49.5 ${\textcolor{green}{-11}}$ \\ 
  & 82.6 & XCiT-M-224 & 65.8 & 65.2 ${\textcolor{green}{-1}}$ & 71.4 ${\textcolor{red}{+6}}$ & 65.4 ${\textcolor{green}{-0}}$ & 58.3 ${\textcolor{green}{-7}}$ & 55.7 ${\textcolor{green}{-10}}$ & \textbf{55.4} ${\textcolor{green}{-10}}$ & 66.9 ${\textcolor{red}{+1}}$ & 63.1 ${\textcolor{green}{-3}}$ & 57.3 ${\textcolor{green}{-8}}$ & 56.4 ${\textcolor{green}{-9}}$ \\ 
  & 84.3 & XCiT-M-224-d & 63.9 & 61.6 ${\textcolor{green}{-2}}$ & 69.9 ${\textcolor{red}{+6}}$ & 61.0 ${\textcolor{green}{-3}}$ & 55.4 ${\textcolor{green}{-8}}$ & 52.8 ${\textcolor{green}{-11}}$ & \textbf{50.4} ${\textcolor{green}{-13}}$ & 66.4 ${\textcolor{red}{+3}}$ & 59.5 ${\textcolor{green}{-4}}$ & 53.6 ${\textcolor{green}{-10}}$ & 52.3 ${\textcolor{green}{-12}}$ \\ 
  & 84.4 & CnvNxt-B & 63.1 & 72.3 ${\textcolor{red}{+9}}$ & 92.1 ${\textcolor{red}{+29}}$ & 62.8 ${\textcolor{green}{-0}}$ & 55.5 ${\textcolor{green}{-8}}$ & 52.1 ${\textcolor{green}{-11}}$ & 53.7 ${\textcolor{green}{-9}}$ & 88.7 ${\textcolor{red}{+26}}$ & 60.8 ${\textcolor{green}{-2}}$ & 53.6 ${\textcolor{green}{-9}}$ & \textbf{50.6} ${\textcolor{green}{-12}}$ \\ 
  & 78.0 & BiT-s & 75.3 & 77.7 ${\textcolor{red}{+2}}$ & 79.8 ${\textcolor{red}{+5}}$ & 59.8 ${\textcolor{green}{-15}}$ & 68.9 ${\textcolor{green}{-6}}$ & \textbf{51.2} ${\textcolor{green}{-24}}$ & 60.1 ${\textcolor{green}{-15}}$ & 65.8 ${\textcolor{green}{-10}}$ & 71.2 ${\textcolor{green}{-4}}$ & 56.0 ${\textcolor{green}{-19}}$ & 84.0 ${\textcolor{red}{+9}}$ \\ 
  & 85.1 & EffNetv2-M & 59.0 & 59.4 ${\textcolor{red}{+0}}$ & 70.5 ${\textcolor{red}{+12}}$ & 56.8 ${\textcolor{green}{-2}}$ & 48.2 ${\textcolor{green}{-11}}$ & \textbf{42.9} ${\textcolor{green}{-16}}$ & 57.4 ${\textcolor{green}{-2}}$ & 59.9 ${\textcolor{red}{+1}}$ & 54.7 ${\textcolor{green}{-4}}$ & 50.2 ${\textcolor{green}{-9}}$ & 43.5 ${\textcolor{green}{-16}}$ \\ 
  & 84.9 & EffNetb7 & 60.1 & 63.4 ${\textcolor{red}{+3}}$ & 75.7 ${\textcolor{red}{+16}}$ & 56.0 ${\textcolor{green}{-4}}$ & 57.9 ${\textcolor{green}{-2}}$ & 47.6 ${\textcolor{green}{-13}}$ & 63.4 ${\textcolor{red}{+3}}$ & 66.6 ${\textcolor{red}{+6}}$ & 58.4 ${\textcolor{green}{-2}}$ & 52.7 ${\textcolor{green}{-7}}$ & \textbf{44.4} ${\textcolor{green}{-16}}$ \\ 
  & 77.7 & EffNet-B0 & 69.3 & 69.9 ${\textcolor{red}{+1}}$ & 77.3 ${\textcolor{red}{+8}}$ & 68.2 ${\textcolor{green}{-1}}$ & 75.9 ${\textcolor{red}{+7}}$ & 68.6 ${\textcolor{green}{-1}}$ & 65.7 ${\textcolor{green}{-4}}$ & 67.8 ${\textcolor{green}{-1}}$ & 77.0 ${\textcolor{red}{+8}}$ & \textbf{51.7} ${\textcolor{green}{-18}}$ & 63.8 ${\textcolor{green}{-6}}$ \\ 
  & 80.4 & ResNet50 & 68.3 & 70.0 ${\textcolor{red}{+2}}$ & 76.6 ${\textcolor{red}{+8}}$ & 64.5 ${\textcolor{green}{-4}}$ & 81.0 ${\textcolor{red}{+13}}$ & 75.9 ${\textcolor{red}{+8}}$ & 73.0 ${\textcolor{red}{+5}}$ & 97.6 ${\textcolor{red}{+29}}$ & 65.9 ${\textcolor{green}{-2}}$ & \textbf{51.7} ${\textcolor{green}{-17}}$ & 55.0 ${\textcolor{green}{-13}}$ \\ 
 \hline 
{\multirow{1}{*}{{JFT}}} & 86.8 & EffNetb7-ns & 53.8 & 49.9 ${\textcolor{green}{-4}}$ & 62.5 ${\textcolor{red}{+9}}$ & 52.7 ${\textcolor{green}{-1}}$ & 79.5 ${\textcolor{red}{+26}}$ & 53.2 ${\textcolor{green}{-1}}$ & 82.4 ${\textcolor{red}{+29}}$ & 57.0 ${\textcolor{red}{+3}}$ & 55.0 ${\textcolor{red}{+1}}$ & 47.0 ${\textcolor{green}{-7}}$ & \textbf{46.5} ${\textcolor{green}{-7}}$ \\ 
 \hline 
{\multirow{2}{*}{{\shortstack[l]{clip\\+12k}}}} & 87.2 & ViT-B-384-l2b & 37.3 & 33.7 ${\textcolor{green}{-4}}$ & 35.6 ${\textcolor{green}{-2}}$ & 40.5 ${\textcolor{red}{+3}}$ & 43.6 ${\textcolor{red}{+6}}$ & 36.9 ${\textcolor{green}{-0}}$ & 36.7 ${\textcolor{green}{-1}}$ & 31.6 ${\textcolor{green}{-6}}$ & 35.0 ${\textcolor{green}{-2}}$ & 29.5 ${\textcolor{green}{-8}}$ & \textbf{29.3} ${\textcolor{green}{-8}}$ \\ 
  & 87.0 & ViT-B-384-oai & 38.7 & 33.1 ${\textcolor{green}{-6}}$ & 32.9 ${\textcolor{green}{-6}}$ & 40.7 ${\textcolor{red}{+2}}$ & 45.9 ${\textcolor{red}{+7}}$ & 37.4 ${\textcolor{green}{-1}}$ & 38.4 ${\textcolor{green}{-0}}$ & 31.2 ${\textcolor{green}{-8}}$ & 33.7 ${\textcolor{green}{-5}}$ & 29.2 ${\textcolor{green}{-9}}$ & \textbf{29.1} ${\textcolor{green}{-10}}$ \\ 
 \hline 
{\multirow{2}{*}{{clip}}} & 86.6 & ViT-B-384-l2b & 54.2 & 52.5 ${\textcolor{green}{-2}}$ & 57.2 ${\textcolor{red}{+3}}$ & 51.0 ${\textcolor{green}{-3}}$ & 40.2 ${\textcolor{green}{-14}}$ & 40.4 ${\textcolor{green}{-14}}$ & \textbf{38.4} ${\textcolor{green}{-16}}$ & 54.0 ${\textcolor{green}{-0}}$ & 44.0 ${\textcolor{green}{-10}}$ & 40.0 ${\textcolor{green}{-14}}$ & 39.5 ${\textcolor{green}{-15}}$ \\ 
  & 86.2 & ViT-B-384-oai & 56.7 & 55.0 ${\textcolor{green}{-2}}$ & 59.0 ${\textcolor{red}{+2}}$ & 53.9 ${\textcolor{green}{-3}}$ & 40.6 ${\textcolor{green}{-16}}$ & 40.8 ${\textcolor{green}{-16}}$ & 41.4 ${\textcolor{green}{-15}}$ & 56.0 ${\textcolor{green}{-1}}$ & 45.6 ${\textcolor{green}{-11}}$ & 41.3 ${\textcolor{green}{-15}}$ & \textbf{40.3} ${\textcolor{green}{-16}}$ \\ 
 \hline 
{\multirow{2}{*}{\shortstack[l]{clip\\z. shot}}} & 74.3 & clip-ViT-L-336 & ---- & ---- & ---- & ---- & ---- & ---- & ---- & ---- & ---- & 64.4 & \textbf{51.8} \\ 
  & 66.6 & clip-ViT-B-224 & ---- & ---- & ---- & ---- & ---- & ---- & ---- & ---- & ---- & 71.4 & \textbf{60.0} \\ 
 \hline 
\end{tabular}
\end{center}

\label{tab:overview-fpr-subsampled}
\end{smaller}
\vskip \belowtablevskip
\end{table*}

\renewcommand{\arraystretch}{.9}
\tabcolsep=0.0310321cm

\begin{table}[htb]
\rowcolors{2}{white}{gray!10}
    \centering
    \caption{Comparing the cleaned and original datasets in terms of \FPR. The best method per model and dataset is marked bold.}
    \label{tab:clean-vs-pre-vit}
    \tiny
\begin{center}
\begin{tabular}{c c c | c  c  c  c  c  c  c  c  c  c  c  c  c  c  c  c  c  c  c  c  c  c  c }
 &  & &   &   &   &   &   &   &   &   &   &   &   & fpr &   &   &   &   &   &   &   &   &   &   &   \\ 
 model &acc. & method & Pl-f & Pl-c & Spc-f & Spc-c & IN-f & IN-c & txt-f & txt-43 & txt-c & OpO-f & OpO-c & iNat-f & iNat-c & IN1K-f & IN1K-c & OS-f & OS-c & SBe-f & SBe-c & SBh-f & SBh-c & CO-f & CO-c \\ 
 \hline 
 &  & MSP & 60.5 & 37.9 & 65.8 & 41.9 & 63.0 & 58.3 & 54.2 & 52.3 & 43.4 & 28.2 & 27.2 & 10.5 & 8.9 & 83.0 & 61.3 & 71.0 & 32.6 & 70.7 & 33.8 & 77.6 & 53.4 & 48.5 & 38.2 \\ 
  &  & MaxL & 50.6 & 27.5 & 65.2 & 33.7 & 46.0 & 40.8 & 36.5 & 33.5 & 21.2 & 12.2 & 12.2 & 3.5 & 2.1 & 75.2 & 54.9 & 58.9 & 21.2 & $\bm{59.0}$ & 16.6 & 66.7 & 43.8 & 33.2 & 22.6 \\ 
  &  & ViM & 49.9 & 26.1 & 53.8 & 22.1 & 38.2 & 35.3 & $\bm{24.0}$ & $\bm{21.2}$ & $\bm{12.2}$ & 12.5 & 11.7 & $\bm{1.5}$ & $\bm{0.5}$ & 78.8 & 57.4 & $\bm{54.8}$ & 14.4 & 60.2 & 17.9 & $\bm{59.9}$ & $\bm{32.2}$ & $\bm{26.8}$ & $\bm{16.0}$ \\ 
  &  & Maha & 57.5 & 34.6 & $\bm{47.5}$ & $\bm{15.7}$ & $\bm{35.2}$ & $\bm{29.1}$ & 28.7 & 25.7 & 15.6 & $\bm{9.5}$ & $\bm{9.5}$ & 2.0 & 0.8 & 74.5 & $\bm{52.8}$ & $\bm{54.8}$ & $\bm{12.3}$ & 65.9 & 17.9 & 64.6 & 41.8 & 29.0 & 16.5 \\ 
  &  & E+R & 53.3 & 30.7 & 60.2 & 32.0 & 43.8 & 38.5 & 34.5 & 31.6 & 20.1 & 10.5 & 10.1 & 2.8 & 1.6 & 83.0 & 63.8 & $\bm{54.8}$ & 17.8 & 63.1 & $\bm{15.9}$ & 68.8 & 50.5 & 32.5 & 21.7 \\ 
 ViT-B-384-21k & 86.0 & Ener & $\bm{49.1}$ & $\bm{25.5}$ & 64.2 & 30.8 & 43.2 & 38.5 & 35.0 & 32.4 & 20.8 & 11.0 & 10.6 & 3.2 & 1.8 & 75.8 & 57.4 & 58.9 & 17.4 & 60.2 & 16.6 & 64.6 & 42.3 & 31.5 & 21.2 \\ 
  &  & KL-M & 64.4 & 43.1 & 68.5 & 39.5 & 57.5 & 53.7 & 50.7 & 48.8 & 38.5 & 26.8 & 25.8 & 8.5 & 6.8 & 82.4 & 62.1 & 69.4 & 29.2 & 71.9 & 33.8 & 77.1 & 51.4 & 48.8 & 37.7 \\ 
  &  & KNN & 69.4 & 50.3 & 81.5 & 59.9 & 50.2 & 46.0 & 38.2 & 36.2 & 25.3 & 36.5 & 37.5 & 40.8 & 41.0 & 91.5 & 86.0 & 64.5 & 42.4 & 67.9 & 24.5 & 75.0 & 71.2 & 54.2 & 46.2 \\ 
  &  & RMaha & 55.8 & 30.1 & 52.0 & 20.3 & 42.8 & 36.6 & 37.8 & 35.1 & 23.3 & 12.2 & 12.2 & 2.0 & 0.8 & $\bm{72.7}$ & 53.2 & 61.3 & 16.9 & 68.7 & 19.9 & 71.4 & 44.2 & 37.8 & 23.1 \\ 
  &  & RCos & 56.8 & 33.3 & 67.8 & 35.5 & 40.8 & 35.3 & 34.5 & 31.4 & 19.1 & 17.0 & 16.8 & 7.2 & 6.0 & 83.0 & 67.2 & 64.5 & 24.2 & 62.7 & 17.2 & 72.4 & 51.0 & 39.5 & 28.8 \\ 
  &  & Cos & 57.0 & 32.7 & 67.0 & 33.7 & 42.5 & 36.6 & 36.2 & 33.2 & 20.8 & 15.5 & 15.2 & 6.2 & 5.0 & 83.0 & 66.0 & 66.1 & 25.4 & 62.7 & 19.9 & 72.4 & 51.4 & 40.0 & 30.2 \\ 
 \hline 
 &  & MSP & 59.5 & 39.9 & 66.2 & 40.7 & 67.8 & 62.8 & 51.2 & 49.3 & 38.5 & 35.2 & 36.1 & 15.2 & 14.1 & 89.7 & 65.5 & 69.4 & 39.4 & 79.5 & 35.8 & 81.8 & 58.7 & 53.2 & 44.3 \\ 
  &  & MaxL & 50.1 & 28.1 & 65.5 & 35.5 & 50.7 & 45.3 & 38.0 & 35.7 & 24.0 & 18.0 & 18.2 & 6.8 & 5.2 & 81.2 & 59.6 & 62.1 & 23.7 & 65.9 & 21.9 & 74.0 & 48.1 & 39.2 & 28.8 \\ 
  &  & ViM & $\bm{48.4}$ & 27.5 & 58.0 & 26.2 & 41.0 & 36.9 & $\bm{23.0}$ & $\bm{20.1}$ & $\bm{12.8}$ & 14.5 & 13.6 & 3.0 & 1.8 & 79.4 & 57.4 & $\bm{58.1}$ & 19.1 & $\bm{61.0}$ & $\bm{17.2}$ & $\bm{64.1}$ & $\bm{38.5}$ & $\bm{29.0}$ & $\bm{20.3}$ \\ 
  &  & Maha & 58.3 & 32.0 & $\bm{49.5}$ & $\bm{16.3}$ & $\bm{39.8}$ & $\bm{34.6}$ & 27.0 & 24.7 & 13.2 & $\bm{10.2}$ & $\bm{10.1}$ & $\bm{2.0}$ & $\bm{0.5}$ & 78.2 & $\bm{54.0}$ & $\bm{58.1}$ & $\bm{14.4}$ & 69.9 & 21.9 & 68.2 & 43.8 & 32.2 & 20.8 \\ 
  &  & E+R & 53.1 & 27.5 & 64.8 & 36.6 & 50.7 & 45.3 & 38.8 & 35.9 & 25.3 & 17.0 & 17.1 & 7.2 & 5.7 & 87.3 & 69.4 & 60.5 & 24.6 & 65.5 & 24.5 & 75.5 & 54.8 & 38.0 & 27.8 \\ 
 ViT-B-224-21k & 84.5 & Ener & 48.9 & $\bm{26.8}$ & 65.2 & 34.9 & 46.8 & 41.7 & 35.8 & 33.2 & 22.2 & 14.0 & 13.9 & 6.8 & 5.2 & 78.8 & 60.4 & 63.7 & 23.3 & 63.1 & 20.5 & 68.8 & 47.6 & 37.0 & 25.9 \\ 
  &  & KL-M & 64.4 & 43.8 & 69.0 & 38.4 & 62.3 & 58.3 & 50.5 & 48.5 & 37.8 & 34.2 & 35.3 & 12.5 & 10.7 & 87.9 & 63.4 & 66.9 & 35.2 & 77.5 & 41.1 & 83.9 & 60.1 & 53.8 & 45.3 \\ 
  &  & KNN & 70.4 & 53.6 & 83.5 & 65.7 & 56.5 & 52.1 & 37.2 & 35.7 & 26.0 & 44.5 & 45.7 & 39.5 & 39.2 & 90.9 & 86.8 & 72.6 & 49.6 & 75.9 & 29.8 & 81.8 & 75.0 & 65.0 & 62.7 \\ 
  &  & RMaha & 60.2 & 37.9 & 51.0 & 17.4 & 47.5 & 41.7 & 39.0 & 36.2 & 24.0 & 16.0 & 15.8 & 2.2 & $\bm{0.5}$ & $\bm{77.0}$ & 54.9 & 62.1 & 20.3 & 77.5 & 25.8 & 76.0 & 46.2 & 40.8 & 25.5 \\ 
  &  & RCos & 61.7 & 39.9 & 71.2 & 42.4 & 45.2 & 41.1 & 36.2 & 33.8 & 22.6 & 23.0 & 23.6 & 12.8 & 11.2 & 83.6 & 70.2 & 71.0 & 30.9 & 71.1 & 22.5 & 78.6 & 57.7 & 45.0 & 35.4 \\ 
  &  & Cos & 60.0 & 39.2 & 70.5 & 41.9 & 51.5 & 46.9 & 35.8 & 33.2 & 22.6 & 22.0 & 22.6 & 11.5 & 10.2 & 85.5 & 71.1 & 70.2 & 31.8 & 71.5 & 22.5 & 77.6 & 59.6 & 48.0 & 38.2 \\ 
 \hline 
 &  & MSP & 58.8 & 37.9 & 65.0 & 44.2 & 62.0 & 59.9 & 53.5 & 51.5 & 44.1 & 34.5 & 34.5 & 19.8 & 17.5 & 80.6 & 61.7 & 70.2 & 33.1 & 67.9 & 35.8 & 76.6 & 54.3 & 48.0 & 38.2 \\ 
  &  & MaxL & 53.8 & 30.7 & 61.3 & 36.6 & 48.5 & 45.6 & 50.7 & 48.8 & 41.3 & 25.2 & 25.0 & 13.2 & 11.2 & 75.2 & 47.7 & 64.5 & 22.9 & 69.5 & 27.2 & 64.6 & 42.8 & 35.2 & 24.5 \\ 
  &  & ViM & 49.9 & 25.5 & 65.0 & 32.6 & 59.0 & 53.4 & 37.8 & 36.2 & 25.7 & $\bm{14.2}$ & $\bm{14.1}$ & $\bm{1.8}$ & $\bm{1.0}$ & 92.1 & 86.8 & 66.9 & 29.2 & 67.9 & 32.5 & 82.3 & 74.0 & 44.5 & 37.7 \\ 
  &  & Maha & 58.8 & 36.6 & 67.8 & 38.4 & 71.0 & 68.0 & 47.0 & 45.6 & 36.5 & 22.2 & 22.3 & 3.5 & 2.6 & 93.3 & 87.7 & 70.2 & 44.1 & 71.5 & 49.0 & 89.6 & 84.6 & 56.5 & 51.9 \\ 
  &  & E+R & $\bm{46.9}$ & $\bm{21.6}$ & $\bm{59.2}$ & $\bm{30.8}$ & $\bm{42.0}$ & $\bm{38.8}$ & 49.0 & 46.6 & 37.8 & 21.8 & 21.5 & 9.0 & 7.3 & 75.2 & 48.1 & $\bm{55.6}$ & $\bm{17.8}$ & $\bm{65.5}$ & 23.8 & $\bm{62.5}$ & 37.5 & $\bm{27.8}$ & $\bm{20.3}$ \\ 
 Swinv2-B-256-21k & 86.3 & Ener & 54.8 & 30.7 & 62.0 & 35.5 & 43.0 & 40.1 & 54.8 & 52.8 & 45.5 & 30.8 & 29.9 & 13.8 & 12.0 & $\bm{73.9}$ & $\bm{44.3}$ & 62.1 & 21.2 & 70.7 & 30.5 & 64.6 & $\bm{34.6}$ & 31.8 & 22.2 \\ 
  &  & KL-M & 64.2 & 47.7 & 70.0 & 44.8 & 66.0 & 64.1 & 52.5 & 50.9 & 42.0 & 33.2 & 33.7 & 20.0 & 17.8 & 86.7 & 70.6 & 67.7 & 39.4 & 69.9 & 36.4 & 82.3 & 66.3 & 56.2 & 47.6 \\ 
  &  & KNN & 60.2 & 35.3 & 73.5 & 44.8 & 63.2 & 58.6 & 38.8 & 37.3 & 27.4 & 20.2 & 20.1 & 7.8 & 6.5 & 95.2 & 90.2 & 66.1 & 34.7 & 70.3 & 29.8 & 84.4 & 79.8 & 53.0 & 46.7 \\ 
  &  & RMaha & 57.5 & 35.3 & 62.7 & 32.0 & 67.8 & 64.7 & 44.5 & 42.9 & 34.0 & 17.5 & 17.4 & 4.2 & 3.1 & 87.9 & 71.1 & 66.9 & 33.1 & 71.9 & 40.4 & 87.5 & 70.7 & 51.0 & 42.5 \\ 
  &  & RCos & 56.3 & 32.7 & 65.8 & 36.6 & 55.2 & 50.5 & 36.0 & 33.8 & 22.9 & 14.8 & 14.4 & 4.2 & 2.9 & 87.3 & 69.8 & 61.3 & 26.7 & 69.1 & $\bm{21.9}$ & 82.8 & 62.5 & 45.0 & 34.9 \\ 
  &  & Cos & 52.1 & 26.1 & 67.0 & 37.8 & 57.5 & 53.4 & $\bm{34.5}$ & $\bm{32.7}$ & $\bm{22.2}$ & $\bm{14.2}$ & $\bm{14.1}$ & 4.0 & 2.6 & 89.7 & 74.5 & 62.1 & 28.4 & 69.5 & 26.5 & 84.9 & 67.3 & 47.0 & 37.7 \\ 
 \hline 
 &  & MSP & 67.2 & 52.9 & 72.8 & 54.7 & 73.5 & 72.2 & 64.8 & 63.5 & 55.9 & 49.5 & 48.9 & 30.0 & 27.9 & 90.9 & 66.8 & 78.2 & 41.5 & 76.7 & 47.0 & 83.9 & 65.9 & 62.3 & 53.3 \\ 
  &  & MaxL & 61.5 & 41.8 & 72.8 & 46.5 & 59.5 & 57.0 & 63.5 & 61.7 & 53.1 & 43.8 & 43.5 & 29.0 & 27.2 & 83.0 & 60.0 & 70.2 & 37.3 & 76.3 & 33.1 & 78.1 & $\bm{53.4}$ & 54.5 & 46.2 \\ 
  &  & ViM & $\bm{50.1}$ & 28.1 & 68.8 & 43.6 & 63.7 & 61.5 & 50.2 & 48.3 & 38.9 & $\bm{20.5}$ & $\bm{19.8}$ & $\bm{3.8}$ & $\bm{1.8}$ & 92.7 & 86.4 & 66.1 & 32.6 & $\bm{62.7}$ & 29.1 & 79.7 & 72.6 & 45.5 & 38.7 \\ 
  &  & Maha & 52.6 & 29.4 & 66.0 & 40.1 & 66.2 & 63.8 & 48.2 & 46.1 & 37.5 & 22.2 & 22.0 & 5.8 & 3.7 & 90.3 & 79.1 & 66.1 & 33.5 & 65.1 & 32.5 & 82.3 & 68.8 & 46.5 & 38.7 \\ 
  &  & E+R & 56.0 & 32.7 & 73.0 & 45.9 & $\bm{51.7}$ & $\bm{49.5}$ & 58.2 & 55.8 & 45.5 & 34.8 & 33.7 & 23.2 & 21.9 & $\bm{81.2}$ & 62.6 & $\bm{63.7}$ & 34.7 & 69.1 & $\bm{22.5}$ & $\bm{73.4}$ & $\bm{53.4}$ & 44.0 & $\bm{36.3}$ \\ 
 Deit3-B-384-21k & 86.7 & Ener & 62.7 & 43.1 & 76.5 & 53.5 & 53.2 & 51.1 & 67.0 & 64.9 & 56.6 & 47.2 & 47.0 & 39.8 & 38.6 & 81.8 & $\bm{56.6}$ & 66.9 & 36.0 & 75.1 & 31.8 & 75.5 & $\bm{53.4}$ & 51.7 & 42.5 \\ 
  &  & KL-M & 67.9 & 50.3 & 72.0 & 52.3 & 69.5 & 68.9 & 59.0 & 57.4 & 49.7 & 47.8 & 47.3 & 28.2 & 26.1 & 91.5 & 71.9 & 75.8 & 41.9 & 72.7 & 45.0 & 85.9 & 69.7 & 62.5 & 54.7 \\ 
  &  & KNN & 53.3 & 28.1 & 72.0 & 45.9 & 59.2 & 56.0 & 44.8 & 42.1 & 32.6 & $\bm{20.5}$ & 20.1 & 9.0 & 7.0 & 92.7 & 81.3 & 65.3 & 33.5 & 63.1 & 27.2 & 82.3 & 70.2 & 45.0 & 38.7 \\ 
  &  & RMaha & 51.6 & 28.1 & $\bm{64.2}$ & $\bm{39.0}$ & 62.0 & 59.5 & 46.2 & 43.4 & 34.0 & 21.5 & 21.2 & 6.8 & 4.7 & 90.3 & 71.9 & 64.5 & 30.1 & 64.3 & 29.1 & 80.7 & 62.0 & $\bm{43.5}$ & $\bm{36.3}$ \\ 
  &  & RCos & 52.8 & 27.5 & 67.8 & 41.3 & 57.0 & 54.7 & $\bm{42.2}$ & $\bm{39.7}$ & $\bm{29.5}$ & 21.2 & 20.7 & 6.5 & 4.7 & 90.3 & 68.9 & 65.3 & $\bm{28.0}$ & 65.1 & 24.5 & 81.8 & 59.1 & $\bm{43.5}$ & 37.3 \\ 
  &  & Cos & 52.8 & $\bm{26.8}$ & 68.0 & 41.3 & 58.5 & 55.7 & 42.8 & 40.2 & 29.9 & 21.5 & 20.9 & 6.5 & 4.7 & 90.3 & 69.8 & 65.3 & $\bm{28.0}$ & 64.7 & 25.8 & 82.8 & 59.6 & 43.8 & 37.3 \\ 
 \hline 
 &  & MSP & 65.4 & 48.4 & 73.2 & 53.5 & 73.5 & 72.5 & 66.8 & 64.6 & 59.4 & 47.2 & 45.9 & 36.2 & 35.0 & 89.7 & 70.2 & 75.8 & 51.7 & 75.1 & 49.0 & 82.8 & 67.3 & 61.5 & 53.3 \\ 
  &  & MaxL & 61.0 & 43.8 & 72.0 & 48.3 & 61.0 & 58.6 & 64.5 & 62.5 & 55.6 & 41.2 & 39.9 & 36.5 & 35.5 & 84.8 & 58.7 & 63.7 & 41.1 & 69.9 & 36.4 & 76.0 & 56.2 & 54.5 & 42.0 \\ 
  &  & ViM & 54.8 & 34.0 & 74.8 & 50.0 & 68.2 & 66.7 & 53.2 & 51.5 & 43.1 & 25.5 & 25.5 & $\bm{7.5}$ & $\bm{5.2}$ & 92.7 & 91.5 & 68.5 & 39.0 & 67.9 & 41.1 & 84.9 & 79.3 & 52.5 & 47.6 \\ 
  &  & Maha & 53.3 & 33.3 & 71.0 & 45.9 & 70.2 & 68.9 & 52.5 & 50.9 & 42.7 & 26.0 & 26.1 & 8.0 & 5.7 & 92.7 & 85.5 & 70.2 & 39.0 & 69.9 & 43.0 & 84.9 & 74.0 & 53.2 & 48.1 \\ 
  &  & E+R & 57.0 & 34.6 & 73.0 & 47.7 & $\bm{52.2}$ & $\bm{50.5}$ & 59.0 & 56.8 & 48.6 & 33.8 & 33.4 & 38.8 & 37.6 & 83.6 & 56.6 & $\bm{60.5}$ & 36.9 & $\bm{65.1}$ & $\bm{26.5}$ & 72.9 & 52.9 & 49.0 & $\bm{37.3}$ \\ 
 Deit3-B-224-21k & 85.7 & Ener & 63.5 & 47.1 & 74.2 & 51.2 & 55.0 & 52.8 & 63.5 & 61.4 & 54.2 & 44.2 & 43.8 & 53.0 & 52.5 & $\bm{81.2}$ & $\bm{53.2}$ & 63.7 & 38.6 & 67.9 & 33.1 & $\bm{70.8}$ & $\bm{51.0}$ & 53.2 & 42.5 \\ 
  &  & KL-M & 68.1 & 51.6 & 74.5 & 55.8 & 71.8 & 70.9 & 63.0 & 61.1 & 55.6 & 48.2 & 46.5 & 34.2 & 32.9 & 91.5 & 75.7 & 74.2 & 55.5 & 75.1 & 47.7 & 84.9 & 69.2 & 63.7 & 55.7 \\ 
  &  & KNN & 56.3 & 35.9 & 75.8 & 51.2 & 62.7 & 60.2 & 47.8 & 45.3 & 36.8 & 24.2 & 23.9 & 14.8 & 12.8 & 93.9 & 83.8 & 70.2 & 42.8 & 65.9 & 29.1 & 85.4 & 73.6 & 51.0 & 45.8 \\ 
  &  & RMaha & $\bm{52.1}$ & $\bm{31.4}$ & $\bm{67.8}$ & $\bm{41.9}$ & 66.8 & 65.7 & 52.0 & 50.4 & 42.7 & $\bm{23.2}$ & $\bm{23.1}$ & 7.8 & 5.7 & 90.9 & 76.6 & 69.4 & $\bm{34.3}$ & 67.5 & 37.1 & 84.4 & 67.8 & 49.8 & 42.5 \\ 
  &  & RCos & 53.3 & 32.0 & 71.0 & 44.8 & 60.2 & 58.6 & $\bm{47.2}$ & $\bm{44.8}$ & $\bm{36.1}$ & 23.8 & $\bm{23.1}$ & 9.0 & 7.0 & 89.7 & 73.2 & 65.3 & 35.6 & 66.7 & 29.8 & 84.9 & 65.9 & $\bm{47.5}$ & 38.2 \\ 
  &  & Cos & 53.6 & 32.7 & 71.8 & 45.9 & 60.5 & 59.2 & 47.8 & 45.3 & 37.2 & 24.2 & 23.6 & 9.2 & 7.3 & 89.1 & 74.9 & 67.7 & 36.9 & 65.9 & 31.1 & 84.9 & 66.8 & 48.0 & 39.6 \\ 
 \hline 
 &  & MSP & 55.8 & 38.6 & 64.0 & 42.4 & 54.5 & 49.5 & 47.5 & 45.0 & 36.1 & 26.5 & 26.4 & 16.0 & 14.1 & 80.6 & 60.0 & 64.5 & 28.8 & 69.9 & 29.1 & 69.8 & 54.3 & 41.2 & 30.7 \\ 
  &  & MaxL & 55.1 & 35.9 & 65.0 & 39.5 & 52.2 & 46.6 & 46.0 & 43.7 & 34.0 & 24.8 & 24.5 & 13.0 & 11.2 & 78.8 & 54.9 & 60.5 & 24.2 & 65.9 & 25.2 & 63.0 & 48.1 & $\bm{36.5}$ & $\bm{26.4}$ \\ 
  &  & ViM & $\bm{43.5}$ & $\bm{17.0}$ & 64.0 & 34.3 & 60.2 & 55.7 & 41.2 & 38.6 & 27.1 & 14.0 & 13.6 & $\bm{4.5}$ & 3.1 & 83.0 & 78.3 & 67.7 & 37.7 & 63.5 & 35.1 & 78.6 & 66.8 & 42.2 & 35.4 \\ 
  &  & Maha & 48.9 & 26.1 & 67.2 & 40.1 & 65.2 & 61.5 & 46.0 & 43.2 & 32.6 & 20.0 & 19.3 & 8.2 & 6.8 & 87.9 & 84.7 & 69.4 & 45.3 & 68.3 & 45.7 & 83.3 & 75.0 & 49.8 & 42.0 \\ 
  &  & E+R & 52.8 & 34.0 & 69.2 & 43.0 & $\bm{45.0}$ & $\bm{41.4}$ & 43.0 & 40.5 & 30.6 & 24.2 & 24.2 & 12.8 & 11.5 & $\bm{77.0}$ & $\bm{51.9}$ & 62.1 & $\bm{21.2}$ & $\bm{63.1}$ & 21.2 & 60.9 & $\bm{40.9}$ & $\bm{36.5}$ & 27.4 \\ 
 ViT-B-384-l2b-12k & 87.2 & Ener & 60.0 & 41.2 & 71.0 & 45.9 & 49.5 & 46.0 & 50.0 & 48.0 & 39.6 & 29.2 & 28.3 & 16.2 & 14.9 & $\bm{77.0}$ & 52.8 & 62.1 & 22.5 & 67.9 & 27.2 & $\bm{60.4}$ & 42.3 & 40.2 & 31.1 \\ 
  &  & KL-M & 55.8 & 37.9 & 64.2 & 40.7 & 56.5 & 52.8 & 43.8 & 41.0 & 33.0 & 27.5 & 27.4 & 19.0 & 17.0 & 84.2 & 70.6 & 63.7 & 33.1 & 71.9 & 33.1 & 76.6 & 62.0 & 46.5 & 37.7 \\ 
  &  & KNN & 46.7 & 25.5 & 67.8 & 38.4 & 54.8 & 49.8 & $\bm{34.8}$ & $\bm{31.6}$ & 19.8 & 15.8 & 14.9 & 7.0 & 5.7 & 89.7 & 80.4 & 59.7 & 29.2 & 65.1 & 21.2 & 78.1 & 67.8 & 41.8 & 32.5 \\ 
  &  & RMaha & 49.4 & 22.9 & 63.5 & 34.3 & 60.5 & 56.6 & 42.2 & 39.7 & 28.5 & 16.5 & 15.8 & 6.0 & 4.4 & 86.7 & 77.9 & 64.5 & 33.9 & 65.1 & 32.5 & 83.3 & 65.4 & 44.5 & 34.0 \\ 
  &  & RCos & 46.4 & 22.9 & $\bm{60.5}$ & $\bm{29.7}$ & 50.0 & 45.3 & 35.2 & 32.2 & 19.4 & $\bm{12.2}$ & $\bm{11.7}$ & $\bm{4.5}$ & $\bm{2.9}$ & 84.2 & 66.0 & 58.9 & 22.9 & 65.1 & $\bm{19.2}$ & 76.0 & 55.8 & 38.0 & 26.9 \\ 
  &  & Cos & 46.9 & 22.9 & $\bm{60.5}$ & $\bm{29.7}$ & 50.0 & 45.3 & 35.0 & 31.9 & $\bm{19.1}$ & $\bm{12.2}$ & $\bm{11.7}$ & $\bm{4.5}$ & $\bm{2.9}$ & 84.2 & 66.8 & $\bm{58.1}$ & 23.3 & 65.1 & 19.9 & 75.5 & 56.2 & 38.0 & 26.9 \\ 
 \hline 
 &  & MSP & 57.0 & 39.9 & 65.0 & 41.3 & 55.0 & 52.8 & 50.7 & 48.8 & 40.3 & 30.2 & 30.4 & 17.0 & 14.9 & 80.0 & 61.7 & 65.3 & 27.5 & 70.3 & 27.8 & 72.9 & 51.9 & 45.0 & 37.7 \\ 
  &  & MaxL & 55.6 & 36.6 & 65.0 & 39.5 & 48.0 & 45.3 & 47.0 & 44.8 & 35.1 & 26.5 & 26.4 & 14.5 & 12.3 & 75.8 & $\bm{51.9}$ & 61.3 & 19.5 & 66.3 & 23.8 & 65.6 & 45.2 & 36.8 & 28.8 \\ 
  &  & ViM & $\bm{53.1}$ & $\bm{27.5}$ & 64.0 & 33.1 & 62.7 & 61.2 & 37.8 & 35.7 & 24.7 & 16.0 & 16.3 & $\bm{5.5}$ & $\bm{3.9}$ & 86.7 & 83.8 & 62.9 & 34.3 & 65.5 & 29.1 & 78.6 & 69.2 & 44.0 & 39.2 \\ 
  &  & Maha & 60.0 & 37.9 & 66.2 & 37.8 & 68.5 & 67.0 & 41.0 & 39.1 & 27.4 & 22.5 & 22.8 & 11.2 & 9.9 & 87.9 & 84.7 & 68.5 & 47.0 & 74.7 & 44.4 & 82.3 & 76.4 & 54.8 & 50.0 \\ 
  &  & E+R & 55.6 & 37.3 & 67.0 & 39.0 & $\bm{43.2}$ & $\bm{40.1}$ & 46.8 & 44.8 & 34.4 & 23.8 & 23.4 & 12.5 & 10.7 & 75.8 & $\bm{51.9}$ & $\bm{58.1}$ & $\bm{17.4}$ & 62.2 & 22.5 & 60.4 & $\bm{40.9}$ & $\bm{34.0}$ & 25.5 \\ 
 ViT-B-384-oai-12k & 87.0 & Ener & 56.3 & 38.6 & 68.8 & 40.7 & 45.0 & 41.4 & 50.5 & 48.5 & 39.2 & 26.8 & 26.6 & 14.5 & 12.8 & $\bm{73.9}$ & $\bm{51.9}$ & 60.5 & 18.6 & 64.7 & 25.2 & $\bm{58.9}$ & 41.8 & 35.0 & $\bm{24.5}$ \\ 
  &  & KL-M & 61.7 & 43.1 & 67.0 & 41.9 & 55.2 & 53.7 & 48.0 & 46.4 & 37.8 & 29.0 & 29.1 & 16.2 & 14.1 & 84.2 & 67.7 & 66.9 & 30.1 & 70.3 & 30.5 & 77.6 & 58.7 & 48.2 & 40.6 \\ 
  &  & KNN & 57.5 & 34.6 & 67.8 & 39.0 & 47.2 & 44.0 & $\bm{30.2}$ & $\bm{27.9}$ & $\bm{15.3}$ & 15.0 & 15.2 & 8.0 & 6.3 & 89.1 & 77.0 & 62.1 & 28.0 & $\bm{61.0}$ & $\bm{17.2}$ & 75.5 & 60.6 & 40.2 & 33.0 \\ 
  &  & RMaha & 58.0 & 35.3 & 62.3 & 32.0 & 60.8 & 58.3 & 40.8 & 38.6 & 27.1 & 16.8 & 17.1 & 8.0 & 6.5 & 86.1 & 73.6 & 61.3 & 31.8 & 71.5 & 29.1 & 80.2 & 61.5 & 48.5 & 39.2 \\ 
  &  & RCos & 54.8 & 29.4 & $\bm{61.5}$ & $\bm{29.1}$ & 46.2 & 41.7 & 33.0 & 30.6 & 18.4 & $\bm{12.0}$ & $\bm{12.5}$ & 6.5 & 5.0 & 84.8 & 64.7 & 58.9 & 22.9 & 63.9 & $\bm{17.2}$ & 70.8 & 52.4 & 37.5 & 27.4 \\ 
  &  & Cos & 55.3 & 30.7 & $\bm{61.5}$ & $\bm{29.1}$ & 46.0 & 41.4 & 32.5 & 30.0 & 17.7 & $\bm{12.0}$ & $\bm{12.5}$ & 6.5 & 5.0 & 84.8 & 66.0 & 58.9 & 22.0 & 63.9 & 18.5 & 70.8 & 51.4 & 37.0 & 27.4 \\ 
 \hline 
 &  & MSP & 63.7 & 50.3 & 67.8 & 48.8 & 78.2 & 76.1 & 55.2 & 52.5 & 44.1 & 43.5 & 44.3 & 37.2 & 35.8 & 83.6 & 70.2 & 72.6 & 53.4 & 73.9 & 49.0 & 77.6 & 69.7 & 60.2 & 54.2 \\ 
  &  & MaxL & 62.0 & 44.4 & 71.2 & 51.7 & 73.2 & 71.2 & 55.0 & 52.0 & 43.4 & 38.2 & 39.4 & 37.2 & 36.3 & 83.0 & $\bm{67.2}$ & 71.0 & 51.3 & 75.5 & 49.7 & 77.6 & 69.7 & 60.2 & 52.8 \\ 
  &  & ViM & $\bm{50.9}$ & $\bm{23.5}$ & 66.2 & 37.2 & $\bm{59.5}$ & $\bm{57.6}$ & 40.2 & 37.3 & 25.7 & $\bm{18.0}$ & $\bm{18.8}$ & 11.0 & 9.4 & 84.8 & 77.9 & 65.3 & 41.1 & $\bm{61.8}$ & $\bm{21.9}$ & $\bm{74.5}$ & 68.8 & 48.2 & 40.1 \\ 
  &  & Maha & 56.3 & 32.7 & 65.0 & $\bm{36.6}$ & 63.0 & 61.2 & 39.0 & 36.2 & 26.0 & 19.0 & 19.8 & $\bm{10.8}$ & $\bm{8.9}$ & 84.2 & 77.4 & $\bm{62.1}$ & 40.3 & 63.1 & 28.5 & 76.0 & 69.7 & 49.5 & 40.6 \\ 
  &  & E+R & 59.5 & 38.6 & 78.0 & 58.7 & 70.8 & 68.6 & 57.5 & 54.7 & 45.8 & 38.8 & 39.9 & 45.2 & 44.6 & 84.8 & 71.1 & 74.2 & 50.8 & 74.3 & 47.7 & 78.1 & 70.7 & 60.2 & 57.5 \\ 
 ViT-B-384-l2b & 86.6 & Ener & 62.5 & 44.4 & 79.0 & 61.6 & 72.2 & 70.2 & 60.5 & 57.6 & 49.3 & 42.8 & 44.3 & 52.0 & 51.7 & 86.1 & 71.5 & 77.4 & 53.4 & 76.3 & 51.7 & 79.2 & 70.2 & 64.8 & 61.3 \\ 
  &  & KL-M & 64.7 & 48.4 & 67.2 & 44.8 & 69.0 & 68.9 & 49.2 & 46.1 & 37.5 & 39.0 & 39.7 & 33.2 & 31.9 & $\bm{81.8}$ & 71.5 & 66.1 & 45.8 & 72.7 & 48.3 & 81.2 & 70.2 & 59.0 & 53.8 \\ 
  &  & KNN & 56.3 & 33.3 & 69.5 & 43.6 & 63.2 & 63.4 & 37.8 & 35.1 & 24.7 & 21.2 & 21.5 & 15.8 & 14.1 & 87.3 & 82.1 & 66.1 & 43.6 & 64.7 & 33.8 & 83.3 & 78.4 & 52.8 & 45.3 \\ 
  &  & RMaha & 58.3 & 32.0 & 64.5 & 39.5 & 64.0 & 61.5 & 40.0 & 37.3 & 26.4 & 19.0 & 19.8 & 12.2 & 10.4 & 83.0 & 71.9 & 64.5 & 39.4 & 62.7 & 33.1 & 78.6 & 68.3 & 49.8 & 41.5 \\ 
  &  & RCos & 54.8 & 32.0 & $\bm{64.2}$ & 37.8 & 64.0 & 61.8 & $\bm{35.2}$ & $\bm{32.2}$ & $\bm{20.5}$ & 20.5 & 21.5 & 13.5 & 11.7 & $\bm{81.8}$ & 73.6 & 65.3 & $\bm{39.0}$ & 64.3 & 32.5 & 79.2 & $\bm{64.4}$ & $\bm{48.0}$ & $\bm{39.6}$ \\ 
  &  & Cos & 55.1 & 33.3 & 65.2 & 37.8 & 63.7 & 61.8 & 37.0 & 34.0 & 22.9 & 19.5 & 20.4 & 13.0 & 11.2 & 84.2 & 75.3 & 65.3 & 39.4 & 64.7 & 30.5 & 80.2 & 67.3 & 48.8 & 40.1 \\ 
 \hline 
 &  & MSP & 64.7 & 49.7 & 69.2 & 54.1 & 78.2 & 76.1 & 54.2 & 52.5 & 44.4 & 52.2 & 52.2 & 40.8 & 39.7 & 84.8 & 74.9 & 76.6 & 52.5 & 74.3 & 51.0 & 86.5 & 72.1 & 63.2 & 56.6 \\ 
  &  & MaxL & 63.5 & 49.7 & 71.8 & 54.7 & 73.2 & 70.9 & 55.2 & 53.6 & 45.1 & 49.2 & 49.5 & 37.8 & 36.8 & 85.5 & 72.8 & 75.8 & 48.7 & 75.1 & 49.7 & 85.9 & 73.1 & 59.5 & 54.2 \\ 
  &  & ViM & $\bm{50.9}$ & $\bm{28.1}$ & 66.5 & 37.8 & 66.5 & 64.4 & 41.2 & 38.6 & 29.5 & 24.0 & $\bm{23.6}$ & $\bm{12.8}$ & $\bm{11.2}$ & 87.3 & 77.9 & 68.5 & 38.1 & $\bm{63.1}$ & 34.4 & $\bm{79.2}$ & 68.3 & 50.0 & 41.5 \\ 
  &  & Maha & 53.1 & 29.4 & 65.2 & $\bm{37.2}$ & 65.0 & 62.1 & 41.2 & 38.9 & 29.2 & $\bm{23.2}$ & 24.2 & 14.0 & 12.3 & 86.1 & 73.2 & $\bm{66.1}$ & 37.7 & 67.1 & 33.8 & 81.8 & 65.9 & 51.2 & 41.5 \\ 
  &  & E+R & 63.2 & 48.4 & 75.2 & 56.4 & 71.0 & 69.3 & 60.0 & 58.4 & 51.4 & 45.8 & 46.7 & 38.5 & 38.4 & 86.1 & 76.6 & 73.4 & 48.7 & 72.3 & 50.3 & 84.9 & 72.6 & 61.8 & 57.1 \\ 
 ViT-B-384-oai & 86.2 & Ener & 65.7 & 52.3 & 76.2 & 58.1 & 73.8 & 72.5 & 62.7 & 61.1 & 54.5 & 51.7 & 52.4 & 44.5 & 44.1 & 86.7 & 75.7 & 78.2 & 52.5 & 75.1 & 53.0 & 84.9 & 74.5 & 64.5 & 59.0 \\ 
  &  & KL-M & 66.2 & 51.0 & 72.0 & 54.7 & 71.5 & 69.6 & 49.2 & 47.5 & 39.2 & 48.8 & 48.6 & 38.8 & 37.3 & 87.3 & 75.3 & 71.0 & 47.5 & 72.7 & 47.7 & 85.4 & 69.2 & 61.0 & 52.8 \\ 
  &  & KNN & 53.3 & 30.7 & 73.5 & 47.1 & 64.8 & 62.5 & 40.5 & 38.1 & 27.8 & 28.7 & 29.3 & 21.5 & 20.1 & 88.5 & 83.4 & 69.4 & 44.1 & 67.1 & $\bm{32.5}$ & 85.4 & 74.5 & 55.8 & 49.5 \\ 
  &  & RMaha & 55.1 & 32.0 & $\bm{64.5}$ & 38.4 & 67.8 & 66.0 & 43.2 & 41.0 & 31.6 & 26.0 & 26.9 & 15.8 & 14.1 & $\bm{81.8}$ & $\bm{69.4}$ & $\bm{66.1}$ & $\bm{36.0}$ & 68.3 & 33.8 & 81.8 & $\bm{63.0}$ & $\bm{48.5}$ & $\bm{37.7}$ \\ 
  &  & RCos & 55.3 & 30.1 & 65.8 & 39.5 & $\bm{63.2}$ & $\bm{60.2}$ & $\bm{37.2}$ & $\bm{34.6}$ & $\bm{23.6}$ & 24.5 & 25.3 & 16.8 & 15.1 & 83.6 & 71.1 & 67.7 & $\bm{36.0}$ & 68.3 & 33.1 & 83.9 & 66.8 & 51.0 & 42.0 \\ 
  &  & Cos & 55.3 & 30.1 & 67.8 & 40.1 & 64.5 & 61.5 & 39.2 & 36.7 & 25.7 & 25.5 & 26.4 & 16.8 & 15.1 & 85.5 & 74.5 & 67.7 & 37.3 & 67.5 & $\bm{32.5}$ & 85.4 & 68.3 & 51.2 & 42.5 \\ 
 \hline 
\end{tabular}
\end{center}

\end{table}
\renewcommand{\arraystretch}{.9}
\tabcolsep=0.0310321cm

\begin{table}[htb]
\rowcolors{2}{white}{gray!10}
    \centering
    \caption{Comparing the cleaned and original datasets in terms of \FPR. The best method per model and dataset is marked bold.}
    \label{tab:clean-vs-all-pre-conv}
    \tiny
    \begin{center}
\begin{tabular}{c c c | c  c  c  c  c  c  c  c  c  c  c  c  c  c  c  c  c  c  c  c  c  c  c }
 &  & &   &   &   &   &   &   &   &   &   &   &   & fpr &   &   &   &   &   &   &   &   &   &   &   \\ 
 model &acc. & method & Pl-f & Pl-c & Spc-f & Spc-c & IN-f & IN-c & txt-f & txt-43 & txt-c & OpO-f & OpO-c & iNat-f & iNat-c & IN1K-f & IN1K-c & OS-f & OS-c & SBe-f & SBe-c & SBh-f & SBh-c & CO-f & CO-c \\ 
 \hline 
 &  & MSP & 58.3 & 35.9 & 63.5 & 40.1 & 61.3 & 58.3 & 51.2 & 48.8 & 38.5 & 32.0 & 30.2 & 18.0 & 16.2 & 80.6 & 56.6 & 70.2 & 30.5 & 71.5 & 30.5 & 77.1 & 51.0 & 46.5 & 36.8 \\ 
  &  & MaxL & 59.5 & 37.9 & 61.0 & 37.2 & 53.8 & 49.8 & 47.2 & 44.8 & 31.9 & 28.2 & 25.8 & 18.5 & 17.0 & 76.4 & 49.8 & 66.9 & 22.5 & 64.7 & 25.2 & 67.2 & 35.6 & 40.5 & 28.8 \\ 
  &  & ViM & $\bm{49.6}$ & $\bm{24.2}$ & $\bm{56.2}$ & 30.2 & $\bm{44.2}$ & $\bm{40.5}$ & $\bm{29.0}$ & $\bm{26.0}$ & $\bm{14.6}$ & $\bm{12.0}$ & $\bm{10.9}$ & $\bm{1.8}$ & $\bm{0.8}$ & 86.1 & 75.7 & $\bm{57.3}$ & 24.6 & 65.1 & 25.2 & 73.4 & 56.2 & $\bm{35.5}$ & $\bm{26.4}$ \\ 
  &  & Maha & 55.3 & 28.8 & 57.8 & 29.1 & 59.0 & 56.3 & 32.0 & 29.5 & 18.1 & 15.5 & 14.4 & 3.0 & 1.8 & 84.2 & 75.3 & 65.3 & 35.2 & 68.7 & 41.7 & 83.3 & 63.5 & 45.2 & 34.9 \\ 
  &  & E+R & 57.5 & 33.3 & 60.2 & 33.1 & 49.0 & 44.0 & 48.2 & 46.1 & 34.4 & 26.8 & 24.7 & 18.2 & 17.0 & $\bm{71.5}$ & 50.2 & 63.7 & $\bm{17.8}$ & 65.9 & 23.2 & $\bm{60.4}$ & 34.1 & 39.5 & 29.7 \\ 
 CnvNxt-B-21k & 86.3 & Ener & 63.0 & 41.2 & 63.5 & 39.5 & 49.5 & 45.0 & 55.2 & 53.9 & 43.8 & 36.0 & 34.0 & 26.0 & 25.1 & 72.1 & $\bm{48.5}$ & 66.9 & 21.6 & 67.9 & 27.8 & 60.9 & $\bm{31.2}$ & 41.0 & 30.7 \\ 
  &  & KL-M & 65.7 & 42.5 & 69.5 & 44.2 & 59.8 & 57.6 & 49.0 & 46.4 & 36.5 & 33.2 & 32.9 & 19.0 & 16.4 & 84.2 & 66.0 & 68.5 & 37.3 & 73.5 & 37.7 & 82.8 & 63.0 & 55.5 & 45.8 \\ 
  &  & KNN & 66.9 & 47.7 & 69.5 & 41.9 & 48.2 & 43.4 & 31.5 & 28.7 & 17.0 & 19.5 & 19.0 & 7.8 & 6.3 & 90.3 & 79.1 & 63.7 & 28.4 & $\bm{60.6}$ & 19.2 & 81.8 & 67.3 & 45.5 & 37.3 \\ 
  &  & RMaha & 57.8 & 32.0 & 56.5 & $\bm{27.3}$ & 54.5 & 50.5 & 33.5 & 30.6 & 18.4 & 16.2 & 15.5 & 2.8 & 1.3 & 80.0 & 56.6 & 60.5 & 24.2 & 68.7 & 27.8 & 76.6 & 54.3 & 40.5 & 27.8 \\ 
  &  & RCos & 59.0 & 30.7 & 62.7 & 32.0 & 50.0 & 45.6 & 32.0 & 29.0 & 16.3 & 15.2 & 14.9 & 4.5 & 2.9 & 84.8 & 61.3 & 63.7 & 23.7 & 62.2 & $\bm{17.2}$ & 75.0 & 51.0 & 40.2 & 28.3 \\ 
  &  & Cos & 59.3 & 32.7 & 63.5 & 32.6 & 49.2 & 44.3 & 32.0 & 29.2 & 16.3 & 14.8 & 14.7 & 4.0 & 2.6 & 85.5 & 64.3 & 64.5 & 23.3 & 62.2 & $\bm{17.2}$ & 76.6 & 51.9 & 41.5 & 29.7 \\ 
 \hline 
 &  & MSP & 63.7 & 40.5 & 69.0 & 46.5 & 68.0 & 65.7 & 52.2 & 50.4 & 42.0 & 38.0 & 38.0 & 15.8 & 13.6 & 89.7 & 68.9 & 70.2 & 33.9 & 73.1 & 39.1 & 80.7 & 55.8 & 49.2 & 41.5 \\ 
  &  & MaxL & 61.0 & 37.9 & 67.8 & 41.3 & 58.8 & 56.0 & 47.5 & 45.8 & 36.5 & 32.2 & 32.1 & 12.2 & 9.9 & 84.2 & 61.7 & 65.3 & 25.8 & 68.7 & 30.5 & 73.4 & 48.6 & 41.8 & 33.0 \\ 
  &  & ViM & $\bm{48.6}$ & $\bm{24.2}$ & $\bm{53.0}$ & $\bm{25.6}$ & $\bm{47.5}$ & $\bm{43.7}$ & $\bm{27.8}$ & $\bm{24.9}$ & $\bm{15.3}$ & $\bm{11.5}$ & $\bm{10.3}$ & $\bm{2.5}$ & $\bm{1.0}$ & 84.8 & 68.5 & $\bm{53.2}$ & $\bm{16.1}$ & $\bm{64.3}$ & $\bm{21.2}$ & $\bm{69.8}$ & 53.8 & $\bm{34.2}$ & $\bm{25.0}$ \\ 
  &  & Maha & 55.6 & 31.4 & 56.8 & 29.7 & 64.2 & 61.8 & 31.0 & 28.7 & 17.4 & 15.8 & 15.2 & 3.8 & 2.3 & 84.8 & 75.7 & 64.5 & 28.4 & 71.1 & 37.7 & 79.2 & 66.8 & 42.5 & 31.6 \\ 
  &  & E+R & 57.8 & 34.6 & 67.5 & 38.4 & 51.2 & 47.2 & 45.0 & 43.4 & 33.7 & 28.5 & 27.7 & 8.2 & 5.7 & $\bm{82.4}$ & 59.1 & 64.5 & 23.3 & 68.7 & 29.8 & 71.9 & 45.2 & 37.0 & 29.2 \\ 
 CnvNxt-T-21k & 84.1 & Ener & 61.0 & 40.5 & 67.8 & 39.5 & 51.5 & 47.6 & 47.5 & 46.1 & 36.8 & 31.8 & 31.2 & 12.0 & 9.4 & $\bm{82.4}$ & $\bm{58.7}$ & 64.5 & 24.6 & 69.5 & 29.8 & 71.9 & $\bm{43.3}$ & 38.8 & 31.1 \\ 
  &  & KL-M & 70.4 & 52.3 & 72.8 & 49.4 & 71.2 & 71.5 & 51.0 & 49.1 & 41.0 & 42.5 & 42.4 & 19.5 & 17.2 & 89.7 & 73.2 & 70.2 & 43.2 & 73.5 & 53.0 & 83.9 & 67.3 & 56.0 & 47.2 \\ 
  &  & KNN & 72.3 & 52.3 & 73.0 & 47.1 & 59.8 & 55.7 & 36.0 & 33.2 & 21.5 & 28.7 & 28.8 & 20.0 & 18.3 & 90.3 & 84.3 & 71.8 & 36.0 & 69.5 & 26.5 & 82.3 & 74.0 & 49.0 & 41.0 \\ 
  &  & RMaha & 59.3 & 41.8 & 59.0 & 32.0 & 63.7 & 61.2 & 38.0 & 35.7 & 25.3 & 21.2 & 21.5 & 5.2 & 3.4 & 84.8 & 67.2 & 65.3 & 26.3 & 74.3 & 31.8 & 79.2 & 62.0 & 43.5 & 34.0 \\ 
  &  & RCos & 63.0 & 39.9 & 65.2 & 36.0 & 62.0 & 58.3 & 38.2 & 36.2 & 25.3 & 24.2 & 24.5 & 10.8 & 8.9 & 86.7 & 71.5 & 67.7 & 30.5 & 69.1 & 32.5 & 81.8 & 65.4 & 44.0 & 35.4 \\ 
  &  & Cos & 64.4 & 41.2 & 67.0 & 39.5 & 62.0 & 58.3 & 37.8 & 35.7 & 24.7 & 26.0 & 26.4 & 11.8 & 10.2 & 86.7 & 73.2 & 66.9 & 32.2 & 70.3 & 31.8 & 80.7 & 67.8 & 44.5 & 36.8 \\ 
 \hline 
 &  & MSP & 74.3 & 56.9 & 72.5 & 54.7 & 83.2 & 82.2 & 72.8 & 71.8 & 68.1 & 52.5 & 52.2 & 28.7 & 26.1 & 86.7 & 76.2 & 79.0 & 50.0 & 78.3 & 59.6 & 90.6 & 76.0 & 64.2 & 57.1 \\ 
  &  & MaxL & 69.6 & 47.7 & 65.8 & 41.3 & 80.2 & 79.3 & 66.5 & 65.4 & 59.7 & 43.5 & 43.5 & 17.0 & 14.1 & 83.6 & 74.0 & 77.4 & 36.9 & 78.3 & 57.0 & 87.0 & 68.3 & 58.2 & 50.0 \\ 
  &  & ViM & $\bm{54.1}$ & $\bm{28.1}$ & 47.0 & 22.7 & 38.2 & 34.6 & $\bm{5.5}$ & $\bm{5.1}$ & $\bm{2.1}$ & $\bm{14.2}$ & $\bm{12.2}$ & $\bm{2.2}$ & $\bm{1.0}$ & 85.5 & 67.2 & $\bm{64.5}$ & $\bm{13.6}$ & $\bm{73.9}$ & 33.1 & 79.2 & 57.7 & $\bm{39.2}$ & $\bm{24.1}$ \\ 
  &  & Maha & 64.4 & 41.8 & 52.5 & 25.0 & $\bm{36.8}$ & $\bm{33.3}$ & 6.5 & 5.9 & $\bm{2.1}$ & 19.0 & 16.0 & 4.5 & 2.9 & 84.2 & 68.1 & 69.4 & 17.8 & 80.3 & 39.1 & 83.9 & 62.5 & 48.0 & 31.6 \\ 
  &  & E+R & 64.2 & 44.4 & $\bm{46.5}$ & 29.1 & 75.8 & 74.8 & 67.8 & 67.3 & 64.6 & 44.2 & 43.8 & 10.2 & 8.4 & 77.0 & 57.9 & 74.2 & 29.7 & 82.7 & 62.9 & $\bm{70.3}$ & $\bm{48.6}$ & 55.5 & 45.3 \\ 
 BiT-m & 82.3 & Ener & 70.9 & 51.0 & 66.5 & 42.4 & 79.8 & 79.0 & 69.0 & 68.4 & 62.8 & 43.5 & 42.9 & 15.2 & 12.8 & 84.8 & 76.6 & 77.4 & 36.4 & 80.7 & 57.6 & 85.9 & 66.8 & 59.0 & 50.0 \\ 
  &  & KL-M & 72.6 & 54.2 & 74.8 & 54.1 & 74.0 & 73.5 & 64.0 & 63.3 & 58.7 & 45.0 & 45.1 & 28.2 & 25.6 & 83.0 & 74.9 & 79.0 & 43.6 & 76.3 & 55.0 & 89.6 & 71.6 & 61.5 & 51.9 \\ 
  &  & KNN & 69.4 & 47.7 & 58.8 & 32.6 & 42.2 & 39.5 & 11.2 & 10.5 & 4.9 & 19.0 & 16.0 & 4.5 & 2.6 & 93.3 & 88.9 & 76.6 & 21.2 & 83.9 & 38.4 & 87.0 & 80.3 & 54.5 & 37.3 \\ 
  &  & RMaha & 65.2 & 45.8 & 49.5 & $\bm{20.3}$ & 56.2 & 54.4 & 23.8 & 22.3 & 13.9 & 23.0 & 22.0 & 4.0 & 2.6 & $\bm{73.3}$ & $\bm{55.3}$ & 72.6 & 19.5 & 80.7 & 37.1 & 78.6 & 56.7 & 49.2 & 31.6 \\ 
  &  & RCos & 66.2 & 39.9 & 62.5 & 36.0 & 64.5 & 61.8 & 31.2 & 29.5 & 18.4 & 24.2 & 23.4 & 6.0 & 4.4 & 83.6 & 72.8 & 73.4 & 28.4 & 74.3 & 33.8 & 82.8 & 65.9 & 49.8 & 35.8 \\ 
  &  & Cos & 63.7 & 37.3 & 58.8 & 30.2 & 50.5 & 46.9 & 16.5 & 14.5 & 6.6 & 18.5 & 17.1 & 4.2 & 2.6 & 83.0 & 71.5 & 70.2 & 21.6 & 74.7 & $\bm{29.1}$ & 83.9 & 65.4 & 48.5 & 33.5 \\ 
 \hline 
 &  & MSP & 61.5 & 39.9 & 67.0 & 45.3 & 65.2 & 62.5 & 52.2 & 50.4 & 42.4 & 36.8 & 36.7 & 19.5 & 17.5 & 85.5 & 62.6 & 76.6 & 40.3 & 76.7 & 34.4 & 83.9 & 54.8 & 51.7 & 41.0 \\ 
  &  & MaxL & 63.2 & 45.8 & 68.8 & 45.9 & 61.5 & 58.6 & 52.8 & 50.4 & 41.7 & 35.0 & 34.8 & 20.8 & 19.1 & 86.1 & $\bm{59.1}$ & 77.4 & 34.3 & 77.1 & 35.8 & 86.5 & $\bm{51.9}$ & 50.2 & 40.6 \\ 
  &  & ViM & 63.2 & 39.2 & 65.8 & 34.3 & 47.2 & 42.1 & $\bm{21.8}$ & $\bm{19.0}$ & $\bm{10.8}$ & $\bm{20.5}$ & $\bm{19.0}$ & $\bm{3.8}$ & $\bm{2.3}$ & 87.9 & 86.8 & 69.4 & 34.3 & 73.5 & 23.2 & $\bm{81.8}$ & 76.0 & 47.8 & 38.7 \\ 
  &  & Maha & 69.1 & 47.1 & 68.2 & 39.0 & 58.0 & 53.1 & 27.5 & 25.5 & 16.0 & 28.7 & 28.0 & 7.8 & 6.0 & 87.9 & 88.9 & $\bm{66.9}$ & 41.9 & 73.5 & 31.1 & 83.9 & 82.2 & 56.8 & 47.2 \\ 
  &  & E+R & 94.3 & 93.5 & 93.2 & 90.7 & 89.2 & 89.3 & 87.2 & 88.7 & 89.2 & 87.5 & 87.0 & 89.2 & 89.3 & 95.8 & 93.6 & 91.9 & 80.1 & 90.0 & 90.1 & 94.8 & 86.5 & 90.0 & 89.6 \\ 
 EffNetv2-M-21k & 85.6 & Ener & 69.6 & 54.9 & 75.0 & 54.7 & 64.5 & 63.1 & 65.5 & 63.5 & 56.9 & 40.2 & 39.4 & 26.2 & 24.5 & 90.3 & 63.8 & 79.8 & 39.0 & 83.9 & 47.7 & 90.1 & 56.7 & 55.2 & 45.8 \\ 
  &  & KL-M & 65.9 & 45.8 & 70.2 & 47.1 & 64.5 & 63.1 & 51.7 & 50.4 & 42.4 & 37.8 & 38.6 & 23.0 & 21.4 & 84.2 & 69.4 & 76.6 & 42.4 & 75.5 & 32.5 & 83.9 & 63.0 & 54.2 & 43.4 \\ 
  &  & KNN & 83.5 & 71.9 & 75.5 & 52.9 & 45.5 & 41.4 & 26.0 & 24.7 & 14.6 & 40.8 & 38.3 & 13.2 & 11.5 & 95.2 & 95.3 & 74.2 & 40.3 & 88.0 & 47.7 & 93.8 & 86.1 & 65.0 & 52.4 \\ 
  &  & RMaha & 70.1 & 53.6 & 61.5 & 34.3 & 59.8 & 56.0 & 34.8 & 33.0 & 24.0 & 27.0 & 26.6 & 9.0 & 7.0 & 84.8 & 75.3 & $\bm{66.9}$ & 35.2 & 77.5 & 29.8 & 82.3 & 71.2 & 53.0 & 39.6 \\ 
  &  & RCos & 63.2 & 39.9 & 63.0 & 38.4 & 57.2 & 53.7 & 39.0 & 36.7 & 27.1 & 25.0 & 25.3 & 6.8 & 5.5 & 84.2 & 71.5 & 68.5 & 33.9 & $\bm{72.7}$ & 25.8 & 85.4 & 66.3 & 47.5 & 35.8 \\ 
  &  & Cos & $\bm{60.5}$ & $\bm{37.3}$ & $\bm{59.8}$ & $\bm{29.7}$ & $\bm{41.8}$ & $\bm{37.5}$ & 23.8 & 22.0 & 12.8 & 22.0 & 21.5 & 5.2 & 3.9 & $\bm{83.6}$ & 71.1 & 67.7 & $\bm{26.3}$ & 73.9 & $\bm{18.5}$ & 83.9 & 63.9 & $\bm{45.2}$ & $\bm{34.0}$ \\ 
 \hline 
 &  & MSP & 54.6 & 34.0 & 69.5 & 53.5 & 69.2 & 67.3 & 56.0 & 54.2 & 48.3 & 42.5 & 41.6 & 38.8 & 37.1 & 84.8 & 69.4 & 75.8 & 62.3 & 73.9 & 53.6 & 83.3 & 69.7 & 63.0 & 55.2 \\ 
  &  & MaxL & $\bm{46.9}$ & $\bm{31.4}$ & $\bm{67.8}$ & 49.4 & 66.0 & 64.4 & 53.2 & 51.2 & 44.8 & $\bm{33.0}$ & $\bm{33.2}$ & 35.2 & 33.2 & 81.8 & $\bm{64.3}$ & 77.4 & 58.5 & 71.5 & 55.6 & 79.7 & $\bm{62.5}$ & 59.8 & 51.4 \\ 
  &  & ViM & 96.5 & 93.5 & 92.0 & 85.5 & 77.5 & 77.0 & 87.5 & 86.9 & 85.4 & 91.8 & 91.8 & 83.0 & 82.8 & 90.9 & 95.3 & 73.4 & 69.9 & 80.7 & 60.3 & 87.0 & 88.5 & 75.5 & 76.9 \\ 
  &  & Maha & 95.8 & 92.2 & 89.0 & 80.2 & 76.5 & 75.7 & 82.8 & 82.0 & 79.2 & 88.2 & 88.3 & 76.5 & 76.2 & 90.9 & 93.6 & 75.8 & 67.4 & 79.9 & 59.6 & 87.5 & 88.5 & 73.5 & 73.1 \\ 
  &  & E+R & 57.8 & 37.9 & 76.2 & 61.6 & 67.5 & 67.6 & 59.8 & 57.9 & 51.7 & 41.5 & 42.1 & 50.7 & 49.3 & 81.8 & 71.9 & 80.6 & 65.7 & 71.9 & 55.6 & 81.2 & 66.8 & 60.8 & 56.6 \\ 
 EffNetb7-ns & 86.8 & Ener & 52.3 & 39.9 & 75.5 & 62.2 & 77.2 & 77.3 & 64.8 & 63.0 & 58.0 & 48.0 & 48.1 & 56.8 & 55.6 & $\bm{79.4}$ & 67.2 & 83.9 & 77.5 & 79.9 & 71.5 & 82.3 & 64.4 & 70.2 & 66.0 \\ 
  &  & KL-M & 62.5 & 37.9 & 74.0 & 54.7 & 62.3 & 59.9 & 53.2 & 51.7 & 45.5 & 40.8 & 40.5 & 32.8 & 30.8 & 87.3 & 78.3 & 69.4 & 55.1 & 69.9 & 51.0 & 85.4 & 73.1 & 62.5 & 53.3 \\ 
  &  & KNN & 63.7 & 41.8 & 83.5 & 67.4 & 65.2 & 63.8 & 43.2 & 41.0 & 29.5 & 45.5 & 46.5 & 36.5 & 35.0 & 90.3 & 91.9 & 77.4 & 55.9 & 67.5 & $\bm{36.4}$ & 83.3 & 79.3 & 59.5 & 57.1 \\ 
  &  & RMaha & 80.5 & 63.4 & 70.2 & $\bm{48.3}$ & 71.5 & 70.6 & 64.8 & 63.5 & 57.3 & 47.2 & 46.2 & $\bm{20.0}$ & $\bm{17.8}$ & 82.4 & 71.1 & 73.4 & $\bm{45.3}$ & 73.9 & 45.7 & 83.3 & 69.2 & 59.2 & 50.0 \\ 
  &  & RCos & 57.8 & 32.0 & 73.8 & 54.7 & $\bm{61.3}$ & 59.5 & 41.0 & 38.9 & 27.4 & 36.5 & 36.1 & 22.8 & 20.9 & 88.5 & 80.0 & $\bm{66.1}$ & 46.2 & $\bm{65.1}$ & $\bm{36.4}$ & 79.7 & 71.6 & $\bm{54.2}$ & $\bm{46.7}$ \\ 
  &  & Cos & 57.0 & $\bm{31.4}$ & 75.2 & 55.2 & 61.5 & $\bm{58.9}$ & $\bm{40.0}$ & $\bm{37.8}$ & $\bm{26.0}$ & 35.0 & 35.3 & 24.2 & 22.5 & 89.1 & 83.0 & 68.5 & 47.0 & 65.5 & 37.1 & $\bm{78.1}$ & 72.6 & 54.5 & 48.1 \\ 
 \hline 
\end{tabular}
\end{center}

\end{table}
\renewcommand{\arraystretch}{.9}
\tabcolsep=0.0310321cm

\begin{table}[htb]
\rowcolors{2}{white}{gray!10}
    \centering
    \caption{Comparing the cleaned and original datasets in terms of \FPR. The best method per model and dataset is marked bold.}
    \label{tab:clean-vs-all-1k-vit}
    \tiny
        \begin{center}
\begin{tabular}{c c c | c  c  c  c  c  c  c  c  c  c  c  c  c  c  c  c  c  c  c  c  c  c  c }
 &  & &   &   &   &   &   &   &   &   &   &   &   & fpr &   &   &   &   &   &   &   &   &   &   &   \\ 
 model &acc. & method & Pl-f & Pl-c & Spc-f & Spc-c & IN-f & IN-c & txt-f & txt-43 & txt-c & OpO-f & OpO-c & iNat-f & iNat-c & IN1K-f & IN1K-c & OS-f & OS-c & SBe-f & SBe-c & SBh-f & SBh-c & CO-f & CO-c \\ 
 \hline 
 &  & MSP & 71.6 & 58.8 & 78.2 & 61.6 & 82.8 & 81.9 & 62.5 & 60.9 & 54.2 & 54.0 & 54.1 & 46.8 & 45.7 & 92.7 & 81.3 & 76.6 & 64.0 & 77.1 & 55.6 & 87.0 & 78.4 & 66.0 & 63.2 \\ 
  &  & MaxL & 68.9 & $\bm{52.9}$ & 75.5 & 54.7 & 80.8 & 80.6 & 55.5 & 54.2 & 46.2 & 44.5 & 44.8 & 37.8 & 36.8 & 90.9 & 85.5 & 75.8 & 57.2 & 75.1 & 51.7 & 85.9 & 82.2 & 63.2 & 61.3 \\ 
  &  & ViM & 73.3 & 62.1 & 73.5 & 55.8 & 82.8 & 81.9 & 54.2 & 54.2 & 46.9 & 49.0 & 49.2 & 39.5 & 38.9 & 84.8 & 80.4 & $\bm{73.4}$ & 59.3 & 76.7 & 61.6 & 84.9 & 77.4 & 67.2 & 62.3 \\ 
  &  & Maha & 70.4 & 56.2 & 62.5 & 35.5 & 79.2 & 78.0 & $\bm{47.0}$ & $\bm{45.8}$ & $\bm{39.6}$ & $\bm{36.0}$ & $\bm{35.1}$ & 16.5 & 14.9 & 78.2 & 65.1 & 77.4 & 44.1 & 77.5 & 57.0 & 81.8 & 63.9 & 61.3 & $\bm{50.5}$ \\ 
  &  & E+R & $\bm{68.6}$ & $\bm{52.9}$ & 70.8 & 46.5 & 80.8 & 79.6 & 50.5 & 49.3 & 41.7 & 39.0 & 38.9 & 27.8 & 26.9 & 87.3 & 79.1 & $\bm{73.4}$ & 53.8 & 75.1 & 55.0 & 82.3 & 77.9 & 60.8 & 57.5 \\ 
 ViT-B-384 & 81.1 & Ener & 69.6 & 53.6 & 75.2 & 55.2 & 80.0 & 78.6 & 50.5 & 49.3 & 41.3 & 42.2 & 42.4 & 35.5 & 34.2 & 89.1 & 86.8 & 75.0 & 56.4 & 75.1 & 53.0 & 83.3 & 82.7 & 63.5 & 62.7 \\ 
  &  & KL-M & 72.8 & 60.8 & 72.2 & 52.3 & $\bm{77.8}$ & $\bm{76.4}$ & 59.5 & 58.2 & 52.8 & 51.0 & 51.4 & 36.2 & 35.0 & 86.7 & 76.6 & 76.6 & 58.1 & 76.3 & 59.6 & 83.3 & 75.5 & 63.7 & 57.5 \\ 
  &  & KNN & 74.8 & 64.7 & 81.5 & 64.0 & 78.5 & 77.0 & 48.8 & 47.2 & 41.3 & 49.0 & 49.7 & 55.2 & 54.3 & 92.7 & 88.5 & 75.8 & 66.9 & $\bm{72.7}$ & $\bm{45.7}$ & 89.1 & 85.6 & 66.5 & 66.0 \\ 
  &  & RMaha & 68.9 & 53.6 & $\bm{59.5}$ & $\bm{33.1}$ & 79.2 & 78.3 & 48.8 & 47.5 & 41.3 & 38.8 & 38.0 & $\bm{13.0}$ & $\bm{11.2}$ & $\bm{77.6}$ & $\bm{59.1}$ & 78.2 & $\bm{41.1}$ & 79.5 & 60.3 & $\bm{78.6}$ & $\bm{62.0}$ & $\bm{60.5}$ & 51.9 \\ 
  &  & RCos & 73.3 & 60.8 & 77.2 & 58.7 & 79.8 & 79.3 & 51.7 & 50.7 & 42.7 & 46.0 & 46.5 & 45.2 & 43.9 & 92.7 & 86.8 & 75.8 & 57.6 & 74.7 & 47.7 & 87.0 & 82.2 & 64.5 & 63.2 \\ 
  &  & Cos & 71.4 & 58.8 & 76.0 & 56.4 & 79.5 & 79.0 & 49.8 & 48.5 & 41.0 & 44.8 & 45.1 & 41.8 & 40.2 & 91.5 & 86.0 & 75.0 & 55.9 & 73.5 & $\bm{45.7}$ & 86.5 & 80.8 & 63.0 & 61.3 \\ 
 \hline 
 &  & MSP & 68.6 & 52.3 & 78.0 & 64.0 & 83.8 & 84.1 & 66.0 & 64.9 & 58.0 & 58.2 & 57.3 & 47.8 & 46.7 & 91.5 & 77.9 & 80.6 & 59.7 & 75.5 & 57.0 & 88.0 & 77.4 & 67.8 & 63.7 \\ 
  &  & MaxL & 69.4 & 54.9 & 76.5 & 62.2 & 79.2 & 79.3 & 56.2 & 55.8 & 48.6 & 58.5 & 57.3 & 45.5 & 44.6 & 90.9 & 79.6 & 77.4 & 59.3 & 79.5 & 62.3 & 85.9 & 78.8 & 69.5 & 65.6 \\ 
  &  & ViM & 63.2 & 45.1 & 76.5 & 54.7 & $\bm{75.8}$ & $\bm{75.1}$ & 48.5 & 46.9 & 38.2 & 31.8 & 32.3 & 22.5 & 21.1 & 90.9 & 87.7 & 73.4 & 44.1 & 67.5 & 41.1 & 81.8 & 83.2 & 55.2 & 50.0 \\ 
  &  & Maha & 59.8 & 41.2 & 73.2 & 50.0 & 77.5 & 77.0 & 51.7 & 50.1 & 41.0 & 31.2 & 31.0 & 20.2 & 18.5 & 90.3 & 77.9 & 74.2 & 43.6 & 67.9 & 39.7 & 80.7 & 75.0 & 55.5 & 48.1 \\ 
  &  & E+R & 67.2 & 49.0 & 81.5 & 66.3 & 76.0 & 75.7 & 47.8 & 46.6 & 38.9 & 49.8 & 49.7 & 44.8 & 43.9 & 90.9 & 84.7 & 75.8 & 57.2 & 75.9 & 53.6 & 82.8 & 81.2 & 68.2 & 65.6 \\ 
 Swinv2-B-256 & 84.6 & Ener & 76.8 & 66.7 & 82.5 & 73.8 & 76.2 & 76.4 & 56.2 & 55.2 & 50.7 & 64.0 & 63.0 & 54.8 & 54.6 & $\bm{87.9}$ & 84.3 & 83.9 & 66.1 & 84.3 & 67.5 & 86.5 & 80.8 & 74.0 & 71.2 \\ 
  &  & KL-M & 72.6 & 56.9 & 75.0 & 58.7 & 79.5 & 80.3 & 63.0 & 62.2 & 55.6 & 51.7 & 50.8 & 45.0 & 44.1 & 91.5 & 77.0 & 77.4 & 56.8 & 72.3 & 53.0 & 83.9 & 76.4 & 65.2 & 60.4 \\ 
  &  & KNN & 64.7 & 47.7 & 79.5 & 60.5 & 79.2 & 78.3 & 49.5 & 47.7 & 39.2 & 37.5 & 38.6 & 40.5 & 39.2 & 93.3 & 90.6 & 78.2 & 53.0 & $\bm{66.7}$ & 39.7 & 83.3 & 84.6 & 59.8 & 55.2 \\ 
  &  & RMaha & $\bm{58.8}$ & $\bm{39.2}$ & $\bm{69.5}$ & $\bm{44.2}$ & 77.0 & 76.4 & 50.7 & 49.1 & 38.9 & $\bm{30.0}$ & $\bm{29.6}$ & $\bm{17.5}$ & $\bm{15.7}$ & $\bm{87.9}$ & $\bm{69.4}$ & $\bm{72.6}$ & $\bm{38.6}$ & 67.9 & $\bm{39.1}$ & $\bm{79.2}$ & $\bm{70.2}$ & $\bm{53.2}$ & $\bm{45.3}$ \\ 
  &  & RCos & 62.7 & 42.5 & 75.5 & 53.5 & 76.8 & 75.7 & $\bm{44.8}$ & $\bm{42.9}$ & $\bm{33.7}$ & 32.5 & 33.2 & 24.8 & 23.2 & 90.3 & 81.3 & 74.2 & 44.1 & 67.1 & 41.1 & 81.2 & 77.9 & 55.8 & 48.1 \\ 
  &  & Cos & 63.5 & 43.1 & 76.2 & 55.2 & 77.2 & 76.7 & 47.0 & 45.0 & 36.1 & 33.5 & 34.0 & 28.7 & 27.4 & 90.9 & 83.4 & 75.0 & 47.0 & 67.5 & $\bm{39.1}$ & 81.2 & 80.3 & 56.8 & 50.9 \\ 
 \hline 
 &  & MSP & 64.7 & 52.3 & 76.5 & 60.5 & 80.8 & 79.9 & 58.0 & 55.8 & 50.0 & 57.2 & 57.6 & 41.5 & 40.7 & 90.3 & 75.7 & 77.4 & 61.4 & 76.7 & 55.0 & 86.5 & 72.6 & 59.2 & 54.2 \\ 
  &  & MaxL & 70.4 & 58.2 & 80.0 & 62.8 & 82.2 & 82.5 & 54.8 & 52.5 & 47.9 & 65.0 & 64.7 & 50.0 & 49.9 & 91.5 & 78.3 & 82.3 & 69.1 & 81.9 & 60.3 & 85.4 & 75.0 & 65.5 & 63.7 \\ 
  &  & ViM & $\bm{58.0}$ & $\bm{34.6}$ & 70.5 & 44.2 & 71.2 & 69.3 & 42.0 & 39.9 & 31.9 & 34.2 & 35.1 & $\bm{16.8}$ & $\bm{15.1}$ & $\bm{82.4}$ & 70.2 & $\bm{71.0}$ & 41.9 & $\bm{65.5}$ & $\bm{35.8}$ & $\bm{77.1}$ & 70.2 & $\bm{51.5}$ & 45.3 \\ 
  &  & Maha & 63.5 & 46.4 & 71.0 & 45.9 & 76.5 & 74.1 & 59.0 & 57.6 & 50.7 & 37.2 & 39.1 & 20.8 & 19.3 & 88.5 & 80.9 & 72.6 & 45.8 & 67.5 & $\bm{35.8}$ & 83.3 & 77.4 & 55.5 & 48.1 \\ 
  &  & E+R & 93.6 & 90.8 & 94.5 & 93.6 & 90.8 & 92.6 & 81.5 & 81.0 & 78.5 & 91.5 & 90.8 & 91.0 & 91.4 & 91.5 & 85.5 & 94.4 & 91.5 & 94.0 & 94.0 & 90.1 & 84.6 & 89.2 & 87.7 \\ 
 Deit3-B-384 & 85.1 & Ener & 88.4 & 81.7 & 91.5 & 86.0 & 86.0 & 87.1 & 66.5 & 65.4 & 61.5 & 83.8 & 82.3 & 83.2 & 83.0 & 94.5 & 91.9 & 88.7 & 85.2 & 92.8 & 84.8 & 90.1 & 87.0 & 84.5 & 84.9 \\ 
  &  & KL-M & 69.9 & 54.9 & 77.2 & 58.1 & 74.5 & 73.1 & 56.5 & 54.4 & 49.3 & 51.5 & 51.9 & 39.2 & 38.6 & 89.7 & 77.0 & 76.6 & 55.9 & 73.5 & 51.0 & 84.9 & 72.1 & 57.8 & 53.8 \\ 
  &  & KNN & 66.4 & 51.0 & 84.8 & 69.8 & 82.5 & 81.6 & 64.2 & 62.7 & 55.2 & 50.5 & 53.3 & 59.8 & 59.5 & 93.3 & 93.6 & 75.0 & 63.6 & 67.5 & 43.0 & 83.9 & 85.1 & 65.5 & 66.0 \\ 
  &  & RMaha & 61.5 & 43.8 & 70.2 & 44.2 & 76.0 & 73.8 & 57.0 & 55.5 & 48.3 & 35.0 & 36.7 & 17.2 & 15.7 & 85.5 & 74.0 & 71.8 & 42.8 & 67.1 & 38.4 & 82.3 & 70.2 & 53.2 & 45.8 \\ 
  &  & RCos & 70.4 & 51.6 & $\bm{69.0}$ & $\bm{41.9}$ & $\bm{66.2}$ & $\bm{63.1}$ & $\bm{39.5}$ & $\bm{37.5}$ & $\bm{29.2}$ & $\bm{29.0}$ & $\bm{29.6}$ & $\bm{16.8}$ & 15.4 & 85.5 & $\bm{69.8}$ & 71.8 & $\bm{37.3}$ & 67.1 & $\bm{35.8}$ & 82.3 & $\bm{66.8}$ & $\bm{51.5}$ & $\bm{41.5}$ \\ 
  &  & Cos & 64.0 & 47.1 & 79.0 & 58.7 & 79.5 & 77.3 & 57.0 & 55.2 & 46.2 & 40.8 & 42.7 & 38.2 & 37.3 & 91.5 & 89.8 & 73.4 & 54.2 & 66.3 & 40.4 & 83.3 & 80.3 & 59.5 & 55.7 \\ 
 \hline 
 &  & MSP & 65.2 & 47.7 & 73.2 & 58.1 & 88.8 & 88.0 & 58.0 & 56.0 & 51.0 & 52.5 & 51.9 & 41.5 & 40.2 & 87.9 & 74.5 & 78.2 & 56.4 & 75.1 & 57.6 & 86.5 & 76.4 & 63.5 & 62.7 \\ 
  &  & MaxL & 64.4 & 46.4 & 74.0 & 58.1 & 86.8 & 85.4 & 55.0 & 52.8 & 47.2 & 56.0 & 55.7 & 46.8 & 45.7 & 92.7 & 79.1 & 78.2 & 61.9 & 79.5 & 60.9 & 86.5 & 78.8 & 66.0 & 65.1 \\ 
  &  & ViM & $\bm{57.8}$ & $\bm{36.6}$ & $\bm{67.2}$ & $\bm{40.7}$ & 80.8 & 79.3 & $\bm{44.0}$ & $\bm{42.6}$ & $\bm{35.8}$ & $\bm{35.5}$ & $\bm{35.1}$ & $\bm{18.8}$ & $\bm{17.5}$ & $\bm{81.2}$ & 72.3 & 73.4 & 47.5 & 72.3 & 47.7 & 81.8 & 74.0 & 58.2 & 51.9 \\ 
  &  & Maha & 67.4 & 49.0 & 75.2 & 52.3 & 85.8 & 84.1 & 65.2 & 64.6 & 60.8 & 41.5 & 42.4 & 28.5 & 27.2 & 86.1 & 81.3 & 72.6 & 50.8 & 72.7 & 47.0 & 82.3 & 80.3 & 62.7 & 58.0 \\ 
  &  & E+R & 83.2 & 75.8 & 84.5 & 72.7 & 84.8 & 86.1 & 69.5 & 68.1 & 63.5 & 79.8 & 80.2 & 78.2 & 78.1 & 95.8 & 94.0 & 87.9 & 81.8 & 88.8 & 80.1 & 92.7 & 89.9 & 81.2 & 82.5 \\ 
 Deit3-B-224 & 83.8 & Ener & 75.3 & 63.4 & 82.5 & 74.4 & 85.8 & 85.8 & 63.7 & 62.5 & 58.3 & 74.2 & 74.5 & 69.8 & 69.5 & 95.2 & 87.7 & 91.1 & 79.7 & 88.4 & 78.8 & 90.6 & 86.5 & 81.2 & 78.8 \\ 
  &  & KL-M & 73.1 & 53.6 & 72.8 & 55.2 & 83.2 & 81.6 & 58.2 & 56.8 & 51.7 & 50.5 & 50.5 & 38.5 & 37.1 & 87.9 & 77.0 & 71.8 & 52.5 & 75.1 & 50.3 & 85.4 & 76.0 & 63.2 & 62.7 \\ 
  &  & KNN & 78.5 & 66.0 & 89.2 & 79.1 & 87.8 & 86.7 & 63.2 & 61.7 & 57.3 & 63.2 & 65.2 & 74.2 & 74.2 & 93.3 & 94.9 & 76.6 & 75.8 & 73.9 & 51.7 & 86.5 & 86.1 & 73.8 & 74.1 \\ 
  &  & RMaha & 64.7 & 45.8 & 68.0 & 43.6 & 85.5 & 84.1 & 62.3 & 61.7 & 58.3 & 37.8 & 38.3 & 21.8 & 20.1 & 84.2 & 74.5 & $\bm{71.0}$ & $\bm{44.5}$ & $\bm{71.5}$ & $\bm{43.7}$ & $\bm{81.2}$ & 73.6 & 59.8 & 54.2 \\ 
  &  & RCos & 68.4 & 49.7 & 70.0 & 45.9 & $\bm{74.0}$ & $\bm{71.2}$ & 45.0 & 43.4 & 37.2 & 38.8 & 39.4 & 25.0 & 23.8 & 83.0 & $\bm{68.9}$ & 73.4 & 47.0 & 73.1 & 44.4 & 81.8 & $\bm{68.3}$ & $\bm{56.2}$ & $\bm{49.1}$ \\ 
  &  & Cos & 72.8 & 53.6 & 82.2 & 66.3 & 84.2 & 82.8 & 57.0 & 55.2 & 49.7 & 52.0 & 53.3 & 51.0 & 50.4 & 92.1 & 92.3 & 73.4 & 64.8 & 72.3 & 48.3 & 83.3 & 84.6 & 64.8 & 62.7 \\ 
 \hline 
\end{tabular}
\end{center}
\end{table}
\renewcommand{\arraystretch}{.9}
\tabcolsep=0.0310321cm

\begin{table}[htb]
\rowcolors{2}{white}{gray!10}
    \centering
    \caption{Comparing the cleaned and original datasets in terms of \FPR. The best method per model and dataset is marked bold.}
    \label{tab:clean-vs-all-1k-conv}
    \tiny
        \begin{center}
\begin{tabular}{c c c | c  c  c  c  c  c  c  c  c  c  c  c  c  c  c  c  c  c  c  c  c  c  c }
 &  & &   &   &   &   &   &   &   &   &   &   &   & fpr &   &   &   &   &   &   &   &   &   &   &   \\ 
 model &acc. & method & Pl-f & Pl-c & Spc-f & Spc-c & IN-f & IN-c & txt-f & txt-43 & txt-c & OpO-f & OpO-c & iNat-f & iNat-c & IN1K-f & IN1K-c & OS-f & OS-c & SBe-f & SBe-c & SBh-f & SBh-c & CO-f & CO-c \\ 
 \hline 
 &  & MSP & 70.1 & 58.8 & 79.8 & 69.2 & 86.0 & 84.1 & 61.5 & 59.8 & 54.2 & 59.0 & 59.8 & 49.0 & 48.0 & 92.1 & 84.3 & 79.0 & 64.8 & 75.9 & 58.3 & 88.5 & $\bm{78.4}$ & 64.8 & 63.7 \\ 
  &  & MaxL & 70.9 & 60.1 & 80.0 & 68.6 & $\bm{82.8}$ & 81.6 & 58.0 & 56.3 & 50.0 & 58.5 & 59.2 & 47.2 & 46.0 & 90.9 & 85.1 & 79.0 & 62.3 & 75.9 & 59.6 & 87.0 & 80.8 & 66.2 & 63.7 \\ 
  &  & ViM & 64.7 & 48.4 & 74.0 & 52.9 & 84.0 & $\bm{81.2}$ & 58.2 & 56.8 & 49.3 & 43.0 & $\bm{42.9}$ & 26.8 & 25.6 & 89.7 & 84.3 & 71.8 & $\bm{47.0}$ & 71.9 & $\bm{47.7}$ & 83.9 & 79.8 & $\bm{56.5}$ & $\bm{50.5}$ \\ 
  &  & Maha & 63.7 & 47.7 & 75.0 & 54.7 & 85.8 & 83.5 & 65.2 & 64.1 & 57.3 & 46.8 & 47.3 & 29.2 & 28.2 & 87.9 & 85.5 & 72.6 & 50.0 & 71.5 & 51.0 & $\bm{82.8}$ & 82.2 & 60.0 & 53.8 \\ 
  &  & E+R & 70.6 & 60.1 & 82.0 & 67.4 & 83.0 & 82.2 & 61.5 & 60.1 & 53.5 & 59.0 & 59.8 & 49.8 & 49.1 & 92.7 & 87.7 & 78.2 & 64.4 & 76.3 & 60.9 & 85.4 & 82.2 & 69.2 & 68.9 \\ 
 XCiT-M-224 & 82.6 & Ener & 77.8 & 70.6 & 82.2 & 72.7 & 85.0 & 84.5 & 62.5 & 61.7 & 55.9 & 64.0 & 64.7 & 59.0 & 57.7 & 92.1 & 87.7 & 83.9 & 69.9 & 82.3 & 66.9 & 87.5 & 83.7 & 74.0 & 71.7 \\ 
  &  & KL-M & 72.3 & 59.5 & 78.8 & 70.3 & 83.8 & 82.2 & 61.5 & 60.1 & 55.2 & 60.5 & 60.6 & 52.0 & 51.2 & 89.1 & 82.6 & 77.4 & 64.0 & 79.5 & 56.3 & 88.0 & $\bm{78.4}$ & 64.2 & 59.0 \\ 
  &  & KNN & 65.9 & 51.0 & 82.5 & 65.7 & 84.2 & 82.8 & 55.0 & 53.6 & 45.5 & 50.7 & 51.6 & 49.5 & 48.8 & 93.3 & 93.2 & 74.2 & 61.0 & 71.5 & 48.3 & 83.3 & 83.7 & 63.5 & 62.7 \\ 
  &  & RMaha & 64.4 & 45.8 & $\bm{72.2}$ & $\bm{50.6}$ & 85.5 & 83.2 & 64.2 & 63.0 & 55.9 & $\bm{42.2}$ & $\bm{42.9}$ & $\bm{25.2}$ & $\bm{24.0}$ & $\bm{87.3}$ & $\bm{81.3}$ & $\bm{71.0}$ & 47.9 & 71.9 & 51.0 & 83.3 & 79.3 & 58.0 & 50.9 \\ 
  &  & RCos & 62.5 & $\bm{45.1}$ & 76.2 & 57.0 & 83.2 & $\bm{81.2}$ & 55.0 & 53.6 & 45.8 & 42.8 & 43.8 & 32.0 & 31.1 & 90.9 & 86.0 & $\bm{71.0}$ & 52.1 & 71.5 & $\bm{47.7}$ & 85.4 & 78.8 & 58.2 & 51.4 \\ 
  &  & Cos & $\bm{62.0}$ & $\bm{45.1}$ & 77.5 & 58.1 & 83.2 & $\bm{81.2}$ & $\bm{54.2}$ & $\bm{52.5}$ & $\bm{44.8}$ & 44.5 & 45.4 & 33.8 & 32.9 & 91.5 & 87.7 & 71.8 & 53.0 & $\bm{71.1}$ & $\bm{47.7}$ & 84.4 & 81.2 & 58.8 & 53.3 \\ 
 \hline 
 &  & MSP & 72.8 & 58.2 & 74.2 & 61.6 & 87.5 & 87.7 & 60.2 & 58.7 & 53.1 & 55.5 & 55.2 & 49.8 & 49.1 & 91.5 & 79.6 & 79.0 & 64.4 & 75.5 & 55.6 & 83.3 & 76.9 & 65.2 & 61.3 \\ 
  &  & MaxL & 70.1 & 56.2 & 73.5 & 59.9 & 86.2 & 85.8 & 52.2 & 50.7 & 44.4 & 55.5 & 55.4 & 40.0 & 38.9 & 88.5 & 80.9 & 78.2 & 61.9 & 75.9 & 58.3 & 83.3 & 75.5 & 64.2 & 60.8 \\ 
  &  & ViM & 69.4 & 49.0 & 70.2 & $\bm{43.6}$ & 84.0 & $\bm{82.5}$ & 49.0 & 47.5 & 38.2 & $\bm{33.5}$ & $\bm{32.9}$ & $\bm{18.8}$ & $\bm{17.0}$ & 87.3 & 77.9 & $\bm{72.6}$ & $\bm{47.5}$ & 70.3 & 41.1 & $\bm{82.8}$ & 73.1 & 59.2 & 51.9 \\ 
  &  & Maha & 71.4 & 52.3 & 73.2 & 51.2 & 88.2 & 87.1 & 55.5 & 53.9 & 45.5 & 41.5 & 41.3 & 26.0 & 24.5 & 87.9 & 79.6 & 74.2 & 53.0 & 71.5 & 44.4 & 83.9 & 74.0 & 62.5 & 56.6 \\ 
  &  & E+R & 74.1 & 58.8 & 81.8 & 69.8 & 85.5 & 84.8 & 54.8 & 53.4 & 45.5 & 58.5 & 58.4 & 48.8 & 47.5 & 91.5 & 87.7 & 78.2 & 68.6 & 75.5 & 59.6 & 86.5 & 81.7 & 67.8 & 67.9 \\ 
 XCiT-M-224-d & 84.3 & Ener & 77.0 & 63.4 & 80.8 & 73.8 & 86.8 & 86.1 & 58.8 & 57.1 & 50.7 & 66.5 & 66.3 & 50.0 & 49.1 & 90.9 & 86.4 & 81.5 & 68.2 & 85.5 & 72.2 & 88.5 & 81.7 & 74.2 & 70.8 \\ 
  &  & KL-M & 75.3 & 63.4 & 76.0 & 60.5 & $\bm{83.0}$ & 83.5 & 56.2 & 54.4 & 46.9 & 56.2 & 56.0 & 45.2 & 44.6 & 90.3 & 76.6 & 75.8 & 58.9 & 73.5 & 49.0 & $\bm{82.8}$ & 74.5 & 63.2 & 57.5 \\ 
  &  & KNN & 73.1 & 54.9 & 82.2 & 65.7 & 86.0 & 84.8 & $\bm{47.0}$ & $\bm{45.0}$ & $\bm{35.4}$ & 41.5 & 42.4 & 41.8 & 40.7 & 93.9 & 90.2 & 79.8 & 58.5 & 69.1 & 40.4 & 84.4 & 81.7 & 62.3 & 59.9 \\ 
  &  & RMaha & 71.6 & 51.0 & $\bm{69.8}$ & 46.5 & 87.8 & 86.4 & 54.8 & 53.1 & 45.1 & 39.2 & 38.9 & 22.5 & 20.9 & $\bm{86.7}$ & $\bm{72.3}$ & 74.2 & 50.4 & 71.1 & 43.0 & 84.4 & $\bm{71.2}$ & 61.0 & 54.7 \\ 
  &  & RCos & $\bm{67.2}$ & $\bm{44.4}$ & 73.0 & 51.2 & 84.8 & 84.1 & 48.0 & 46.1 & 36.8 & 36.5 & 36.7 & 25.2 & 23.8 & 92.1 & 80.9 & 75.8 & 51.3 & 69.1 & 41.1 & 83.3 & 75.0 & $\bm{58.2}$ & $\bm{50.5}$ \\ 
  &  & Cos & 68.6 & 48.4 & 75.0 & 53.5 & 84.8 & 83.2 & 48.0 & 46.1 & 36.1 & 39.0 & 38.9 & 28.5 & 26.9 & 91.5 & 82.6 & 76.6 & 50.4 & $\bm{67.5}$ & $\bm{39.7}$ & 83.9 & 76.9 & 58.8 & 52.8 \\ 
 \hline 
 &  & MSP & 69.4 & 55.6 & $\bm{66.8}$ & 50.6 & 91.5 & 91.3 & 69.5 & 67.3 & 61.8 & 62.7 & 63.0 & 39.2 & 37.9 & 85.5 & 74.9 & 79.8 & 62.3 & 78.7 & 59.6 & 83.3 & 72.6 & 68.2 & 64.2 \\ 
  &  & MaxL & 77.8 & 66.7 & 72.2 & 60.5 & 94.0 & 93.9 & 75.8 & 75.1 & 69.8 & 78.0 & 78.3 & 49.8 & 49.1 & $\bm{84.8}$ & 74.0 & 88.7 & 74.6 & 88.8 & 79.5 & 87.5 & 75.5 & 77.0 & 73.1 \\ 
  &  & ViM & 66.7 & 44.4 & 74.2 & 51.7 & 84.8 & 83.8 & 51.2 & 49.3 & 39.9 & 37.2 & 37.8 & 32.8 & 31.3 & 87.9 & 80.0 & $\bm{70.2}$ & 48.3 & 73.5 & 43.7 & 85.4 & 77.4 & 59.0 & 52.4 \\ 
  &  & Maha & 66.2 & 43.8 & 74.8 & 55.2 & 87.8 & 87.1 & 54.5 & 52.8 & 44.1 & 37.8 & 39.1 & 39.5 & 37.6 & 87.9 & 80.4 & 71.8 & 50.4 & 71.5 & 41.7 & 84.4 & 76.0 & 60.2 & 55.2 \\ 
  &  & E+R & 92.1 & 88.9 & 84.2 & 86.0 & 95.8 & 95.5 & 89.8 & 89.8 & 87.8 & 92.5 & 92.7 & 76.8 & 76.5 & 87.3 & 86.0 & 97.6 & 89.8 & 94.4 & 96.7 & 89.6 & 83.2 & 92.8 & 92.5 \\ 
 CnvNxt-B & 84.4 & Ener & 94.3 & 94.8 & 86.2 & 90.7 & 96.2 & 96.1 & 93.2 & 93.6 & 92.4 & 95.8 & 95.9 & 85.8 & 85.6 & 89.1 & 85.5 & 98.4 & 93.6 & 96.8 & 98.0 & 90.6 & 85.1 & 96.2 & 95.8 \\ 
  &  & KL-M & 74.3 & 57.5 & 79.0 & 62.8 & 85.2 & 85.8 & 61.3 & 59.5 & 52.4 & 53.5 & 54.1 & 48.0 & 46.0 & 88.5 & 82.6 & 73.4 & 58.1 & 71.9 & 51.0 & $\bm{82.8}$ & 79.8 & 63.7 & 60.8 \\ 
  &  & KNN & 74.8 & 56.9 & 82.0 & 65.1 & 84.5 & $\bm{83.2}$ & $\bm{48.8}$ & $\bm{46.6}$ & $\bm{36.1}$ & 43.8 & 44.8 & 46.8 & 46.2 & 93.3 & 88.9 & 76.6 & 60.6 & 71.9 & 44.4 & 90.1 & 83.2 & 62.3 & 59.0 \\ 
  &  & RMaha & $\bm{64.7}$ & 42.5 & 70.8 & 50.6 & 87.8 & 87.1 & 54.2 & 52.5 & 43.8 & 35.2 & 36.4 & 29.8 & 27.7 & $\bm{84.8}$ & $\bm{71.5}$ & 71.8 & 47.0 & 70.3 & 42.4 & 83.9 & $\bm{71.2}$ & 57.5 & 52.8 \\ 
  &  & RCos & 65.9 & $\bm{41.8}$ & 70.0 & $\bm{48.8}$ & $\bm{84.2}$ & $\bm{83.2}$ & 49.0 & 46.9 & 36.5 & $\bm{34.8}$ & $\bm{35.9}$ & $\bm{26.5}$ & $\bm{25.1}$ & 89.7 & 79.1 & 75.8 & $\bm{44.1}$ & $\bm{69.9}$ & $\bm{40.4}$ & 84.4 & 73.1 & $\bm{55.2}$ & $\bm{49.1}$ \\ 
  &  & Cos & 67.4 & 43.1 & 72.8 & 53.5 & 85.5 & 84.8 & 49.2 & 47.2 & 37.2 & 35.8 & 36.4 & 33.0 & 31.9 & 89.7 & 83.4 & 76.6 & 44.5 & $\bm{69.9}$ & 42.4 & 85.9 & 78.8 & 58.0 & 53.3 \\ 
 \hline 
 &  & MSP & 78.8 & 65.4 & 84.2 & 74.4 & 97.0 & 96.8 & 76.0 & 74.8 & 71.9 & 71.8 & 72.8 & 60.8 & 59.8 & 89.1 & 90.6 & $\bm{79.8}$ & 73.3 & 80.7 & 60.3 & 87.0 & 88.9 & 75.2 & 73.6 \\ 
  &  & MaxL & 75.3 & 60.1 & 85.0 & 74.4 & 98.0 & 98.1 & 72.2 & 70.8 & 68.4 & 71.0 & 72.3 & 75.0 & 75.2 & 89.7 & 92.8 & 82.3 & 77.5 & $\bm{79.1}$ & 64.9 & 88.0 & 91.3 & 77.8 & 80.2 \\ 
  &  & ViM & 69.1 & 58.2 & 80.8 & 66.3 & $\bm{64.8}$ & $\bm{62.8}$ & $\bm{6.2}$ & $\bm{5.6}$ & $\bm{3.5}$ & 44.5 & 42.4 & 53.5 & 53.5 & 94.5 & 94.9 & 87.1 & 62.7 & 87.1 & $\bm{57.6}$ & 97.4 & 91.3 & 75.5 & 67.5 \\ 
  &  & Maha & 80.5 & 72.5 & 83.2 & 70.3 & 70.8 & 67.6 & 17.0 & 16.1 & 13.9 & 62.5 & 60.3 & 61.3 & 60.6 & 93.9 & 97.0 & 96.0 & 71.6 & 93.6 & 79.5 & 95.8 & 90.4 & 82.0 & 74.1 \\ 
  &  & E+R & $\bm{56.0}$ & $\bm{39.2}$ & 76.8 & 61.6 & 90.8 & 90.6 & 51.0 & 49.3 & 45.1 & 53.8 & 53.3 & 48.2 & 48.6 & 83.6 & 87.2 & 83.9 & 66.1 & 81.9 & 70.9 & 86.5 & 88.9 & 73.2 & 71.7 \\ 
 BiT-s & 78.0 & Ener & 75.6 & 61.4 & 85.2 & 75.6 & 97.8 & 97.7 & 73.0 & 71.6 & 68.1 & 70.2 & 71.2 & 79.0 & 80.2 & 89.7 & 92.8 & 84.7 & 78.4 & 79.9 & 76.2 & 90.1 & 94.2 & 79.5 & 81.6 \\ 
  &  & KL-M & 76.5 & 62.7 & 70.0 & 52.9 & 87.8 & 87.7 & 52.2 & 50.9 & 46.2 & 51.0 & 51.6 & 31.5 & 29.8 & 80.6 & 72.3 & $\bm{79.8}$ & 58.9 & 82.3 & 62.3 & $\bm{83.3}$ & 72.1 & 69.2 & 60.8 \\ 
  &  & KNN & 84.0 & 74.5 & 85.2 & 73.3 & 65.5 & 63.1 & 13.0 & 13.1 & 9.4 & 61.3 & 59.2 & 79.2 & 79.9 & 94.5 & 96.2 & 91.9 & 75.8 & 93.6 & 78.8 & 97.9 & 96.2 & 84.5 & 77.4 \\ 
  &  & RMaha & 80.0 & 69.3 & $\bm{56.0}$ & $\bm{34.3}$ & 77.8 & 77.7 & 24.2 & 22.5 & 15.3 & $\bm{42.2}$ & $\bm{40.8}$ & $\bm{24.2}$ & $\bm{22.5}$ & $\bm{77.0}$ & $\bm{68.9}$ & 83.1 & $\bm{47.9}$ & 87.6 & 67.5 & 84.9 & $\bm{63.5}$ & 70.5 & $\bm{56.1}$ \\ 
  &  & RCos & 90.1 & 85.6 & 87.8 & 83.7 & 93.5 & 93.5 & 66.2 & 65.1 & 60.8 & 75.5 & 77.2 & 87.8 & 88.3 & 86.7 & 92.8 & 87.9 & 88.6 & 89.6 & 76.2 & 89.1 & 93.8 & 85.8 & 84.0 \\ 
  &  & Cos & 68.9 & 53.6 & 74.5 & 57.6 & 67.2 & 64.4 & 11.8 & 10.5 & 7.6 & 45.0 & 42.1 & 39.0 & 38.6 & 87.3 & 90.2 & 87.1 & 58.5 & 90.4 & $\bm{57.6}$ & 91.1 & 89.4 & $\bm{68.2}$ & $\bm{56.1}$ \\ 
 \hline 
 &  & MSP & 66.9 & 50.3 & 72.0 & 52.9 & 87.5 & 87.4 & 54.8 & 53.4 & 47.2 & 53.2 & 54.3 & 45.2 & 44.4 & 85.5 & 77.0 & 77.4 & 56.4 & 74.3 & 53.6 & 84.9 & 71.2 & 59.2 & 54.2 \\ 
  &  & MaxL & 69.1 & 52.3 & 73.0 & 54.1 & 86.5 & 87.1 & 53.8 & 52.0 & 45.5 & 56.2 & 56.8 & 45.2 & 44.4 & 87.3 & 77.4 & 79.8 & 56.4 & 75.1 & 51.0 & 85.9 & 73.1 & 60.5 & 55.7 \\ 
  &  & ViM & 68.1 & 51.6 & 74.8 & 52.9 & 80.2 & 79.0 & 47.8 & 45.6 & 37.5 & 37.8 & 38.3 & 33.0 & 32.4 & 90.9 & 90.2 & 69.4 & 55.1 & 71.9 & 48.3 & 82.3 & 85.1 & 61.8 & 61.3 \\ 
  &  & Maha & 66.9 & 48.4 & 68.2 & 41.9 & 75.5 & 73.8 & 40.2 & 38.1 & 29.5 & 30.8 & 31.0 & 18.2 & 17.0 & 86.1 & 79.1 & 69.4 & 45.3 & 69.1 & 41.1 & $\bm{76.0}$ & 74.5 & 53.2 & 48.6 \\ 
  &  & E+R & 69.9 & 49.7 & 78.5 & 59.3 & 79.8 & 79.0 & 45.5 & 43.4 & 35.1 & 47.8 & 47.8 & 43.2 & 42.8 & 88.5 & 88.9 & 74.2 & 58.9 & 70.7 & 50.3 & 85.9 & 82.2 & 65.8 & 64.6 \\ 
 EffNetv2-M & 85.1 & Ener & 76.8 & 64.7 & 80.8 & 69.8 & 86.5 & 87.7 & 61.5 & 60.3 & 54.9 & 66.0 & 66.8 & 63.5 & 63.4 & 87.9 & 83.0 & 82.3 & 72.5 & 78.7 & 62.3 & 89.1 & 78.4 & 71.2 & 72.6 \\ 
  &  & KL-M & 71.1 & 56.9 & 70.8 & 50.6 & 81.5 & 81.6 & 54.5 & 53.6 & 46.9 & 49.8 & 49.5 & 41.0 & 40.2 & 84.8 & 71.5 & 73.4 & 51.7 & 71.1 & 51.0 & 82.8 & 71.2 & 61.8 & 53.8 \\ 
  &  & KNN & 69.9 & 51.6 & 74.8 & 51.7 & 80.5 & 79.3 & 43.5 & 41.0 & 32.6 & 38.5 & 39.9 & 35.5 & 34.5 & 89.7 & 84.7 & 75.8 & 51.3 & 69.1 & 44.4 & 83.3 & 79.3 & 55.8 & 52.4 \\ 
  &  & RMaha & 64.4 & 43.8 & $\bm{62.7}$ & $\bm{34.9}$ & 75.5 & 74.1 & 38.8 & 36.5 & 27.8 & $\bm{27.8}$ & $\bm{28.3}$ & $\bm{15.0}$ & $\bm{13.6}$ & $\bm{81.2}$ & $\bm{65.5}$ & $\bm{67.7}$ & $\bm{36.9}$ & $\bm{67.5}$ & 39.1 & 76.6 & $\bm{65.4}$ & 50.5 & $\bm{42.9}$ \\ 
  &  & RCos & $\bm{59.3}$ & $\bm{35.9}$ & 65.8 & 39.5 & $\bm{74.8}$ & $\bm{72.8}$ & $\bm{34.5}$ & $\bm{32.2}$ & $\bm{22.9}$ & 29.5 & 29.6 & 18.2 & 16.7 & 86.7 & 72.3 & 70.2 & 39.4 & 67.9 & $\bm{38.4}$ & 80.2 & 67.3 & $\bm{50.2}$ & 43.4 \\ 
  &  & Cos & 65.9 & 43.8 & 71.2 & 47.7 & 80.2 & 79.3 & 43.0 & 40.8 & 31.6 & 33.8 & 34.8 & 27.0 & 25.8 & 87.9 & 77.4 & 73.4 & 44.9 & 68.7 & 44.4 & 81.8 & 72.6 & 54.5 & 49.5 \\ 
 \hline 
 &  & MSP & 66.7 & 59.5 & 68.5 & 50.6 & 86.5 & 86.4 & 55.5 & 53.6 & 46.5 & 52.2 & 52.7 & 44.2 & 43.1 & 84.2 & 77.9 & 77.4 & 57.6 & 73.9 & 57.6 & 83.9 & 73.1 & 60.8 & 56.6 \\ 
  &  & MaxL & 69.4 & 60.8 & 76.8 & 60.5 & 85.8 & 85.8 & 57.5 & 55.5 & 47.6 & 57.0 & 57.6 & 50.5 & 49.6 & 84.8 & 79.1 & 82.3 & 63.1 & 78.3 & 60.3 & 85.4 & 73.6 & 62.3 & 59.0 \\ 
  &  & ViM & 74.6 & 63.4 & 79.8 & 64.5 & 73.5 & 72.5 & 61.5 & 60.6 & 55.6 & 49.8 & 50.3 & 41.2 & 40.7 & 89.1 & 91.5 & $\bm{67.7}$ & 61.0 & 77.5 & 49.7 & 84.4 & 84.6 & 66.5 & 63.7 \\ 
  &  & Maha & 73.3 & 59.5 & 73.5 & 52.3 & 73.5 & 73.5 & 59.8 & 58.7 & 53.1 & 41.8 & 42.7 & 29.2 & 28.2 & 87.9 & 87.7 & 71.0 & 54.2 & 77.5 & 45.0 & 86.5 & 83.2 & 63.2 & 57.1 \\ 
  &  & E+R & 76.5 & 65.4 & 82.0 & 65.7 & 80.2 & 80.6 & 54.2 & 52.3 & 42.4 & 58.0 & 59.2 & 56.8 & 56.1 & 91.5 & 88.9 & 76.6 & 69.1 & 79.1 & 58.3 & 85.4 & 78.8 & 70.2 & 68.4 \\ 
 EffNetb7 & 84.9 & Ener & 83.5 & 77.1 & 87.0 & 76.2 & 87.2 & 88.0 & 65.8 & 64.1 & 56.6 & 73.8 & 74.2 & 72.2 & 71.8 & 89.1 & 85.5 & 87.1 & 77.5 & 87.1 & 72.2 & 89.6 & 77.4 & 76.8 & 76.4 \\ 
  &  & KL-M & 69.4 & 57.5 & 70.2 & 48.8 & 79.0 & 78.3 & 50.2 & 48.3 & 42.4 & 44.0 & 45.1 & 38.0 & 36.3 & 86.1 & 75.7 & 69.4 & 50.8 & 73.5 & 54.3 & $\bm{82.8}$ & 73.6 & 61.0 & 53.3 \\ 
  &  & KNN & 75.1 & 59.5 & 77.8 & 58.1 & 77.0 & 76.7 & 47.5 & 45.0 & 35.4 & 42.2 & 44.3 & 44.2 & 43.1 & 87.9 & 86.0 & 71.8 & 58.1 & $\bm{69.5}$ & 43.0 & 84.9 & 81.2 & 59.8 & 56.6 \\ 
  &  & RMaha & 71.4 & 56.9 & $\bm{63.2}$ & $\bm{37.2}$ & 74.8 & 73.5 & 51.0 & 49.3 & 42.4 & 31.0 & 32.1 & 20.8 & 19.6 & $\bm{82.4}$ & $\bm{64.3}$ & 71.0 & 42.4 & 75.5 & 43.7 & $\bm{82.8}$ & $\bm{66.3}$ & 56.2 & $\bm{45.3}$ \\ 
  &  & RCos & $\bm{63.0}$ & $\bm{39.9}$ & 65.5 & 39.5 & $\bm{72.0}$ & $\bm{70.2}$ & $\bm{39.0}$ & $\bm{37.0}$ & $\bm{27.8}$ & $\bm{29.5}$ & $\bm{30.7}$ & $\bm{18.2}$ & $\bm{17.0}$ & 83.0 & 74.0 & 73.4 & $\bm{39.8}$ & 69.9 & $\bm{35.1}$ & 84.4 & 68.8 & $\bm{53.2}$ & $\bm{45.3}$ \\ 
  &  & Cos & 70.4 & 51.6 & 71.8 & 50.0 & 76.5 & 74.4 & 51.0 & 48.8 & 39.2 & 34.5 & 36.1 & 30.8 & 29.5 & 84.8 & 78.7 & 72.6 & 47.0 & $\bm{69.5}$ & 48.3 & 84.4 & 73.1 & 56.5 & 51.9 \\ 
 \hline 
 &  & MSP & 77.0 & 67.3 & 80.2 & 69.8 & 95.2 & 94.5 & 66.8 & 65.4 & 62.5 & 65.0 & 64.9 & 54.0 & 53.3 & 92.1 & 79.6 & $\bm{79.8}$ & 67.8 & $\bm{79.1}$ & 64.2 & 89.1 & 75.5 & 68.0 & 63.2 \\ 
  &  & MaxL & 80.7 & 70.6 & 83.5 & 73.8 & 93.8 & 93.2 & 66.5 & 65.1 & 60.1 & 67.0 & 66.8 & 61.5 & 60.8 & 88.5 & $\bm{73.2}$ & 83.9 & 69.9 & 80.7 & 66.9 & $\bm{85.9}$ & 71.2 & 69.2 & 62.7 \\ 
  &  & ViM & 84.2 & 75.2 & 79.5 & 65.7 & 74.5 & 71.5 & 24.8 & 23.6 & 17.0 & 56.8 & 55.2 & 54.8 & 53.8 & 90.3 & 93.2 & 81.5 & 74.6 & 83.5 & $\bm{53.0}$ & 91.1 & 90.9 & 77.2 & 73.1 \\ 
  &  & Maha & 90.1 & 85.0 & 83.0 & 72.7 & 77.8 & 76.1 & 44.2 & 44.2 & 37.8 & 70.5 & 70.9 & 65.2 & 64.0 & 92.7 & 95.3 & 89.5 & 87.3 & 90.0 & 69.5 & 94.8 & 92.8 & 84.5 & 83.5 \\ 
  &  & E+R & 82.5 & 72.5 & 83.5 & 71.5 & 84.0 & 82.5 & 50.0 & 49.6 & 44.1 & 64.8 & 64.1 & 63.5 & 62.9 & 90.9 & 74.5 & 86.3 & 68.6 & 85.9 & 71.5 & $\bm{85.9}$ & 72.1 & 72.0 & 61.8 \\ 
 EffNet-B0 & 77.7 & Ener & 86.9 & 77.8 & 86.2 & 79.1 & 91.5 & 91.9 & 72.8 & 72.4 & 69.4 & 77.2 & 76.9 & 75.8 & 75.7 & 87.9 & 74.9 & 89.5 & 78.0 & 85.9 & 85.4 & 87.0 & $\bm{70.7}$ & 76.8 & 70.3 \\ 
  &  & KL-M & 80.7 & 68.0 & 77.0 & 64.5 & 90.8 & 90.9 & 61.8 & 61.1 & 58.0 & 59.8 & 59.5 & 45.2 & 44.1 & $\bm{86.7}$ & 83.0 & 81.5 & 66.1 & 82.7 & 70.9 & 88.0 & 82.2 & 70.5 & 63.2 \\ 
  &  & KNN & 94.1 & 90.8 & 87.8 & 83.1 & $\bm{70.2}$ & $\bm{67.0}$ & 30.2 & 29.0 & 22.2 & 74.0 & 72.8 & 75.5 & 75.2 & 98.2 & 96.6 & 96.8 & 82.6 & 96.0 & 84.1 & 97.9 & 92.8 & 86.8 & 79.7 \\ 
  &  & RMaha & 83.2 & 69.3 & 71.8 & 57.6 & 89.0 & 88.0 & 55.0 & 54.2 & 50.3 & 61.8 & 61.7 & 46.5 & 45.2 & 87.9 & 83.0 & 86.3 & 71.2 & 83.1 & 69.5 & 90.1 & 88.9 & 77.0 & 70.3 \\ 
  &  & RCos & 77.5 & 62.1 & 64.8 & 48.8 & 89.5 & 89.3 & 53.0 & 51.7 & 45.5 & 56.2 & 56.0 & 38.2 & 37.1 & $\bm{86.7}$ & 80.0 & 83.1 & 66.9 & 82.7 & 67.5 & 88.0 & 84.1 & 71.8 & 64.2 \\ 
  &  & Cos & $\bm{71.6}$ & $\bm{54.2}$ & $\bm{59.0}$ & $\bm{41.9}$ & 73.2 & 72.2 & $\bm{23.0}$ & $\bm{21.7}$ & $\bm{14.2}$ & $\bm{41.5}$ & $\bm{38.6}$ & $\bm{24.2}$ & $\bm{22.5}$ & 88.5 & 79.1 & 80.6 & $\bm{49.2}$ & 83.5 & 61.6 & $\bm{85.9}$ & 80.8 & $\bm{65.0}$ & $\bm{54.7}$ \\ 
 \hline 
 &  & MSP & $\bm{73.1}$ & $\bm{58.2}$ & 79.5 & 69.2 & 96.0 & 95.5 & 63.7 & 62.5 & 58.3 & 62.3 & 62.5 & 51.0 & 49.9 & 92.7 & 83.0 & 83.1 & 67.4 & $\bm{83.1}$ & 60.9 & 90.1 & 81.2 & 66.5 & 65.1 \\ 
  &  & MaxL & 75.8 & 64.1 & 79.2 & 67.4 & 96.5 & 96.1 & 67.5 & 66.0 & 62.2 & 65.2 & 64.9 & 54.0 & 53.0 & 92.1 & 83.4 & 83.1 & 67.4 & 83.5 & 62.3 & 90.1 & 80.3 & 68.8 & 68.9 \\ 
  &  & ViM & 86.9 & 79.7 & 84.8 & 75.6 & 89.0 & 88.3 & 40.8 & 39.9 & 33.7 & 64.8 & 66.6 & 74.0 & 73.6 & 88.5 & 87.7 & $\bm{81.5}$ & 77.1 & 83.9 & 55.6 & 88.0 & 89.9 & 75.2 & 75.0 \\ 
  &  & Maha & 94.1 & 90.2 & 89.8 & 84.3 & 79.8 & 78.3 & 53.2 & 53.1 & 49.7 & 77.5 & 78.5 & 85.8 & 85.4 & 92.7 & 89.8 & 87.1 & 83.5 & 89.2 & 73.5 & 90.1 & 92.3 & 86.2 & 85.8 \\ 
  &  & E+R & 99.8 & 100.0 & 99.2 & 99.4 & 92.0 & 92.2 & 98.2 & 98.1 & 98.3 & 96.5 & 97.0 & 99.8 & 99.7 & 98.2 & 99.1 & 92.7 & 97.9 & 91.6 & 92.7 & 96.9 & 98.6 & 95.8 & 98.1 \\ 
 ResNet50 & 80.4 & Ener & 83.7 & 78.4 & 81.5 & 72.1 & 96.0 & 95.5 & 76.0 & 75.3 & 74.0 & 72.0 & 72.6 & 68.5 & 67.6 & 90.9 & 84.3 & 88.7 & 74.6 & 85.5 & 68.2 & 88.0 & 81.7 & 74.5 & 74.1 \\ 
  &  & KL-M & 75.3 & 62.1 & 75.0 & 62.8 & 93.2 & 92.2 & 58.8 & 57.1 & 51.7 & 57.2 & 57.1 & 44.5 & 43.6 & 91.5 & 77.4 & 82.3 & 62.3 & $\bm{83.1}$ & 60.3 & 84.9 & 77.4 & 66.5 & 62.3 \\ 
  &  & KNN & 84.2 & 73.9 & 82.2 & 69.2 & $\bm{68.5}$ & $\bm{65.0}$ & $\bm{24.0}$ & $\bm{23.1}$ & $\bm{16.0}$ & 58.2 & 57.3 & 58.8 & 58.7 & 89.1 & 88.9 & 84.7 & 70.3 & 89.2 & 67.5 & 89.6 & 87.5 & 78.8 & 70.8 \\ 
  &  & RMaha & 93.1 & 89.5 & 70.8 & 62.2 & 79.5 & 79.3 & 75.5 & 75.6 & 73.6 & 75.0 & 76.4 & 83.2 & 82.8 & 86.1 & $\bm{66.4}$ & 85.5 & 76.7 & 88.4 & 79.5 & $\bm{79.7}$ & $\bm{73.1}$ & 81.8 & 75.5 \\ 
  &  & RCos & 79.8 & 64.7 & $\bm{65.2}$ & $\bm{47.1}$ & 82.5 & 79.9 & 39.0 & 36.7 & 27.8 & 47.5 & 48.6 & 30.8 & 29.2 & $\bm{80.0}$ & 70.6 & 82.3 & 49.2 & 84.3 & 60.9 & 80.2 & $\bm{73.1}$ & 65.8 & 53.3 \\ 
  &  & Cos & 76.3 & 60.8 & 67.2 & $\bm{47.1}$ & 78.0 & 75.7 & 32.2 & 30.0 & 21.2 & $\bm{43.0}$ & $\bm{42.9}$ & $\bm{25.2}$ & $\bm{23.8}$ & $\bm{80.0}$ & 74.0 & 83.1 & $\bm{44.9}$ & $\bm{83.1}$ & $\bm{52.3}$ & 81.2 & 73.6 & $\bm{63.0}$ & $\bm{52.8}$ \\ 
 \hline 
\end{tabular}
\end{center}

\end{table}

\renewcommand{\arraystretch}{1.}
\FloatBarrier
\section{Results on NINCO classes with and without overlap with IN-21K}\label{sec:overlap21K}
Since the classes of \dsetname{} can be distinguished by whether they belong to an IN-21k class or not, we present results on both of these groups here. We note that they should be taken with care, since the groups differ both in size (9 vs. 55 classes) and difficulty of the individual classes. Most models and methods perform better on the classes \textit{with} IN-21k overlap, and ViT+Maha is the best OOD-detector in both cases. While RMaha and (Relative) Cosine yield the most consistent improvements over MSP in both cases, ViM performs comparably better on the classes without overlap. Pretraining \textit{only} on IN-21k yields the best OOD-detectors in both cases.
\renewcommand{\arraystretch}{1.}
\tabcolsep=0.1071cm
\begin{table*}[htb]
    \caption{Mean \FPR{} for classes without 21k overlap.}
    \centering
    \small
    \begin{smaller}
\begin{center}
\begin{tabular}{l l l l l l l l l l l l l l }
pre & acc. & model & MSP & MaxL & Ener & KL-M & Maha & RMaha & ViM & E+R & KNN & Cos & MCM/RCos \\ 
 \hline 
{\multirow{9}{*}{{21k}}} & 86.0 & ViT-B-384 & 56.5 & 41.8 ${\textcolor{green}{-15}}$ & 39.6 ${\textcolor{green}{-17}}$ & 51.7 ${\textcolor{green}{-5}}$ & \textbf{31.7} ${\textcolor{green}{-25}}$ & 36.9 ${\textcolor{green}{-20}}$ & 32.2 ${\textcolor{green}{-24}}$ & 40.9 ${\textcolor{green}{-16}}$ & 67.3 ${\textcolor{red}{+11}}$ & 46.7 ${\textcolor{green}{-10}}$ & 42.2 ${\textcolor{green}{-14}}$ \\ 
  & 84.5 & ViT-B-224 & 64.8 & 50.6 ${\textcolor{green}{-14}}$ & 48.3 ${\textcolor{green}{-17}}$ & 60.2 ${\textcolor{green}{-5}}$ & \textbf{34.1} ${\textcolor{green}{-31}}$ & 43.7 ${\textcolor{green}{-21}}$ & 34.8 ${\textcolor{green}{-30}}$ & 50.2 ${\textcolor{green}{-15}}$ & 68.5 ${\textcolor{red}{+4}}$ & 54.8 ${\textcolor{green}{-10}}$ & 53.5 ${\textcolor{green}{-11}}$ \\ 
  & 86.3 & Swinv2-B-256 & 66.3 & 58.7 ${\textcolor{green}{-8}}$ & 59.1 ${\textcolor{green}{-7}}$ & 62.0 ${\textcolor{green}{-4}}$ & 40.1 ${\textcolor{green}{-26}}$ & 42.7 ${\textcolor{green}{-24}}$ & \textbf{34.3} ${\textcolor{green}{-32}}$ & 50.3 ${\textcolor{green}{-16}}$ & 54.8 ${\textcolor{green}{-11}}$ & 47.5 ${\textcolor{green}{-19}}$ & 47.3 ${\textcolor{green}{-19}}$ \\ 
  & 86.7 & Deit3-B-384 & 72.9 & 71.1 ${\textcolor{green}{-2}}$ & 73.3 ${\textcolor{red}{+0}}$ & 68.6 ${\textcolor{green}{-4}}$ & \textbf{43.0} ${\textcolor{green}{-30}}$ & 43.6 ${\textcolor{green}{-29}}$ & 44.1 ${\textcolor{green}{-29}}$ & 64.4 ${\textcolor{green}{-9}}$ & 49.3 ${\textcolor{green}{-24}}$ & 47.2 ${\textcolor{green}{-26}}$ & 46.8 ${\textcolor{green}{-26}}$ \\ 
  & 85.7 & Deit3-B-224 & 75.1 & 72.8 ${\textcolor{green}{-2}}$ & 72.6 ${\textcolor{green}{-3}}$ & 69.5 ${\textcolor{green}{-6}}$ & 47.7 ${\textcolor{green}{-27}}$ & 48.7 ${\textcolor{green}{-26}}$ & \textbf{47.1} ${\textcolor{green}{-28}}$ & 67.5 ${\textcolor{green}{-8}}$ & 56.3 ${\textcolor{green}{-19}}$ & 52.9 ${\textcolor{green}{-22}}$ & 53.5 ${\textcolor{green}{-22}}$ \\ 
  & 86.3 & CnvNxt-B & 61.4 & 60.0 ${\textcolor{green}{-1}}$ & 67.0 ${\textcolor{red}{+6}}$ & 57.6 ${\textcolor{green}{-4}}$ & 31.0 ${\textcolor{green}{-30}}$ & 37.4 ${\textcolor{green}{-24}}$ & \textbf{27.5} ${\textcolor{green}{-34}}$ & 61.6 ${\textcolor{red}{+0}}$ & 47.0 ${\textcolor{green}{-14}}$ & 40.6 ${\textcolor{green}{-21}}$ & 39.7 ${\textcolor{green}{-22}}$ \\ 
  & 84.1 & CnvNxt-T & 62.9 & 57.2 ${\textcolor{green}{-6}}$ & 54.4 ${\textcolor{green}{-9}}$ & 61.6 ${\textcolor{green}{-1}}$ & 34.7 ${\textcolor{green}{-28}}$ & 42.2 ${\textcolor{green}{-21}}$ & \textbf{30.6} ${\textcolor{green}{-32}}$ & 52.9 ${\textcolor{green}{-10}}$ & 53.3 ${\textcolor{green}{-10}}$ & 49.1 ${\textcolor{green}{-14}}$ & 46.2 ${\textcolor{green}{-17}}$ \\ 
  & 82.3 & BiT-m & 69.7 & 62.2 ${\textcolor{green}{-7}}$ & 63.9 ${\textcolor{green}{-6}}$ & 62.6 ${\textcolor{green}{-7}}$ & 40.9 ${\textcolor{green}{-29}}$ & 42.1 ${\textcolor{green}{-28}}$ & \textbf{31.5} ${\textcolor{green}{-38}}$ & 60.2 ${\textcolor{green}{-10}}$ & 39.1 ${\textcolor{green}{-31}}$ & 36.0 ${\textcolor{green}{-34}}$ & 42.1 ${\textcolor{green}{-28}}$ \\ 
  & 85.6 & EffNetv2-M & 55.9 & 51.8 ${\textcolor{green}{-4}}$ & 56.3 ${\textcolor{red}{+0}}$ & 55.7 ${\textcolor{green}{-0}}$ & 48.6 ${\textcolor{green}{-7}}$ & 46.9 ${\textcolor{green}{-9}}$ & 40.9 ${\textcolor{green}{-15}}$ & 96.5 ${\textcolor{red}{+41}}$ & 55.3 ${\textcolor{green}{-1}}$ & \textbf{33.8} ${\textcolor{green}{-22}}$ & 42.4 ${\textcolor{green}{-14}}$ \\ 
 \hline 
{\multirow{12}{*}{{none}}} & 81.1 & ViT-B-384 & 70.0 & 64.5 ${\textcolor{green}{-5}}$ & 61.1 ${\textcolor{green}{-9}}$ & 65.0 ${\textcolor{green}{-5}}$ & 56.6 ${\textcolor{green}{-13}}$ & \textbf{56.2} ${\textcolor{green}{-14}}$ & 62.8 ${\textcolor{green}{-7}}$ & 59.7 ${\textcolor{green}{-10}}$ & 66.3 ${\textcolor{green}{-4}}$ & 63.0 ${\textcolor{green}{-7}}$ & 63.5 ${\textcolor{green}{-6}}$ \\ 
  & 84.6 & Swinv2-B-256 & 72.4 & 67.7 ${\textcolor{green}{-5}}$ & 68.2 ${\textcolor{green}{-4}}$ & 68.2 ${\textcolor{green}{-4}}$ & 58.9 ${\textcolor{green}{-13}}$ & \textbf{56.9} ${\textcolor{green}{-15}}$ & 57.6 ${\textcolor{green}{-15}}$ & 65.8 ${\textcolor{green}{-7}}$ & 67.8 ${\textcolor{green}{-5}}$ & 62.2 ${\textcolor{green}{-10}}$ & 60.5 ${\textcolor{green}{-12}}$ \\ 
  & 85.1 & Deit3-B-384 & 70.4 & 75.1 ${\textcolor{red}{+5}}$ & 85.4 ${\textcolor{red}{+15}}$ & 64.4 ${\textcolor{green}{-6}}$ & 59.3 ${\textcolor{green}{-11}}$ & 57.4 ${\textcolor{green}{-13}}$ & 51.5 ${\textcolor{green}{-19}}$ & 91.2 ${\textcolor{red}{+21}}$ & 70.7 ${\textcolor{red}{+0}}$ & 65.1 ${\textcolor{green}{-5}}$ & \textbf{49.2} ${\textcolor{green}{-21}}$ \\ 
  & 83.8 & Deit3-B-224 & 76.4 & 77.1 ${\textcolor{red}{+1}}$ & 83.3 ${\textcolor{red}{+7}}$ & 69.5 ${\textcolor{green}{-7}}$ & 62.3 ${\textcolor{green}{-14}}$ & 60.0 ${\textcolor{green}{-16}}$ & 57.9 ${\textcolor{green}{-18}}$ & 83.9 ${\textcolor{red}{+8}}$ & 75.7 ${\textcolor{green}{-1}}$ & 69.4 ${\textcolor{green}{-7}}$ & \textbf{55.8} ${\textcolor{green}{-21}}$ \\ 
  & 82.6 & XCiT-M-224 & 79.5 & 79.1 ${\textcolor{green}{-0}}$ & 82.4 ${\textcolor{red}{+3}}$ & 76.1 ${\textcolor{green}{-3}}$ & 71.6 ${\textcolor{green}{-8}}$ & 69.7 ${\textcolor{green}{-10}}$ & \textbf{69.2} ${\textcolor{green}{-10}}$ & 78.5 ${\textcolor{green}{-1}}$ & 76.6 ${\textcolor{green}{-3}}$ & 73.3 ${\textcolor{green}{-6}}$ & 73.0 ${\textcolor{green}{-7}}$ \\ 
  & 84.3 & XCiT-M-224-d & 72.6 & 71.7 ${\textcolor{green}{-1}}$ & 78.8 ${\textcolor{red}{+6}}$ & 66.6 ${\textcolor{green}{-6}}$ & 63.4 ${\textcolor{green}{-9}}$ & 60.8 ${\textcolor{green}{-12}}$ & \textbf{60.0} ${\textcolor{green}{-13}}$ & 75.3 ${\textcolor{red}{+3}}$ & 69.6 ${\textcolor{green}{-3}}$ & 62.7 ${\textcolor{green}{-10}}$ & 60.9 ${\textcolor{green}{-12}}$ \\ 
  & 84.4 & CnvNxt-B & 74.1 & 82.3 ${\textcolor{red}{+8}}$ & 94.5 ${\textcolor{red}{+20}}$ & 63.9 ${\textcolor{green}{-10}}$ & 59.3 ${\textcolor{green}{-15}}$ & 56.8 ${\textcolor{green}{-17}}$ & \textbf{56.2} ${\textcolor{green}{-18}}$ & 90.8 ${\textcolor{red}{+17}}$ & 65.7 ${\textcolor{green}{-8}}$ & 59.2 ${\textcolor{green}{-15}}$ & 58.0 ${\textcolor{green}{-16}}$ \\ 
  & 78.0 & BiT-s & 74.2 & 74.5 ${\textcolor{red}{+0}}$ & 76.5 ${\textcolor{red}{+2}}$ & 58.2 ${\textcolor{green}{-16}}$ & 83.2 ${\textcolor{red}{+9}}$ & \textbf{56.8} ${\textcolor{green}{-17}}$ & 64.4 ${\textcolor{green}{-10}}$ & 71.3 ${\textcolor{green}{-3}}$ & 81.3 ${\textcolor{red}{+7}}$ & 66.8 ${\textcolor{green}{-7}}$ & 77.2 ${\textcolor{red}{+3}}$ \\ 
  & 85.1 & EffNetv2-M & 70.0 & 69.5 ${\textcolor{green}{-1}}$ & 74.4 ${\textcolor{red}{+4}}$ & 65.3 ${\textcolor{green}{-5}}$ & 52.1 ${\textcolor{green}{-18}}$ & \textbf{51.4} ${\textcolor{green}{-19}}$ & 59.6 ${\textcolor{green}{-10}}$ & 61.7 ${\textcolor{green}{-8}}$ & 60.3 ${\textcolor{green}{-10}}$ & 56.6 ${\textcolor{green}{-13}}$ & 53.0 ${\textcolor{green}{-17}}$ \\ 
  & 84.9 & EffNetb7 & 69.0 & 70.5 ${\textcolor{red}{+2}}$ & 81.3 ${\textcolor{red}{+12}}$ & 62.5 ${\textcolor{green}{-7}}$ & 55.5 ${\textcolor{green}{-14}}$ & \textbf{50.4} ${\textcolor{green}{-19}}$ & 59.2 ${\textcolor{green}{-10}}$ & 71.0 ${\textcolor{red}{+2}}$ & 61.7 ${\textcolor{green}{-7}}$ & 58.0 ${\textcolor{green}{-11}}$ & 50.4 ${\textcolor{green}{-19}}$ \\ 
  & 77.7 & EffNet-B0 & 75.0 & 75.9 ${\textcolor{red}{+1}}$ & 84.0 ${\textcolor{red}{+9}}$ & 68.7 ${\textcolor{green}{-6}}$ & 71.0 ${\textcolor{green}{-4}}$ & 66.8 ${\textcolor{green}{-8}}$ & 62.2 ${\textcolor{green}{-13}}$ & 75.0 ${\textcolor{red}{+0}}$ & 85.8 ${\textcolor{red}{+11}}$ & \textbf{58.7} ${\textcolor{green}{-16}}$ & 62.8 ${\textcolor{green}{-12}}$ \\ 
  & 80.4 & ResNet50 & 76.0 & 76.6 ${\textcolor{red}{+1}}$ & 77.5 ${\textcolor{red}{+1}}$ & 69.0 ${\textcolor{green}{-7}}$ & 77.0 ${\textcolor{red}{+1}}$ & 66.4 ${\textcolor{green}{-10}}$ & 75.1 ${\textcolor{green}{-1}}$ & 94.8 ${\textcolor{red}{+19}}$ & 64.0 ${\textcolor{green}{-12}}$ & 57.6 ${\textcolor{green}{-18}}$ & \textbf{56.6} ${\textcolor{green}{-19}}$ \\ 
 \hline 
{\multirow{1}{*}{{JFT}}} & 86.8 & EffNetb7-ns & 71.3 & 64.8 ${\textcolor{green}{-7}}$ & 67.5 ${\textcolor{green}{-4}}$ & 66.5 ${\textcolor{green}{-5}}$ & 83.7 ${\textcolor{red}{+12}}$ & 72.0 ${\textcolor{red}{+1}}$ & 85.2 ${\textcolor{red}{+14}}$ & 65.8 ${\textcolor{green}{-6}}$ & 70.3 ${\textcolor{green}{-1}}$ & 64.2 ${\textcolor{green}{-7}}$ & \textbf{63.8} ${\textcolor{green}{-7}}$ \\ 
 \hline 
{\multirow{2}{*}{{\shortstack[l]{clip\\+12k}}}} & 87.2 & ViT-B-384-l2b & 53.7 & 51.1 ${\textcolor{green}{-3}}$ & 55.9 ${\textcolor{red}{+2}}$ & 52.7 ${\textcolor{green}{-1}}$ & 37.8 ${\textcolor{green}{-16}}$ & 40.2 ${\textcolor{green}{-14}}$ & \textbf{31.7} ${\textcolor{green}{-22}}$ & 47.3 ${\textcolor{green}{-6}}$ & 41.1 ${\textcolor{green}{-13}}$ & 37.3 ${\textcolor{green}{-16}}$ & 37.0 ${\textcolor{green}{-17}}$ \\ 
  & 87.0 & ViT-B-384-oai & 56.0 & 51.8 ${\textcolor{green}{-4}}$ & 54.6 ${\textcolor{green}{-1}}$ & 53.6 ${\textcolor{green}{-2}}$ & 40.9 ${\textcolor{green}{-15}}$ & 39.8 ${\textcolor{green}{-16}}$ & 36.9 ${\textcolor{green}{-19}}$ & 50.4 ${\textcolor{green}{-6}}$ & 36.6 ${\textcolor{green}{-19}}$ & \textbf{33.8} ${\textcolor{green}{-22}}$ & 34.1 ${\textcolor{green}{-22}}$ \\ 
 \hline 
{\multirow{2}{*}{{clip}}} & 86.6 & ViT-B-384-l2b & 65.8 & 63.5 ${\textcolor{green}{-2}}$ & 62.5 ${\textcolor{green}{-3}}$ & 59.0 ${\textcolor{green}{-7}}$ & 49.6 ${\textcolor{green}{-16}}$ & 50.4 ${\textcolor{green}{-15}}$ & \textbf{46.1} ${\textcolor{green}{-20}}$ & 61.0 ${\textcolor{green}{-5}}$ & 53.8 ${\textcolor{green}{-12}}$ & 49.5 ${\textcolor{green}{-16}}$ & 48.3 ${\textcolor{green}{-18}}$ \\ 
  & 86.2 & ViT-B-384-oai & 65.8 & 64.1 ${\textcolor{green}{-2}}$ & 67.7 ${\textcolor{red}{+2}}$ & 62.4 ${\textcolor{green}{-3}}$ & 52.4 ${\textcolor{green}{-13}}$ & 54.7 ${\textcolor{green}{-11}}$ & \textbf{48.1} ${\textcolor{green}{-18}}$ & 65.4 ${\textcolor{green}{-0}}$ & 57.1 ${\textcolor{green}{-9}}$ & 53.9 ${\textcolor{green}{-12}}$ & 53.4 ${\textcolor{green}{-12}}$ \\ 
 \hline 
{\multirow{2}{*}{\shortstack[l]{clip\\z. shot}}} & 74.3 & clip-ViT-L-336 & ---- & ---- & ---- & ---- & ---- & ---- & ---- & ---- & ---- & \textbf{55.7} & 55.8 \\ 
  & 66.6 & clip-ViT-B-224 & ---- & ---- & ---- & ---- & ---- & ---- & ---- & ---- & ---- & \textbf{56.9} & 62.8 \\ 
 \hline 
\end{tabular}
\end{center}

\end{smaller}
\vskip \belowtablevskip
\end{table*}

\renewcommand{\arraystretch}{1.}
\tabcolsep=0.107cm
\begin{table*}[htb]
    \caption{Mean \FPR{} for classes with 21k overlap.}
    \centering
    \small
    \begin{smaller}
\begin{center}
\begin{tabular}{l l l l l l l l l l l l l l }
pre & acc. & model & MSP & MaxL & Ener & KL-M & Maha & RMaha & ViM & E+R & KNN & Cos & MCM/RCos \\ 
 \hline 
{\multirow{9}{*}{{21k}}} & 86.0 & ViT-B-384 & 51.1 & 37.2 ${\textcolor{green}{-14}}$ & 36.5 ${\textcolor{green}{-15}}$ & 50.1 ${\textcolor{green}{-1}}$ & \textbf{26.8} ${\textcolor{green}{-24}}$ & 30.2 ${\textcolor{green}{-21}}$ & 32.7 ${\textcolor{green}{-18}}$ & 38.1 ${\textcolor{green}{-13}}$ & 61.9 ${\textcolor{red}{+11}}$ & 45.9 ${\textcolor{green}{-5}}$ & 45.5 ${\textcolor{green}{-6}}$ \\ 
  & 84.5 & ViT-B-224 & 56.8 & 45.8 ${\textcolor{green}{-11}}$ & 45.7 ${\textcolor{green}{-11}}$ & 56.7 ${\textcolor{green}{-0}}$ & \textbf{31.6} ${\textcolor{green}{-25}}$ & 35.7 ${\textcolor{green}{-21}}$ & 39.0 ${\textcolor{green}{-18}}$ & 49.3 ${\textcolor{green}{-8}}$ & 68.9 ${\textcolor{red}{+12}}$ & 54.6 ${\textcolor{green}{-2}}$ & 54.4 ${\textcolor{green}{-2}}$ \\ 
  & 86.3 & Swinv2-B-256 & 48.6 & 38.2 ${\textcolor{green}{-10}}$ & 36.9 ${\textcolor{green}{-12}}$ & 55.0 ${\textcolor{red}{+6}}$ & 66.5 ${\textcolor{red}{+18}}$ & 55.7 ${\textcolor{red}{+7}}$ & 58.2 ${\textcolor{red}{+10}}$ & \textbf{35.3} ${\textcolor{green}{-13}}$ & 63.1 ${\textcolor{red}{+15}}$ & 52.0 ${\textcolor{red}{+3}}$ & 48.3 ${\textcolor{green}{-0}}$ \\ 
  & 86.7 & Deit3-B-384 & 60.0 & 53.5 ${\textcolor{green}{-6}}$ & 53.6 ${\textcolor{green}{-6}}$ & 59.0 ${\textcolor{green}{-1}}$ & 55.7 ${\textcolor{green}{-4}}$ & 49.6 ${\textcolor{green}{-10}}$ & 59.0 ${\textcolor{green}{-1}}$ & 49.5 ${\textcolor{green}{-10}}$ & 54.1 ${\textcolor{green}{-6}}$ & 48.6 ${\textcolor{green}{-11}}$ & \textbf{47.8} ${\textcolor{green}{-12}}$ \\ 
  & 85.7 & Deit3-B-224 & 63.1 & 57.0 ${\textcolor{green}{-6}}$ & 55.8 ${\textcolor{green}{-7}}$ & 64.5 ${\textcolor{red}{+1}}$ & 62.0 ${\textcolor{green}{-1}}$ & 54.7 ${\textcolor{green}{-8}}$ & 65.0 ${\textcolor{red}{+2}}$ & 53.1 ${\textcolor{green}{-10}}$ & 59.1 ${\textcolor{green}{-4}}$ & 54.4 ${\textcolor{green}{-9}}$ & \textbf{53.1} ${\textcolor{green}{-10}}$ \\ 
  & 86.3 & CnvNxt-B & 44.9 & 38.0 ${\textcolor{green}{-7}}$ & 39.4 ${\textcolor{green}{-5}}$ & 54.4 ${\textcolor{red}{+10}}$ & 52.6 ${\textcolor{red}{+8}}$ & 43.2 ${\textcolor{green}{-2}}$ & 43.8 ${\textcolor{green}{-1}}$ & \textbf{37.0} ${\textcolor{green}{-8}}$ & 52.6 ${\textcolor{red}{+8}}$ & 44.8 ${\textcolor{green}{-0}}$ & 43.0 ${\textcolor{green}{-2}}$ \\ 
  & 84.1 & CnvNxt-T & 53.3 & 46.4 ${\textcolor{green}{-7}}$ & 44.0 ${\textcolor{green}{-9}}$ & 60.6 ${\textcolor{red}{+7}}$ & 48.9 ${\textcolor{green}{-4}}$ & 46.4 ${\textcolor{green}{-7}}$ & \textbf{38.5} ${\textcolor{green}{-15}}$ & 42.7 ${\textcolor{green}{-11}}$ & 57.1 ${\textcolor{red}{+4}}$ & 51.5 ${\textcolor{green}{-2}}$ & 49.7 ${\textcolor{green}{-4}}$ \\ 
  & 82.3 & BiT-m & 67.5 & 62.0 ${\textcolor{green}{-6}}$ & 63.0 ${\textcolor{green}{-4}}$ & 65.3 ${\textcolor{green}{-2}}$ & 51.5 ${\textcolor{green}{-16}}$ & 45.6 ${\textcolor{green}{-22}}$ & \textbf{42.2} ${\textcolor{green}{-25}}$ & 56.6 ${\textcolor{green}{-11}}$ & 61.1 ${\textcolor{green}{-6}}$ & 54.2 ${\textcolor{green}{-13}}$ & 56.5 ${\textcolor{green}{-11}}$ \\ 
  & 85.6 & EffNetv2-M & 49.9 & 47.8 ${\textcolor{green}{-2}}$ & 53.8 ${\textcolor{red}{+4}}$ & 54.4 ${\textcolor{red}{+4}}$ & 65.3 ${\textcolor{red}{+15}}$ & 52.4 ${\textcolor{red}{+2}}$ & 55.6 ${\textcolor{red}{+6}}$ & 88.7 ${\textcolor{red}{+39}}$ & 69.5 ${\textcolor{red}{+20}}$ & \textbf{47.3} ${\textcolor{green}{-3}}$ & 51.9 ${\textcolor{red}{+2}}$ \\ 
 \hline 
{\multirow{12}{*}{{none}}} & 81.1 & ViT-B-384 & 69.4 & 68.2 ${\textcolor{green}{-1}}$ & 69.3 ${\textcolor{green}{-0}}$ & 67.0 ${\textcolor{green}{-2}}$ & 60.6 ${\textcolor{green}{-9}}$ & \textbf{57.2} ${\textcolor{green}{-12}}$ & 70.4 ${\textcolor{red}{+1}}$ & 66.8 ${\textcolor{green}{-3}}$ & 74.8 ${\textcolor{red}{+5}}$ & 69.6 ${\textcolor{red}{+0}}$ & 70.8 ${\textcolor{red}{+1}}$ \\ 
  & 84.6 & Swinv2-B-256 & 69.5 & 67.6 ${\textcolor{green}{-2}}$ & 72.9 ${\textcolor{red}{+3}}$ & 67.4 ${\textcolor{green}{-2}}$ & 64.7 ${\textcolor{green}{-5}}$ & \textbf{60.6} ${\textcolor{green}{-9}}$ & 67.9 ${\textcolor{green}{-2}}$ & 69.3 ${\textcolor{green}{-0}}$ & 69.5 ${\textcolor{green}{-0}}$ & 63.7 ${\textcolor{green}{-6}}$ & 62.3 ${\textcolor{green}{-7}}$ \\ 
  & 85.1 & Deit3-B-384 & 66.8 & 72.4 ${\textcolor{red}{+6}}$ & 87.9 ${\textcolor{red}{+21}}$ & 64.6 ${\textcolor{green}{-2}}$ & 64.8 ${\textcolor{green}{-2}}$ & 59.7 ${\textcolor{green}{-7}}$ & 61.3 ${\textcolor{green}{-5}}$ & 90.0 ${\textcolor{red}{+23}}$ & 75.0 ${\textcolor{red}{+8}}$ & 67.5 ${\textcolor{red}{+1}}$ & \textbf{58.1} ${\textcolor{green}{-9}}$ \\ 
  & 83.8 & Deit3-B-224 & 69.3 & 71.1 ${\textcolor{red}{+2}}$ & 82.1 ${\textcolor{red}{+13}}$ & 68.2 ${\textcolor{green}{-1}}$ & 70.1 ${\textcolor{red}{+1}}$ & 64.9 ${\textcolor{green}{-4}}$ & 64.4 ${\textcolor{green}{-5}}$ & 83.0 ${\textcolor{red}{+14}}$ & 81.2 ${\textcolor{red}{+12}}$ & 73.6 ${\textcolor{red}{+4}}$ & \textbf{62.9} ${\textcolor{green}{-6}}$ \\ 
  & 82.6 & XCiT-M-224 & 71.5 & 72.3 ${\textcolor{red}{+1}}$ & 78.6 ${\textcolor{red}{+7}}$ & 71.1 ${\textcolor{green}{-0}}$ & 65.4 ${\textcolor{green}{-6}}$ & \textbf{62.5} ${\textcolor{green}{-9}}$ & 64.2 ${\textcolor{green}{-7}}$ & 76.0 ${\textcolor{red}{+4}}$ & 71.1 ${\textcolor{green}{-0}}$ & 66.1 ${\textcolor{green}{-5}}$ & 64.9 ${\textcolor{green}{-7}}$ \\ 
  & 84.3 & XCiT-M-224-d & 67.6 & 65.2 ${\textcolor{green}{-2}}$ & 72.2 ${\textcolor{red}{+5}}$ & 66.9 ${\textcolor{green}{-1}}$ & 66.9 ${\textcolor{green}{-1}}$ & \textbf{62.1} ${\textcolor{green}{-6}}$ & 62.7 ${\textcolor{green}{-5}}$ & 72.0 ${\textcolor{red}{+4}}$ & 70.6 ${\textcolor{red}{+3}}$ & 64.9 ${\textcolor{green}{-3}}$ & 62.9 ${\textcolor{green}{-5}}$ \\ 
  & 84.4 & CnvNxt-B & 63.2 & 69.7 ${\textcolor{red}{+7}}$ & 88.2 ${\textcolor{red}{+25}}$ & 68.7 ${\textcolor{red}{+6}}$ & 66.8 ${\textcolor{red}{+4}}$ & 61.2 ${\textcolor{green}{-2}}$ & 67.0 ${\textcolor{red}{+4}}$ & 85.1 ${\textcolor{red}{+22}}$ & 71.2 ${\textcolor{red}{+8}}$ & 61.7 ${\textcolor{green}{-2}}$ & \textbf{58.7} ${\textcolor{green}{-5}}$ \\ 
  & 78.0 & BiT-s & 79.6 & 82.3 ${\textcolor{red}{+3}}$ & 83.9 ${\textcolor{red}{+4}}$ & 70.0 ${\textcolor{green}{-10}}$ & 83.6 ${\textcolor{red}{+4}}$ & \textbf{65.3} ${\textcolor{green}{-14}}$ & 75.0 ${\textcolor{green}{-5}}$ & 78.9 ${\textcolor{green}{-1}}$ & 83.5 ${\textcolor{red}{+4}}$ & 73.0 ${\textcolor{green}{-7}}$ & 85.3 ${\textcolor{red}{+6}}$ \\ 
  & 85.1 & EffNetv2-M & 64.5 & 64.6 ${\textcolor{red}{+0}}$ & 74.6 ${\textcolor{red}{+10}}$ & 62.4 ${\textcolor{green}{-2}}$ & 64.2 ${\textcolor{green}{-0}}$ & 55.5 ${\textcolor{green}{-9}}$ & 74.7 ${\textcolor{red}{+10}}$ & 70.9 ${\textcolor{red}{+6}}$ & 65.1 ${\textcolor{red}{+1}}$ & 60.0 ${\textcolor{green}{-4}}$ & \textbf{54.6} ${\textcolor{green}{-10}}$ \\ 
  & 84.9 & EffNetb7 & 66.4 & 68.7 ${\textcolor{red}{+2}}$ & 81.6 ${\textcolor{red}{+15}}$ & 62.7 ${\textcolor{green}{-4}}$ & 70.2 ${\textcolor{red}{+4}}$ & 55.3 ${\textcolor{green}{-11}}$ & 74.9 ${\textcolor{red}{+8}}$ & 77.2 ${\textcolor{red}{+11}}$ & 67.7 ${\textcolor{red}{+1}}$ & 61.0 ${\textcolor{green}{-5}}$ & \textbf{54.3} ${\textcolor{green}{-12}}$ \\ 
  & 77.7 & EffNet-B0 & 71.6 & 71.9 ${\textcolor{red}{+0}}$ & 78.9 ${\textcolor{red}{+7}}$ & 72.8 ${\textcolor{red}{+1}}$ & 85.3 ${\textcolor{red}{+14}}$ & 75.2 ${\textcolor{red}{+4}}$ & 77.3 ${\textcolor{red}{+6}}$ & 75.1 ${\textcolor{red}{+4}}$ & 87.1 ${\textcolor{red}{+16}}$ & \textbf{61.8} ${\textcolor{green}{-10}}$ & 70.9 ${\textcolor{green}{-1}}$ \\ 
  & 80.4 & ResNet50 & 71.8 & 73.9 ${\textcolor{red}{+2}}$ & 78.0 ${\textcolor{red}{+6}}$ & 69.0 ${\textcolor{green}{-3}}$ & 87.3 ${\textcolor{red}{+16}}$ & 70.0 ${\textcolor{green}{-2}}$ & 79.2 ${\textcolor{red}{+7}}$ & 97.9 ${\textcolor{red}{+26}}$ & 80.2 ${\textcolor{red}{+8}}$ & 63.8 ${\textcolor{green}{-8}}$ & \textbf{63.0} ${\textcolor{green}{-9}}$ \\ 
 \hline 
{\multirow{1}{*}{{JFT}}} & 86.8 & EffNetb7-ns & 61.8 & \textbf{54.2} ${\textcolor{green}{-8}}$ & 60.5 ${\textcolor{green}{-1}}$ & 64.1 ${\textcolor{red}{+2}}$ & 88.0 ${\textcolor{red}{+26}}$ & 68.2 ${\textcolor{red}{+6}}$ & 89.8 ${\textcolor{red}{+28}}$ & 61.1 ${\textcolor{green}{-1}}$ & 74.3 ${\textcolor{red}{+13}}$ & 65.4 ${\textcolor{red}{+4}}$ & 63.7 ${\textcolor{red}{+2}}$ \\ 
 \hline 
{\multirow{2}{*}{{\shortstack[l]{clip\\+12k}}}} & 87.2 & ViT-B-384-l2b & 49.6 & 46.8 ${\textcolor{green}{-3}}$ & 49.4 ${\textcolor{green}{-0}}$ & 52.1 ${\textcolor{red}{+3}}$ & 55.0 ${\textcolor{red}{+5}}$ & 48.5 ${\textcolor{green}{-1}}$ & 48.1 ${\textcolor{green}{-2}}$ & 44.5 ${\textcolor{green}{-5}}$ & 46.2 ${\textcolor{green}{-3}}$ & \textbf{40.6} ${\textcolor{green}{-9}}$ & 40.7 ${\textcolor{green}{-9}}$ \\ 
  & 87.0 & ViT-B-384-oai & 47.7 & 42.3 ${\textcolor{green}{-5}}$ & 42.3 ${\textcolor{green}{-5}}$ & 48.9 ${\textcolor{red}{+1}}$ & 60.4 ${\textcolor{red}{+13}}$ & 49.8 ${\textcolor{red}{+2}}$ & 55.1 ${\textcolor{red}{+7}}$ & 40.9 ${\textcolor{green}{-7}}$ & 46.3 ${\textcolor{green}{-1}}$ & 40.2 ${\textcolor{green}{-7}}$ & \textbf{39.9} ${\textcolor{green}{-8}}$ \\ 
 \hline 
{\multirow{2}{*}{{clip}}} & 86.6 & ViT-B-384-l2b & 61.2 & 61.3 ${\textcolor{red}{+0}}$ & 66.4 ${\textcolor{red}{+5}}$ & 57.3 ${\textcolor{green}{-4}}$ & 53.2 ${\textcolor{green}{-8}}$ & 50.5 ${\textcolor{green}{-11}}$ & 52.6 ${\textcolor{green}{-9}}$ & 63.6 ${\textcolor{red}{+2}}$ & 57.5 ${\textcolor{green}{-4}}$ & 51.0 ${\textcolor{green}{-10}}$ & \textbf{49.2} ${\textcolor{green}{-12}}$ \\ 
  & 86.2 & ViT-B-384-oai & 64.7 & 65.1 ${\textcolor{red}{+0}}$ & 70.1 ${\textcolor{red}{+5}}$ & 61.7 ${\textcolor{green}{-3}}$ & 56.3 ${\textcolor{green}{-8}}$ & \textbf{53.6} ${\textcolor{green}{-11}}$ & 58.3 ${\textcolor{green}{-6}}$ & 67.7 ${\textcolor{red}{+3}}$ & 62.0 ${\textcolor{green}{-3}}$ & 57.0 ${\textcolor{green}{-8}}$ & 54.4 ${\textcolor{green}{-10}}$ \\ 
 \hline 
{\multirow{2}{*}{\shortstack[l]{clip\\z. shot}}} & 74.3 & clip-ViT-L-336 & ---- & ---- & ---- & ---- & ---- & ---- & ---- & ---- & ---- & 75.2 & \textbf{68.9} \\ 
  & 66.6 & clip-ViT-B-224 & ---- & ---- & ---- & ---- & ---- & ---- & ---- & ---- & ---- & 82.8 & \textbf{82.6} \\ 
 \hline 
\end{tabular}
\end{center}

\end{smaller}
\vskip \belowtablevskip
\end{table*}

\end{document}